%% file: main.tex
\documentclass[11pt]{article}
\usepackage{mathrsfs}
\usepackage[centertags]{amsmath}
\usepackage{amsfonts}
\usepackage{amssymb}
\usepackage{amsthm}
\usepackage{cases}
\usepackage{epsfig}
\usepackage {graphicx}
\usepackage{color}
\usepackage{CJK}
\usepackage{hyperref}
\usepackage{algorithm}
\usepackage{algorithmicx}
\usepackage{algpseudocode}
\usepackage{natbib}
\usepackage{comment}
\usepackage{pdfpages}
\usepackage{multirow}
\usepackage{tcolorbox}
\usepackage{booktabs}
\usepackage{siunitx}
\usepackage{multirow}
\usepackage[normalem]{ulem}

\usepackage[shortlabels]{enumitem}

\usepackage{anysize}

\marginsize{1.1in}{1.1in}{0.55in}{0.8in} %

\newtheorem{theorem}{Theorem}

\newtheorem{lemma}{Lemma}
\newtheorem{condition}{Condition}

\newtheorem{remark}{Remark}

\renewcommand{\theequation}{
\arabic{equation}}

\def\bTheta{\boldsymbol{\Theta}}

\def\hA{\hat{A}}
\def\calF{\mathcal{F}}
\def\calC{\mathcal{C}}

\def\hddz{\hat{\ddot{z}}}
\def\calI{\mathcal{I}}

\def\hv{\hat{v}}
\def\hd{\hat{d}}

\def\tbW{\tilde{\bW}}

\def\bDelta{\boldsymbol{\Delta}}
\def\tdbW{\tilde{\dot{\bW}}}

\def\cd{\check{d}}
\def\bE{\mathbf{E}}
\def\tcalS{\bar{\mathcal{S}}}

\renewcommand{\hat}{\widehat}
\renewcommand{\tilde}{\widetilde}
\renewcommand{\ldots}{\cdots}
\renewcommand{\dots}{\cdots}

\usepackage{letltxmacro}
\makeatletter
\let\oldr@@t\r@@t
\def\r@@t#1#2{%
\setbox0=\hbox{$\oldr@@t#1{#2\,}$}\dimen0=\ht0
\advance\dimen0-0.2\ht0
\setbox2=\hbox{\vrule height\ht0 depth -\dimen0}%
{\box0\lower0.4pt\box2}}
\LetLtxMacro{\oldsqrt}{\sqrt}
\renewcommand*{\sqrt}[2][\ ]{\oldsqrt[#1]{#2}}
\makeatother

\begin{document}
\include{definition}

\title{MOSAIC: Minimax-Optimal Sparsity-Adaptive Inference for Change Points in Dynamic Networks%
\thanks{
Yingying Fan is Centennial Chair in Business Administration and Professor, Data Sciences and Operations Department, Marshall School of Business, University of Southern California, Los Angeles, CA 90089 (E-mail: \textit{fanyingy@marshall.usc.edu}). %
Jingyuan Liu is Professor, 
Department of Statistics and Data Science, 
and Wang Yanan Institute for Studies in Economics, 
Xiamen University, China (E-mail: \textit{jingyuan@xmu.edu.cn}). %
Jinchi Lv is Kenneth King Stonier Chair in Business Administration and Professor, Data Sciences and Operations Department, Marshall School of Business, University of Southern California, Los Angeles, CA 90089 (E-mail: \textit{jinchilv@marshall.usc.edu}). %
Ao Sun is Postdoctoral Scholar, Data Sciences and Operations Department, Marshall School of Business, University of Southern California, Los Angeles, CA 90089 (E-mail: \textit{ao.sun@marshall.usc.edu}). %
}
\date{September 7, 2025}
\author{Yingying Fan$^1$, Jingyuan Liu$^2$, Jinchi Lv$^1$ and Ao Sun$^1$
\medskip\\
University of Southern California$^1$ and Xiamen University$^2$
\\
} 
}

\maketitle

\begin{abstract}
We propose a new inference framework, named MOSAIC, for change-point detection in dynamic networks with the simultaneous low-rank and sparse-change structure. We establish the minimax rate of detection boundary, which relies on the sparsity of changes. We then develop an eigen-decomposition-based test with screened signals that approaches the minimax rate in theory, with only a minor logarithmic loss. For practical implementation of MOSAIC, we adjust the theoretical test by a novel residual-based technique, resulting in a pivotal statistic that converges to a standard normal distribution via the martingale central limit theorem under the null hypothesis and achieves full power under the alternative hypothesis. We also analyze the minimax rate of testing boundary for dynamic networks without the low-rank structure, which almost aligns with the results in high-dimensional mean-vector change-point inference. We showcase the effectiveness of MOSAIC and verify our theoretical results with several simulation examples and a real data application.
\end{abstract}

\textit{Running title}: MOSAIC


\section{Introduction} \label{intro}


Dynamic networks have been exploited extensively in economics \citep{graham2015methods, bonhomme2019distributional}, finance \citep{bonaccolto2022dynamic}, medical science \citep{simpson2013analyzing, wang2021network}, social science \citep{ji2022co, chen2022monitoring}, and beyond. These networks are commonly described as sequences of undirected graphs, $\mathcal{G}_t = (\calV, \mathcal{E}_t)$ with $t = 1,\ldots, T$, where $\calV = \{1, \ldots, n\}$ denotes a fixed set of nodes, and  $\mathcal{E}_t$ represents a time-varying set of edges. 
Denote the adjacency matrix at time $t$ as $\bX^{(t)} = (x_{i,j}^{(t)})_{1 \le i,j \le n}$ with mean matrix $\bTheta^{(t)} = (\theta_{i,j}^{(t)})_{1 \le i, j \le n}$, where $x_{i,j}^{(t)} = 1$ if an edge exists between nodes $i$ and $j$, and $x_{i,j}^{(t)} = x_{j,i}^{(t)}  \stackrel{i.i.d.}{\sim} \operatorname{Bernoulli}(\theta_{i,j}^{(t)})$
for $(i,j) \in \Omega$ with $\Omega:=\{(i,j): 1 \le i < j \le n \}$ and $t = 1,\ldots, T$. Here, we assume that $T \ll n$.

An intriguing question when modeling dynamic networks is whether a change point $\tau^*$ exists, in the sense that $\bTheta^{(t)} = \bTheta_{1}=(\theta_{1,i,j})_{1 \le i, j \le n}$ for $1 \le t \le \tau^*$ and $\bTheta^{(t)} = \bTheta_{2}=(\theta_{2,i,j})_{1 \le i, j \le n}$ for $\tau^*+1 \le t \le T$. For example, large-scale protests may create significant changes in the activist networks. Likewise, when social media platforms change their recommendation algorithms, user engagement patterns may shift abruptly, altering the structure of the interaction networks. Previous studies have explored this problem mainly under the low-rank network structure, i.e., $\Rank(\bTheta^{(t)}) = K_t^*$ with $t \in [T]$ for some fixed constant $K_t^*$.
Specifically, \cite{wang2013locality} proposed a class of locality statistics for a special class of low-rank networks---the dynamic stochastic block model with fixed membership and altered connectivity matrices. The proposed statistics converge to Gumbel distributions, which contain many unknown parameters, and its performance relies heavily on the model assumptions. 
Under similar settings, \cite{cribben2017estimating} employed a stationary bootstrap method to test the significance of change points, which is computationally expensive and  without any theoretical guarantee. For general dynamic networks with the low-rank structure, \cite{enikeeva2021change} provided a theoretical test statistic for detecting change points based on the operator norm. However, their work focuses mainly on the theoretical exploration of the minimax detection boundary, with no asymptotic distribution of the proposed statistic or guidance of practical implementation. 

Another notable issue with existing testing procedures is their reliance on relatively strong signals to achieve non-trivial power. One key reason is that they do not account for sparsity in change points, which is characterized by the number of nonzero entries in the difference matrix $\bDelta := \bTheta_1 - \bTheta_2$. The sparsity plays a crucial role in high-dimensional learning and inference. In high-dimensional mean-vector change-point inference, \cite{liu2021minimax} emphasized that the sparsity level influences significantly the optimal signal strength, and suggested an approximate linear minimax detection boundary with respect to the sparsity level. For network inference, the sparse-change assumption is also reasonable. For instance, when certain political activists are shadow-banned on a social media platform, their connections decrease merely in a localized part of the network, while the major structure remains largely unchanged. 
Under such assumptions, the detection boundaries in the existing literature could be suboptimal. 

In this paper, we bridge such gap and propose a new framework of minimax-optimal sparsity-adaptive inference for change points (MOSAIC) for a broad class of dynamic networks that possess simultaneous low-rank and sparse-change structure. We rigorously quantify the detection difficulty of change points based on three key characteristics of dynamic networks: the change magnitude calculated from $\bDelta$, the change sparsity, and the network ranks. To obtain a sharp detection boundary, we carefully derive the required minimax rate of change magnitude, which relies on the change sparsity and network ranks. Then we establish the new inference framework MOSAIC for the detection of change points in dynamic networks. MOSAIC comprises a theoretical test that attains the minimax-optimality, and a novel residual-based pivotal empirical test that is readily applicable in practice.

The novelty of our work is threefold. First, to the best of our knowledge, MOSAIC is the first inference framework for network analysis that \textit{simultaneously} accounts for the low-rankness and change sparsity. The minimax-optimality for the detection boundary is established rigorously as a function of the change-sparsity level, where the required signal strength is \textit{much weaker} than when considering \textit{only} the low-rankness or change sparsity. We also develop a theoretical test based on the eigen-decompositions of mean matrices that achieves the minimax rate. Second, from the practical perspective, given that no tests are directly implementable even solely under the low-rank assumption, we adjust the theoretical test statistic by a newly advocated residual-based technique coupled with a signal selection procedure, and propose a pivotal MOSAIC test statistic that is \textit{tuning-free} and \textit{robust} to misspecification of network ranks. The MOSAIC statistic converges to a standard normal distribution under the null hypothesis according to the martingale central limit theorem. The newly suggested test statistic attains nearly the minimax-optimality under the low-rank and sparse-change structure. Third, we establish parallel results \textit{without} the low-rank assumption, closely aligning with findings in high-dimensional mean-vector change-point inference.

The rest of the paper is organized as follows. Section \ref{bound_result} investigates the minimax-optimal rate of the detection boundary for dynamic network change-point detection and introduces a theoretical MOSAIC test. We suggest an empirical MOSAIC test and establish its size and power analysis in Section \ref{sec:inference}. Section \ref{new.sec.mosaic.withoutlowrank} further examines the testing problem without the low-rank structure. We present several simulation examples and a real data application in Sections \ref{new.sec.simustud} and \ref{sec.realdata}, respectively. Section \ref{new.sec.discu} discusses some implications and extensions of our work. The Supplementary Material provides all the proofs and technical details.

\subsection{Notation} \label{notation}

For a set $\calA$, denote by $|\calA|$ its cardinality. Given a vector $\bx = (x_1, \ldots, x_p)^\top \in \R^p$, write the 
$\ell_q$-norm as $\| \bx\|_q := (\sum_{j=1}^p |x_j|^q)^{1/q}$ for $1 \le q < \infty$, the $\ell_{\infty}$-norm as $\| \bx\|_{\infty} := \max_{1 \le j \le p} |x_j|$, and the $\ell_0$-norm as $\| \bx\|_{0} := \sum_{1 \le j \le p} \I(x_i \ne 0)$. $\mathbf{1}_n$ and $\mathbf{I}_n$ denote an $n$-dimensional vector of all components being $1$'s and an $n \times n$ identity matrix, respectively. For a symmetric matrix $\bX = (x_{i,j})_{1 \le i,j \le n}$, denote by $d_k(\bX)$ the $k$th largest eigenvalue, $\|\bX \|_F=\sqrt{\sum_{1 \le i,j \le n} x_{i,j}^2}$ the Frobenius norm, $\| \bX\|_2=d_1(\bX)$ the operator norm, $\|\bX\|_0=\sum_{1 \le i,j \le n} \I(x_{i,j} \ne 0)$ the $\ell_0$-norm, and $\| \bX\|_{\infty}:= \max_{1 \le i,j \le n} |x_{i,j}|$ the entrywise maximum norm. For each matrix $\bU$,  $|\text{rowsupp}(\bU)|$ represents the number of nonzero rows of $\bU$. For $a, b \in \R$, $a \lesssim b$, $a \gtrsim b$, and $a \asymp b$ refer to $a \le c_1 b$, $a \ge c_2 b$, and $a = c_3b$, respectively, for some positive constants $c_1$, $c_2$, and $c_3$. $a \gg b$ means that $b/a \to 0$. The function $\log(\cdot)$ stands for the natural logarithm, and $\log_c(\cdot)$ stands for the logarithm with base $c>0$. For an integer $k$, $[k]$ represents the set $\{1,\ldots,k\}$.

\section{Minimax-optimal rate of detection boundary} \label{bound_result}

In this section, we characterize the minimax-optimal detection boundary for MOSAIC under the simultaneous low-rank and sparse-change structure with one change point.

\subsection{Formulation of the testing problem} \label{new.sec2.1}

Let us first describe the parameter spaces for the testing problem. For the rank-$K^*_t$ mean matrix $\bTheta^{(t)}$ introduced in Section \ref{intro}, let
$\bTheta^0 = {\bTheta^{(t)}}_{t=1}^T$, $K^* = \max_{t \in [T]} K_t^*$,
and $\rho_n := \max_{t \in [T]}\max_{1 \le i,j \le n} (\theta_{i,j}^{(t)})$ denote the network sparsity parameter, which controls the maximum edge probability and is assumed to vanish as the network size $n$ increases. We stress that $\rho_n$ is distinct from the change-sparsity parameter to be introduced later. We further impose an eigen-structure of $\bTheta^{(t)}$ in Condition \ref{con:eigenvalue} below, which is commonly adopted in network analysis with the low-rank assumption; see, e.g., \cite{fan2022asymptotic}, \cite{fan2022simple}, \cite{han2019universal}, and \cite{fan2022simplerc}. Specifically, for each $\bTheta^{(t)} \in \bTheta^0$, consider its 
eigen-decomposition $\bTheta^{(t)} = \bV^{(t)} \bD^{(t)} (\bV^{(t)})^\top$, where $\bD^{(t)} = \diag\left(d_1^{(t)},\ldots, d_{K^*_t}^{(t)} \right)$ is a diagonal matrix of nonzero eigenvalues $d_k^{(t)}$'s with $|d_1^{(t)}| \ge \ldots \ge|d^{(t)}_{K^*_t}|>0$, and $\bV^{(t)} = (\bv_1^{(t)}, \ldots, \bv^{(t)}_{K^*_t})$ is an $n \times K^*_t$ orthonormal matrix collecting the corresponding eigenvectors $\bv_k^{(t)}$'s with $1 \leq k \leq K^*_t$. To exploit the low-rank structure  for the testing problem, we impose the condition below on the spiked eigenvalues. 


\begin{condition}\label{con:eigenvalue}
For $1 \leq t \leq T$ and $1 \le k\le K^*_t-1$, if $|d_k^{(t)}| \ne |d_{k+1}^{(t)}|$ it holds that  $|d_k^{(t)}|/ |d_{k+1}^{(t)}| \ge 1+c_0$ for some $c_0 > 0$. Moreover,  $\min_{t \in [T]}|d^{(t)}_{1}| \gtrsim q_n/\sqrt{T}$ with $q_n: = \sqrt{n\rho_n}$.
\end{condition}


Condition \ref{con:eigenvalue} above requires enough separation between  eigenvalues of different magnitude to avoid the technical complication caused by the eigenvalue multiplicity. The condition for the largest  eigenvalue $d_1^{(t)}$ ensures that it can be well estimated using the avaiable data. These eigenvalue assumptions differentiate fundamentally the technical framework of network change-point detection from that of high-dimensional mean-vector change-point detection such as in \cite{jirak2015uniform} and \cite{liu2021minimax}. Based on the quantities introduced above and Condition \ref{con:eigenvalue}, we define the entire parameter space for dynamic network change-point detection as 
\begin{equation}\label{par-space}
\begin{aligned}
\calF(\rho_n, K^*)
    = \Big\{ \bTheta^0:  &\max_{1 \le t \le T} \| \bTheta^{(t)} \|_{\infty} \lesssim \rho_n, \, \bTheta^{(t)}\in \mathbb{S}^{n\times n}, \, 1 \le \Rank(\bTheta^{(t)}) \le K^*, \\ 
    &\text{ and }\bTheta^{(t)} \text{ satisfies Condition \ref{con:eigenvalue}}\Big\},
\end{aligned}
\end{equation}
where $\mathbb{S}^{n \times n}$ represents the set containing all symmetric $n \times n$ matrices with diagonal entries $0$ and off-diagonal entries in $[0,1]$.
The parameter space of mean matrices $\calF(\rho_n, K^*)$ introduced in (\ref{par-space}) above encompasses a broad class of network models, including the stochastic block models and their various generalizations, latent space models, and graphon models with low-dimensional representations. 
Let us further define the null space as 
\begin{equation}\label{def:null_space}
	\calN(\rho_n, K^*) := \left\{\bTheta^0 \in \calF(\rho_n, K^*): \bTheta^{(t)} \equiv \bTheta \text{ with } \bTheta\in \mathbb S^{n\times n}, \, {\max_{t \in [T]}\| \bV^{(t)}\|_{\infty} \lesssim \frac{1}{\sqrt{n}}}\right\}.
\end{equation}
For each $\bTheta^0\in \calN(\rho_n, K^*)$, the corresponding mean matrices $\bTheta^{(t)}$'s are invariant with respect to $t$.

For the alternative hypothesis with a single change point at some time $\tau^* \in [T]$, it is crucial to reconcile the low-rank structure with the assumption of sparse changes. To accommodate the low-rank structure, we adopt the concept of the change sparsity from \cite{yang2016rate} and \cite{ma2020adaptive}. Specifically, for the difference matrix $\bDelta=\bTheta_1-\bTheta_2$, denote by $\bDelta = \bU_{\bDelta} \bD_{\bDelta} \bU_{\bDelta}^\top$ its 
eigen-decomposition with $ \bD_{\bDelta}$ collecting the nonzero eigenvalues and $\bU_{\bDelta}$ the corresponding eigenector matrix. Then the \textit{change sparsity} is defined as 
\begin{equation} \label{new.eq.FL021}
s^*:=|\text{rowsupp}(\bU_{\bDelta})|, 
\end{equation}
which counts the number of nonzero rows of $\bU_{\bDelta}$. Thus, the difference matrix $\bDelta$ contains at most $s^*\times s^* = (s^*)^2$ nonzero entries. The low-rank constraint $\Rank(\bDelta)\leq \Rank(\bTheta_1) + \Rank(\bTheta_2)\leq 2K^*$ restricts that the size of $\bD_{\bDelta}$ is at most $2K^*$. We emphasize the distinction between the change sparsity $s^*$ and the network sparsity $\rho_n$: $\rho_n$ captures the probability of connectivity between two edges, while $s^*$ measures the number of changing edges.

To characterize the detection boundary and obtain a ``testable" alternative space, we further define the signal strength of the testing problem via the change magnitude at $\tau^*$. Let $\vech(\bDelta) \in \R^{n(n-1)/2}$ be the vectorization of the lower half matrix of $\bDelta$, and $\| \vech(\bDelta)\|_2^2$ the \textit{change magnitude}. The difficulty of the dynamic network change-point detection problem can be captured by the change magnitude $\| \vech(\bDelta)\|_2^2$, network change-sparsity $s^*$, and network rank $K^*$. We allow the spiked eigen-structure in the alternative parameter space to change with the sparsity level $s^*$ and define 
\begin{equation}\label{rmk: cond1}
    \calF(\rho_n, K^*, s^*) := \begin{cases}
 &\left\{\bTheta^0 \in \calF(\rho_n, K^*), \, \min_{t \in [T]} |d^{(t)}_{K_t^*}| \gtrsim  \sqrt{\rho_n  / T} \right\} \ \text{ if } s^* \in [\sqrt{c_{\xi} n}, \sqrt{n}],\\
   &\left\{\bTheta^0 \in \calF(\rho_n, K^*), \, \min_{t \in [T]} |d^{(t)}_{K_t^*}| \gtrsim \sqrt{\rho_n s^* / T}\right\}  \ \text{ otherwise}
\end{cases} 
\end{equation}
for a positive constant $c_{\xi}\in (0,1)$ given in Theorem \ref{low:minimax_lower}.
Consequently, we define the simple alternative hypothesis at a given change point $\tau^*$ as 
\begin{equation}
	\label{low:single_alter_indi}
	\begin{aligned}
		& \calA^{(\tau^*)}(\rho_n, K^*, s^*, \epsilon) := \Big\{ \bTheta^0 \in \calF(\rho_n, K^*, s^*): \bTheta^{(t)} \equiv \bTheta_1, \, \forall 1 \le t \le \tau^*;\\ 
		& \quad \bTheta^{(t)} \equiv \bTheta_{2}, \, \forall \tau^*+1\le t  \le T \text{ for some } \bTheta_{1}, \bTheta_{2}\in \mathbb S^{n\times n};\\
		&  \quad  |\text{rowsupp}(\bU_{\Delta})| \le  s^*; \,  \frac{\tau^*(T - \tau^*)}{ T} \| \vech(\boldsymbol{\Delta}) \|_2^2 \ge \epsilon^2; \, {\max_{t \in [T]}\| \bV^{(t)}\|_{\infty} \lesssim \frac{1}{\sqrt{s^*}}}
		\Big\},
	\end{aligned} 
\end{equation} 
where we refer to $\epsilon^2$ as the \textit{signal strength} that represents the lower bound of change magnitude after being adjusted for the time range $T$. We stress that $\epsilon^2$ is a key characteristic in investigating the detection boundary in network change-point problems. Recall that $\vech(\boldsymbol{\Delta})$ has at most $(s^*)^2$ nonzero entries, with the absolute value of each entry being bounded from above by $2\rho_n$. Hence, the lower bound condition in \eqref{low:single_alter_indi} involving $\epsilon^2$ implies that 
\begin{equation}\label{eq:epsilon_condition}
T(s^*)^2(\rho_n)^2\gtrsim \epsilon^2,	
\end{equation} 
which imposes an implicit assumption on the time span $T$ that will be made throughout the paper.

Following the change-point literature, assume that the change point does not occur at the boundaries; that is, $\tau^* \in\{hT, hT+1,\ldots, (1-h)T\}$ with $0<h<1/2$ and $hT$ an integer. Then we define the composite alternative hypothesis space as 
\begin{equation}\label{low:com_alter_space}
	\calA(\rho_n, K^*, s^*, \epsilon) := \cup_{\tau^*\in\{hT, hT+1, \ldots, (1-h)T\}} \calA^{(\tau^*)}(\rho_n, K^*, s^*, \epsilon).
\end{equation}
This alternative space traces explicitly the impacts of $s^*$ and $K^*$, differentiating our minimax-optimality analysis from existing ones which have been developed either under only the low-rank assumption (e.g., \cite{enikeeva2021change,wang2021optimal}) or for the mean-vector change-point detection problem (e.g., \cite{liu2021minimax}). Consequently, the hypothesis testing for a single network change point can be formulated as
\begin{equation}
	\label{sparse:test_single}
	H_0(\rho_n, K^*): \bTheta^0 \in \calN(\rho_n, K^*) 
	\ \text{ versus } \ H_1(\rho_n, K^*, s^*, \epsilon): \bTheta^0 \in \calA(\rho_n, K^*, s^*, \epsilon).
\end{equation}

For the hypothesis testing problem given in \eqref{sparse:test_single}, we aim to find the minimax-optimal detection boundary $\epsilon_n^2$ of signal strength $\epsilon^2$ in \eqref{low:single_alter_indi} that potentially depends on the change sparsity $s^*$ and network sparsity $\rho_n$. That is, the separation between $H_0(\rho_n,K^*)$ and $H_1(\rho_n, K^*, s^*, \epsilon)$ is infeasible when signal strength  $\epsilon^2$ falls below the rate of $\epsilon_n^2$; conversely, when $\epsilon^2$ exceeds $\epsilon_n^2$, there must exist a test that can successfully distinguish $H_1(\rho_n, K^*, s^*, \epsilon)$ from $H_0(\rho_n,K^*)$ with nontrivial power. Formally, for each fixed significance level $\alpha \in (0,1)$ and power $\eta \in(\alpha, 1)$, the \textit{minimax-optimal detection boundary} is said to be $\epsilon_n$ if for any test $\psi$ of asymptotic significance level $\alpha$, there exists some small constant $\underline{c}>0$ such that 
\begin{equation}
	\label{sparse:samll}
	\lim_{n \to \infty} \inf_{v \in \calA(\rho_n, K^*, s^*, \underline{c} \epsilon_n)} \Pr_v\left\{ \psi \text{ rejects } H_0\right\} < \eta,
\end{equation}
while there exist a test $\psi_0$ of asymptotic significance level $\alpha$ and some large constant $\overline{c}>0$ such that 
\begin{equation}
	\label{sparse:gtr}
	\lim_{n \to \infty} \inf_{v \in \calA(\rho_n, K^*, s^*, \overline{c}\epsilon_n)} \Pr_v\left\{ {\psi_0} \text{ rejects } H_0\right\} \ge \eta,
\end{equation}
where $\Pr_v$ represents the probability measure with respect to parameter $v$. As in the literature, we refer to \eqref{sparse:samll} as the lower bound result and \eqref{sparse:gtr} as the upper bound result. 



\subsection{Lower bound result} \label{new.sec2.2}

We formally state the lower bound result of the minimax-optimal boundary $\epsilon_n^2$ for detecting a single network change point in the theorem below.

\begin{theorem}[Lower bound]
	\label{low:minimax_lower}
Let $c_{\xi} \in (0,1)$ be a small constant depending on $\xi:= \eta-\alpha$, and define 
		\begin{equation}
		\label{low:minimax_rate}
		\epsilon^2_n = \begin{cases}
			\rho_n s^* &\text{ if } s^* \ge \sqrt{n},\\
			 \rho_n \Big( \big(s^* \log(\frac{{ c_{\xi}} n}{(s^*)^2 })\big) \vee 1 \Big)   &\text{ if } s^* < \sqrt{n}.
		\end{cases}
	\end{equation}
	Then no asymptotic level-$\alpha$ test for hypotheses \eqref{sparse:test_single} possesses nontrivial power in the sense of \eqref{sparse:samll} with constant $\underline{c}$ defined in \eqref{equ:thm1:final_c} in the Supplementary Material. 
\end{theorem}


\begin{figure}[t]
    \centering
    \includegraphics[width=0.7\linewidth]{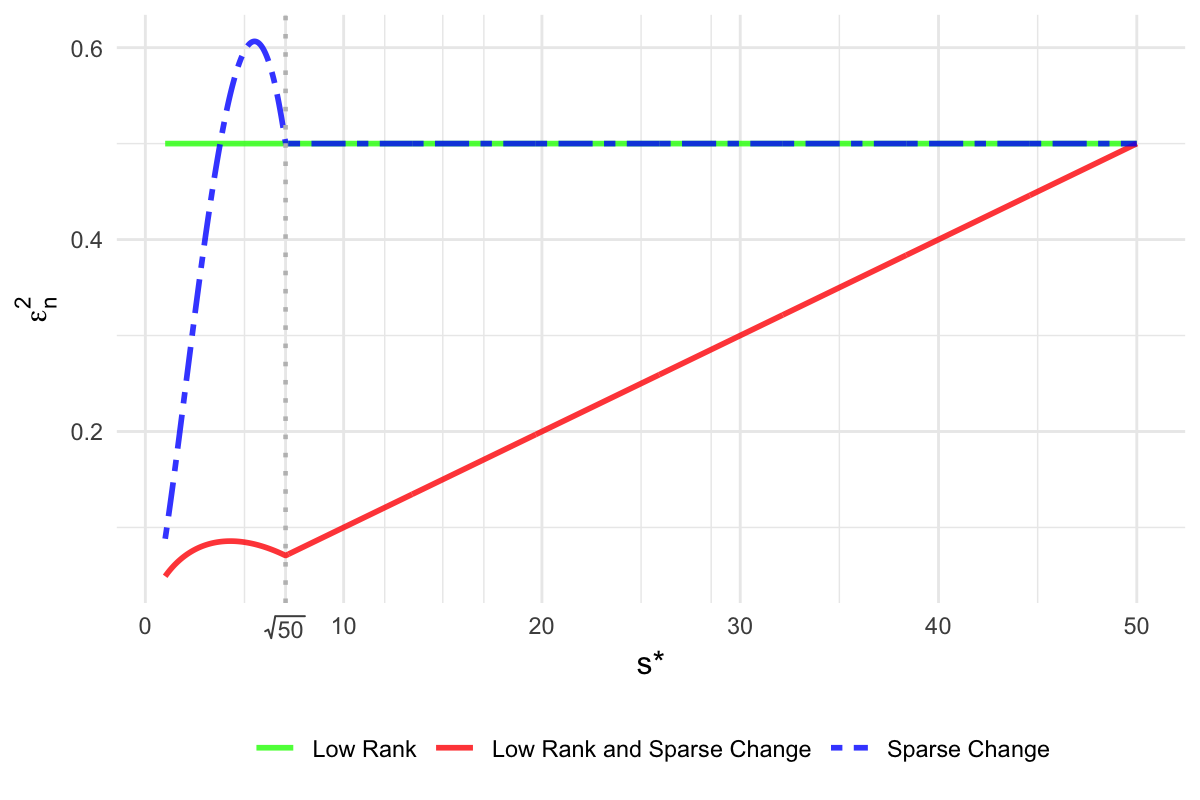}
    \caption{A comparison of three lower bound results 
	when $\rho=0.01$ and $n=50$.}
    \label{fig:lower_bound_plot}
\end{figure}

Constraint \eqref{eq:epsilon_condition} and the choice \eqref{low:minimax_rate} above entail that time span $T$ should satisfy 
	\begin{equation}\label{low:time_span_con}
		T \gtrsim  
		\begin{cases}
			\rho_n^{-1} (s^{*})^{-1} \ \text{ if } s^* \ge \sqrt{n},\\
			\rho_n^{-1} (s^*)^{-1} \big(\log(\frac{{ c_{\xi}} n}{(s^*)^2})  \vee  (s^*)^{-1} \big) \ \text{ if } s^* < \sqrt{n}.
		\end{cases}
	\end{equation}
It is seen that the lower bound $\epsilon_n^2$ in (\ref{low:minimax_rate}) depends on the change sparsity $s^*$, with a phase transition at $s^*=\sqrt{n}$, as illustrated by the red curve in Figure \ref{fig:lower_bound_plot}. Under the low-rank (but no sparsity) assumption, \cite{enikeeva2021change} derived a lower bound of order $\rho_nn$; see the green line in Figure \ref{fig:lower_bound_plot}. In addition to Theorem \ref{low:minimax_lower}, we will also 
establish the lower bound under the sparse-change assumption (without the low-rank structure) in Theorem \ref{thm:minimax_lower} later in Section \ref{subsec:sparse_lower}; it aligns closely with the minimax result for high-dimensional mean-vector change-point detection \citep{liu2021minimax}. Such bound is illustrated by the blue dash-dotted curve in Figure \ref{fig:lower_bound_plot}. 
Comparing the three lower bounds depcited in Figure \ref{fig:lower_bound_plot}, it is evident that the simultaneous low-rank and sparse-change structure relaxes significantly the required signal strength condition compared to settings that assume only low-rank structure or only sparse-change structure. The regime $\sqrt{c_{\xi} n} \leq s^* \leq \sqrt{n}$ corresponds to the case where $\log\left(\frac{c_{\xi} n}{(s^*)^{2}}\right)$ becomes negative, rendering the lower bound $\rho_n s^* \log\left(\frac{c_{\xi} n}{(s^*)^{2}}\right)$ uninformative. To address this, we provide a trivial lower bound of $\rho_n$ for this situation;  we do not claim that the result in this regime is optimal and will leave the optimality study for the future.

\subsection{Upper bound result on a theoretical MOSAIC test} \label{upper-section}

To establish the upper bound result, we tentatively assume that the change sparsity $s^*$ and network sparsity $\rho_n$ are both known. 
We also want to exclude the easy scenario where under the alternative hypothesis, the numbers of spiked eigenvalues of $\bTheta_1$ and $\bTheta_2$ are drastically different 
so that a simple estimate of these numbers is sufficient for change point detection. For such reason, we modify the eigen-structure assumption by replacing Condition \ref{con:eigenvalue} with Condition \ref{con:upper} below.


\begin{condition} \label{con:upper} 
There exists some constant $K$ with $0 < K \leq \min_{t \in [T]}K^*_t$ such that i)
 for $1 \leq t \leq T$ and $1 \le k\le K$, if $|d_k^{(t)}| \ne |d_{k+1}^{(t)}|$ it holds that $|d_k^{(t)}|/ |d_{k+1}^{(t)}| \ge 1+c_0$ with some small constant $c_0 > 0$; ii) $\min_{t\in [T]}|d^{(t)}_{K}| \gtrsim \max\left(\frac{q_n\log^{3+2\epsilon_0}(n)}{\sqrt{T}}, \frac{q_n^2}{T(\log\log(n))^{\epsilon_0^\prime}}\right)$ for some small constants $\epsilon_0, \epsilon_0^\prime > 0$; and iii) $\max_{t \in [T]}|d^{(t)}_{K+1}| = o(\frac{q_n}{\sqrt{T}})$. Moreover, assume that   $n \ge \log^8(n)$, $\sqrt{T}q_n\ge \log^4(n)$, and $ \max_{t \in [T]}\| \bV^{(t)}_{K^*_t}\|_{\infty} \lesssim 1/\sqrt{n}$.
\end{condition}

The difference between Conditions \ref{con:upper} and 
\ref{con:eigenvalue} can be small in the regime of $s^* \asymp n$ and $K_t^* = K$ for all $t$. To understand this, let us recall parameter space defined in \eqref{rmk: cond1}. The localization condition in both Conditions \ref{con:upper} and \ref{con:eigenvalue} reduces to $\max_{1\leq t\leq T}\| \bV^{(t)}_{K^*_t}\|_{\infty} \lesssim 1/\sqrt{n}$ in this regime.  When $q_n \lesssim \sqrt{T} \log^{3+2\epsilon_0}(n) (\log\log(n))^{\epsilon_0^\prime}$, the spiked eigenvalue structure in Condition \ref{con:upper} reduces to $\min_{t\in [T]}|d^{(t)}_{K_t^*}| \gtrsim \frac{q_n\log^{3+2\epsilon_0}(n)}{\sqrt{T}}$, so  that it matches that in  \eqref{rmk: cond1} up to a logarithmic factor $\log^{3+2\epsilon_0}(n)$.  

On the other hand, when $s^* \ll n$ but $s^* \not\in (\sqrt{c_{\xi}n}, \sqrt{n}]$, in view of \eqref{rmk: cond1}, the largest $K$ eigenvalues (in magnitude) needed for the lower bound result (see Theorem \ref{low:minimax_lower}) should be bounded from below by $\sqrt{\rho_n s^* / T}$, and the localization condition therein is ${\max_{t \in [T]}\| \bV^{(t)}\|_{\infty} \lesssim \frac{1}{\sqrt{s^*}}}$, while the corresponding bounds in Condition \ref{con:upper} replace $s^*$ with $n$. The stronger localization condition that depends on $n$ in Condition \ref{con:upper} is imposed for technical reasons to ensure that the asymptotic decompositions for eigenvectors in \cite{fan2022asymptotic} and \cite{han2019universal} can be applied. Indeed, Condition \ref{con:upper} ensures an accurate characterization of the asymptotic behavior of the eigen-decompositions of adjacency matrices, which plays a crucial role in constructing test statistic $\psi_0$ satisfying \eqref{sparse:gtr}; see Lemmas \ref{lem:gen1} and \ref{lem:gen2} in Section \ref{new.sec.D} of the Supplementary Material for technical details.

The constant $K$ introduced in Condition \ref{con:upper} above restricts that the number of true spiked eigenvalues of $\bTheta^{(t)}$ stays the same across all $t\in [T]$ so that a simple counting of the number is insufficient for change-point detection. Such condition is weaker than assuming that $K_t^*$'s are all identical and all $K_t^*$ nonzero eigenvalues of $\bTheta^{(t)}$ are large enough in magnitude.  


Using Condition \ref{con:upper}, we modify the parameter space as
\begin{equation} \label{new.eq.FL022}
\begin{aligned}
\calF^\prime(\rho_n, K^*)
    = \Big\{ \bTheta^0:  &\max_{1 \le t \le T} \| \bTheta^{(t)} \|_{\infty} \lesssim \rho_n, \, \bTheta^{(t)}\in \mathbb{S}^{n\times n}, \, 1 \le \Rank(\bTheta^{(t)}) \le K^*, \\ 
    & \text{ and }\bTheta^{(t)} \text{ satisfies Condition \ref{con:upper}}\Big\}.
\end{aligned}
\end{equation}
The null and alternative spaces can then be defined in the same manner as in \eqref{def:null_space} and \eqref{low:com_alter_space}, respectively, except that $\calF(\rho_n, K^*)$ is replaced with $\calF^\prime(\rho_n, K^*)$. Similarly, we define the corresponding terms $\calN(\rho_n, K^*)$, $\calA(\rho_n, K^*, s^*, \epsilon)$, $H_0(\rho_n, K^*)$, and $H_1(\rho_n, K^*, s^*, \epsilon)$ with a slight abuse of notation.



We now aim to construct a theoretical MOSAIC test that possesses nontrivial power when the signal strength $\epsilon^2$ exceeds the detection boundary. Without loss of generality, assume that $(T+1)/2$ is an integer. For each time point $\tau\in \{hT,hT+1,\ldots,(T+1)/2\}$, we investigate whether the network change lies in $\{\tau, \tau+1, \ldots, T-\tau+1\}$ by advocating the eigen-decomposition-based cumulative sums (CUSUM) test. Specifically, let us first take the CUSUM-type averages,  $\bar{\bX}^{(\tau,1)}$ and $\bar{\bX}^{(\tau,2)}$, of observed adjacency matrices $\bX^{(t)}$'s over the left interval $\{1,\ldots, \tau\}$ and the right interval $\{T-\tau+1, \ldots, T\}$, respectively, as
\begin{equation}\label{left-right-mean}
	\begin{aligned}
		&\bar{\bX}^{(\tau,1)} := \frac{\sum_{t=1}^{\tau} \bX^{(t)}}{\tau} \ \text{ and } \ \bar{\bX}^{(\tau,2)} := \frac{\sum_{t=1}^{\tau} \bX^{(T-t+1)}}{\tau}.
	\end{aligned}
\end{equation}
In light of the low-rank structure of $\bTheta^{(t)}$, we then apply the eigen-decomposition to $\bar\bX^{(\tau,1)}$ and $\bar\bX^{(\tau,2)}$ to obtain the smoothed versions of the above statistics, denoted as $ED(\bar{\bX}^{(\tau,1)})$ and $ED(\bar{\bX}^{(\tau,2)})$, respectively, using the first $K$ 
eigen pairs. 
Based on the smoothed statistics, the behavior of difference matrix $\bDelta$ at point $\tau$ can be captured by
\begin{equation}\label{low:zij}
\bZ^{(\tau)} = \{z_{i,j}^{(\tau)}\}_{1 \le i,j \le n} := \sqrt{\frac{\tau}{2  \rho_n}} \left(ED(\bar{\bX}^{(\tau, 1)}) - ED(\bar{\bX}^{(\tau, 2)}) \right),
\end{equation}
where the normalization factor $\sqrt{\tau/(2\rho_n)}$ is to ensure that the variances of $z_{i,j}^{(t)}$'s do not vanish as $n$ or $T$ increases.
We remark that the eigen-decomposition process alters the structure of independent sums in standard CUSUM-type statistics, making the entries in  (\ref{low:zij}) above \textit{dependent}. Such dependency sets the subsequent analysis apart from the conventional CUSUM-based techniques for vector change-point detection.

Intuitively, a large magnitude of $\bZ^{(\tau)}$ implies a potential change point in $\{\tau, \tau+1, \ldots, T-\tau+1\}$. Since the mean and variance of a Bernoulli random variable are of the same order, using $\ell_2$-type metrics for evaluating the magnitude of $\bZ^{(\tau)}$ would introduce inevitable biases from the squared $z_{i,j}^{(\tau)}$ terms, leading to size distortion of  tests built on it. Additionally, under the alternative hypothesis, squaring $z_{i,j}^{(\tau)}$ causes the error variance to be of the same order as the signal, thereby obscuring the distinction between the signal and noise. Hence, we instead use a \textit{product-type} metric between two independent copies of $\bZ^{(\tau)}$ to ensure that the expected value of statistic is zero under the null hypothesis, and that the signal is not masked by noise under the alternative hypothesis. Further, to incorporate the sparse-change structure, we introduce a ``\textit{strong-signal edge set}" through screening as the power enhancement component, so that noises and weak signals can be mostly screened out in the final test statistic. To avoid the issue caused by spurious correlation, we construct the screening set in theoretical MOSAIC test  using an independent copy of $\bZ^{(\tau)}$. 

Specifically, we adopt a threefold order-preserved data-splitting approach across the time horizon, which is used widely in the change-point detection literature, prior to constructing the test statistic. With a slight abuse of notation, denote by $\bX^{(t)}$ with $t= 1,\ldots, 3T$ the observed dynamic networks, and define 
\begin{equation}
	\label{sparse:iidcopy}
	x_{i,j}^{(t)} = x_{i,j}^{(3t-2)}, \, \dot{x}_{i,j}^{(t)} = x_{i,j}^{(3t-1)}, \, \ddot{x}_{i,j}^{(t)} = x_{i,j}^{(3t)}
\end{equation}
for $t = 1,\ldots, T$, so that $x_{i,j}^{(t)}$, $\dot{x}_{i,j}^{(t)}$, and $\ddot{x}_{i,j}^{(t)}$ are independent samples of the originally observed adjacency matrices under the null hypothesis. Correspondingly, we can obtain three independent copies $z_{i,j}^{(\tau)}$, $\dot{z}_{i,j}^{(\tau)}$, and $\ddot{z}_{i,j}^{(\tau)}$ of the smoothed CUSUM-type statistics via \eqref{low:zij}. Then for each given $\tau$, we define 
\begin{equation}
	\label{low:A_stat}
	A^{(\tau)}_\calS =  \sum_{(i,j)\in\calS}  z_{i,j}^{(\tau)}\dot{z}_{i,j}^{(\tau)},
\end{equation}
where  \begin{equation} \label{new.eq.FL023}
\calS=\{(i,j): 1\leq i<j\leq n, \, |\ddot{z}^{(\tau)}_{i,j}| \ge  d(s^*)\}
\end{equation}
represents the signal edge set identified using the third independent sample, and $d(s^*)$ is the hard threshold for screening to be determined in Theorem \ref{low:minimax_upper} later. By construction, $A^{(\tau)}_\calS$ in (\ref{low:A_stat}) above consists of zero-mean random noises under the null hypothesis, benefiting from the use of products $z_{i,j}^{(\tau)}\dot{z}_{i,j}^{(\tau)}$ rather than squared terms $(z_{i,j}^{(\tau)})^2$. The screened signal edge set $\calS$ introduced in (\ref{new.eq.FL023}) above is particularly important for power enhancement when the change-sparsity level $s^*$ is small, as it prevents excessive noise aggregation that could otherwise compromise power.

Moreover, since the hypothesis testing problem \eqref{sparse:test_single} aim at detecting only on the existence of a change point rather than its exact location, it is unnecessary to examine every possible position in $\{hT, hT+1, \ldots, (T+1)/2\}$. Thus, we take a smaller set of time points $\calT := \{hT,  2hT, 2^2hT, \ldots, 2^{|\calT|}h(T+1)\}$ with finite cardinality $|\calT| = \log_2\{(2h)^{-1}\}$ so that if a true change point $\tau^*$ exists, it has to lie within $(\tau,T-\tau+1)$ for some $\tau\in\calT$. 
Based on these considerations, we are ready to construct a final theoretical MOSAIC test as 
\begin{equation}
	\label{low:phi_test}
	\psi_n = \I \left( {  \max( |A_{\calS}|/s^*, |A_{\Omega}|/n)} > r_n/s^*\right), 
\end{equation}
where  $A_{\calS} := \max_{\tau \in \calT}A^{(\tau)}_\calS$,  $A_{\Omega} := \max_{\tau \in \calT}A^{(\tau)}_\Omega$, and  $r_n = \log(en)s^*$. The theorem below formally states that the suggested theoretical MOSAIC test $\psi_n$ in \eqref{low:phi_test} above achieves the upper bound described in \eqref{sparse:gtr}. 

\begin{theorem}[Upper bound]
	\label{low:minimax_upper}
     Set $c_d \ge  \sqrt{8c_0(1+\epsilon)/(1-\epsilon)^2}$ for some $\epsilon \in (0,1)$ and some constant $c_0 > 0$, 
     {$$d^2(s^*) \in [{c^2_d} n^{-1}\log(en/s^*),{c^2_d}(s^*)^{-1}\log(en/s^*)]$$} in \eqref{low:A_stat}--\eqref{new.eq.FL023}, and the detection boundary as  
	\begin{equation}\label{equ:upper_sep}
      \epsilon_n^2 =  \tilde c_d \rho_n s^* \log(e n)
	\end{equation}
 with $\tilde c_d$ a positive constant depending only on $c_d$.    
 Then $\psi_n$ defined in \eqref{low:phi_test} is an asymptotic level-$\alpha$ test under $H_0(\rho_n, K^*)$ and has power $\eta > \alpha$ under  $H(\rho_n,K^*, s^*, \epsilon_n)$, where $H_0(\rho_n, K^*)$ and $H(\rho_n,K^*, s^*, \epsilon_n)$ are defined in this section using Condition \ref{con:upper}. 
\end{theorem} 

Theorem \ref{low:minimax_upper} above ensures the validity of the suggested theoretical MOSAIC test $\psi_n$ in \eqref{low:phi_test}, which has size $\alpha$ under null hypothesis $H_0(\rho_n, K^*)$ and achieves a power greater than $\alpha$ under alternative hypothesis $H_1(\rho_n,K^*, s^*,\epsilon_n)$ with $\epsilon_n$ given in \eqref{equ:upper_sep}. From Theorem \ref{low:minimax_upper}, we observe that the required detection boundary for the upper bound differs from the one for the lower bound in Theorem \ref{low:minimax_lower} only by a logarithmic factor in $n$ for any $s^*\not\in [\sqrt{c_{\xi}n},\sqrt n)$. Hence, the rate of $\epsilon_n$ given in \eqref{low:minimax_rate} in Theorem \ref{low:minimax_lower} is minimax-optimal up to a logarithmic factor loss for any $s^*\not\in [\sqrt{c_{\xi}n},\sqrt n)$.  The constant $c_0$ depends only on the eigenvector localization condition $\max_{t \in [T]}\| \bV^{(t)}_{K^*_t}\|_{\infty} \lesssim 1/\sqrt{n}$; if we can find some constant $C$ such that $\max_{t \in [T]}\| \bV^{(t)}_{K^*_t}\|_{\infty} \le C/\sqrt{n}$, we can set $c_0 = 2C$.

In constructing the theoretical MOSAIC test $\psi_n$ in \eqref{low:phi_test}, the set $\calS$ defined in \eqref{low:A_stat} plays a pivotal role in attaining the detection boundary in \eqref{equ:upper_sep} across different sparsity levels $s^*$. Notably, no minimum signal strength condition is imposed on individual $\delta_{ij}$’s. Consequently, the function of $\calS$ in \eqref{low:A_stat} should be interpreted primarily as noise reduction rather than feature selection. Indeed, because $\psi_n$ relies on the CUSUM principle, mitigating noise accumulation in the summation is more crucial than perfectly recovering all nonzero signals $\delta_{ij}\neq 0$.  


Since the base of the logarithm is $e$ in Theorem \ref{low:minimax_upper}, it is seen that $d^2(s^*)$ is nonnegative. We can replace the base with any positive constant $c > 1$, and the detection boundary can correspondingly be expressed as $\epsilon_n^2 = \tilde c_d \rho_n s^* \log_{c}(c n)$. However, we cannot choose $c < 1$ as this would make $d^2(s^*)$ negative.

 

\begin{remark}\label{remark:may11}
When $s \asymp n$, i.e., the network change is dense, feature screening is unnecessary and we can just use $A_{\Omega}$ to construct our test. By slightly modifying the proof of Theorem \ref{low:minimax_upper}, we can relax condition \eqref{equ:upper_sep} to $\epsilon_n^2 \gg \rho_n n$. Such requirement nearly matches the lower bound in \eqref{low:minimax_rate} for the dense case of $s \asymp n$. This also explains the $A_{\Omega}$ component in the test \eqref{low:phi_test}.   
\end{remark}

To highlight the benefits of imposing the low-rank structure and incorporating the eigen-decomposition into the CUSUM-type statistics, we examine the asymptotic behavior of $z_{i,j}^{(\tau)}$ defined in \eqref{low:zij}. Let us rewrite the observed adjacency matrices as 
\begin{equation}	\label{equ:model}
	\bX^{(t)} = \bTheta^{(t)} + \bW^{(t)}
\end{equation}
with $t=1,\ldots, T$, where $\bW^{(t)} =(w_{i,j}^{(\tau)})_{1\le i,j\le n}= \bX^{(t)} - \bTheta^{(t)}$ represents the error matrix. We further compute the standardized mean error matrix over the ``stationary time zones" as 
$$
\tilde{\bW}^{(\tau,l)}:= (\tilde{w}_{i,j}^{(\tau,l)})_{1\le i,j\le n}=\frac{\sqrt{\tau}}{q_n} \bar{\bW}^{(\tau, l)}
$$
with $l =1,2$, where $\bar{\bW
}^{(\tau,l)}= (\bar{w}_{i,j}^{(\tau)}) :=\bar{\bX}^{(\tau, l)} - \bTheta_l$  and $\sqrt{\tau}/ q_n$ is a normalization term that guarantees $ \Var(\tilde{w}_{i,j}^{(\tau)}) = 1/n$. If $\tau \le \tau^* \le (T+1)/2$, $z_{i,j}^{(\tau)}$ can be expressed as 
\begin{equation}\label{equ: Nov19:01}
z^{(\tau)}_{i,j} = \sqrt{\frac{\tau}{2\rho_n}}\delta_{i,j} + \sqrt{\frac{n}{2}}\sum_{l=1}^2(-1)^{l+1}\left(\be_i^\top \tilde{\bW}^{(\tau,l)} \bb^{(l)}_j + \be_j^\top \tilde{\bW}^{(\tau,l)}\bb^{(l)}_i \right)+ O\left(\frac{1}{\sqrt{n}}\right) 
\end{equation}
uniformly in $1 \le i, j \le n $ with high probability, where $\delta_{i,j}$ is the $(i,j)$th entry of $\bDelta$, $\be_i \in \mathbb{R}^n$ is a basis vector with the $i$th component being $1$ and $0$ otherwise, and $\bb_j$ is a constant vector with $\| \bb_j\|_{\infty} \lesssim 1/n$. See Lemma \ref{lem:F} in Section \ref{new.sec.D} of the Supplementary Material for details. This entails that 
\begin{equation}\label{varZ}
\Var(z_{i,j}^{(\tau)}) \asymp \Var(\sqrt{n} \be_i^\top \tilde{\bW}^{(\tau,1)} \bb_j) \lesssim n \sum_{j=1}^n \Var(\tilde{w}^{(\tau,1)}_{i,j}) b^2_j(i) =O(1/n).
\end{equation}

In contrast, if we do not exploit the low-rank approximations $ED(\bar{\bX}^{(\tau,1)})$ and $ED(\bar{\bX}^{(\tau,1)})$, the test statistic would be constructed based on the average adjancency matrices $\bar{\bX}^{(\tau,1)}$ and $\bar{\bX}^{(\tau,2)}$ directly, following the standard CUSUM testing idea \citep{wang2021optimal,liu2021minimax}; 
see \eqref{nolow:zij} for the formal definition. 
Direct calculations show that for such standard CUSUM test statistic, the variance of the counterpart of $z_{i,j}^{(\tau)}$  (i.e., $e^{(\tau)}_{i,j}$ in \eqref{nolow:zij}) is of a much larger magnitude $O(1)$. This result confirms the advantage of leveraging the low-rank structure in dynamic network change-point testing problems.

\section{Practical implementation of MOSAIC} \label{sec:inference}

We have systematically studied in Section \ref{bound_result} the minimax-optimal detection boundary for MOSAIC, and advocated a theoretical MOSAIC test $\psi_n$ for hypotheses in \eqref{sparse:test_single} when both 
$s^*$ and $\rho_n$ are known. Such oracle test admits an upper bound result unveiling that the detection boundary is indeed minimax-optimal for any $s^*\not\in [\sqrt{c_{\xi}n}, \sqrt{n})$ (up to some logarithmic terms). However, the fact that $s^*$ and $\rho_n$ are unknown in practice renders $\psi_n$ inapplicable. In this section, we modify the theoretical MOSAIC test using a novel residual-based technique, and obtain a pivotal test statistic that enjoys an asymptotic standard normal distribution under the null 
for a given change position. The unknown change-point and sparsity parameters are addressed through a new data-driven edge selection procedure, for which we further show that the suggested empirical MOSAIC test exhibits compelling power performance under the alternative.

\subsection{A new residual-based CUSUM for constructing empirical MOSAIC
} \label{new.sec3.1}

We first construct a pivotal empirical MOSAIC statistic for testing whether there is a change point at given time $\tau \in \calT$. Such a test will serve as a building block for testing the existence of an unknown change position which will be discussed in the next subsection.

Recall the theoretical MOSAIC test $\psi_n$ introduced in (\ref{low:phi_test}). Even with known $s^*$ and $\rho_n$, deriving the null distribution of $\psi_n$ is challenging, if not impossible, mainly because of the complicated dependency structure among entries of $\bZ^{(\tau)}$ in \eqref{low:zij}.  To overcome this difficulty, in constructing our empirical test, we modify the error matrix estimation in the theoretical MOSAIC test by swapping the mean matrix estimator prior to time $\tau$ with the one after time $\tau$. Different from the theoretical MOSAIC test,
we only need two independent samples for constructing empirical MOSAIC. With a slightly abuse of notation, assume that we observe dynamic networks $\bX^{(t)}$ with $t= 1,\ldots, 2T$, and define 
\begin{equation*}
	x_{i,j}^{(t)} = x_{i,j}^{(2t-1)}, \, \dot{x}_{i,j}^{(t)} = x_{i,j}^{(2t)},
\end{equation*}
for $t = 1,\ldots, T$, so that $x_{i,j}^{(t)}$ and $\dot{x}_{i,j}^{(t)}$ are independent samples of entries in the originally observed adjacency matrices  under the null hypothesis.

We replace $\bZ^{(\tau)} = (z_{i,j}^{(\tau)})$ and $\dot{\bZ}^{(\tau)} = (\dot{z}_{i,j}^{(\tau)})$ in the theoretical MOSAIC test \eqref{low:A_stat} with 
\begin{equation}\label{low:omega}
\begin{aligned}
    \hat{\bW}^{(\tau)}=& (\hat{w}_{i,j}^{(\tau)})_{1 \le i,j \le n} := \bar{\bX}^{(\tau, 1)} - ED(\bar{\bX}^{(\tau, 2)}),\\
     \hat{\dot{\bW}}^{(\tau)}=& (\hat{\dot{w}}_{i,j}^{(\tau)})_{1 \le i,j \le n} := \bar{\dot{\bX}}^{(\tau, 1)} - ED(\bar{\dot{\bX}}^{(\tau, 2)}),
\end{aligned}
\end{equation}
where we recall that $\bar{\bX}^{(\tau, l)}$, $l=1,2$, are defined in \eqref{left-right-mean}, and $\bar{\dot{\bX}}^{(\tau, l)}$ is an independent copy of $\bar{\bX}^{(\tau, l)}$ defined by data splitting discussed in the last paragraph. For a given edge set $\calS$  independent of $\hat{w}_{i,j}^{(\tau)}$ or $ \hat{\dot{w}}_{i,j}^{(\tau)}$, we define 
the residual-based CUSUM statistic at each given $\tau$ as 
\begin{equation}\label{Dec2:A-tau}
    \hA_{{ \calS}}^{(\tau)}
     := \sum_{(i,j) \in { \calS}} \hat{w}_{i,j}^{(\tau)} \hat{\dot{w}}_{i,j}^{(\tau)}, 
\end{equation}
where we use the convention $ \hA_{{ \calS}}^{(\tau)} = 0$ when $\calS = \emptyset$. 
The advantage of this approach is that $\hat{\bW}^{(\tau)}$ and $\hat{\dot{\bW}}^{(\tau)}$ both invoke only one eigen-decomposition, making the derivation of the asymptotic null distribution much more manageable.

Intuitively, under $H_0(\rho_n, K^*)$, the mean matrices of $\bar{\bX}^{(\tau, 1)}$ and $\bar{\bX}^{(\tau, 2)}$ (i.e., $\bTheta_1$ and $\bTheta_2$, respectively) are identical. Provided that the eigen-decomposition $ED(\bar{\bX}^{(\tau, 2)})$ in \eqref{low:omega} yields a reasonable estimate of $\bTheta_2 = \bTheta_1$,  the test $\hA_{{ \calS}}^{(\tau)}$ with any pre-chosen deterministic edge set  $\calS$ would behave similarly to the summation of independent random variables constructed from entries in de-meaned matrices $\bar{\bW
}^{(\tau,l)}= (\bar{w}_{i,j}^{(\tau)}) :=\bar{\bX}^{(\tau, l)} - \bTheta_l$ and $\bar{\dot\bW
}^{(\tau,l)}= (\bar{\dot w}_{i,j}^{(\tau,l)}) :=\bar{\dot \bX}^{(\tau, l)} - \bTheta_l$, $l=1,2$. Formally, we prove that $ \hA_{\calS}^{(\tau)}$ defined in (\ref{Dec2:A-tau}) admits an asymptotic decomposition 
\begin{equation}\label{Feb13:equ:03}
    \begin{aligned}
     \hA_{\calS}^{(\tau)} :=& \sum_{(i,j) \in \calS}\bar{w}_{i,j}^{(\tau,1)} \bar{\dot{w}}_{i,j}^{(\tau,1)} + \sum_{(i,j) \in \calS} \delta_{i,j}^2 +  O_p\left(\frac{\rho_n \sqrt{|\calS|}}{T\sqrt{n}}\right)\\
    :=& \Upsilon^{(\tau)}_n + \sum_{(i,j) \in \calS} \delta_{i,j}^2 
 + O_p\left(\frac{\rho_n\sqrt{|\calS|}}{T\sqrt{n}}\right).
\end{aligned}
\end{equation}
Consequently, under the null hypothesis $H_0(\rho_n, K^*)$ with $\delta_{i,j}=0$ for all $i,j\in[n]$, the behavior of $\hA^{(\tau)}_{\calS}$ can be characterized by $\Upsilon^{(\tau)}_n$, a sum of products of independent mean-zero errors. Such representation allows us to establish the asymptotic null distribution of $\hA^{(\tau)}_{{\calS}}$ in (\ref{Dec2:A-tau}), as stated in the theorem below.

\begin{theorem}
	\label{Thm:null}
	Under $H_0(\rho_n, K^*)$, assume that Condition \ref{con:upper} holds, 
        \begin{equation} \label{con:nu1}
		\begin{aligned}
            &\sum_{(i,j) \in \calS} \theta_{i,j}^2 { \gg 1/T^2},\\
           & T(\sum_{(i,j) \in \calS} \theta_{i,j}^2)^{2} { \gg  \sum_{(i,j)\in\calS} \theta_{i,j}^3},
		\end{aligned}
        \end{equation}
         and further there exists a sequence $\eta_n \to 0$ such that $\rho_n^2|\calS|/(\eta_n^2 n \sigma_{\calS}^2) \to 0$. Then we have 
\begin{equation}\label{null-A}
    \sup_{t \in \R}\left|\Pr_0 \left\{ \frac{{ \tau} \hA^{(\tau)}_{\calS}}{\sqrt{ \sigma_{\calS}^2}}  \ge t\right\} - \Pr\left\{N(0,1) \ge t\right\} \right| = { o(1)}, 
\end{equation}
where
$\sigma_{\calS}^2 = \sum_{(i,j)\in\calS}\theta_{i,j}^2$,     
and $N(0,1)$ represents the standard normal distribution. 
\end{theorem} 

In Theorem \ref{Thm:null} above, the mild moment conditions in \eqref{con:nu1} are imposed to verify the Lindeberg-type conditions and the variance conditions in the Berry--Esseen bound for martingale differences;  see, e.g., Theorem 2 of \cite{haeusler1988rate} and Theorem 1.1 of \cite{mourrat2013rate}.  A similar set of conditions as \eqref{con:nu1} is also imposed for change-point inference in high-dimensional categorical data; see Assumptions (A.3)--(A.4) in \cite{wang2018change}.   To gain insight on the existence of $\eta_n$, let us consider the special case where $\theta_{i,j} \asymp \rho_n$. The condition involving $\eta_n$ reduces to $n \eta_n^2 \to \infty$, suggesting that any sequence satisfying $1/\sqrt{n} \ll \eta_n \to 0$ would be sufficient.

Under these conditions, Theorem \ref{Thm:null} establishes the asymptotic normal distribution of the empirical residual-based CUSUM statistic in (\ref{Dec2:A-tau}) under the null hypothesis for any given edge set $\calS$ and any given change point $\tau \in \calT$. 

Under the alternative hypothesis in \eqref{low:com_alter_space} where a change occurs at time point $\tau^*$, there exists a $\tau\in\calT$ such that $\tau<\tau^*<T-\tau+1$, in view of the construction of $\calT$ in \eqref{low:phi_test}. Hence, even though $ED(\bar{\bX}^{(\tau, 2)})$ still estimates $\bTheta_2$, it is different from $\bTheta_1$. In this case, the second term in decomposition \eqref{Feb13:equ:03}, consisting of mean differences $\delta_{i,j}$ with  $(i,j)\in\calS$, is expected to dominate the first term $\Upsilon^{(\tau)}_n$ when the cumulative change signal is strong enough in $\calS$. Thus, constructing an informative edge set $\calS$ is essential in achieving high detection power under the alternative hypothesis. This motivates the pivotal empirical MOSAIC statistic for detecting the existence of change point at some unknown time location $\tau$ to be introduced in the next subsection.

\subsection{Empirical MOSAIC statistic} \label{power1}

For $\hA_{\calS}$ defined in the last section to be practically implementable with power, we need to estimate the unknown $\sigma_{\calS}^2$ and choose an informative signal edge set $\calS$. Additionally,
to adapt to the unknown change point location, 
we propose the following aggregated statistic  
\begin{equation}\label{Dec2:A}
    \hA_{\hat\calS} := \max_{\tau \in \calT} \frac{\tau\hA_{\hat\calS}^{(\tau)}}{\sqrt{ \hsigma^2_{\hat\calS}}},
\end{equation}
where $\hat\calS$ is a data-driven edge set and $\hsigma^2_{\hat\calS}$ is a sample estimate of $\sigma^2_{\hat\calS}$ that will be introduced later. We will discuss the detailed implementation next.  

For any given edge set $\calS$, we estimate $\sigma^2_{\calS}$ in \eqref{null-A} using the left-most and right-most $hT$ stationary networks as 
\begin{equation}\label{def:estimate_sigma}
    \hsigma_{\calS}^2 = \sum_{(i,j)\in\calS} \frac12\left\{(\hat{\theta}^{(hT,1)}_{i,j})^2 + (\hat{\theta}^{(hT,2)}_{i,j})^2\right\},
\end{equation}
where $\hat{\theta}^{(\tau,1)}_{i,j}$ and $\hat{\theta}^{(\tau,2)}_{i,j}$ represent the $(i,j)$th entry of the smoothed mean matrices  $ED( (\bar{\bX}^{(\tau,1)} + \bar{\dot{\bX}}^{(\tau,1)})/2)$ and  $ED((\bar{\bX}^{(\tau,2)}+\bar{\dot{\bX}}^{(\tau,2)})/2)$ in \eqref{low:zij}, respectively.
Lemma \ref{lem:svd_var} in Section \ref{new.sec.B.3} of the Supplementary Material shows that $\hsigma_{\calS}^2$ is a consistent estimator of $\sigma_{\calS}^2$.

We next discuss the selection of edge set $\calS$ for achieving high detection power. Recall from \eqref{low:A_stat} that with oracle information on model parameters, we would want to choose $\calS=\{(i,j): 1\leq i<j\leq n, \, |\ddot{z}^{(\tau)}_{i,j}| \ge d(s^*)\}$, where $\ddot{z}^{(\tau)}_{i,j}$ is the $(i,j)$th entry of the third independent copy of $\bZ^{(\tau)}$ \eqref{low:zij} as in \eqref{new.eq.FL023}. To mimic such oracle choice of $\calS$, we need to estimate unknown network sparsity parameter $\rho_n$ involved in the definition of $\ddot{z}^{(\tau)}_{i,j}$. To this end, we propose to use the estimator   $\hat{\rho}_n=\max\{\|ED((\bar{\bX}^{(hT,1)}+\bar{\dot{\bX}}^{(hT,1)})/2)\|_{\infty}, \, \|ED((\bar{\bX}^{(hT,2)})+\bar{\dot{\bX}}^{(hT,2)})/2\|_{\infty}\}$, that is, the maximum of estimated connectivity probabilities using only the left and right $hT$ stationary networks.
Lemma \ref{lemma:rho} in Section \ref{new.sec.B.3} of the Supplementary Material shows that $\hat{\rho}_n$ is indeed a consistent estimate of $\rho_n$. Equipped with such an estimate and using the full sample $\{\bX^{(t)}, t \in [2T]\}$, we can construct an estimate $\hz^{(\tau)}_{i,j}$  at $\tau=hT$ by replacing $\rho_n$ in $\ddot{z}^{(\tau)}_{i,j}$ \eqref{low:zij} with $\hat{\rho}_n$. Specifically,
\begin{equation*}
\hat{\bZ}^{(hT)} = \{\hz_{i,j}^{(hT)}\}_{1 \le i,j \le n} := \sqrt{\frac{hT}{2  \hat{\rho}_n}} \left(ED\left(\frac{\bar{\bX}^{(\tau, 1)}) +\bar{\dot{\bX}}^{(\tau, 1)}}{2}\right) - ED\left(\frac{\bar{\bX}^{(\tau, 2)}) +\bar{\dot{\bX}}^{(\tau, 2)}}{2}\right)\right).
\end{equation*}

We thus construct a data-driven signal edge set as
\begin{equation} \label{new.eq.FL019}
\hat{\calS} := \{(i,j): |\hz^{(hT)}_{i,j}| > \cd_n,(i,j) \in \Omega\},
\end{equation}
where $\cd_n = 4 { c_d}\sqrt{\log(en)/n}$ is the hard threshold that is taken slightly larger than its theoretical counterpart (Cf. Theorem  \ref{low:minimax_upper}) to offset the potential estimation error of $\hat{\rho}_n$. 
We emphasize that $\hat{\calS}$ is \textit{not} constructed from a third independent sample, and hence our empirical test suffers less from the power loss caused by excess sample splitting. 

Comparing \eqref{new.eq.FL019} with the oracle set \eqref{new.eq.FL023}, it is seen that the latter depends on the sparsity level $s^*$ which is unknown in practice.  To account for this, we propose the final empirical MOSAIC statistic as 
\begin{equation} \label{finalMOSAIC}
   \hat{A}= \max\left( |\hat{A}_{\hat{\calS}}|, |\hat{A}_{{\Omega}}| \right), 
\end{equation}
where $\hat\calS$ is defined in \eqref{new.eq.FL019} and $\hat{A}_{{\Omega}}$ is obtained by setting $\calS = \Omega$ in \eqref{Dec2:A}. 

We next formally justify that $\hat{A}$ proposed above is minimax optimal up to some logarithmic factor for nearly all values of $s^*$.  The key is to measure the quality of $\hat{\calS}$ in retaining the signals in the underlying true signal set $\calS^* := \{(i,j) \in \Omega : \delta_{i,j} \ne 0\}$. To this end, we partition $\calS^*$ into three non-overlapping sets according to the signal strength. Let $y_n = \sqrt{\frac{\rho_n}{\tau^*}}\check{d}_n$.  Define the strong signal set $\calS_1 = \{(i,j): (i,j) \in \calS^*,  |\delta_{i,j}| \ge c_h y_n\}$ with some constant $c^2_h \ge 8h/(1-h)$ and the weak signal set $\calS_2 = \{ (i,j): (i,j) \in \calS^*, 0< |\delta_{i,j}| /{y_n} \leq c_n\}$ with some positive sequence $c_n \rightarrow 0$ as $n \to \infty$. Let $\calS_3 := \calS^* \cap (\calS_1 \cup \calS_2)^c$ be the subset collecting moderate change signals. 

\begin{condition}\label{con:modset}
    The 
    moderate signal set $\calS_3$ satisfies $|\calS_3|  \ll s^*$. 
\end{condition}

Condition \ref{con:modset} above is needed because the screened edge set $\hcalS$ is not constructed from an independent sample. To understand this, recall that the asymptotic decomposition in  \eqref{Feb13:equ:03} plays a pivotal role in deriving the asymptotic distribution in \eqref{null-A}, and the independence of $\calS$ with sample observations in $\hat A_{\calS}^{(\tau)}$ is critical for establishing such a decomposition. To derive a similar decomposition as in  \eqref{Feb13:equ:03}, we need to decouple the dependence of $\hat\calS$ with other data used in constructing $\hat A_{\hat\calS}$. To this end, we form a ``bridge'' statistic based on an edge set $\tilde\calS$ constructed from an independent sample from data splitting similar to the idea in the theoretical MOSAIC statistic presented in the last section. 
Since $\hat\calS$ are $\tilde\calS$ are constructed from independent copies of identically distributed data,  their difference should be small and mainly on the not-so-strong edges. Indeed, we prove in Lemma \ref{lem:bridge} in Section \ref{new.sec.B.3} of the Supplementary Material that the set difference between $\hat\calS$ and $\tilde\calS$ can be upper bounded by $|\calS_3|$ with high probability; see the formal statements and proofs therein for details.  Based on such a result, we obtain that
\begin{equation}\label{eq:Ahat-Shat-error}
    |\hat{A}^{(\tau)}_{\hcalS} - \hat{A}^{(\tau)}_{\tilde{\calS}}| = O_p\left(|\calS_3| \max_{(i,j) \in \calS_3} |\hat{w}^{(\tau)}_{i,j}  \hat{\dot{w}}^{(\tau)}_{i,j}|\right) = O_p\left( \frac{|\calS_3|\rho_n\log(n)}{T} \right),
\end{equation}
where the penultimate bound is from $\max_{(i,j) \in \calS_3} |\hat{w}^{(\tau)}_{i,j}|= O_p\left(\sqrt{\rho_n \log(n)/T} \right)$ (see Lemma \ref{lem:bridge}). Under Condition \ref{con:modset}, the discrepancy can be bounded by $o_p(\bar{\epsilon}^2_n/T)$  with $\bar{\epsilon}_n$ as defined in Theorem \ref{Thm:power}, and thus is asymptotically negligible and ensures that $\hat{A}^{(\tau)}_{\calS}$ and $\hat{A}^{(\tau)}_{\tilde{\calS}}$ enjoy the same asymptotic properties. Since $\tilde{\calS}$ is constructed using an independent data set, a similar decomposition as in  \eqref{Feb13:equ:03} can be established for the bridge statistic $\hat{A}^{(\tau)}_{\tilde{\calS}}$, based on which the asymptotic size and power of $\hat{A}^{(\tau)}_{\hcalS}$ can be established.

We stress that similar to the theoretical MOSAIC statistic, we do not require any minimum signal strength condition on individual change magnitude in our analysis here. To understand why,  
note that the strong signal set $\calS_1$ must be non-empty and satisfies the aggregated signal condition because of the following result 
\begin{equation} \label{new.eq.FL020}
 \begin{aligned}
  T \sum_{(i,j) \in \calS_1} \delta_{i,j}^2 
  = &T \sum_{(i,j) \in \calS^*} \delta_{i,j}^2 - T \sum_{(i,j) \in \calS_2 \cup \calS_3} \delta_{i,j}^2 \\
    =& T \sum_{(i,j) \in \calS^*} \delta_{i,j}^2 - T |\calS_2| \max_{(i,j) \in \calS_2} \delta_{i,j}- T |\calS_3| \max_{(i,j) \in \calS_3} \delta_{i,j}\\
  \asymp &\overline{\epsilon}^2_n - o\left( \frac{\rho_n (s^*)^2 \log(en)}{n} \right) -  O\left( \frac{\rho_n s^*\log(en)}{n} \right)\\
  =&\overline{\epsilon}^2_n - o\left(\rho_n s^* \log(en) \right)
  = \overline{\epsilon}^2_n(1-o(1)).
 \end{aligned}   
\end{equation}
Lemma \ref{lem:selection} in Section \ref{new.sec.B.3} of the Supplementary Material ensures a weak sure screening property in the sense that $\hcalS$ contains $\calS_1$ with probability tending to one. Meanwhile, set $\calS_1$ has sufficient signals to reject the null hypothesis, and thus the empirical MOSAIC statistic has good power. As discussed before, this suggests that the function of $\hcalS$ is more for noise reduction than for feature selection.      

The theorem below guarantees that the suggested empirical MOSAIC statistic in (\ref{finalMOSAIC}) is a valid level-$\alpha$ test and achieves full power under the alternative. 

\begin{theorem} \label{Thm:power}
Assume that Condition \ref{con:upper} and moment conditions in \eqref{con:nu1}  hold with $\calS = \Omega$. Further assume that time span $T$ and network sparsity $\rho_n$ satisfy that 
$T \rho_n\gg \log(n)$ and 
     \begin{equation}\label{Jan5:1:omega}
          \rho_n \log^{2}(n)n \ll  T \sigma^2_{\Omega}, 
    \end{equation}
    where we redefine $\sigma_{\calS}^2 := \sum_{(i,j)\in\calS} \{(\theta_{1,i,j})^2 + (\theta_{2,i,j})^2\}/2$ for a given edge set $\calS$. 
  \begin{itemize}
      \item[i)](Asymptotic size). Under $H_0(\rho_n, K^*)$ and assuming in addition that there exists a sequence $\eta_n \to 0$ such that $\rho_n^2 n\ll\eta_n^2 \sigma^2_{\Omega}$, we have 
\begin{equation}\label{Feb27:size}
    \Pr_0\left\{\hat{A} >  z_{\alpha/(2|\calT|)} \right\} \le \alpha + o(1)
\end{equation}
as $n\rightarrow \infty$, where $z_{\alpha/|\calT|}$ is the upper $\alpha/(2|\calT|)$-quantile of the standard normal distribution with $\alpha\in (0,1)$ the significance level. 
\item[ii)](Asymptotic power). Under $H_1(\rho_n, K^*, s^*,  \bar{\epsilon}_n)$  with $ \bar{\epsilon}_n \ge c\epsilon_n$, $\epsilon_n$ defined in \eqref{equ:upper_sep}, and $c > 0$ some large enough constant, and assuming in addition that Condition \ref{con:modset} holds and 
 \begin{equation}\label{July:24:omega}
           \rho_n \log^{2}(n)\max\left\{|\calS^*|, n\log(n) \right\} \ll  nT \sigma^2_{\calS^*},
    \end{equation}
 we have 
\begin{equation}\label{Feb14:power}
 \Pr_1\left\{\hat{A} > z_{\alpha/(2|\calT|)} \right\} \to 1
\end{equation} 
as $n\rightarrow\infty$.
  \end{itemize}
\end{theorem}   

Theorem \ref{Thm:power} above justifies that the size of the empirical MOSAIC test $\hat{A}$ is asymptotically valid with asymptotic power one if the signal strength $\bar\epsilon_n$ is above the minimax detection boundary. The conditions required in this theorem are relatively mild. 
The additional time span condition of $T \rho_n \gg \log(n)$ is required for ensuring the consistency of $\hat{\rho}_n$ for estimating $\rho_n$, and is also used for establishing \eqref{eq:Ahat-Shat-error}. The extra moment condition \eqref{Jan5:1:omega} is necessary for the consistency of variance estimation, i.e,  $\hsigma^2_{\Omega}/\sigma^2_{\Omega} \to 1$ with probability tending to one under both null and alternative hypotheses.  In the special case of $\theta_{i,j} \asymp \rho_n$, this extra moment condition \eqref{Jan5:1:omega} reduces to  
$$
\frac{\log^2(n)}{Tn} \ll \rho_n,
$$
which is guaranteed to hold because of the  
assumption $n\rho_n = q_n^2 \ge T^{-1}\log^8(n)$ in Condition \ref{con:upper}. 

Under the null hypothesis, $\hcalS$ is an empty set with asymptotic probability one. Consequently, $\hat{A} = |\hat{A}_{\Omega}|$ with asymptotic probability one. Hence, it suffices to consistently estimate $\hsigma^2_{\Omega}$ for yielding the valid asymptotic size. The additional condition involving $\eta_n$  entails that there exists a sequence $\eta_n \to 0$ such that $\rho_n^2|\Omega|/(\eta_n^2 n \sigma_{\Omega}^2) \to 0$. This is required for establishing the asymptotic null distribution using the Berry--Esseen bound as presented in Theorem \ref{Thm:null}. When $\theta_{i,j} \asymp \rho_n$, the additional condition reduces to $n \eta_n^2 \to \infty$. 

Under the alternative hypothesis, based on Lemma \ref{lem:selection} in the Supplementary Material, we have  
$$
\Pr_1\left\{ \hsigma^2_{\hcalS} \leq \hsigma^2_{\calS^*} \right\} \geq \Pr_1\left\{ \hcalS \subset \calS^* \right\} \to 1,
$$  
which provides an upper bound for $\hsigma^2_{\hcalS}$ with probability approaching one. The additional condition \eqref{July:24:omega} ensures that this upper bound, $\hsigma^2_{\calS^*}$, converges to its population quantity $\sigma^2_{\calS^*}$.  This allows us to lower bound $\hA_{\hat\calS}$ as $\sigma^{-1}_{\calS^*}\max_{\tau}(\tau \widehat A_{\hcalS}^{(\tau)})$ in studying the power.
Additionally, condition \eqref{Jan5:1:omega} is required, as we also need to investigate the asymptotic property of $\widehat A_{\Omega}$ since it is part of the empirical MOSAIC statistic \eqref{finalMOSAIC}; this explains the necessity for imposing the global conditions that can ensure the consistency of $\hsigma^2_{\Omega}$ as an estimator of $\sigma^2_{\Omega}$.  
When $\theta_{l,i,j} \asymp \rho_n$, the additional condition \eqref{July:24:omega} reduces to
$$
\max\left\{ \frac{\log^2(n)}{nT}, \frac{\log^3(n)}{T |\calS^*|}\right\} \ll \rho_n.
$$
When $|\calS^*| \ge n/\log(n)$, such condition is implied by Condition \ref{con:upper} as discussed above. However, when $|\calS^*| < n/\log(n)$, the lower bound on $\rho_n$ depends on the cardinality of $\calS^*$ and time span $T$. A smaller $T|\calS^*|$ requires a larger $\rho_n$, which implies that denser networks are needed for change-point detection when the change locations are scarce (i.e., small $|\calS^*|$).  


\begin{remark}
If an independent copy of data is available for constructing $\hat\calS$, Condition \ref{con:modset} can be removed and the extra time span condition of $T \rho_n \gg \log(n)$ can be relaxed to $T q_n^2 \gg \log(n)$ while the results in Theorem \ref{Thm:power} continue to hold. 
\end{remark}

\section{MOSAIC without low-rank structure} \label{new.sec.mosaic.withoutlowrank}

In this section, we further investigate the problem of dynamic network change-point detection \textit{without} the low-rank structure. Several previous works in the network inference literature did not impose the low-rank structure; see, for example, \cite{wang2021optimal} and \cite{xia2022hypothesis}, among others.  
We explore the minimax-optimal rate \textit{without} the low-rank structure to better appreciate the role of low-rankness in network change-point detection. The main results in this section align closely with those in high-dimensional mean-vector change-point inference  \citep{liu2021minimax}, with the major difference that the data is generated from the Bernoulli distribution in our setting instead of Gaussian distribution. 

\subsection{Formulation of the testing problem} \label{new.sec4.1}

Parallel to \eqref{par-space}, \eqref{def:null_space}, and \eqref{low:single_alter_indi}, the parameter spaces of networks without the low-rank structure assumption are now defined as 
\begin{equation} \label{new.eq.FL017}
    \calF(\rho_n) = \left\{ \bTheta^0:  \max_{1 \le t \le T} \| \bTheta^{(t)} \|_{\infty} \lesssim \rho_n, \, \bTheta^{(t)}\in \mathbb{S}^{n\times n}\right \},
\end{equation}
\begin{equation*} \label{new.eq.FL018}
    \calN(\rho_n) := \left\{\bTheta^0 \in \calF(\rho_n): \bTheta^{(t)} \equiv \bTheta \text{ with } \bTheta\in \mathbb S^{n\times n}, \, \forall t\in [T]\right\},
\end{equation*}
and
\begin{equation}
\label{def:single_alter_indi}
   \begin{aligned}
	& \calA^{(\tau^*)}(\rho_n, m^*, \epsilon) :=\Big\{ \bTheta^0 \in \calF(\rho_n): \bTheta^{(t)} \equiv \bTheta_1, \, \forall 1 \le t \le \tau^*; \, 
	\bTheta^{(t)} \equiv \bTheta_{2}, \, \forall \tau^* \le t  \le T; \\
	& \quad \| \bDelta \|_0 \le  m^*; \, \frac{\tau^*(T - \tau^*)}{ T} \|\vech(\bDelta) \|_2^2 \ge \epsilon^2 \text{ for some } \bTheta_{1}, \bTheta_{2}\in \mathbb S^{n\times n}
	\Big\},
\end{aligned} 
\end{equation}
respectively, where $\bDelta = \bTheta_1-\bTheta_2$.
Different from the low-rank setting, the change sparsity parameter in this section, denoted as $m^*$, is directly defined as the number of nonzero entries in the difference matrix $\bDelta$. We note that $m^*$ here plays the same role as $ (s^*)^2$ in the previous low-rank setting. The composite alternative space is defined as 
\begin{equation}\label{def:com_alter_space}
    \calA(\rho_n, m^*, \epsilon) := \cup_{\tau^*=hT, hT+1, \ldots, (1-h)T} \calA^{(\tau^*)}(\rho_n, m^*, \epsilon).
\end{equation}
The problem for testing a single change point can be formulated as
\begin{equation}
	\label{def:test_single}
	H_0(\rho_n): \bTheta^0 \in \calN(\rho_n) 
 \ \text{ versus } \ H_1(\rho_n, m^*, \epsilon): \bTheta^0 \in \calA(\rho_n, m^*, \epsilon).
\end{equation}
 

\subsection{Lower bound result without low-rank structure} \label{subsec:sparse_lower}

Parallel to Theorem \ref{low:minimax_lower}, the main result on the lower bound of the detection boundary without the low-rank structure is summarized in the theorem below.

\begin{theorem}[Lower bound]
	\label{thm:minimax_lower}
Let $c_{\xi}^\prime \in (0,1)$ be a small constant depending on $\xi = \eta - \alpha$, and assume that time span $T$ satisfies
 \begin{equation}\label{equ:time_span_con}
     T \gtrsim
  \begin{cases}
      \rho_n^{-1} \frac{\sqrt{p_n}}{m^*} \ \text{ if } m^* \ge \sqrt{p_n},\\
      \rho_n^{-1} \Big( \log(\frac{{ c^\prime_{\xi}} p_n}{(m^*)^2}) { \vee  (m^*)^{-1}}\Big)  \ \text{ if } m^* < \sqrt{p_n},
  \end{cases}
 \end{equation}
with $p_n=n(n-1)/2$. 
Set 
	\begin{equation}
		\label{equ:minimax_rate}
			\epsilon^2_n = \begin{cases}
			\rho_n \sqrt{p_n} & \text{ if } m^* \ge \sqrt{p_n},\\
			 \rho_n \Big( \big(m^* \log(\frac{{ c^\prime_{\xi}} p_n}{(m^*)^2 })\big) {\vee 1}\Big)  &  \text{ if } m^* < \sqrt{p_n}.
		\end{cases}
	\end{equation}
Then no level-$\alpha$ test under $H_0(\rho_n)$ possesses power greater than $\eta$ under $H_1(\rho_n, m^*, {\underline{c}^\prime} \epsilon_n)$, where ${ \underline{c}^\prime} \le \sqrt{\log(1+\xi)}/2$.
\end{theorem}
 The regime of $\sqrt{c^\prime_{\xi} p_n} \leq m^* \leq \sqrt{p_n}$ is a special case when $\log(c^\prime_{\xi} p_n / (m^*)^2)$ becomes negative and thus the lower bound $\rho_n m^* \log(\frac{{ c^\prime_{\xi}} p_n}{(m^*)^2 })$ becomes uninformative. Similar to the low-rank case, we provide a trivial lower bound of $\rho_n$ that may not be optimal for this situation.
Comparing Theorem \ref{thm:minimax_lower} above to Theorem \ref{low:minimax_lower}, we see that the lower bound under the simultaneous sparse-change and low-rank structure is substantially smaller than that  without the low-rank structure. Such improvement is particularly profound when the change-sparsity level is low, as demonstrated in Figure \ref{fig:lower_bound_plot} in Section \ref{new.sec2.2}.

\subsection{Upper bound result on MOSAIC without low-rank structure} \label{sec:onlysparse}



Building on the methodology established in Section \ref{upper-section}, we can design an analogous CUSUM-based test that attains the minimax-optimal rate with only the sparse-change assumption. In the absence of the low-rank structure, we adapt the formulation of $\bZ^{(\tau)}$ from \eqref{low:zij}, replacing the smoothed matrices obtained through the eigen-decomposition with the original average matrices; that is, 
\begin{equation}\label{nolow:zij}
\bE^{(\tau)} = \{e_{i,j}^{(\tau)}\}_{1 \le i,j \le n} := \sqrt{\frac{\tau}{2  \rho_n}} \left(\bar{\bX}^{(\tau, 1)} - \bar{\bX}^{(\tau, 2)} \right).
\end{equation}

The remaining constructions follow the same steps as in Section \ref{upper-section}. Specifically, we denote two additional independent copies of $\bE^{(\tau)}$ constructed from data splitting as $\dot{\bE}^{(\tau)} = \{\dot{e}_{i,j}^{(\tau)}\}_{1 \le i,j \le n}$ and $\ddot{\bE}^{(\tau)} = \{\ddot{e}_{i,j}^{(\tau)}\}_{1 \le i,j \le n}$, respectively. Denote by $B^{(\tau)}_\calS$ the test constructed analogously to $A^{(\tau)}_\calS$ in \eqref{low:A_stat}, and 
$$
B_{\calS} = \sum_{(i,j) \in \calS} e_{i,j}^{(\tau)} \dot{e}_{i,j}^{(\tau)},
$$
where the screened signal set $\calS$ here is redefined as $\calS=\{(i,j): 1\leq i<j\leq n, \, |\ddot{e}^{(\tau)}_{i,j}| \ge d(m^*)\}$ with $d(m^*) = 3\log(ep_n/(m^*)^2) \I_{\{ m^* < \sqrt{p_n} \}}$. 
Then the MOSAIC test without the low-rank structure takes the form
\begin{equation} \label{def:phi_test}
\phi_n = \I \left(\max_{\tau \in \calT} B_{\calS}^{(\tau)} > r_n\right),
\end{equation}
where the threshold $r_n = c_1\epsilon_n^2/\rho_n$ with $\epsilon_n^2$ defined in \eqref{equ:minimax_rate}, and  $c_1 = c_{\alpha}$ for $m^* \ge \sqrt{p_n}$ and $c_1 = 3 c_{\alpha}$ for $m^* < \sqrt{p_n}$ with $c_{\alpha} = \sqrt{2(\alpha)^{-1} \log_2((2h)^{-1})}$.

\begin{theorem}[Upper bound] \label{thm:minimax_upper}
Assume that time span $T$ satisfies 
\begin{equation}\label{equ:time_span_con1}
     T \gtrsim
  \begin{cases}
      \rho_n^{-1} \frac{\sqrt{p_n}}{m^*} \ \text{ if } m^* \ge \sqrt{p_n},\\
      \rho_n^{-1} \log(\frac{e p_n}{(m^*)^2}) \ \text{ if } m^* < \sqrt{p_n}.
  \end{cases}
 \end{equation}
Then $\phi_n$ defined in \eqref{def:phi_test} is a level-$\alpha$ test under $H_0(\rho_n)$ and has power $\eta > \alpha$ under $\calA(\rho_n ,m^*, { \bar{c}^\prime} \epsilon_n)$, where ${(\bar{c}^\prime)^2} \ge \max\{(64+c_{\eta}), 4c_{\eta}\}$ with $c_{\eta} \ge \max\{16 c_{\alpha}, \sqrt{260 (1-\eta)^{-1}}\}$ and $\epsilon^2_n$ satisfies
\begin{equation}
		\label{equ:minimax_rate1}
			\epsilon^2_n = \begin{cases}
			\rho_n \sqrt{p_n} & \text{ if } m^* \ge \sqrt{p_n},\\
			 \rho_n m^* \log(\frac{e p_n}{(m^*)^2 })  &  \text{ if } m^* < \sqrt{p_n}.
		\end{cases}
	\end{equation}
\end{theorem} 

Theorem \ref{thm:minimax_upper} above establishes that the MOSAIC test without the low-rank structure $\phi_n$ in (\ref{def:phi_test}) is a valid asymptotic level-$\alpha$ test that exhibits nontrivial power $\eta > \alpha$. The detection boundary $\epsilon_n$ and the time span condition in this result align precisely with those in \eqref{equ:minimax_rate} and \eqref{equ:time_span_con}, respectively, of Theorem \ref{thm:minimax_lower},  except for the range $m^* \in \left[\sqrt{c_{\xi}^\prime p_n}, \sqrt{p_n}\right]$.  Combining Theorems \ref{thm:minimax_lower} and \ref{thm:minimax_upper}, we conclude that the detection boundary $\epsilon_n^2$ given in \eqref{equ:minimax_rate} is minimax-optimal for any $m^* \not\in \left[\sqrt{c_{\xi}^\prime p_n}, \sqrt{p_n}\right]$. \cite{wang2021optimal} discussed optimal change-point detection without a low-rank structure. They did not explicitly consider the effect of sparsity, and their lower bound is approximately $\rho_n \sqrt{p_n} \log(T)$. Their result matches our lower bound for the dense case $m^* \geq \sqrt{p_n}$ up to a logarithmic factor of $\log T$. The extra $\log(T)$ term in \cite{wang2021optimal} arises because they focus on finding the location of change point, which requires a stronger signal than our problem of testing the existence. However, since their lower bound does not adapt to the sparsity level, it is suboptimal in the sparse case.


\section{Simulation studies} \label{new.sec.simustud}

In this section, we evaluate the finite-sample performance of MOSAIC proposed in Section \ref{power1} using simulated data. We compare with the matrix $\ell_2$-CUSUM test \citep{enikeeva2021change}, which takes into account the low-rank structure.  

\subsection{Simulation with known network ranks} \label{simu:known}

\begin{figure}[t]
\centering
	\includegraphics[width=5cm]{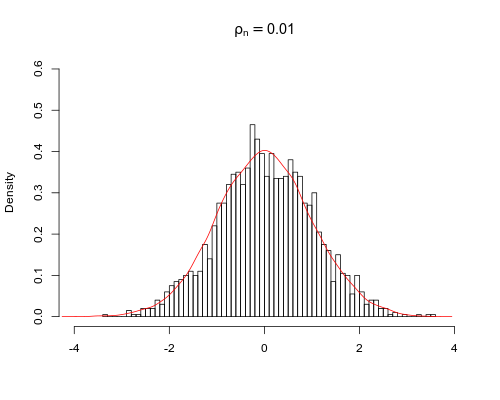}
	\includegraphics[width=5cm]{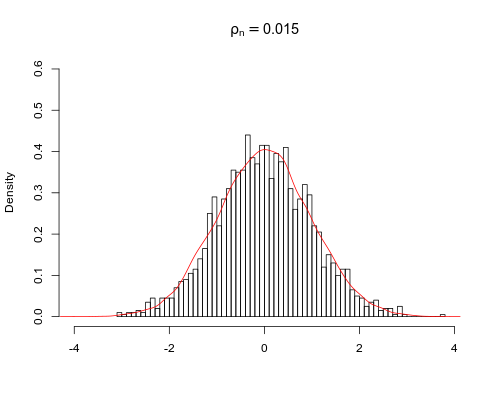}
        \includegraphics[width=5cm]{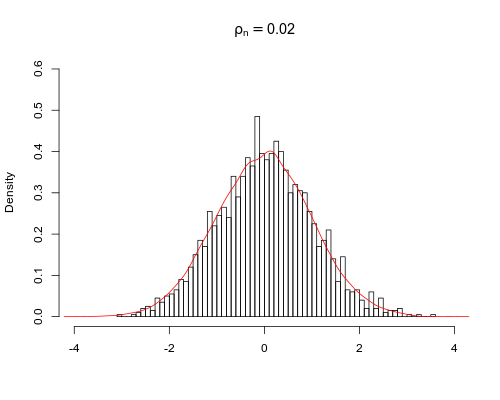}
\caption{The empirical null distribution of MOSAIC statistic $\hat{A}^{(\tau)}_\calS$ across different values of $\rho_n$ when network ranks are known in Section \ref{simu:known}. The red curves represent the standard normal distribution.}
\label{histogram1}
\end{figure}

We start with assuming that the true ranks of relevant networks are known in advance. To obtain the empirical null distribution of the suggested MOSAIC statistic, we consider $T$ homogeneous dynamic networks with a common mean matrix $\bTheta=\rho_n\mathbf{1}_n\mathbf{1}_n^\top + \frac{\rho_n}{2} \bu_n \bu_n^\top$, where $\bu_n = (u_1,\ldots,u_n)^\top$ is obtained by randomly choosing $n/2$ components as $1$ and the remaining ones as $0$. We choose $n=150$ and $T=120$ so that after twofold order-preserved data splitting, each independent copy contains $60$ networks. The network sparsity parameter is set as $\rho_n\in\{0.01,0.015,0.02\}$. Based on $2000$ simulation repetitions, Figure \ref{histogram1} depicts the histograms of the MOSAIC test statistics $\hat{A}^{(\tau)}_{\calS}$ under the null hypothesis $H_0(\rho_n, K^*)$  with working rank $K = K^* =2$, where $\calS$ is chosen randomly as one half of $\{(i,j):1 \le i< j \le n\}$ and $\tau =60$. The reference red curves represent the density of the standard normal distribution. From Figure \ref{histogram1}, we see that the empirical null distribution of MOSAIC statistic aligns well with the target standard normal distribution. The asymptotic normality is further supported by the Shapiro--Wilk test, which yields p-values of $0.339$, $0.797$, and $0.986$, respectively, corresponding to the three different $\rho_n$ values. These results justify the asymptotic null distribution of $\hat{A}^{(\tau)}_{\calS}$ established in \eqref{null-A} of Theorem \ref{Thm:null}.

\begin{table}[htp]
  \centering
  \caption{The empirical power of different methods across varying values of $(\rho_n, s^*, \delta)$ when network ranks are known in Section \ref{simu:known}.}
  \label{tab:full_comparison}
  \sisetup{
    table-format=1.3,
    table-number-alignment=center,
  }
  \begin{tabular}{
    c
    c
    S[table-format=1.3]
    S[table-format=1.3]
    S[table-format=1.3]
    S[table-format=1.3]
    S[table-format=1.3]
    S[table-format=1.3]
  }
    \toprule
    \multirow{2}{*}{$\rho_n$} & 
    \multirow{2}{*}{$s^*$} & 
    \multicolumn{2}{c}{$\delta =0.8$} & 
    \multicolumn{2}{c}{$\delta = 1.0$} & 
    \multicolumn{2}{c}{$\delta = 1.2$} \\
    \cmidrule(lr){3-4} \cmidrule(lr){5-6} \cmidrule(lr){7-8}
    & & {MOSAIC} & {$\ell_2$-CUSUM} & {MOSAIC} & {$\ell_2$-CUSUM} & {MOSAIC} & {$\ell_2$-CUSUM} \\
    \midrule
    0.01 &  0 & 0.070 & 0.000 & 0.070 & 0.000 & 0.070 & 0.000 \\
     &  5 & 0.112 & 0.008 & 0.122 & 0.012 & 0.150 & 0.018 \\
     & 10 & 0.114 & 0.010 & 0.246 & 0.028 & 0.386 & 0.052 \\
     & 15 & 0.258 & 0.046 & 0.438 & 0.082 & 0.684 & 0.238 \\
     & 20 & 0.338 & 0.054 & 0.602 & 0.178 & 0.824 & 0.456 \\
     & 25 & 0.402 & 0.102 & 0.700 & 0.332 & 0.916 & 0.676 \\
     & 30 & 0.486 & 0.174 & 0.814 & 0.482 & 0.946 & 0.828 \\
     & 35 & 0.524 & 0.234 & 0.812 & 0.600 & 0.948 & 0.942 \\
     & 40 & 0.552 & 0.342 & 0.800 & 0.776 & 0.910 & 0.970 \\
    \midrule
    0.02 &  0 & 0.052 & 0.004 & 0.052 & 0.004 & 0.052 & 0.004 \\
     &  5 & 0.064 & 0.002 & 0.090 & 0.006 & 0.132 & 0.014 \\
     & 10 & 0.120 & 0.008 & 0.214 & 0.028 & 0.400 & 0.084 \\
     & 15 & 0.232 & 0.034 & 0.472 & 0.088 & 0.716 & 0.244 \\
     & 20 & 0.308 & 0.068 & 0.634 & 0.192 & 0.886 & 0.490 \\
     & 25 & 0.458 & 0.108 & 0.788 & 0.376 & 0.942 & 0.740 \\
     & 30 & 0.576 & 0.208 & 0.856 & 0.576 & 0.966 & 0.888 \\
     & 35 & 0.632 & 0.324 & 0.900 & 0.726 & 0.978 & 0.964 \\
     & 40 & 0.632 & 0.358 & 0.904 & 0.836 & 0.984 & 0.988 \\
    \midrule
    0.03 &  0 & 0.044 & 0.004 & 0.044 & 0.004 & 0.044 & 0.004 \\
     &  5 & 0.050 & 0.002 & 0.060 & 0.006 & 0.086 & 0.012 \\
     & 10 & 0.114 & 0.004 & 0.238 & 0.020 & 0.416 & 0.052 \\
     & 15 & 0.210 & 0.032 & 0.510 & 0.120 & 0.772 & 0.266 \\
     & 20 & 0.334 & 0.070 & 0.650 & 0.236 & 0.872 & 0.534 \\
     & 25 & 0.458 & 0.118 & 0.826 & 0.414 & 0.986 & 0.818 \\
     & 30 & 0.590 & 0.214 & 0.886 & 0.606 & 0.994 & 0.908 \\
     & 35 & 0.630 & 0.328 & 0.888 & 0.784 & 0.994 & 0.980 \\
     & 40 & 0.720 & 0.448 & 0.938 & 0.890 & 0.994 & 0.998 \\
    \bottomrule
  \end{tabular}
\end{table}

\begin{figure}[t]
\centering
	\includegraphics[width=5cm]{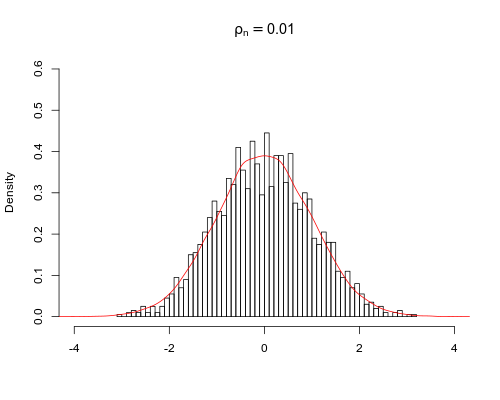}
	\includegraphics[width=5cm]{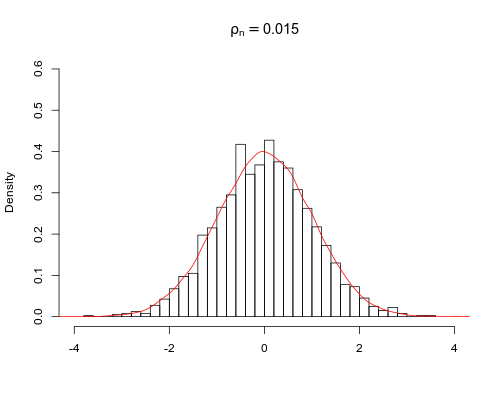}
        \includegraphics[width=5cm]{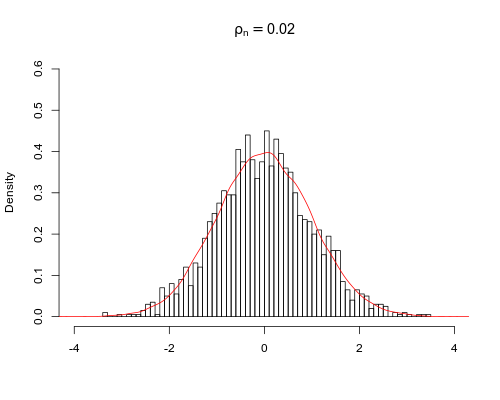}
        \caption{The empirical null distribution of MOSAIC statistic $\hat{A}^{(\tau)}_\calS$ across different values of $\rho_n$ when network ranks are misspecified in Section \ref{new.sec.simu.misspe}. The red curves represent the standard normal distribution.}
        \label{null-mis}
\end{figure}

\begin{table}[htp]
  \centering
  \caption{The empirical powers of different methods across varying values of $(\rho_n, s^*, \delta)$ when network ranks are misspecified in Section \ref{new.sec.simu.misspe}.}
  \label{tab:miss}
  \sisetup{
    table-format=1.3,
    table-number-alignment=center,
  }
  \begin{tabular}{
    c
    c
    S[table-format=1.3]
    S[table-format=1.3]
    S[table-format=1.3]
    S[table-format=1.3]
    S[table-format=1.3]
    S[table-format=1.3]
  }
    \toprule
    \multirow{2}{*}{$\rho_n$} & 
    \multirow{2}{*}{$s^*$} & 
    \multicolumn{2}{c}{$\delta = 0.8$} & 
    \multicolumn{2}{c}{$\delta = 1.0$} & 
    \multicolumn{2}{c}{$\delta = 1.2$} \\
    \cmidrule(lr){3-4} \cmidrule(lr){5-6} \cmidrule(lr){7-8}
    & & {MOSAIC} & {$\ell_2$} & {MOSAIC} & {$\ell_2$-CUSUM} & {MOSAIC} & {$\ell_2$-CUSUM} \\
    \midrule
    0.01 &  0 & 0.070 & 0.000 & 0.070 & 0.000 & 0.070 & 0.000 \\
     &  5 & 0.114 & 0.008 & 0.132 & 0.012 & 0.148 & 0.020 \\
     & 10 & 0.106 & 0.012 & 0.240 & 0.030 & 0.362 & 0.058 \\
     & 15 & 0.242 & 0.046 & 0.428 & 0.096 & 0.654 & 0.242 \\
     & 20 & 0.316 & 0.068 & 0.570 & 0.186 & 0.806 & 0.464 \\
     & 25 & 0.390 & 0.106 & 0.676 & 0.326 & 0.910 & 0.654 \\
     & 30 & 0.478 & 0.180 & 0.784 & 0.488 & 0.946 & 0.818 \\
     & 35 & 0.526 & 0.234 & 0.812 & 0.618 & 0.950 & 0.940 \\
     & 40 & 0.532 & 0.354 & 0.790 & 0.766 & 0.916 & 0.974 \\
    \midrule
    0.02 &  0 & 0.052 & 0.004 & 0.052 & 0.004 & 0.052 & 0.004 \\
     &  5 & 0.072 & 0.000 & 0.080 & 0.008 & 0.118 & 0.020 \\
     & 10 & 0.126 & 0.012 & 0.206 & 0.034 & 0.352 & 0.102 \\
     & 15 & 0.212 & 0.042 & 0.444 & 0.100 & 0.688 & 0.254 \\
     & 20 & 0.288 & 0.072 & 0.606 & 0.210 & 0.860 & 0.486 \\
     & 25 & 0.468 & 0.120 & 0.756 & 0.396 & 0.928 & 0.754 \\
     & 30 & 0.544 & 0.216 & 0.846 & 0.580 & 0.976 & 0.894 \\
     & 35 & 0.612 & 0.334 & 0.876 & 0.752 & 0.978 & 0.964 \\
     & 40 & 0.608 & 0.380 & 0.886 & 0.842 & 0.978 & 0.992 \\
    \midrule
    0.03 &  0 & 0.044 & 0.004 & 0.044 & 0.004 & 0.044 & 0.004 \\
     &  5 & 0.064 & 0.002 & 0.066 & 0.014 & 0.078 & 0.018 \\
     & 10 & 0.110 & 0.008 & 0.218 & 0.022 & 0.380 & 0.082 \\
     & 15 & 0.208 & 0.042 & 0.472 & 0.146 & 0.714 & 0.308 \\
     & 20 & 0.326 & 0.082 & 0.614 & 0.258 & 0.852 & 0.586 \\
     & 25 & 0.438 & 0.146 & 0.790 & 0.436 & 0.978 & 0.830 \\
     & 30 & 0.544 & 0.232 & 0.864 & 0.618 & 0.984 & 0.914 \\
     & 35 & 0.600 & 0.354 & 0.886 & 0.790 & 0.988 & 0.980 \\
     & 40 & 0.700 & 0.480 & 0.934 & 0.898 & 0.994 & 1.000 \\
    \bottomrule
  \end{tabular}
\end{table}

We now evaluate the empirical power of MOSAIC under the alternative hypothesis $H_1(\rho_n, K^*, s^*, \epsilon_n)$. We set the common mean matrix of the first  $\tau^* = 60$ networks as $\bTheta_1 = \rho_n\mathbf{1}_n\mathbf{1}_n^\top$, and that of the remaining $T-\tau^*$ networks as $\bTheta_2 = \bTheta_1 + \delta \sqrt{\rho_n/s^*}\tilde{\bu}_n \tilde{\bu}_n^\top$, where $\tilde{\bu}_n = (\tilde{u}_1,\ldots,\tilde{u}_n)^\top$ is obtained by randomly choosing $s^*$ components as $1$ and the remaining ones as $0$, and parameter $\delta\in\{0.8, 1.0, 1.2\}$ quantifies the degree of separation between the null and alternative hypotheses. We set $c_d = 1$ in \eqref{new.eq.FL019} throughout all simulations.

Table \ref{tab:full_comparison} summarizes the empirical powers of the MOSAIC and $\ell_2$-CUSUM under different values of $\rho_n$, $s^*$, and $\delta$. From Table \ref{tab:full_comparison}, it is seen that MOSAIC exhibits superior power compared to the $\ell_2$-CUSUM method, especially when the change-sparsity level $s^*$ is small. For example, when $\rho_n=0.03$, $s^*=15$, and $\delta=1.2$, the power of the MOSAIC test is $0.772$, much higher than that of the $\ell_2$-CUSUM, $0.266$. Notably, the $\ell_2$-CUSUM test attains comparable power only when $s^*$ is considerably large, e.g., $s^* \geq 30$. These findings underscore the effectiveness of MOSAIC in scenarios when sparsity plays a critical role in signal recovery.

\subsection{Simulation with misspecified network ranks} \label{new.sec.simu.misspe}

In the second simulation study, we further assess the robustness of MOSAIC when the network ranks are misspecified. Under the null hypothesis, we define $\tilde{\bTheta} = \bTheta + 0.05 \rho_n \bu_n^c (\bu_n^c)^\top$ with $\bu^c := \big((1 - u_{n,i})v_i, i \in [n]\big)^\top$. Here, $u_{n,i}$ is the $i$th component of $\bu_n$ defined in Section \ref{simu:known}, and $v_i$ is a 
Rademacher random variable. The remaining simulation setting and test statistics implementation are identical to those in Section \ref{simu:known} under the null hypothesis. It is seen that the true rank is $3$. Figure \ref{null-mis} depicts the empirical distributions of MOSAIC statistics under the null hypothesis with misspecified working ranks  $(K=2)$. Despite the rank misspecification, the empirical null distributions of MOSAIC still remain approximately normal, as confirmed by the Shapiro--Wilk normality test p-values of $0.119$, $0.970$, and $0.493$, respectively. These results demonstrate the robustness of the MOSAIC test under mild rank misspecification.

To evaluate the power of MOSAIC under misspecified network ranks, we follow the same setup as described in Section \ref{simu:known}, except that an additional term is introduced for $\bTheta_2$. Specifically, set $\bTheta_2 = \bTheta_1 + \delta \sqrt{\rho_n/s^*}\tilde{\bu}_n \tilde{\bu}_n^\top + 0.1\rho_n (\tilde{\bu}_n^c) (\tilde{\bu}_n^c)^\top$ with $\tilde{\bu}^c := \big((1 - \tilde{u}_{n,i})\tilde{v}_i, i \in [n]\big)^\top$. Here, $\tilde{u}_{n,i}$ is the $i$th component of $\tilde{\bu}_n$ defined in Section \ref{simu:known} and $\tilde{v}_i$ is a 
Rademacher random variable. As a result, the true rank of $\bTheta_2$ is $K^* = 3$, whereas we adopt a working rank of $K=2$. Table \ref{tab:miss} reports the empirical powers of the MOSAIC and $\ell_2$-CUSUM across different values of $\rho_n$, $s^*$, and $\delta$. Again, the MOSAIC test demonstrates consistently superior power performance across all settings, with its power increasing rapidly as the change-sparsity level $s^*$ grows. In contrast, the $\ell_2$-CUSUM test exhibits a more gradual improvement in power with increasing $s^*$. Overall, these results indicate that MOSAIC remains effective even with misspecified network ranks, enabling robustness in practical applications.

\section{Real data application} \label{sec.realdata}

In this section, we apply MOSAIC to analyze the dynamic pattern of a collaboration network data set in \citet{gao2023large}, which assembles all articles and author information from $42$ representative journals in the statistics and data science field, during the years of 1992 to 2021. It is well-known that the statistics and data science community has grown remarkably, with a steady rise in publication volume. This motivates researchers to examine the collaboration patterns over recent years. \citet{ji2022co} reported an interesting ``statistics triangle'' comprising three tightly connected subcommunities: Nonparametrics, Biostatistics, and Bayesian Statistics. Then it is of interest to investigate when such tightly knit structures get to emerge and whether the coauthorship network changes over time.

For each year, we construct an undirected coauthorship network in which nodes are authors and an edge indicates at least one coauthored paper in year $t$. The corresponding adjacency matrix $\bX^{(t)}$ has entries $x^{(t)}_{i_1,i_2}=1$ if authors $i_1$ and $i_2$ coauthored in year $t$ and $0$ otherwise, with the convention $x^{(t)}_{i,i}=0$. We focus on authors with at least one publication per year, yielding $1{,}998$ unique authors. The resulting yearly networks are extremely sparse, with an average of $96.067$ observed edges per year.

The MOSAIC is applied to test structural changes in this dynamic network. Guided by \citet{ji2022co}, we set the working rank to $K=3$. We use bandwidth $h=0.1$ and thus candidate change-point set $\mathcal{T}=\{4,8\}$. For comparison, we also implement the network change-point methods of \citet{wang2021optimal} and \citet{enikeeva2021change}. Both of these competing methods fail to detect any change point, whereas MOSAIC yields a statistic of $3.927$, exceeding the threshold $z_{\alpha/(2|\mathcal{T}|)}=z_{0.0125}=2.241$ at level $\alpha=0.05$, indicating a statistically significant change. The maximizing candidate $\tau=8$ corresponds to calendar year 1999.

\begin{figure}[H]
	\centering
	\includegraphics[width=\linewidth,height=0.6\linewidth]{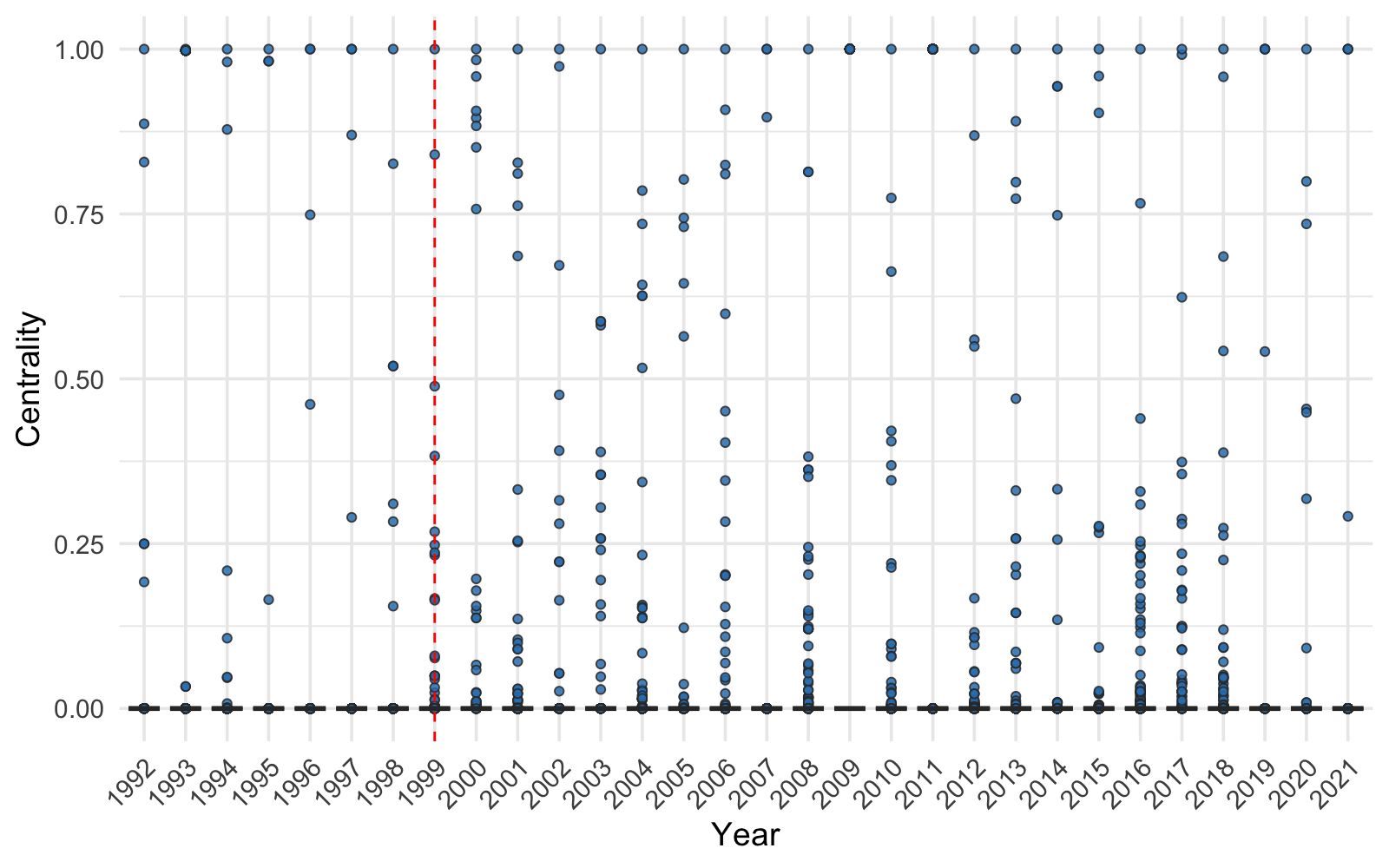}
	\caption{Boxplots of eigenvector centrality over time in the coauthorship network (1992--2021). The red dashed line marks 1999, the year at which MOSAIC attains its maximum.} 
	\label{fig_realdata}
\end{figure}

To characterize how the collaboration pattern shifts, we compute the eigenvector centrality \citep{bonacich1987power} for each node in each year and summarize its evolution with boxplots across time in Figure \ref{fig_realdata}. The centrality series exhibits a clear change around 1999: in the early 1990s, most values lie near zero, indicating weak or isolated positions in the coauthorship network; around 1999 (marked by a vertical red dashed line), the distribution shifts upward, with more authors attaining moderate to high centrality and a thicker upper tail, including repeated values near $1.0$ in the 2000s and 2010s. Such pattern suggests the emergence of a more integrated core of highly connected authors after 1999 while overall sparsity persists, consistent with a structural reorganization of collaboration patterns at the end of the 1990s.

\section{Discussions} \label{new.sec.discu}

We have investigated in this paper the problem of dynamic network change-point detection and suggested a new inference framework, MOSAIC, for this problem, accommodating both low-rank and sparse structures. The minimax-optimal, sparsity-adaptive detection boundary for testing the existence of change points has been established. To enable practical implementation, we have introduced an innovative residual-based adjustment technique coupled with a signal selection procedure. This gives rise to a pivotal test statistic that is tuning-free and robust to rank misspecification. We have also obtained the minimax-optimal results for the setting without the low-rank structure. It would be interesting to extend this work to more general dynamic network model settings with time series dependency over the time horizon, and to incorporate the perspectives of network causal inference and generalized Laplacian matrices (see, e.g., \cite{DuDingFanLv2025,FanFanLvYangYu2025}). These problems are beyond the scope of the current paper and will be interesting topics for future research.

\bibliography{wpref}
\bibliographystyle{chicago}

\newpage
\appendix
\setcounter{page}{1}
\setcounter{section}{0}
\setcounter{equation}{0}
\renewcommand{\theequation}{A.\arabic{equation}}

\begin{center}{\bf \Large Supplementary Material to ``MOSAIC: Minimax-Optimal Sparsity-Adaptive Inference for Change Points in Dynamic Networks''}

\bigskip

Yingying Fan, Jingyuan Liu, Jinchi Lv and Ao Sun
\end{center}

\noindent This Supplementary Material contains all the proofs of main results, some key lemmas, and asymptotic expansions of the empirical spiked eigenvalues and eigenvectors, as well as some additional technical results. All the notation is the same as defined in the main body of the paper.

\section{Proofs of Theorems \ref{low:minimax_lower}--\ref{low:minimax_upper} and some key lemmas} \label{new.sec.A}

To facilitate the technical analyses, let us introduce some additional notation. Let $\mathbf{1}_{\calS} \in \R^{n}$ be a vector with all components on index set $\calS \subset [n]$ being $1$ and the rest being $0$. Define $\bI_{\calS} \in \R^{n \times n}$ as a diagonal matrix with all entries on index set $\calS \subset \{(i,i), \, i \in [n]\}$ being $1$ and the rest being $0$. For a given measure $\nu$ and measurable function $f(\cdot)$, $\E_{x \sim \nu}(f(x)) = \int f(x) \nu(dx)$ represents the expectation of $f(x)$ under measure $\nu$. For two density functions $\phi_0(\cdot)$ and $\phi_1(\cdot)$ corresponding to distributions $\Psi_0$ and $\Psi_1$, respectively, the chi-square divergence is defined as  
\begin{equation} \label{new.eq.FL001}
\chi^2(\phi_0, \phi_1) = \E_{x \sim \Phi_0}({\phi_1^2(x)}/{\phi_0^2(x)}) -1.
\end{equation} 
For any two measures $\nu_1$ and $\nu_2$, denote by $\nu_1 \otimes \nu_2$ the product measure of $\nu_1$ and $\nu_2$. For a matrix $\bX \in  \R^{n \times n}$, define $\diag(\bX) = \diag\{x_{11},\ldots, x_{nn}\}$ by extracting all the diagonal entries of $\bX = (x_{ij})_{1 \leq i,j \leq n}$.

\subsection{Proof of Theorem \ref{low:minimax_lower}} \label{proof:lower}

The main idea of the proof is to construct a suitable prior distribution $\nu$ on the alternative hypothesis parameter space $\mathcal A(\rho_n, K^*, s^*, \underline{c} \epsilon)$, and show that the worst case power can be upper bounded by the total variation distance between the null distribution and the mixture alternative distribution induced by prior $\nu$. To this end, let us consider a single parameter $\bTheta^0_0 \in \calN(\rho_n, K^*)$ in the null space. We also consider a parameter $g:=\bTheta^0_1$ sampled from the prior distribution $\nu$ on the alternative parameter space, where $\supp(\nu) \subset \calA(\rho_n, K^*, s^*, \underline{c}\epsilon)$ and $\supp(\cdot)$ stands for the support of a given distribution. The joint distribution of dynamic networks with parameters in $\bTheta_0^0$ is denoted as $\Pr_0$, and we use $\E_{0}$ to represent the expectation under $\Pr_0$. Given a single parameter $g \sim \nu$ sampled from the alternative space, the joint distribution of dynamic networks with parameters in $g$ is denoted as $\Pr_g$, and we use $\E_{g}$ to represent the expectation under $\Pr_g$. For a given prior $\nu$ and a parameter $h$ drawn from prior distribution $\nu$, denote by $\bar{\Pr} := \E_{g \sim \nu}(\Pr_g) $ the induced average probability measure, and $\bar{\E}$ the expectation under such average probability measure.

For each test $\psi_0$, it holds that 
\begin{equation} \label{new.eq.FL002}
	\begin{aligned}
		& \sup_{g \in \mathcal{A}(\rho_n,K^*,s^*,\underline{c}\epsilon)} \left\{\E_{0}(\psi_0) +\E_{g}(1-\psi_0) \right\} \\
        & \ge \inf_{\psi}\left\{\sup_{g \in \supp(\nu)} (\E_{0}(\psi) +\E_{g}(1-\psi) ) \right\} \\
		& \ge \inf_{\psi}\left\{\E_{0}(\psi) +\bar{\E}(1-\psi) \right\} \\
		& = \inf_{A}\left\{\Pr_0(A) + \overline{\Pr}(A^c)\right\} \\
		& = 1 - \|\mathbb{P}_0 - \bar{\mathbb{P}} \|_{TV},
	\end{aligned}
	\end{equation}
	where the second inequality above is because the supremum can always be lower bounded by the average (i.e., $\sup_g \E_g(1-\psi) \geq \bar{\E}(1-\psi)$), the second last equality above is because each testing procedure $\psi$ corresponds to a rejection area $A$, and $\|\mathbb{P}_0 - \bar{\mathbb{P}} \|_{TV}$ in the last equality above represents the total variation distance between two distributions. Then if test $\psi_0$ has significance level $\alpha$, it follows from (\ref{new.eq.FL002}) that 
    \begin{equation} \label{new.eq.FL003}
    \alpha + \|\mathbb P_0-\bar{\mathbb P}\|_{TV} \geq \inf_{g \in \mathcal A(\rho_n,K^*,s^*,\underline{c}\epsilon)}\mathbb E_g(\psi_0). 
    \end{equation}
    Hence, in view of (\ref{new.eq.FL003}) we have that 
	\begin{equation} \label{new.eq.FL004}
	\inf_{g \in \calA(\rho_n,K^*,s^*,\underline{c}\epsilon)} \Pr_g\left\{\psi_0 \text{ rejects } H_{0}\right\} = \inf_{g \in \calA(\rho,K^*,s^*,\underline{c}\epsilon)}\E_g(\psi_0) \le \alpha + \|\mathbb{P}_0 - \bar{\mathbb{P}} \|_{TV}.
	\end{equation}
 
	We aim to establish that for each given $\eta > \alpha$ and some constant $\underline{c} > 0$, no test of significance level $\alpha$ can satisfy \eqref{sparse:gtr}. To this end, it suffices to show that 
    \begin{equation} \label{new.eq.FL005}
    \|\mathbb{P}_0 - \bar{\mathbb{P}} \|_{TV} \le \eta - \alpha 
    \end{equation}
    for some carefully designed prior distribution $\nu$.
	Since $\|\mathbb{P}_0 - \bar{\mathbb{P}} \|_{TV}^2  \le \chi^2(\mathbb{P}_0, \bar{\mathbb{P}})$ (see, e.g., (2.27) in \cite{tsybakov09}), to establish (\ref{new.eq.FL005}) it is sufficient to prove that 
    \begin{equation} \label{new.eq.FL006}
    \chi^2(\mathbb{P}_0, \bar{\mathbb{P}}) \le \xi :=(\eta - \alpha)^{2}.
    \end{equation}
    For each given $\xi > 0$, with the aid of Lemma \ref{lemma:chi_dis} in Section \ref{new.sec.A.3}, it suffices to find a probability measure $\nu$ such that independent random variables $\gamma = \{\gamma_{i,j}^{(t)}: t \in [T], (i,j) \in \Omega\}$ and $\zeta = \{\zeta_{i,j}^{(t)}: t \in [T], (i,j) \in \Omega\}$ with the joint distribution $\nu \otimes \nu$ satisfy that
	\begin{equation}
        \label{equ:diver}
	    \E_{(\gamma, \zeta) \sim \nu \otimes \nu } \left\{\exp\left(\sum_{t=1}^T \sum_{i < j} \frac{\gamma_{i,j}^{(t)} \zeta_{i,j}^{(t)}}{\rho_n(1-\rho_n)}  \right)\right\} \le 1 + \xi.
	\end{equation}
Thus, the problem reduces to constructing a prior distribution $\nu$ such that $\supp(\nu) \subset \calA(\rho_n,K^*, s^*, \underline{c}\epsilon)$ and condition \eqref{equ:diver} above is satisfied. The sparsity 
parameter $s^*$ is critical in the technical analysis. We will separate the remaining proof into three parts according to the sparsity level.

\medskip

	\textbf{Part I. Dense alternative with $s^* \ge \sqrt{n}$}. Let 
    $\beta^2 = c_{1,\xi}/s^*$ with $c_{1,\xi} $ a positive constant related to $\xi$ that will be determined later. 
    We will construct prior distribution $\nu$ in the following way such that each element in $\boldsymbol{\Theta}^0 \in \calA(\rho_n, K^*, s^*, \underline{c}_1 \epsilon_n)$ (setting $\epsilon = \epsilon_n$ as defined in \eqref{low:minimax_rate}) is independently generated from the sampling process below:
	\begin{enumerate}
		\item[1)] Uniformly sample a subset $\calS \subset [n]$ of cardinality $s^*$ and divide it into $K^*-1$ partitions $\calS_k$ with $k =1,\ldots, K^* - 1$ such that $|\calS_k| = s^*/(K^*-1)$ and $\calS = \cup_{k=1}^{K^* - 1} \calS_k$.
        
		\item[2)] Independent of $\calS$, generate $l \sim \operatorname{Unif}\left\{{ \log_2(h T), \log_2(h T) + 1 },\ldots, \log_2(T/2)\right\}$ with $h \in (0,\frac{1}{2})$ an arbitrarily small constant.
        
		\item[3)] Independent of $l$ and given $\calS_k$ with $k \in [K^*-1]$, sample $K^*-1$ vectors $\bu_{k} := \bu_{\calS_k} = (u_{\calS_k,i})^\top \in \R^{n}$ with $u_{\calS_k,i} \stackrel{i.i.d.}{\sim} \operatorname{Unif}(\{-1,1\})$ for $i \in \calS_k$, and $u_{\calS_k,i}  = 0$ for $i \in S_k^c$, where i.i.d. is short for independent and identically distributed. Then represent these vectors in a compact matrix form $\bU:= [\bu_1, \ldots, \bu_{K^*-1}] \in \R^{n \times (K^*-1)}$.
        
		\item[4)] Given the triplet $(\calS, l, \bU)$ sampled in the previous steps, define $\bTheta_2 = (\rho_n)_{1\le i, j \le n} \in \R^{n \times n}$ and $\bTheta_1 = \bTheta_2 + \rho_n^{1/2} \frac{\beta}{\sqrt{2^l}} \bU \boldsymbol{\Lambda} \bU^\top$, where $\boldsymbol{\Lambda} = \diag\{\lambda_1, \ldots, \lambda_{K^*-1}\}$ satisfies that $1 \le \lambda_k \le 2$ for $ k \in [K^*-1]$ and $\min_{k \in [K^*-2]}\lambda_k/\lambda_{k+1} \ge 1+c_0$ for some $c_0 > 0$. Then based on the representation in Lemma \ref{lemma:chi_dis}, we have $$\gamma_{i,j}^{(t)} = \rho_n^{1/2} \frac{\beta}{\sqrt{2^l}} \sum_{k=1}^{K^*-1} \lambda_k u_{k,i} u_{k,j}$$ for $1 \le t \le 2^l$, and $\gamma_{i,j}^{(t)} = 0$ for $2^l + 1 \le t \le T$. We refer to $\gamma_{i,j}^{(t)}$ as the jump size hereafter. 
	\end{enumerate}
    
	From the above construction process, we see that $$\Rank(\bTheta_2) \le \Rank(\bTheta_1) \le K^*.$$ 
    It is also easy to show that
    $$\| \bTheta_1\|_{\infty} \lesssim \rho_n + \rho_n^{1/2} \beta/\sqrt{T} \lesssim \rho_n,$$ where the last inequality above is due to the fact that
	$\sqrt{\frac{\rho_n \beta^2}{T}} \asymp \sqrt{\frac{\rho_n}{s^* T}}  \lesssim \rho_n$ in light of $T \gtrsim (\rho_n s^*)^{-1}$.
	Observe that for any $\bTheta^0$ sampled from the above process, we have $\tau^* = 2^l$. It follows that the corresponding $\boldsymbol{\Delta} = \bTheta_1-\bTheta_2$ of $\bTheta^0$ satisfies that 
    \begin{equation}\label{equ:june20:thm1:01}
        \begin{aligned}
            \sqrt{\frac{\tau^*(T-\tau^*)}{T} \|\vech(\boldsymbol{\Delta})\|_2^2} =& \sqrt{\frac{\tau^*(T-\tau^*)}{2T}} \|\boldsymbol{\Delta} -\diag(\boldsymbol{\Delta})\|_F\\
            \ge &\sqrt{\frac{\tau^*(T-\tau^*)}{2T}}\left( \|\boldsymbol{\Delta}\|_F -\|\diag(\boldsymbol{\Delta})\|_F\right).
        \end{aligned}
    \end{equation}
    Moreover, it holds that 
	\begin{equation}\label{Feb9:equ02}
    \begin{aligned}
	    \frac{\tau^*(T-\tau^*)}{T}\|\boldsymbol{\Delta} \|_F^2 & =\frac{2^l(T- 2^l)}{T} \frac{\beta^2 \rho_n}{2^l} \| \bU\boldsymbol{\Lambda}\bU^\top\|_F^2  \\
        & \ge  \frac{\lambda_{K^*-1}^2(s^*)^2\beta^2 \rho_n}{2(K^*-1)} \ge \frac{c_{1,\xi}}{2(K^*-1)} \epsilon^2_n,
        \end{aligned}
	\end{equation}
	where in the last inequality above, we have used $\lambda_{K^*-1} \ge 1$ and the fact that the disjointness of $\calS_k$ with $ k \in [K^*-1]$ entails that 
	$$
    \begin{aligned}
        \| \bU\boldsymbol{\Lambda}\bU^\top\|_F^2 =& \tr(\bU \boldsymbol{\Lambda} \bU^\top \bU \boldsymbol{\Lambda} \bU^\top) = \frac{s^*}{K^*-1}  \tr\left(  \bU \boldsymbol{\Lambda}^2 \bU^\top\right)\\
        =& \frac{s^*}{K^*-1}\tr\left(\sum_{k=1}^{K^*-1} \lambda_k^2 \bu_{k} \bu_{k}^\top \right) \ge \frac{(\lambda_{K^*-1} s^*)^2}{K^*-1}
    \end{aligned}
	$$
    since $\bU^\top \bU = s^* \bI_{K^*-1}/(K^*-1)$.
    
    Meanwhile, we can show that 
    \begin{equation}\label{equ:june20:thm1:02}
       \frac{\tau^*(T-\tau^*)}{T} \|\diag(\boldsymbol{\Delta})\|_F^2 \le  c_{1,\xi} \rho_n \lambda_1^2 \le \frac{c_{1,\xi}}{16(K^*-1)} \epsilon^2_n
    \end{equation}
    for sufficiently large $n$. Combining \eqref{equ:june20:thm1:01}--\eqref{equ:june20:thm1:02} leads to 
   \begin{equation}\label{equ:june20:thm1:03}
       \begin{aligned}
         \frac{\tau^*(T-\tau^*)}{T} \|\vech(\boldsymbol{\Delta})\|_2^2 \ge   \frac{c_{1,\xi}}{8(K^*-1)} \epsilon^2_n,
       \end{aligned}
   \end{equation}
   where $\vech(\cdot)$ denotes the matrix vectorization. Then it follows from (\ref{equ:june20:thm1:03}) and Lemma \ref{lem:Feb9:01} in Section \ref{new.sec.A.3} that $\supp(\nu) \subset \calA(\rho_n, K^*, s^*, \underline{c}_1 \epsilon_n)$, where
\begin{equation}\label{equ:March5:cprime1}
\begin{aligned}
   \underline{c}_1 =& \sqrt{\frac{c_{1,\xi}}{8(K^*-1)}}.
\end{aligned}
\end{equation}

	For two independently sampled triplets $(\calS, l, \bU)$ and $(\calT, m, \bV)$ from the process described above, denote by $ \{\gamma_{i,j}^{(t)}\}_{t \in [T], 1 \le i,j\le n}$ and  $\{\zeta_{i,j}^{(t)}\}_{t \in [T], 1 \le i,j\le n}$ the corresponding jump sizes in Step 4, respectively. We will show that 
    condition \eqref{equ:diver} holds for such  $\{\gamma_{i,j}^{(t)}\}_{t \in [T], 1 \le i,j\le n}$ and $\{\zeta_{i,j}^{(t)}\}_{t \in [T], 1 \le i,j\le n}$. Let us define $\boldsymbol{\Gamma}^{(t)} := \{\gamma_{i,j}^{(t)}\}_{1\le i,j \le n}$ and $\boldsymbol{\Delta}^{(t)} := \{\zeta^{(t)}_{i,j}\}_{1\le i, j \le n}$. Then it holds that $$\boldsymbol{\Gamma}^{(t)} = \rho_n^{1/2} \frac{\beta}{\sqrt{2^l}} \bU \boldsymbol{\Lambda} \bU^\top$$ for $1 \le t \le 2^{l}$ and $\boldsymbol{\Gamma}^{(t)} = \mathbf{0}_{n \times n}$ for $2^{l}+1 \le t \le T$, and $$\boldsymbol{\Delta}^{(t)} = \rho_n^{1/2} \frac{\beta}{\sqrt{2^m}} \bV \boldsymbol{\Lambda} \bV^\top$$ for $1 \le t \le 2^{m}$ and $\boldsymbol{\Delta}^{(t)} = \mathbf{0}_{n \times n}$ for $2^{m}+1 \le t \le T$. Hence, we can deduce that 
    \begin{equation}\label{equ:june18:thm1:01}
	\begin{aligned}
		&\sum_{t=1}^T \sum_{i < j}\gamma_{i,j}^{(t)} \zeta_{i,j}^{(t)}  
        = \frac{1}{2} \sum_{t=1}^T\left( \sum_{1 \le i, j\le n}\gamma_{i,j}^{(t)} \zeta_{i,j}^{(t)}-  \sum_{ i \in [n]}\gamma_{i,i}^{(t)} \zeta_{i,i}^{(t)}\right)\\
        & = \frac{1}{2} \sum_{t=1}^T\left( \sum_{i=1}^n \sum_{j=1}^n \gamma_{i,j}^{(t)} \zeta_{j,i}^{(t)}   -  \sum_{ i \in [n]}\gamma_{i,i}^{(t)} \zeta_{i,i}^{(t)}\right)\\
		 & = \frac{1}{2}\sum_{t=1}^T \left( \tr(\boldsymbol{\Gamma}^{(t)}  \boldsymbol{\Delta}^{(t)})  - \tr\left(\diag\left( \boldsymbol{\Gamma}^{(t)} \right) \diag\left( \boldsymbol{\Delta}^{(t)}\right)\right) \right)\\
		& =  (2^m \land 2^l) \frac{\beta^2 \rho_n}{2\sqrt{2^{l+m}}} \left\{\tr(\bU \boldsymbol{\Lambda}\bU^\top \bV\boldsymbol{\Lambda} \bV^\top) - \tr\left(\diag\left(\bU \boldsymbol{\Lambda}\bU^\top \right) \diag\left( \bV \boldsymbol{\Lambda}\bV^\top\right)\right) \right\}\\
		& = \frac{\beta^2 \rho_n}{2^{|l-m|/2+1}} \left(\sum_{k_1=1}^{K^*-1} \sum_{k_2 = 1}^{K^*-1} \lambda_{k_1} \lambda_{k_2}(\bu_{k_1}^\top \bv_{k_2} )_2^2 \right.\\
        & \quad\left.-  \sum_{k_1=1}^{K^*-1} \sum_{k_2 = 1}^{K^*-1}\lambda_{k_1} \lambda_{k_2}\tr\left(\diag\left(\bu_{k_1} \bu_{k_1}^\top \right) \diag\left( \bv_{k_2} \bv_{k_2}^\top\right)\right)\right)\\
		& = \frac{\beta^2 \rho_n}{2^{|l-m|/2+1}} \sum_{k_1=1}^{K^*-1} \sum_{k_2 = 1}^{K^*-1} \left\{ \lambda_{k_1} \lambda_{k_2} \sum_{i \in \calS_{k_1} \cap \calT_{k_2}} \sum_{j \in \calS_{k_1} \cap \calT_{k_2}, j \ne i} u_{k_1, i} v_{k_2, i} u_{k_1, j} v_{k_2, j} \right\}.
	\end{aligned}
	\end{equation}

    Further, it holds that 
	\begin{equation}\label{add:2.1}
		\begin{aligned}
			&\E_{(\gamma, \zeta) \sim \nu \otimes \nu } \left\{\exp\left(\sum_{t=1}^T \sum_{i <j} \frac{\gamma_{i,j}^{(t)} \zeta_{i,j}^{(t)}}{\rho_n(1-\rho_n)}  \right)\right\}\\
			 & = \E \left\{\exp\left( \frac{\beta^2 \rho_n }{2^{1+|l-m|/2}  \rho_n(1-\rho_n)} \sum_{k_1=1}^{K^*-1} \sum_{k_2 = 1}^{K^*-1} \lambda_{k_1} \lambda_{k_2} \right.\right.\\
             &\quad\left.\left. \times \sum_{i \in \calS_{k_1} \cap \calT_{k_2}} \sum_{j \in \calS_{k_1} \cap \calT_{k_2}, j \ne i} u_{k_1, i} v_{k_2, i} u_{k_1, j} v_{k_2, j}  \right) \right\}\\
			  & = \prod_{1 \le k_1, k_2 \le K^*-1}\E \left\{\exp\left( \frac{\lambda_{k_1} \lambda_{k_2}\beta^2  }{2^{1+|l-m|/2}(1-\rho_n)} \right.\right.\\
             &\quad\left.\left. \times \sum_{i \in \calS_{k_1} \cap \calT_{k_2}} \sum_{j \in \calS_{k_1} \cap \calT_{k_2}, j \ne i} u_{k_1, i} v_{k_2, i} u_{k_1, j} v_{k_2, j}  \right) \right\}\\
			   & = \prod_{1 \le k_1, k_2 \le K^*-1}\E \left\{\exp\left( \frac{\lambda_{k_1} \lambda_{k_2}\beta^2  }{2^{1+|l-m|/2}(1-\rho_n)} \sum_{i \in \calS_{k_1} \cap \calT_{k_2}}  \sum_{j \in \calS_{k_1} \cap \calT_{k_2}, j \ne i} u_{k_1, i}  u_{k_1, j}  \right) \right\}\\
               & \le \prod_{1 \le k_1, k_2 \le K^*-1}\E \left\{\exp\left( \frac{\lambda_{k_1} \lambda_{k_2}\beta^2  }{2^{|l-m|/2}} \sum_{i \in \calS_{k_1} \cap \calT_{k_2}}  \sum_{j \in \calS_{k_1} \cap \calT_{k_2}, j \ne i} u_{k_1, i}  u_{k_1, j}  \right) \right\},
		\end{aligned}
	\end{equation}
where the second last equality above is due to $u_{k_1,i} \stackrel{i.i.d.}{\sim} \operatorname{Unif}(\{-1,1\})$, $v_{k_2,i} \stackrel{i.i.d.}{\sim}\operatorname{Unif}(\{-1,1\})$, and the fact that all $u_{{k_1},i}$'s and $v_{{k_2},i}$'s are independent of each other, leading to $u_{{k_1},i} v_{{k_2},i} \stackrel{i.i.d.}{\sim} \operatorname{Unif}(\{-1,1\})$, and the last inequality above has used the fact of $1 - \rho_n \ge 1/2$. 

We now aim to bound each summand in \eqref{add:2.1} above. To this end, we define $\kappa^{l,m}_{k_1,k_2}:= \frac{\lambda_{k_1} \lambda_{k_2}\beta^2  }{2^{|l-m|/2}}$, set 
$$\mathbf{1}_{k_1,k_2} := \mathbf{1}_{\calS_{k_1} \cap \calT_{k_2}} \ \text{ and } \ \bI_{k_1,k_2} := \bI_{\calS_{k_1} \cap \calT_{k_2}}, $$
and choose $c^2_{1,\xi} \le (K^*-1)/(128\lambda_1^4)$, where $\mathbf{1}_{\calS_{k_1} \cap \calT_{k_2}}$ and $\bI_{\calS_{k_1} \cap \calT_{k_2}}$ are defined at the begining of this section. Then it follows that 
\begin{equation}\label{Jan26:01}
  \begin{aligned}
  & 64 (\kappa^{l,m}_{k_1,k_2})^2\| \mathbf{1}_{k_1,k_2} \mathbf{1}_{k_1,k_2}^\top- \bI_{k_1,k_2}\|_2^2\\
    & =\frac{64 \lambda^2_{k_1} \lambda^2_{k_2}\beta^4}{2^{|l-m|}} \| \mathbf{1}_{k_1,k_2} \mathbf{1}_{k_1, k_2}^\top- \bI_{k_1 k_2}\|_2^2 \\
    &\le 64\lambda_{k_1}^2 \lambda_{k_2}^2 c^2_{1,\xi} (s^*)^{-2} \| \mathbf{1}_{k_1,k_2} \mathbf{1}_{k_1,k_2}^\top- \bI_{k_1,k_2}\|_F^2 \\
    &\le  64 \lambda_1^4 c^2_{1,\xi}/(K^*-1) \le 1/2,
\end{aligned}  
\end{equation}
where the first inequality above has used the facts that the matrix Frobenius norm is an upper bound of the matrix operator norm and $2^{|l-m|} \ge 1$, and the second inequality above is due to $\| \mathbf{1}_{k_1,k_2} \mathbf{1}_{k_1,k_2}^\top- \bI_{k_1,k_2}\|_F^2 \le (s^*)^2/(K^*-1)$. 
Moreover, since 
$$
(8 \| \mathbf{1}_{k_1,k_2} \mathbf{1}_{k_1,k_2}^\top- \bI_{k_1,k_2}\|_F)^{-1} \ge \frac{\sqrt{K^*-1}}{8 s^*},
$$
with the choice of $c_{1,\xi}$ it holds that 
$$
\begin{aligned}
\kappa^{l,m}_{k_1,k_2} & \le \lambda_1^2 c_{1,\xi} (s^*)^{-1} \le (8\| \mathbf{1}_{k_1,k_2} \mathbf{1}_{k_1,k_2}^\top- \bI_{k_1,k_2} \|_F)^{-1} \\
&\le (8\| \mathbf{1}_{k_1,k_2} \mathbf{1}_{k_1,k_2}^\top- \bI_{k_1,k_2} \|_2)^{-1},
\end{aligned}
$$
where we have utilized $1 \le \lambda_k \le 2$ with $ k \in [K^*-1]$. 

Let $\bu := (u_i)_{i \in [n]}$ be a Rademacher random vector with $u_i \stackrel{i.i.d.}{\sim} \operatorname{Unif}(\{-1,1\})$. Then an application of Lemma \ref{lem: rademacher} in Section \ref{new.sec.A.3} gives that 
\begin{equation}\label{Jan27:01}
	\begin{aligned}
	&	\E \left\{\exp\left( \frac{\lambda_{k_1} \lambda_{k_2}\beta^2  }{2^{|l-m|/2}} \sum_{i \in \calS_{k_1} \cap \calT_{k_2}}  \sum_{j \in \calS_{k_1} \cap \calT_{k_2}, j \ne i} u_{k_1, i}  u_{k_1, j}  \right) \right\}\\
	& =\E \left\{\exp\left( \kappa^{l,m}_{k_1,k_2}  \bu^\top \left( \mathbf{1}_{k_{1},k_2} \mathbf{1}_{k_{1},k_2}^\top- \bI_{k_1,k_2} \right) \bu  \right) \right\}\\
	& \le \exp\left\{ \frac{8 (\kappa^{l,m}_{k_1,k_2})^2 \|\mathbf{1}_{k_1,k_2} \mathbf{1}_{k_1,k_2}^\top- \bI_{k_1,k_2}\|_F^2}{1 - 64 { (\kappa^{l,m}_{k_1,k_2})^2}\|\mathbf{1}_{k_1,k_2} \mathbf{1}_{k_1,k_2}^\top- \bI_{k_1,k_2}\|_2^2}\right\}  \\
	& \le \exp\left\{ 16 (\kappa^{l,m}_{k_1,k_2})^2 |\calS_{k_1} \cap \calT_{k_2}|^2 \right\} \\
    & \le e^{\tilde{c}_1 \lambda_1^4c_{1,\xi}^2}
	\le (1+\xi)^{1/(K^*-1)^2},
	\end{aligned}
\end{equation}
where the the second inequality above is due to \eqref{Jan26:01} and the definitions of $\mathbf{1}_{k_{1},k_2}$ and $\bI_{k_1,k_2}$, the third inequality above is because of $|\calS_{k_1} \cap \calT_{k_2}| \le s^*/(K^*-1)$, the second last inequality above is entailed by 
$$
 16 (\kappa^{l,m}_{k_1,k_2})^2 |\calS_{k_1} \cap \calT_{k_2}|^2  \le 16 \lambda_1^4 c^2_{1,\xi} (s^*)^{-2} \frac{(s^*)^{2}}{(K^*-1)^2} \le \tilde{c}_1 \lambda_1^4c_{1,\xi}^2,
$$
and the last inequality above holds for $c_{1,\xi}  = \sqrt{\log(1+\xi)/(\tilde{c}_1\lambda_1^4(K^*-1)^2)}$ and $\tilde{c}_1 \ge 16/(K^*-1)^2$. 

Thus, in view of \eqref{add:2.1} and \eqref{Jan27:01}, we see that when we choose 
$$
c_{1,\xi} = \sqrt{\log(1+\xi)/(\tilde{c}_1\lambda_1^4(K^*-1)^2)} \land (128\lambda_1^4)^{-1} (K^*-1),
$$
condition \eqref{sparse:samll} holds for the dense case of $s^* \ge \sqrt{n}$ with $\underline{c} = \underline{c}_1$ as defined in \eqref{equ:March5:cprime1}.

\medskip
    
	\textbf{Part II. Sparse alternative with $s^* < \sqrt{n}$ and $\log\left(\frac{c_{\xi}n}{(s^*)^2}\right) \ge 0$}. We now consider the sparse case of $s^* \le \sqrt{c_{\xi}n}$ and $\log\left(\frac{c_{\xi}n}{(s^*)^2}\right) \ge 0$ for a small constant $c_{\xi}\in (0,1)$, whose dependence on $\xi$ will be made clear later. In this regime, let us set $\beta^2 = \frac{1}{2} (s^*)^{-1}\log(\frac{c_{\xi}n}{(s^*)^2})$. 
	We will construct similar prior distribution $\nu$ as specified in the dense case from \textit{Part I}, with some slight modifications. To be specific, in the third step therein, we define $u_{k,i} := u_{\calS_k,i} = 1$ for all $i \in \calS_k$, and in the final step therein, we choose $\boldsymbol{\Lambda} = \diag\{1,\ldots, 1\} \in \R^{(K^*-1) \times (K^*-1)}$. Then it holds that 
     $$
     \begin{aligned}
    \frac{\tau^*(T-\tau^*)}{T}\|\bDelta\|_F^2&=\frac{2^l(T- 2^l)}{T} \frac{\beta^2 \rho_n}{2^l} \| \bU \bU^\top\|_F^2  \ge  \frac{(s^*)^2\beta^2 \rho_n}{2(K^*-1)} \\
    &\ge \frac{1}{4(K^*-1)} \epsilon_n^2.
    \end{aligned}
    $$
    We can also show that 
    $$
    \begin{aligned}
    \frac{\tau^*(T-\tau^*)}{T}\|\diag(\bDelta)\|_F^2 & \le \beta^2 \rho_n s^* = \frac{1}{2} \log(\frac{c_{\xi}n}{(s^*)^2}) \\
    &\le \frac{1}{16(K^*-1)} \epsilon_n^2
    \end{aligned}
    $$
    for $s^* \ge 8(K^*-1)$. Using similar arguments as for \eqref{Feb9:equ02} and Lemma \ref{lem:Feb9:01}, one can verify that $\supp(\nu) \subset \calA(\rho_n, K^*, s^*,\underline{c}_2 \epsilon_n)$ for 
    \begin{equation}\label{equ:june20:thm1:04}
           \underline{c}_2 = \sqrt{16(K^*-1)}. 
    \end{equation}
    
    Moreover, similar to \eqref{add:2.1}, we can deduce that 
	\begin{equation}\label{add:108:1}
		\begin{aligned}
			& \E_{(\gamma, \zeta) \sim \nu \otimes \nu } \left\{\exp\left(\sum_{t=1}^T \sum_{i < j} \frac{\gamma_{i,j}^{(t)} \zeta_{i,j}^{(t)}}{\rho_n(1-\rho_n)}  \right)\right\}\\
            \le & \prod_{1 \le k_1, k_2 \le K^*-1}\E \left\{\exp\left( \frac{\beta^2 \rho_n }{2^{1+|l-m|/2} \rho_n(1-\rho_n)} (\bu_{k_1}^\top \bv_{k_2})^2  \right) \right\}\\
			\le & \prod_{1 \le k_1, k_2 \le K^*-1}\E \left\{\exp\left( 2\beta^2  (\bu_{\calS_{k_1}}^\top \bv_{\calT_{k_2}})^2  \right) \right\} \\
			\le & \left( \E \left\{\exp\left( 2\beta^2  (|\calS \cap \calT|)^2  \right) \right\} \right)^{(K^*-1)^2}
		\end{aligned}
	\end{equation}
	for sufficient large $n$. Observe that $|\calS \cap \calT| \sim \text{hypergeometric distribution } \operatorname{Hyp}(n,s^*,s^*)$, and is no larger than the binomial distribution $Y:=\operatorname{Bin}(s^*,s^*/n)$ in the convex ordering sense (see, e.g., (24) in \cite{liu2021minimax}), that is, 
	\begin{equation}\label{eq:004}
		\begin{aligned}
			\E \left\{\exp\left( 2\beta^2  (|\calS \cap \calT|)^2  \right) \right\} \le& \E\left\{\exp\left(  Y^2 2\beta^2  \right) \right\}.
		\end{aligned} 
	\end{equation}
    
	To upper bound \eqref{eq:004} above, note that for $1 \le x \le s^*$, we have
	$$
	\begin{aligned}
		\log(e^{2\beta^2})  =\frac{1}{s^*} \log\left(\frac{c_{\xi} n}{(s^*)^2}\right) \le \frac{1}{x} \log\left(\frac{c_{\xi} n  }{(s^*)^2}\right).
	\end{aligned}
	$$
	Then it holds that 
    $$e^{2\beta^2} \le (\frac{c_{\xi} n}{(s^*)^2})^{1/x}.$$ 
    By setting $c_{\xi} \le 2 \log(1+\xi)/(K^*-1)^2$, we can obtain that 
	$$
	\begin{aligned}
		\E\left\{\exp\left(  Y^2 2\beta^2  \right) \right\} & = \sum_{x=0}^{s^*} (e^{2 \beta^2})^{x^2 } \binom{s^*}{x}  \left(\frac{s^*}{n}\right)^{x}\left(1-\frac{s^*}{n}\right)^{s^*-x}\\
		& \le\left(1-\frac{s^*}{n}\right)^{s^*}+\sum_{x=1}^{s^*}\binom{s^*}{x} \left(\frac{c_{\xi} n }{(s^*)^2}\right)^{x} \left(\frac{s^*}{n}\right)^{x}\left(1-\frac{s^*}{n}\right)^{s^*-x} \\
		& = \sum_{x=0}^{s^*}\binom{s^*}{x} \left(\frac{c_{\xi} }{s^*}\right)^{x}\left(1-\frac{s^*}{n}\right)^{s^*-x} \\
		& = \left(1-\frac{s^*}{n} + \frac{c_{\xi} }{s^*}\right)^{s^*} \le \left(1+ \frac{c_{\xi} }{s^*}\right)^{s^*} \\
        &\le e^{c_{\xi}} \le (1+\xi)^{1/(K^*-1)^2}.
	\end{aligned}
	$$
	Hence, this result along with \eqref{add:108:1} shows that when we choose $\underline{c}_2$ as in \eqref{equ:june20:thm1:04} and 
    \begin{equation}\label{equ:july10:c_xi2}
    c_{\xi} \le  \log(1+\xi)/(K^*-1)^2,    
    \end{equation}
     condition \eqref{sparse:samll} holds for the sparse case of $s^* \le \sqrt{c_{\xi}n}$ and $\log\left(\frac{c_{\xi}n}{(s^*)^2}\right) \ge 0$ too.
    
	
    
	\textbf{Part III. Sparse alternative with $s^* < \sqrt{n}$ and $\log\left(\frac{{  c_{\xi}}n}{(s^*)^2}\right) < 0$}. For the remaining sparse case of $s^* < \sqrt{n}$ and $\log\left(\frac{{ c_{\xi}}n}{(s^*)^2}\right) < 0$, the lower bound $\epsilon_n^2$ defined in \textit{Part II} becomes negative, which is useless.   For a given $s^*$, let us consider an arbitrary $\sqrt{n} > s^*$ to obtain a uniform (and relatively simple) lower bound for the sparse alternative. We will construct the prior distribution $\nu$ in the following way:  (1) sample an $l \sim \operatorname{Unif}\{\log_2(hT),$ $ \log_2(hT) + 1, \ldots, \log_2(T/2)\}$, and (2) let  $\bTheta_2 = (\rho_n)_{1 \le i,j \le n}$ and  $\bTheta_1 = \bTheta_2+\sqrt{\frac{c_{\xi}\rho_n}{2^{l} (s^*)^2}} \bU \bU^\top$ with change-point $\tau^* =  2^l$. Here, $\bU$ is defined the same as in \textit{Part II}. Letting $\epsilon^2_n = \rho_n/(K^*-1)$, the signal strength is given by 
	$$
    \begin{aligned}
	\frac{\tau^*(T-\tau^*)}{T}\|\bDelta \|_F^2&=\frac{2^{l}(T-2^{l})}{T} \frac{c_{\xi}\rho_n}{2^{l} (s^*)^2} \|\bU^\top \bU\|_F^2 \ge \frac{c_{\xi}\rho_n}{2(K^*-1)} \\
    &= \frac{c_{\xi}}{2} \epsilon_n^2.
    \end{aligned}
	$$
    It also holds that 
    $$
    \begin{aligned}
    \frac{\tau^*(T-\tau^*)}{T}\|\diag(\bDelta) \|_F^2&=\frac{2^{l}(T-2^{l})}{T} \frac{c_{\xi}\rho_n}{2^{l} (s^*)^2} s^* \le c_{\xi}\rho_n (s^*)^{-1} \\
    & \le  \frac{c_{\xi}}{8}\epsilon_n^2
    \end{aligned}
    $$
for $s^* \ge 8(K^*-1)$.  From \eqref{equ:june20:thm1:01}, we see that the signal strength parameter is given by 
$$\frac{\tau^*(T-\tau^*)}{T} \|\vech(\boldsymbol{\Delta})\|_2^2 \ge 16^{-1} c_{\xi} \epsilon_n^2 \ge \underline{c}_3^2 \rho_n $$
with
    \begin{equation}\label{equ:june20:thm1:06}
  \underline{c}_3=  \sqrt{ c_{\xi}/(16(K^*-1))}.  
    \end{equation}
    
Further, it holds that 
$$\theta_{i,j}^{(t)} \lesssim \rho_n + \sqrt{\rho_n (T(s^*)^2)^{-1}} \lesssim \rho_n$$
in light of the time span condition of $T \gtrsim \rho_n^{-1} (s^*)^{-2}$. Meanwhile, the eigenvalues also satisfy the eigenvalue conditions by Lemma \ref{lem:Feb9:01}. These results entail that $$\supp(\nu) \subset \calA(\rho_n, K^*, s^*, \underline{c}_3\epsilon_n)$$ 
for $s^* < \sqrt{n}$. Within this sampling process, the two sampling triplets are $(\calS, l, \bU)$ and $(\calT, m, \bV)$ with $\calT = \calS$ and $\bV = \bU$. It follows from \eqref{add:108:1} that 
     \begin{equation*}
		\begin{aligned}
			& \E_{(\gamma, \zeta) \sim \nu \otimes \nu } \left\{\exp\left(\sum_{t=1}^T \sum_{i < j} \frac{\gamma_{i,j}^{(t)} \zeta_{i,j}^{(t)}}{\rho_n(1-\rho_n)}  \right)\right\}\\
            \le & \prod_{1 \le k_1, k_2 \le K^*-1}\E \left\{\exp\left( \frac{c_{\xi} \rho_n }{2^{|l-m|/2+1} (s^*)^2 \rho_n(1-\rho_n)} (\bu_{k_1}^\top \bv_{k_2})^2  \right) \right\}\\
			\le & \prod_{1 \le k_1, k_2 \le K^*-1}\exp\left(\frac{c_{\xi}}{2^{|l-m|/2}(K^*-1)^2} \right)  \\
            \le& \prod_{1 \le k_1, k_2 \le K^*-1}\exp\left(\frac{c_{\xi}}{(K^*-1)^2} \right) \le e^{c_{\xi}} \le 1+\xi,
		\end{aligned}
	\end{equation*}
    where we have used the fact $\bu_{k_1}^\top \bv_{k_2} \le s^*/(K^*-1)$ for all $1 \le k_1, k_2 \le K^*-1$ in the second inequality above, and by noting
    \begin{equation*}
     c_{\xi} \le \log(1+\xi)/(K^*-1)^2 \le \log(1+\xi)    
    \end{equation*}
   as $K^* \ge 2$. This shows that condition \eqref{sparse:samll} holds for the sparse case of $s^* < \sqrt{n}$ and $\log\left(\frac{c_{\xi}n}{(s^*)^2}\right) < 0$ with rate $ \epsilon_n^2 \sim \rho_n$.
	
	Therefore, combining the above three parts (i.e., \textit{Parts I--III}), choosing 
    \begin{equation}\label{equ:thm1:final_c}
     \underline{c} =    \underline{c}_1\I\left( s^* \ge \sqrt{n}\right) + \underline{c}_2 \I\left( s^* <  \sqrt{c_{\xi} n}\right) + \underline{c}_3 \I\left(\sqrt{c_{\xi} n} < s^* \le \sqrt{n} \right) 
    \end{equation}
     with $\underline{c}_1$, $\underline{c}_2$, and $\underline{c}_3$ given in \eqref{equ:March5:cprime1}, \eqref{equ:june20:thm1:04}, and \eqref{equ:june20:thm1:06}, respectively, and letting $c_{\xi}$ be as given in \eqref{equ:july10:c_xi2} yield the desired conclusion. This completes the proof of Theorem \ref{low:minimax_lower}.

\subsection{Proof of Theorem \ref{low:minimax_upper}} \label{new.sec.A.2}

We aim to prove that the decision function $\psi_n$ achieves the desired size $\alpha$ under the null hypothesis $H_0(\rho_n, K^*)$, while rejecting the null hypothesis with probability no smaller than $1-\eta$ under the alternative hypothesis $H_1(\rho_n, K^*, s^*, \epsilon_n)$, where $\epsilon_n$ satisfies \eqref{equ:upper_sep}. We will first establish the size under null hypothesis $H_0(\rho_n, K^*)$. By resorting to Lemma \ref{lem:H0inequality} in Section \ref{new.sec.A.3}, we have 
$$|A_{\calS}^{(\tau)}| = O_p(s^*) \text{ and }  |A_{\Omega}^{(\tau)}| = O_p(n).$$ 
This implies that 
	$$
    \begin{aligned}
	\Pr_0\{\psi_n  =1\} & \le \sum_{\tau \in \calT} \Pr_0\{ |A_{\calS}^{(\tau)}|/ s^* > r_n/(2|\calT| s^*) \} +  \sum_{\tau \in \calT} \Pr_0\{| A_{\Omega}^{(\tau)}|/n > r_n/(2|\calT|s^*) \} \\
    &\le 2|\calT| o(1) = o(1) \le \alpha
    \end{aligned}
	$$
	since the cardinality $|\calT|$ is fixed. As a result, the decision function $\psi_n$ achieves the desired size $\alpha$ under null hypothesis $H_0(\rho_n, K^*)$.
    
    We next establish the power of the testing procedure under the alternative hypothesis $H_1(\rho_n, K^*, s^*, \epsilon_n)$. By the construction of $\calT$, there exists some $\tau \in \calT$ such that   $\tau \le \min( \tau^*, T-\tau^*+1) \le 2\tau$ and $\tau \le T/2$, where $\tau^*$ is the true change point. For $s^* \ll n$, an application of Lemma \ref{lem:H1inequality} in Section \ref{new.sec.A.3} leads to 
	\begin{equation}\label{add:equ:h1a}
		\Pr_1\left\{  A_{\calS}^{(\tau) }  > r_n \right\} \ge 1-o(1).
	\end{equation}
It follows directly from (\ref{add:equ:h1a}) that 
    $$
    \Pr_1\{\psi_n  =1\}  \ge \Pr_1\{|A_{\calS}^{(\tau)}|/s^* > r_n/s^*\}  \ge \Pr_1\left\{  A_{\calS}^{(\tau)} > r_n \right\} \ge 1-\eta.
    $$ 
    For $s^* \asymp n$ and $r_n =\log(en) n$, based on Lemma \ref{lem:H1inequality}, we have
     $$
    \Pr_1\{\psi_n  =1\} \ge \Pr_1\{|A_{\Omega}^{(\tau)}|/n > r_n/n\} \ge \Pr_1\left\{  A_{\Omega}^{(\tau) } > r_n \right\} \ge 1-\eta.
    $$ 
   This shows that the decision function $\psi_n$ rejects the null hypothesis with probability at least $1-\eta$ under the alternative hypothesis $H_1(\rho_n, K^*, s^*, \epsilon_n)$, which concludes the proof of Theorem \ref{low:minimax_upper}.

\subsection{Lemmas \ref{lemma:chi_dis}--\ref{lem:H1inequality} and their proofs} \label{new.sec.A.3}

Since graph $\mathcal{G}_t = (\calV, \mathcal{E}_t)$ is undirected without self-loops, the corresponding adjacency matrix is symmetric, and thus we consider only the upper triangular entries of each adjacency matrix when we define the distribution of the network at each time point $t$.

\begin{lemma} \label{lemma:chi_dis}
Consider dynamic networks $\{\bX^{(t)}\}_{t\in [T]}$ 
with independent adjacency matrices $\bX^{(t)}$ across time $t$. Denote by $\Phi_0$ the joint distribution of upper triangular entries of $\{\bX^{(t)}\}_{t\in [T]}$ with each mean matrix $\bTheta^{(t)} = (\theta_{i,j}^{(t)})_{1 \le i,j \le n}$ satisfying that $\theta^{(t)}_{i,j} = \rho$ with $\rho\in (0,1)$ for all $t \in [T]$, and 
$\phi_0$ the density function of $\Phi_0$. Let $\nu$ be the joint distribution of independent random variables $\{\gamma_{i,j}^{(t)}: t \in [T], (i,j) \in \Omega\}$. Conditional on sequence $\{ \gamma_{i,j}^{(t)}\}_{(t,i,j)}$, define $\phi_{\gamma}$ as the joint density function of upper triangular entries of dynamic networks $\{\bX^{(t)}\}_{t\in [T]}$, where each mean matrix $\bTheta^{(t)}$ takes the form $\theta^{(t)}_{i,j} = \theta^{(t)}_{j,i} = \rho + \gamma_{i,j}^{(t)}$ with $|\theta^{(t)}_{i,j}| \le 1$ for $i \le j$. Denote by $\phi_1 = \E_{\gamma \sim \nu} \phi_{\gamma}(x)$ the marginal (joint) density of upper triangular entries of dynamic networks $\{\bX^{(t)}\}_{t\in [T]}$, and $\Phi_1$ the corresponding distribution. Then it holds that 
	$$
	\chi^2(\phi_0, \phi_1):= \E_{x \sim \Phi_0}({\phi_1^2(x)}/{\phi_0^2(x)}) -1 \le \E_{(\gamma, \zeta) \sim \nu \otimes \nu} \left\{  \exp\left(\sum_{t=1}^T \sum_{i < j}  \frac{\gamma_{i,j}^{(t)} \zeta_{ij}^{(t)}}{\rho(1-\rho)} \right) \right\} -1,
	$$
	where $\{\zeta_{i,j}^{(t)}\}_{(t,i,j)}$ is an independent copy of $\{\gamma_{i,j}^{(t)}\}_{(t,i,j)}$.
\end{lemma}

\noindent\textit{Proof}. By definition in (\ref{new.eq.FL001}), the chi-square divergence is given by 
$$\chi^2(\phi_0, \phi_1) = \E_{x \sim \Phi_0}({\phi_1^2(x)}/{\phi_0^2(x)}) -1.$$
 Recall that $\{\zeta_{i,j}^{(t)}\}_{t\in [T], 1 \le i < j\le n}$ and $\{\gamma_{i,j}^{(t)}\}_{t\in [T], 1 \le i< j\le n}$ are independent copies across $t$, $i$, and $j$, and that $\phi_1(x) = \int \phi_{\gamma}(x)\nu(d\gamma) = \mathbb E_{\gamma\sim \nu}(\phi_{\gamma}(x))$. Then an application of Fubini's Theorem leads to 
	$$
	\begin{aligned}
		\E_{x \sim \Phi_0}\left(\frac{\phi_1^2(x)}{\phi^2_0(x)}\right) =& \E_{x \sim \Phi_0} \left\{\frac{(\E_{\gamma \sim \nu} \phi_{\gamma}(x))^2}{\phi^2_0(x)} \right\}\\
		=&\E_{x \sim \Phi_0} \left\{\frac{\left(\E_{\gamma \sim \nu} \phi_{\gamma}(x)\right) \left(\E_{\zeta \sim \nu} \phi_{\zeta}(x)\right)}{\phi^2_0(x)} \right\} \\
		=&\E_{(\gamma, \zeta) \sim \nu \otimes \nu } \left\{ \E_{x \sim \Phi_0} \left( \frac{\phi_{\gamma}(x) \phi_{\zeta}(x)}{\phi_0^2(x)} \right) \right\}.
	\end{aligned}
	$$
	
    We next investigate the inner expectation in the last step above. Since all of the upper triangular  entries of network $\bX^{(t)}$ follow the Bernoulli distributions, it holds that 
    $$\phi_0(x) = \prod_{t=1}^T \prod_{i < j}\rho^{x^{(t)}_{i,j}}(1-\rho)^{1-x^{(t)}_{i,j}}$$ and $$\phi_{\gamma}(x) = \prod_{t=1}^T \prod_{i < j}(\rho+\gamma^{(t)}_{i,j})^{x^{(t)}_{i,j}}(1-\rho-\gamma_{i,j}^{(t)})^{1-x^{(t)}_{i,j}}.$$ 
	Then we can deduce that 
	$$
 \begin{aligned}
		&\E_{x \sim \Phi_0} \left( \frac{\phi_{\gamma}(x) \phi_{\zeta}(x)}{\phi_0^2(x)} \right)\\
  =& \E_{x \sim \Phi_0} \left(\prod_{t=1}^T \prod_{i < j}\left( \frac{(\gamma_{i,j}^{(t)} + \rho)^{x^{(t)}_{i,j}} (1 - \gamma_{i,j}^{(t)} - \rho)^{1 - x_{i,j}^{(t)}}}{\rho^{x_{i,j}^{(t)}}(1-\rho)^{1-x_{i,j}^{(t)}}} \cdot \frac{(\zeta_{i,j}^{(t)} + \rho)^{x_{i,j}^{(t)}} (1 - \zeta_{i,j}^{(t)} - \rho)^{1 - x_{i,j}^{(t)}}}{\rho^{x_{i,j}^{(t)}}(1-\rho)^{1-x_{i,j}^{(t)}}}\right)\right) \\
		=& \prod_{t=1}^T \prod_{i < j} \left( \frac{(\gamma_{i,j}^{(t)} + \rho)(\zeta_{i,j}^{(t)} + \rho)}{\rho} +\frac{(1 - \gamma_{i,j}^{(t)} - \rho)(1 - \zeta_{i,j}^{(t)} - \rho)}{1 - \rho} \right) \\
		=& \prod_{t=1}^T \prod_{i< j} \left(1 + \frac{\gamma_{i,j}^{(t)} \zeta_{i,j}^{(t)}}{\rho(1-\rho)} \right) \le \prod_{t=1}^T \prod_{i < j} \exp\left(  \frac{\gamma_{i,j}^{(t)} \zeta_{i,j}^{(t)}}{\rho(1-\rho)}  \right)  \\
		=& \exp\left( \sum_{t=1}^T \sum_{i< j} \frac{\gamma_{i,j}^{(t)} \zeta_{i,j}^{(t)}}{\rho(1-\rho)}  \right). 
	\end{aligned}
		$$ 
	
    Therefore, combining the above results yields that 
	$$
	\begin{aligned}
		\chi^2(\phi_0, \phi_1) \leq \E_{(\gamma, \zeta) \sim \nu \otimes \nu } \left\{\exp\left(\sum_{t=1}^T \sum_{i < j} \frac{\gamma_{i,j}^{(t)} \zeta_{i,j}^{(t)}}{\rho(1-\rho)}  \right)\right\}-1,
	\end{aligned}
	$$
	which proves the desired result. This completes the proof of Lemma \ref{lemma:chi_dis}.  

The lemma below establishes the validity of the construction process introduced in the proof of Theorem \ref{low:minimax_lower} in Section \ref{proof:lower}.

\begin{lemma}\label{lem:Feb9:01}
    The realization $\bTheta^0$ generated from the sampling process defined in \textit{Parts I--III} of the proof of Theorem \ref{low:minimax_lower} 
    in Section \ref{proof:lower} satisfies Condition \ref{con:eigenvalue} and eigenvector condition in \eqref{low:single_alter_indi} with asymptotic probability one. Thus, we have $\bTheta^0 \in \calA^{(\tau^*)}(\rho_n, K^*, s^*, \underline{c}\epsilon)$ for $\tau^* = 2^l$ with asymptotic probability one.
\end{lemma}

\noindent\textit{Proof}. Note that there are two different elements in $\bTheta^0$ from the sampling process defined in \textit{Parts I--III} of the proof of Theorem \ref{low:minimax_lower} in Section \ref{proof:lower}, denoted as $\bTheta_1$ and $\bTheta_2$. It is easy to see that $\bTheta_2 = (\rho_n)_{1\le i, j \le n} \in \R^{n \times n}$ satisfies Condition \ref{con:eigenvalue} and eigenvector condition in \eqref{low:single_alter_indi} over all three scenarios in \textit{Parts I--III}. It remains to consider $\bTheta_1$ from the sampling process over those three scenarios.

For the sampling process in \textit{Part I}, let us define 
$$\tilde{\bv}_1 := \sqrt{\rho_n} \mathbf{1}_n \ \text{ and } \ {\tilde{\bv}_{k+1} :=  (\lambda_{k}\rho^{1/2}_n \frac{\beta}{\sqrt{2^l}} )^{1/2} \bu_k},$$ where $\bu_{k}$ with $ k \in [K^*-1]$ is as given in Step 3 in \textit{Part I}, and $\lambda_{k}$ with $ k \in [K^*-1]$ is as given in Step 4 in \textit{Part I}. Some standard calculations lead to 
    \begin{equation}\label{Feb9:equ1}
      \bTheta_1 =  \sum_{k=1}^{K^*} \| \tilde{\bv}_k\|_2^2 \frac{\tilde{\bv}_k}{\| \tilde{\bv}_k\|_2}\left(\frac{\tilde{\bv}_k}{\| \tilde{\bv}_k\|_2}\right)^\top:= \sum_{k=1}^{K^*} d_k \bv_k \bv_k^\top,  
    \end{equation}
    where $d_k := \| \tilde{\bv}_k\|_2^2$ and $\bv_k := \tilde{\bv}_k/\| \tilde{\bv}_k\|_2$ with $k =1,\ldots, K^*$. Indeed, the decomposition in (\ref{Feb9:equ1}) is the asymptotic eigen-decomposition of $\bTheta_1$. To see this, we need only to show the asymptotic orthogonality of pairs $(\bv_1,\bv_k)$ for all $2 \le k \le K^*$ since other pairs are orthogonal by definition. Observe that for each $\epsilon > 0$,
    \begin{equation*}
        \begin{aligned}
            \Pr\left\{\max_{2 \le k \le K^*}|\bv_1^\top \bv_k| \ge \epsilon\right\} & = \sum_{k=2}^{K^*}\Pr\left\{  \frac{\sum_{i \in \calS_k} u_{k,i}}{ \sqrt{s^*/(K^*-1)}} \ge \sqrt{n}\epsilon\right\}\\
            &\le (K^*-1)\exp\left(- \epsilon^2 n /2\right) \to 0,
        \end{aligned}
    \end{equation*}
  where the last inequality above is due to the tail probability bound for the self-normalized independent Rademacher random variables (see, e.g., Theorem 2.14 in \cite{pena2009self}). Hence, the eigenvalues of $\bTheta_1$ satisfy that $d_1 = \rho_nn $ and %
  $$
  \begin{aligned}
  d_k & = \lambda_{k-1} \rho^{1/2}_n \beta/\sqrt{2^l} \| \bu_k\|_2^2 \\
  & = (K^*-1)^{-1}c^{1/2}_{1,\xi}\lambda_{k-1} \sqrt{\frac{\rho_n s^*}{2^l}} \\
  &\asymp \sqrt{\frac{\rho_n s^*}{T}}
  \end{aligned}
  $$
  for $k=2,\ldots, K^*$. 
 Then it holds that 
 \begin{equation}\label{equ:june18:lem2:01}
    \frac{d_{1}}{d_{2}} \gtrsim q_n\sqrt{T}  \ge 1+c_0  \ \text{ and } \ \frac{d_{k}}{d_{k+1}} = \frac{\lambda_{k}}{\lambda_{k+1}} \ge 1+c_0
 \end{equation}
 for $k \ge 2$. 
 On the other hand, the maximum absolute values of components in each eigenvector are bounded by $O(1/\sqrt{s^*})$. These facts show that the sampling process in \textit{Part I} satisfies Condition \ref{con:eigenvalue} and eigenvector condition in \eqref{low:single_alter_indi}.

    We next turn to the sampling process in \textit{Part II}. It is easy to see that $\bTheta_1$ adapts a similar eigen-decomposition as in \eqref{Feb9:equ1}, except that $\bu_k$'s with $k \in [K^*-1]$ are nonrandom vectors and $\lambda_k \equiv 1$. Further, it follows from $s^* \le \sqrt{n} \ll n$ in \textit{Part II} that for all $k >1$,
    \begin{equation}\label{equ:june19:lem2:01}
        \bv_1^\top \bv_k = \frac{ \sum_{i \in \calS_k} u_{k,i}}{ \sqrt{ n \sum_{i \in \calS_k} u^2_{k,i}}}  = \frac{s^*}{(K^*-1)^{1/2}\sqrt{ns^*}} \to 0.
    \end{equation}
    Similar to \eqref{equ:june18:lem2:01}, we can show that the eigenvalues of $\bTheta_1$ satisfy that 
    $$
      \frac{d_{1}}{d_{2}} \gtrsim q_n\sqrt{T}  \ge 1+c_0  \ \text{ and } \ \frac{d_{k}}{d_{k+1}} = 1
    $$
    for $k \ge 2$, thereby satisfying Condition \ref{con:eigenvalue}. We can directly see that the eigenvectors also satisfy the eigenvector condition in \eqref{low:single_alter_indi}. 
    
    The proof for $\bTheta_1$ from the sampling process in \textit{Part III} is nearly identical to that for the scenario of \textit{Part II}, with the key difference being the definition of  $\tilde{\bv}_k, k \ge 2$. Specifically, we set
$$\tilde{\bv}_k = \left( \frac{c_{\xi} \rho_n}{2^l (s^*)^2} \right)^{1/4} \bu_k.$$ 
We can establish the asymptotic orthogonality between $\bv_1$ and $\bv_k$ in the same manner as in \eqref{equ:june19:lem2:01}. Moreover, it holds that 
\[
\frac{d_1}{d_2} = \frac{n \rho_n}{\sqrt{\frac{c_{\xi} \rho_n}{2^l (s^*)^2}}s^*} \gtrsim \sqrt{Tn} q_n \geq 1 + c_0  \ \text{ and } \ \frac{d_{k}}{d_{k+1}} = 1.
\]
This concludes the proof of Lemma \ref{lem:Feb9:01}.

The lemma below provides a decoupling expectation inequality for the Rademacher random vector.

\begin{lemma}\label{lem: rademacher}
    Let $\bu = (u_1,\ldots, u_n)^\top \in \R^n$ be a  Rademacher random vector with $\Pr\{u_i = \pm 1\} = 1/2$, and $\bA$ a symmetric matrix with zero diagonal entries. Then it holds that for $ 0 < \kappa < (8\| \bA\|_2)^{-1}$,
    $$
    \E\left\{\exp\left(  \kappa \bu^\top \bA \bu \right)\right\} \le \exp\left\{ \frac{8 \kappa^2 \|\bA \|_F^2}{1 - 64\kappa^2 \| \bA\|_2^2}\right\}.
    $$
\end{lemma}

\noindent\textit{Proof}. The proof of this lemma can be found in the intermediate steps of the proof of Theorem 8.13 in \cite{foucart2013invitation}. More precisely, see (8.22) and below on Page 216 therein.

The lemma below gives a probability upper bound for $A^{(\tau)}$ under $H_0(\rho_n, K^*)$. 
Let us define two functions $a^2(s) = \log(en/s)$ and $d^2(s) \in [ a^2(s)/n, a^2(s)/s]$.

\begin{lemma}
	\label{lem:H0inequality}
	Under the null hypothesis $H_{0}(\rho_n, K^*)$, and assuming Condition \ref{con:upper} holds, we have 
    $$
    \begin{aligned}
    &\max_{\tau \in \calT}|A_{\calS}^{(\tau)}| = O_p(s^*) \text{ for $s^* \ll n$}, \\
    &\max_{\tau \in \calT}|A_{\calS}^{(\tau)}| = O_p(n) \text{ for $s^* \asymp n$}, \\
    &\max_{\tau \in \calT}|A_{\Omega}^{(\tau)}| = O_p(n),    
    \end{aligned}
    $$
    where $\calS$ is defined in \eqref{new.eq.FL023}, which is a function of $d(s^*)$, and hence $s^*$.
\end{lemma}

\noindent\textit{Proof}. To prove this lemma, it suffices to establish the result for each individual $\tau$. An application of the union bound leads to the conclusion of the lemma uniformly over $\tau \in \calT$ since the cardinality of $\calT$ is of  
constant order. For each given $\tau$, it is sufficient to prove the statement under the assumption that event $F_0$, as defined in \eqref{event:F}, holds. With a slight abuse of notation, we define in this proof $F_0$ as the event such that the asymptotic expansions in \eqref{event:F} hold simultaneously for $z_{i,j}^{(\tau)}$ and $\dot{z}_{i,j}^{(\tau)}$. This is justified because Condition \ref{con:upper} ensures that the mean matrix of the dynamic network satisfies Condition \ref{con:expansion}. Consequently, from Lemma \ref{lem:F} in Section \ref{new.sec.D}, we see that event $F_0$ holds with asymptotic probability one. Then for any $x > 0$, we have that 
 $$
 \begin{aligned}
   \Pr_0\left\{ A_{\calS}^{(\tau)} \le x\right\} =& \Pr_0\left\{ A_{\calS}^{(\tau)} \le x, \I(F_0)=1\right\} + \Pr_0\left\{ A_{\calS}^{(\tau)}  \le x, \I(F_0^c)=1\right\}   \\
   \le & \Pr_0\left\{ A_{\calS}^{(\tau)} \I(F_0) \le x, \I(F_0)\right\}  + \Pr_0\{F_0^c\}\\
   =& \Pr_0\left\{ A_{\calS}^{(\tau)} \I(F_0) \le x \right\}  + o(1).
 \end{aligned}
 $$
Hence, the proof reduces to verifying the statements that for each $\tau$, 
$$
    \begin{aligned}
    &\left|A_{\calS}^{(\tau)}\I(F_0) \right|= O_p(s^*) \ \text{ for $s^* \ll n$}, \\
    &\left|A_{\calS}^{(\tau)}\I(F_0)\right| = O_p(n) \ \text{ for $s^* \asymp n$}, \\
    &\left|A_{\Omega}^{(\tau)} \I(F_0) \right|= O_p(n). 
    \end{aligned}
    $$
We will separately consider three sparsity scenarios of $s^*\ll n$, $s^* \asymp n$, and $\calS = \Omega$.

Let us first consider the scenario of $s^* \ll n$.
Note that \eqref{eq:epsilon_condition} implies $Ts^* \rho_n \geq \log(e n) \gg \log(e n/s^*)$ when $s^* \ll n$, which satisfies the time span condition in Lemma \ref{lem:gen2}. We can decompose statistic $A_{\calS}^{(\tau)}{ \I(F_0)}$ as 
\begin{equation}\label{Feb9:equ:03}
		\begin{aligned}
			& A_{\calS}^{(\tau)}{ \I(F_0)} 
			=\sum_{(i,j) \in \Omega} z^{(\tau)} _{i,j} \dot{z}^{(\tau)} _{i,j} \I\left(|\ddot{z}^{(\tau)}_{i,j}| > d(s^*)\right){ \I(F_0)}	\\
			&=   \sum_{(i,j) \in \Omega} \frac{n}{2}\left\{  \left(\be_i^\top \tilde{\bW}^{(\tau)} \bb_j + \be_j^\top \tilde{\bW}^{(\tau)} \bb_i \right) \left(\be_i^\top \tilde{\dot{\bW}}^{(\tau)} \bb_j + \be_j^\top \tilde{\dot{\bW}}^{(\tau)} \bb_i \right)\right\}\I\left(|\ddot{z}_{i,j}^{(\tau)}| > d(s^*)\right){ \I(F_0)}\\
			&\quad+ O \left(\sum_{(i,j) \in \Omega}   \left(\be_i^\top \tilde{\bW}^{(\tau)} \bb_j+ \be_j^\top \tilde{\bW}^{(\tau)} \bb_i \right) \I\left(|\ddot{z}_{i,j}^{(\tau)}| > d(s^*)\right) { \I(F_0)}\right)\\
			&\quad+ O\left(\sum_{(i,j) \in \Omega} \left (\be_i^\top \tilde{\dot{\bW}}^{(\tau)} \bb_j + \be_j^\top \tilde{\dot{\bW}}^{(\tau)} \bb_i\right) \I\left(|\ddot{z}_{i,j}^{(\tau)}| > d(s^*)\right){ \I(F_0)}\right)\\
            &\quad+  { O}\left(\sum_{(i,j) \in \Omega} \frac{1}{n}\I\left(|\ddot{z}_{i,j}^{(\tau)}| > d(s^*)\right){ \I(F_0)}\right)\\
			&:= e_1 + e_2 + e_3 + e_4. 
		\end{aligned}
	\end{equation}
We will bound each summand on the right-hand side of (\ref{Feb9:equ:03}) above separately.

For the first term $e_1$, it follows from part  4) of Lemma \ref{lem:gen2} in Section \ref{new.sec.D} and the Markov inequality that 
	$$
	\sum_{(i,j) \in \Omega} \be_i^\top \tilde{\bW}^{(\tau)} \bb_j  \be_i^\top \tilde{\dot{\bW}}^{(\tau)} \bb_j  \I\left(|\ddot{z}_{i,j}^{(\tau)}| > d(s^*)\right){ \I(F_0)} = O_p\left(\frac{\sqrt{s^*}}{n}\right).
	$$
	Similarly, it holds that 
	$$\sum_{(i,j) \in \Omega} \be_i^\top \tilde{\bW}^{(\tau)} \bb_j  \be_j^\top \tilde{\dot{\bW}}^{(\tau)} \bb_i  \I\left(|\ddot{z}_{i,j}^{(\tau)}| > d(s^*)\right){ \I(F_0)} = O_p\left(\frac{\sqrt{s^*}}{n}\right),$$ 
    $$\sum_{(i,j) \in \Omega} \be_j^\top \tilde{\bW}^{(\tau)} \bb_i  \be_i^\top \tilde{\dot{\bW}}^{(\tau)} \bb_j  \I\left(|\ddot{z}_{i,j}^{(\tau)}| > d(s^*)\right){ \I(F_0)} = O_p\left(\frac{\sqrt{s^*}}{n}\right),$$ and 
	$$\sum_{(i,j) \in \Omega} \be_j^\top \tilde{\bW}^{(\tau)} \bb_i  \be_j^\top \tilde{\dot{\bW}}^{(\tau)} \bb_i  \I\left(|\ddot{z}_{i,j}^{(\tau)}| > d(s^*)\right){ \I(F_0)} = O_p\left(\frac{\sqrt{s^*}}{n}\right).$$ 
	Thus, we have that 
	\begin{equation}\label{Oct19:4}
			|e_1| = n O_p\left(\frac{\sqrt{s^*}}{n}\right) = O_p(\sqrt{s^*}) = O_p(s^*). 
	\end{equation}

    	We now turn to terms $e_2$ and $e_3$. With the aid of part 3) of Lemma \ref{lem:gen2}, we can show that
	$$
	\Var\left(\sum_{(i,j) \in \Omega} \be_i^\top \tilde{\bW}^{(\tau)} \bb_j  \I\left(|\ddot{z}_{i,j}^{(\tau)}| > d(s^*)\right){ \I(F_0)} \right) = O\left(\frac{s^*}{n^2} \cdot n^2\right) = O(s^*).
	$$
    An application of the Markov inequality gives that 
	$$
    \sum_{(i,j) \in \Omega} \be_i^\top \tilde{\bW}^{(\tau)} \bb_j \I\left(|\ddot{z}_{i,j}^{(\tau)}| > d(s^*)\right){ \I(F_0)}  = O_p(\sqrt{s^*}).
    $$ Similarly, we can also show that
	 $$\sum_{(i,j) \in \Omega} \be_j^\top \tilde{\bW}^{(\tau)} \bb_i\I\left(|\ddot{z}_{i,j}^{(\tau)}| > d(s^*)\right){ \I(F_0)}   = O_p(\sqrt{s^*}),$$
     $$\sum_{(i,j) \in \Omega} \be_i^\top \tilde{\dot{\bW}}^{(\tau)} \bb_j \I\left(|\ddot{z}_{i,j}^{(\tau)}| > d(s^*)\right){ \I(F_0)}  = O_p( \sqrt{s^*}),$$ 
     and 
     $$\sum_{(i,j) \in \Omega} \be_j^\top \tilde{\dot{\bW}}^{(\tau)} \bb_i \I\left(|\ddot{z}_{i,j}^{(\tau)}| > d(s^*)\right){ \I(F_0)}  = O_p( \sqrt{s^*}).$$ 
     Hence, it holds that 
	 \begin{equation}\label{Oct21:3}
	 	|e_2| = |e_3|  =  { O}_p(\sqrt{s^*}) = { O}_p(s^*).
	 \end{equation}

	 For the last term $e_4$, using part 1) of Lemma \ref{lem:gen2}, we can deduce that 
	$$
    \begin{aligned}
	& \E\left\{\sum_{(i,j) \in \Omega} \I(|\ddot{z}_{i,j}^{(\tau)}| > d(s^*)) { \I(F_0)}\right\} \\
    & = \sum_{(i,j) \in \Omega} \Pr\left\{ |\ddot{z}_{i,j}^{(\tau)}| > d(s^*), { F_0} \right\} \\
    &\lesssim n^2 \frac{s^*}{n} = n s^*.
    \end{aligned}
	$$
	On the other hand, its second moment can be bounded as 
	$$
	\begin{aligned}
	&\E\left( \sum_{(i,j) \in \Omega} \I(|\ddot{z}_{i,j}^{(\tau)}| > d(s^*)){ \I(F_0)} \right)^2\\
    \le& \sum_{(i_1, j_1) \in \Omega}\sum_{(i_1, j_1) \in \Omega, i_1 = i_2} \E\left( \I(|\ddot{z}_{i_1,j_1}^{(\tau)}| > d(s^*))  \I(|\ddot{z}_{i_2,j_2}^{(\tau)}| > d(s^*)){ \I(F_0)} \right)  \\
	&+\sum_{(i_1, j_1) \in \Omega}\sum_{(i_1, j_1) \in \Omega, j_1 = j_2} \E\left( \I(|\ddot{z}_{i_1,j_1}^{(\tau)}| > d(s^*))  \I(|\ddot{z}_{i_2,j_2}^{(\tau)}| > d(s^*)){ \I(F_0)} \right)  \\
	&+\sum_{(i_1, j_1) \in \Omega}\sum_{(i_2, j_2) \in \Omega; i_1 \ne j_1; j_1 \ne j_2} \E\left( \I(|\ddot{z}_{i_1,j_1}^{(\tau)}| > d(s^*))  \I(|\ddot{z}_{i_2,j_2}^{(\tau)}| > d(s^*)){ \I(F_0)} \right)\\
     \le& \sum_{(i_1, j_1) \in \Omega}\sum_{(i_1, j_1) \in \Omega, i_1 = i_2} \Pr\left\{ |\ddot{z}_{i_1,j_1}^{(\tau)}| > d(s^*), { F_0}\right\} 
	+\sum_{(i_1, j_1) \in \Omega}\sum_{(i_1, j_1) \in \Omega, j_1 = j_2} \Pr\left( |\ddot{z}_{i_1,j_1}^{(\tau)}| > d(s^*), { F_0} \right)  \\
	&+\sum_{(i_1, j_1) \in \Omega}\sum_{(i_2, j_2) \in \Omega; i_1 \ne j_1; j_1 \ne j_2} \Pr\left( |\ddot{z}_{i_1,j_1}^{(\tau)}| > d(s^*), |\ddot{z}_{i_2,j_2}^{(\tau)}| > d(s^*),  { F_0} \right)\\
	\lesssim&  n^3 \frac{s^*}{n} +  n^4 \frac{(s^*)^2}{n^2} = O\left((n s^*)^2\right),
	\end{aligned}
	$$
	where the last inequality above has invoked parts 1) and 2) of Lemma \ref{lem:gen2}. Thus, we have that 
	$$\sum_{(i,j) \in \Omega} \I(|\ddot{z}_{i,j}^{(\tau)}| > d(s^*)) { \I(F_0)} = O_p(n s^*),$$ which entails that 
	\begin{equation} \label{Oct21:4}
		|e_4| = \frac{1}{n}  {O}_p(n s^*) =  { O}_p(s^*).
	\end{equation}
    
Therefore, combining the bounds for $|e_1|,|e_2|, |e_3|$, and $|e_4|$ in \eqref{Oct19:4}--\eqref{Oct21:4} above yields that 
$$
\left|A^{(\tau)}{ \I(F_0)}\right| = O_p(s^*).
$$
This completes the proof of the sparse case of Lemma \ref{lem:H0inequality}.

For the scenario of $s \asymp n$, $A^{(\tau)}_{\calS}{ \I(F_0)}$ admits the same decomposition as in \eqref{Feb9:equ:03}. The only difference here is that we use a conservative probability bound 
$\Pr\left\{|\ddot{z}_{i,j}^{(\tau)}| > d(s^*), { F_0} \right\} \leq 1$.
Under this regime, based on part 4) of Lemma \ref{lem:gen2} and following the same calculations as \eqref{Oct19:4}, we have that 
$$|e_1| = n O_p(1/\sqrt{n}) = O_p(\sqrt{n}) = O_p(n).$$ 
Following from part 3) of Lemma \ref{lem:gen2} and the same calculations as \eqref{Oct21:3}, it can be shown that 
$$|e_2| = |e_3| = O_p(\sqrt{n}) = O_p(n).$$ 
Moreover, for $e_4$, it holds that 
$$\sum_{(i,j) \in \Omega} \I(|\ddot{z}_{i,j}^{(\tau)}| > d(s^*)){ \I(F_0)} \leq n^2,$$ 
and thus 
$$|e_4| \le n^{-1} n^2 = O(n),$$
which is a trivial upper bound. Plugging these facts into \eqref{Feb9:equ:03}, we can obtain that
$$\left|A_{\calS}^{(\tau)} { \I(F_0)} \right| = O_p(n)$$ 
for $s^* \asymp n$.

The proof for the scenario of $\calS = \Omega$ is nearly identical to the one above. We can decompose statistic $A_{\calS}^{(\tau)}{ \I(F_0)}$ as 
\begin{equation}\label{July:07:lem4:01}
		\begin{aligned}
			 A_{\Omega}^{(\tau)}{ \I(F_0)}
			=&\sum_{(i,j) \in \Omega} z^{(\tau)} _{i,j} \dot{z}^{(\tau)} _{i,j}{ \I(F_0)}	\\
			&=   \sum_{(i,j) \in \Omega} \frac{n}{2}\left\{  \left(\be_i^\top \tilde{\bW}^{(\tau)} \bb_j + \be_j^\top \tilde{\bW}^{(\tau)} \bb_i \right) \left(\be_i^\top \tilde{\dot{\bW}}^{(\tau)} \bb_j + \be_j^\top \tilde{\dot{\bW}}^{(\tau)} \bb_i \right)\right\}{ \I(F_0)}\\
			&\quad+ O\left(1\right)\sum_{(i,j) \in \Omega}   \left(\be_i^\top \tilde{\bW}^{(\tau)} \bb_j+ \be_j^\top \tilde{\bW}^{(\tau)} \bb_i \right){ \I(F_0)}\\
			&\quad+O\left(1\right)\sum_{(i,j) \in \Omega} \left (\be_i^\top \tilde{\dot{\bW}}^{(\tau)} \bb_j + \be_j^\top \tilde{\dot{\bW}}^{(\tau)} \bb_i\right){ \I(F_0)}
            + O\left(n\right)\\
			&:= r_1 + r_2 + r_3 + O\left(n\right). 
		\end{aligned}
	\end{equation}
Similar to \eqref{Oct19:4}--\eqref{Oct21:3} in the sparse scenario and with the aid of Lemma \ref{lem:gen2} for $\calS = \Omega$, we can bound each summand on the right-hand side of (\ref{July:07:lem4:01}) above as $r_l = O_p(n), \, l\in[3]$. This completes the proof for the dense case of Lemma \ref{lem:H0inequality}.

We next aim to show that $A_{\calS}^{(\tau)}$ is far larger than threshold $r_n = \log(en) s^*$ under $H_1(\rho_n, K^*, s^*, \epsilon_n)$.  

\begin{lemma}
	\label{lem:H1inequality}
	Under Condition \ref{con:upper} and the alternative hypothesis $H_{1}(\rho_n, K^*, s^*, \epsilon_n)$ with $\epsilon_n$ satisfying \eqref{equ:upper_sep}, we have
    $$
    \begin{aligned}
        &\Pr_1\left\{A_{\calS}^{(\tau)} > r_n\right\} \ge 1-o(1) \ \text{ for } s^* \ll n,\\
        &\Pr_1\left\{A_{\Omega}^{(\tau)} > r_n\right\} \ge 1-o(1) \ \text{ for } s^* \asymp n
    \end{aligned}
    $$
    for $\tau \in \calT$ such that $\tau \le  \min(\tau^*, T-\tau^*+1) \le 2\tau$.
\end{lemma}

\noindent\textit{Proof}. To prove this lemma, we will separately consider the sparse scenario $s \ll n$ and dense scenario $s \asymp n$.

Let us first consider the sparse scenario $s \ll n$. Recall that \eqref{eq:epsilon_condition} implies that $Ts^* \rho_n \geq \log(e n) \gg \log(e n/s^*)$ for the sparse scenario, ensuring that the time span condition in Lemma \ref{lem:gen2} is satisfied. Without loss of generality, we assume that $\tau^* \le T/2$. Observe that for each given $\tau \in \calT$ with $\tau \le \tau^* \le T-\tau$, the observations within the ranges $\{1,\ldots, \tau\}$ and $\{T-\tau+1, \ldots, T\}$ are stationary. Denote by $\bTheta_1$ and $\bTheta_2$ the mean matrices in these stationary ranges. In view of Condition \ref{con:upper}, $\bTheta_1$ and $\bTheta_2$ satisfy Condition \ref{con:expansion}. Consequently, from Lemma \ref{lem:F} we see that event $F_1$ as introduced in \eqref{event:F_1} holds with high probability. We can deduce that 
 $$
 \begin{aligned}
   \Pr_1\left\{ A_{\calS}^{(\tau)} > r_n\right\} =& \Pr_1\left\{ A_{\calS}^{(\tau)} > r_n, \I(F_1)=1\right\} + \Pr_1\left\{ A_{\calS}^{(\tau)}  > r_n, \I(F_1^c)=1\right\}   \\
   \ge& \Pr_1\left\{ A_{\calS}^{(\tau)} \I(F_1) > r_n, \I(F_1)=1\right\}  - \Pr_1\{F_1^c\}\\
   =& \Pr_1\left\{ A_{\calS}^{(\tau)} \I(F_1) > r_n \right\}  - o(1),
 \end{aligned}
 $$
 where the last step above is from $r_n > 0$. 
Hence, we need only to prove the desired inequalities for {$A_{\calS}^{(\tau)} \I(F_1)$}.

Recall that $\bDelta = \bU_{\bDelta} \bD_{\bDelta} \bU_{\bDelta}^\top$ and $s^* = |\supp(\bU_{\bDelta})|$. Let us define $\calS:= \{ (i,j): \delta_{i,j} \ne 0\}$ and $\calG := \{i \in [n]: \|\bU_{\bDelta}(i) \|_2 > 0 \}$ collecting the nonzero rows of $\bU_{\bDelta}$, where $\bU_{\bDelta}(i)$ is the $i$th row of $\bU_{\bDelta}$. It holds that 
$$\calS \subset \calG \times \calG:= \tcalS \ \text{ and } \  |\calG| = s^*.$$ 
We refer to ${\tcalS}$ as a ``squared set.''
The statistic $A^{(\tau)} $ can be decomposed as 
	\begin{equation}\label{Oct21:6}
		\begin{aligned}
			A_{\calS}^{(\tau)} { \I(F_1)}
			=&  \sum_{(i,j) \in \tcalS} z^{(\tau)} _{i,j}\dot{z}^{(\tau)} _{i,j} \I\left(|\ddot{z}^{(\tau)}_{i,j}| > d(s^*)\right){ \I(F_1)}  + \sum_{(i,j) \in ({\tcalS})^c} z^{(\tau)} _{i,j}\dot{z}^{(\tau)} _{i,j} \I\left(|\ddot{z}^{(\tau)}_{i,j}| > d(s^*)\right){ \I(F_1)} \\
			:=& A^{(\tau)}_{1} { \I(F_1)} + A^{(\tau)}_{2} { \I(F_1)}.
		\end{aligned}
	\end{equation}
By similar arguments as in the proof of Lemma \ref{lem:H0inequality}, we can obtain that $$A^{(\tau)}_{2}{ \I(F_1)} = O_p(s^*) = o_p(r_n).$$ 
We would like to mention that Lemma \ref{lem:gen2} requires $({\tcalS})^c$ to be a ``squared'' set, which does not hold. However, this can be resolved by leveraging the following decomposition. Specifically, we can write  
$$
\begin{aligned}
    \left| A^{(\tau)}_{2}{ \I(F_1)} \right| =& \Big|\sum_{(i,j) \in \Omega} \left(z^{(\tau)}_{i,j} - \sqrt{\frac{\tau}{2\rho_n}}\delta_{i,j}\right)\left(\dot{z}^{(\tau)}_{i,j} - \sqrt{\frac{\tau}{2\rho_n}}\delta_{i,j}\right)\I\left(|\ddot{z}^{(\tau)}_{i,j}| > d(s^*)\right){ \I(F_1)}  \\
    &- \sum_{(i,j) \in {\tcalS}} \left(z^{(\tau)}_{i,j} - \sqrt{\frac{\tau}{2\rho_n}}\delta_{i,j}\right)\left(\dot{z}^{(\tau)}_{i,j} - \sqrt{\frac{\tau}{2\rho_n}}\delta_{i,j}\right)\I\left(|\ddot{z}^{(\tau)}_{i,j}| > d(s^*)\right){ \I(F_1)}\Big| \\
    \leq& \left|\sum_{(i,j) \in \Omega} \left(z^{(\tau)}_{i,j} - \sqrt{\frac{\tau}{2\rho_n}}\delta_{i,j}\right)\left(\dot{z}^{(\tau)}_{i,j} - \sqrt{\frac{\tau}{2\rho_n}}\delta_{i,j}\right)\I\left(|\ddot{z}^{(\tau)}_{i,j}| > d(s^*)\right){ \I(F_1)}\right| \\
    &+ \left| \sum_{(i,j) \in {\tcalS}} \left(z^{(\tau)}_{i,j} - \sqrt{\frac{\tau}{2\rho_n}}\delta_{i,j}\right)\left(\dot{z}^{(\tau)}_{i,j} - \sqrt{\frac{\tau}{2\rho_n}}\delta_{i,j}\right)\I\left(|\ddot{z}^{(\tau)}_{i,j}| > d(s^*)\right){ \I(F_1)}\right| \\
    =& O_p(s^*),
\end{aligned}
$$
using arguments similar to those in the proof of Lemma \ref{lem:H0inequality}. Then it suffices to show that 
	\begin{equation}\label{Oct23:1}
		\Pr_1\left\{ A_{1}^{(\tau)}{ \I(F_1)}  > r_n\right\} \ge 1-o(1).
	\end{equation}
    
    The statistic $A_{1}^{(\tau)}$ can be further decomposed as
	\begin{equation}\label{equ:may8:02}
		\begin{aligned}
			& A_{1}^{(\tau)} { \I(F_1)}
			= \frac{ \tau}{2\rho_n}\sum_{ (i,j) \in \tcalS}\delta^2_{i,j} \I\left(|\ddot{z}^{(\tau)}_{i,j}| > d(s^*)\right){ \I(F_1)}\\
            &+  \sqrt{\frac{\tau n}{ 4\rho_n}} \sum_{(i,j) \in \tcalS}  \sum_{l=1}^2 \delta_{i,j}(-1)^{l+1}\left(\be_i^\top \tilde{\bW}^{(\tau,l)} \bb^{(l)}_j+ \be_j^\top \tilde{\bW}^{(\tau,l)}  \bb^{(l)}_i  \right)\I\left(|\ddot{z}^{(\tau)}_{i,j}| > d(s^*)\right){ \I(F_1)} \\
			&+ {O}_p\left(\sqrt{\frac{\tau}{2n\rho_n}}\sum_{(i,j) \in \tcalS} \delta_{i,j} \I\left(|\ddot{z}_{i,j}^{(\tau)}| > d(s^*)\right){ \I(F_1)}\right) +{O}_p\left(\frac{1}{n}\sum_{(i,j) \in \tcalS}\I\left(|\ddot{z}_{i,j}^{(\tau)}| > d(s^*)\right){ \I(F_1)} \right)\\
			&+  \frac{n}{2}\sum_{(i,j) \in \tcalS}\Big\{ \sum_{l_1=1}^2 \sum_{l_2=1}^2 (-1)^{l_1+1} (-1)^{l_2+1}\left(\be_i^\top \tilde{\bW}^{(\tau,l_1)} \bb^{(l_1)}_j + \be_j^\top \tilde{\bW}^{(\tau,l_1)} \bb^{(l_1)}_i \right) \\
            &\quad \times \left(\be_i^\top \tilde{\dot{\bW}}^{(\tau,l_2)} \bb^{(l_2)}_j + \be_j^\top \tilde{\dot{\bW}}^{(\tau,l_2)} \bb^{(l_2)}_i \right)  \I\left(|\ddot{z}_{i,j}^{(\tau)}| > d(s^*)\right){ \I(F_1)}\Big\}\\
			&+ {O}_p\left(1\right)\sum_{(i,j) \in \tcalS}  \sum_{l=1}^2 (-1)^{l+1}\left(\be_i^\top \tilde{\bW}^{(\tau,l)} \bb^{(l)}_j+ \be_j^\top \tilde{\bW}^{(\tau,l)} \bb^{(l)}_i \right)\I\left(|\ddot{z}_{i,j}^{(\tau)}| > d(s^*)\right){ \I(F_1)} \\
            &+  \sqrt{\frac{\tau n}{ 4\rho_n}} \sum_{(i,j) \in \tcalS}  \sum_{l=1}^2 \delta_{i,j}(-1)^{l+1}\left(\be_i^\top \tilde{\dot{\bW}}^{(\tau,l)} \bb^{(l)}_j+ \be_j^\top \tilde{\dot{\bW}}^{(\tau,l)}  \bb^{(l)}_i  \right)\I\left(|\ddot{z}^{(\tau)}_{i,j}| > d(s^*)\right){ \I(F_1)}\\
            &+ {O}_p\left(1\right)\sum_{(i,j) \in \tcalS}  \sum_{l=1}^2 (-1)^{l+1}\left(\be_i^\top \tilde{\dot{\bW}}^{(\tau,l)} \bb^{(l)}_j+ \be_j^\top \tilde{\dot{\bW}}^{(\tau,l)} \bb^{(l)}_i \right)\I\left(|\ddot{z}_{i,j}^{(\tau)}| > d(s^*)\right){ \I(F_1)} \\
			:= &\frac{ \tau}{2 \rho_n} \sum_{(i,j) \in \tcalS}\delta^2_{i,j} \I\left(|\ddot{z}^{(\tau)}_{i,j}| > d(s^*)\right){ \I(F_1)} +v_1 + v_2 + v_3 + v_4 + v_5 + v_6 + v_7.
		\end{aligned}
	\end{equation}
We aim to show that the first term on the right-hand side above dominates $r_n$ and the remaining terms $v_1,\ldots, v_7$ are stochastically bounded by $r_n$.

Let us first prove that the screening component $\I\left(|\ddot{z}_{i,j}^{(\tau)}| > d(s^*)\right){ \I(F_1)}$ in the statistic does not reduce the signal strength that much; that is, 
$$\frac{ \tau}{2 \rho_n} \sum_{(i,j) \in \tcalS}\delta^2_{i,j} \I\left(|\ddot{z}^{(\tau)}_{i,j}| > d(s^*)\right){ \I(F_1)} \asymp \epsilon_n^2/\rho_n$$ with asymptotic probability one. To this end, we divide the signal coordinate set $\calS$ into a strong signal set $$\calA_1:= \left\{(i,j) \in \tcalS: |\delta_{i,j}| \ge { 8c_d}\sqrt{\frac{\rho_n}{\tau^* s^*}\log(en)}\right\}$$ and a weak signal set $$\calA_2 := \left\{(i,j) \in \tcalS: |\delta_{i,j}| < { 8c_d}\sqrt{\frac{ \rho_n}{\tau^* s^*} \log(en)}\right\}.$$ 
Notice that $d(s^*) \in [c_d\sqrt{1/n} a(s^*), c_d\sqrt{1/s^*} a(s^*)]$ and $a^2(s^*) = \log(en/s^*)$. Then we have 
$$\tcalS = \calA_1 \cup \calA_2.$$ 

Define  $\bar{e}_{i,j}^{(\tau)} := n \sum_{l=1}^2 (-1)^{l+1}\left(\be_i^\top \tilde{\bW}^{(\tau,l)} \bb^{(l)}_j + \be_j^\top \tilde{\bW}^{(\tau,l)} \bb^{(l)}_i \right)$ and event $$H:= \{  \min_{(i,j) \in \calA_1} |\ddot{z}^{(\tau)}_{i,j}| > d(s^*) \}.$$
We can deduce that 
	\begin{equation}\label{Feb11:equ:01}
		\begin{aligned}
        \Pr(H^c\cap{ F_1})=
		&\Pr_1\left\{ \min_{(i,j) \in \calA_1}|\ddot{z}^{(\tau)}_{i,j}| \le d(s^*), { F_1} \right\}\\
        =& \sum_{(i,j) \in \calA_1}\Pr_1\left\{\left|\sqrt{\frac{\tau}{2\rho_n} } \delta_{i,j}+ \sqrt{\frac{1}{2n}}\bar{e}_{i,j}^{(\tau)} + {O}(\frac{1}{\sqrt{n}})\right| \le d(s^*)\right\} \\
        \le&  \sum_{(i,j) \in \calA_1} \Pr\left\{ \left|\bar{e}_{i,j}^{(\tau)}\right| \ge  \sqrt{\frac{n \tau}{\rho_n}}|\delta_{i,j}| - \sqrt{n}d(s^*) -{O}(1)\right\} \\
        \le&  \sum_{(i,j) \in \calA_1}\Pr\left\{   \left|\bar{e}_{i,j}^{(\tau^*)}\right| \ge  \sqrt{3\log(n)}\right\} \le n^{-1},
	\end{aligned}
		\end{equation}
	where the first inequality above has utilized the fact that for $s^* \ge 1$ and $s^* \ll n$,
	$$
	\frac{1}{4}\sqrt{\frac{n \tau}{ \rho_n}}| \delta_{i,j}| \ge { c_d}\sqrt{\frac{n }{s^*}\log(en)}  \gg { c_d}\sqrt{\frac{n}{s^*}}a(s^*) \ge \sqrt{n}d(s^*),
	$$
   leading to 
    $$
    \sqrt{\frac{n \tau}{\rho_n}}|\delta_{i,j}| - \sqrt{n}d(s^*) -{O}(1) \ge \frac{3}{4} \sqrt{\frac{n \tau}{\rho_n}}|\delta_{i,j}|  \ge \sqrt{3\log(n)},
    $$
    and last inequality above is due to \eqref{add:lem:equ1}. 
    
    Further, since $\tau^* \sum_{(i,j) \in \calS} \delta_{i,j}^2  \ge \epsilon_n^2$, it holds that 
	\begin{equation}\label{eq:002}
    \begin{aligned}
	\frac{\tau}{2 \rho_n}\sum_{(i,j) \in \calA_1 } \delta^2_{i,j} =& \frac{\tau}{2 \rho_n}\left( \sum_{(i,j) \in \tcalS} \delta^2_{i,j}  - \sum_{(i,j) \in \calA_2} \delta^2_{i,j} \right) \\
    =& \frac{\tau}{2 \rho_n}\left( \sum_{(i,j) \in \calS} \delta^2_{i,j}  - \sum_{(i,j) \in \calA_2} \delta^2_{i,j} \right) \\
    \ge& \frac{\tau}{2 \rho_n} \left( \frac{T\epsilon_n^2}{\tau^*(T-\tau^*)} - { 64 c^2_d}(s^*)^2  \frac{\rho_n}{\tau^* s^*} \log(en)\right)\\
    \ge & \frac{\epsilon_n^2}{4\rho_n} -  { 32 c^2_d}s^* \log(en)   \ge 2 r_n,
    \end{aligned}
    \end{equation} 
    where we have used the fact of $1/2 \le \tau/\tau^* \le 1$.
Combining \eqref{Feb11:equ:01} and \eqref{eq:002} above, we can obtain that 
\begin{equation}
    \begin{aligned}
        &\Pr\left\{\frac{ \tau}{2\rho_n} \sum_{ (i,j) \in \tcalS}\delta^2_{i,j} \I\left(|\ddot{z}^{(\tau)}_{i,j}| > d(s^*)\right){ \I(F_1)} \le  r_n\right\}\\
        \le& \Pr\left\{\frac{ \tau}{2\rho_n}\sum_{ (i,j) \in \calA_1}\delta^2_{i,j} \I\left(|\ddot{z}^{(\tau)}_{i,j}| > d(s^*)\right) \le  r_n, H\cap{ F_1}\right\} + \Pr_1(H^c \cap{ F_1})\\
        \le&  \Pr\left\{\frac{ \tau}{2\rho_n}\sum_{ (i,j) \in \calA_1}\delta^2_{i,j} \le  r_n\right\} + \Pr_1(H^c{ \cap F_1}) \le n^{-1}.
    \end{aligned}
\end{equation}
    Hence, it follows that 
    \begin{equation}\label{equ:signal}
    \begin{aligned}
	\Pr\left\{ \frac{ \tau}{2 \rho_n}   \sum_{(i,j) \in \tcalS}\delta^2_{i,j} \I\left(|\ddot{z}^{(\tau)}_{i,j}| > d(s^*)\right) { \I(F_1)}  \ge r_n \right\} \ge 1-n^{-1}.
    \end{aligned}
    \end{equation}

We now focus on bounding the remaining terms. For term $v_1$, by part 3) of Lemma \ref{lem:gen2}  (where we set $\mu_{i,j} = \delta_{i,j}$), we have that 
$$\Var\left(\sum_{(i,j) \in \tcalS} \delta_{i,j}\be_i^\top \tilde{\bW}^{(\tau,l)} \bb^{(l)}_j\right) = O\left(\frac{s^*}{n^2} \sum_{(i,j) \in \tcalS} \delta_{i,j}^2\right)$$ and $$\Var\left(\sum_{(i,j) \in \tcalS} \delta_{i,j}\be_j^\top \tilde{\bW}^{(\tau,l)} \bb^{(l)}_i\right) = O\left(\frac{s^*}{n^2} \sum_{(i,j) \in \tcalS} \delta_{i,j}^2\right).$$ Then an application of the Markov inequality together with $s^* \ll n$ gives that 
	\begin{equation}\label{add:h0:v1}
    \begin{aligned}
		v_1 & = \sqrt{\frac{T n}{\rho_n}}O_p\left(\sqrt{\frac{s^*}{n^2 } \sum_{(i,j)\in \tcalS} \delta_{i,j}^2}\right) = o_p\left(\sqrt{\frac{T}{\rho_n} \sum_{(i,j)\in \tcalS} \delta_{i,j}^2}\right) \\
        &= o_p\left(\sqrt{\frac{\epsilon_n^2}{\rho_n}}\right)=o_p\left(\frac{\epsilon_n^2}{\rho_n}\right) = o_p(r_n).
        \end{aligned}
	\end{equation}
    Similarly, we can show that 
        $$v_6 = o_p(r_n).$$
        
	For term $v_2$, an application of the Cauchy--Schwarz inequality leads to $$\sum_{(i,j)\in \tcalS} \delta_{i,j} \le s^*\sqrt{ \sum_{(i,j)\in \tcalS} \delta^2_{i,j} },$$ 
	which along with $s^* \ll n$ entails that 
	\begin{equation}\label{add:h0:v2}
    \begin{aligned}
		v_2 & = O_p\left(\sqrt{\frac{T(s^*)^2}{n\rho_n} \sum_{(i,j)\in\tcalS} \delta_{i,j}^2} \right) = o_p\left( \sqrt{\frac{Ts^* \sum_{(i,j) \in \tcalS} \delta_{i,j}^2}{\rho_n}  } \right) \\
        &= o_p\left(\sqrt{\frac{\epsilon_n^2 s^*}{\rho_n}}\right)=o_p\left(\frac{\epsilon_n^2}{\rho_n}\right)=o_p(r_n).
        \end{aligned}
	\end{equation}
	For term $v_3$, it follows from $|\tcalS| = (s^*)^2$ that 
	\begin{equation}\label{add:h0:v3}
		v_3 = {O}\left(\frac{(s^*)^2}{n}\right) = o(s^*) = o\left( r_n \right).
	\end{equation}
	For term $v_4$, by resorting to part 4) of Lemma \ref{lem:gen2}, we have that $$\Var\left(\sum_{(i,j) \in \tcalS}\be_i^\top \tilde{\bW}^{(\tau,l)} \bb^{(l)}_j\be_i^\top \tilde{\dot{\bW}}^{(\tau)} \bb^{(l)}_j \right) = O(s^*/n^2).$$ 
	Similar to part 4) of Lemma \ref{lem:gen2}, we can deduce that $$\Var\left(\sum_{(i,j) \in \tcalS}\be_i^\top \tilde{\bW}^{(\tau,l)} \bb^{(l)}_j\be_j^\top \tilde{\dot{\bW}}^{(\tau,l)} \bb_i \right) = O(s^*/n^2),$$ 
    $$\Var\left(\sum_{(i,j) \in \tcalS}\be_j^\top \tilde{\bW}^{(\tau,l)} \bb^{(l)}_i\be_i^\top \tilde{\dot{\bW}}^{(\tau,l)} \bb^{(l)}_j \right) = O(s^*/n^2),$$ 
    and 
    $$\Var\left(\sum_{(i,j) \in \tcalS}\be_j^\top \tilde{\bW}^{(\tau,l)} \bb^{(l)}_i\be_j^\top \tilde{\dot{\bW}}^{(\tau,l)} \bb^{(l)}_i \right) = O(s^*/n^2).$$ 
    
    Combining the above results and applying the Markov inequality yield that 
	\begin{equation}\label{add:h0:v4}
		v_4 = nO_p\left( \sqrt{\frac{s^*}{n^2}}\right)  = o_p\left(r_n\right).
	\end{equation}
	For term $v_5$, it follows from part 3) of Lemma \ref{lem:gen2} that $$\Var\left( \sum_{(i,j) \in \tcalS} \be_i^\top \tbW^{(\tau,l)} \bb^{(l)}_j \I(|\ddot{z}_{i,j}^{(\tau)}| \ge d(s^*)) { \I(F_1)}\right) = O(\frac{(s^*)^3}{n^2}) =  O(s^*).$$ 
	Similarly, it holds that 
	$$\Var\left( \sum_{(i,j) \in \tcalS} \be_j^\top \tbW^{(\tau,l)} \bb^{(l)}_i \I(|\ddot{z}_{i,j}^{(\tau)}| \ge d(s^*)){ \I(F_1)}\right) =  O(s^*).$$ Thus, we can obtain that 
	\begin{equation}\label{add:h0:v5}
		v_5 = O_p(\sqrt{s^*}) = o_p\left(r_n\right).
	\end{equation}
	Similar to $v_5$, we can also show that 
	$$v_7 = o_p\left(r_n\right).$$ 
	Therefore, plugging \eqref{equ:signal}--\eqref{add:h0:v5} into \eqref{equ:may8:02} leads to the conclusion in \eqref{Oct23:1}. This concludes the proof of the sparse scenario of Lemma \ref{lem:H1inequality}.

We now turn to the dense scenario of $s^* \asymp n$.  Similarly, the statistic $A_{\Omega}^{(\tau)}$ can be further decomposed as
	\begin{equation}\label{equ:july07:lem5:01}
		\begin{aligned}
		 A_{\Omega}^{(\tau)}\I(F_1) 
			& = \frac{ \tau}{2\rho_n}\sum_{ (i,j) \in \Omega}\delta^2_{i,j}\I(F_1)\\
            &\quad+  \sqrt{\frac{\tau n}{ 4\rho_n}} \sum_{(i,j) \in \Omega}  \sum_{l=1}^2 \delta_{i,j}(-1)^{l+1}\left(\be_i^\top \tilde{\bW}^{(\tau,l)} \bb^{(l)}_j+ \be_j^\top \tilde{\bW}^{(\tau,l)}  \bb^{(l)}_i  \right)\I(F_1) \\
			&\quad+ O_p\left(\sqrt{\frac{\tau}{2n\rho_n}}\sum_{(i,j) \in \Omega} \delta_{i,j}\I(F_1)\right)\\
			&\quad+  \frac{n}{2}\sum_{(i,j) \in \Omega}\Big\{ \sum_{l_1=1}^2 \sum_{l_2=1}^2 (-1)^{l_1+1} (-1)^{l_2+1}\left(\be_i^\top \tilde{\bW}^{(\tau,l_1)} \bb^{(l_1)}_j + \be_j^\top \tilde{\bW}^{(\tau,l_1)} \bb^{(l_1)}_i \right) \\
            &\quad\quad \times \left(\be_i^\top \tilde{\dot{\bW}}^{(\tau,l_2)} \bb^{(l_2)}_j + \be_j^\top \tilde{\dot{\bW}}^{(\tau,l_2)} \bb^{(l_2)}_i \right)\I(F_1)\Big\}\\
			&\quad+ O_p\left(1\right)\sum_{(i,j) \in \Omega}  \sum_{l=1}^2 (-1)^{l+1}\left(\be_i^\top \tilde{\bW}^{(\tau,l)} \bb^{(l)}_j+ \be_j^\top \tilde{\bW}^{(\tau,l)} \bb^{(l)}_i \right)\I(F_1) \\
            &\quad+  \sqrt{\frac{\tau n}{ 4\rho_n}} \sum_{(i,j) \in \Omega}  \sum_{l=1}^2 \delta_{i,j}(-1)^{l+1}\left(\be_i^\top \tilde{\dot{\bW}}^{(\tau,l)} \bb^{(l)}_j+ \be_j^\top \tilde{\dot{\bW}}^{(\tau,l)}  \bb^{(l)}_i  \right)\I(F_1)\\
            &\quad+ o_p\left(1\right)\sum_{(i,j) \in \Omega}  \sum_{l=1}^2 (-1)^{l+1}\left(\be_i^\top \tilde{\dot{\bW}}^{(\tau,l)} \bb^{(l)}_j+ \be_j^\top \tilde{\dot{\bW}}^{(\tau,l)} \bb^{(l)}_i \right)\I(F_1) +O_p\left(n\right)\\
			& := \frac{ \tau}{2 \rho_n} \sum_{(i,j) \in \Omega}\delta^2_{i,j} \I(F_1)+z_1 + z_2 + z_3 + z_4 + z_5 + z_6 + O_p\left(n\right).
		\end{aligned}
	\end{equation}
    
An application of similar arguments as for \eqref{add:h0:v1}--\eqref{add:h0:v4} gives that $z_l = o_p(r_n), \, l\in [6]$. Using these bounds, we see that \eqref{equ:july07:lem5:01} reduces to
\begin{equation}\label{equ:sep:05}
\begin{aligned}
A_{\Omega}^{(\tau)} \I(F_1) =& \frac{ \tau}{2 \rho_n} \sum_{(i,j) \in \Omega}\delta^2_{i,j}\I(F_1) + o_p(r_n)+O_p(n).
\end{aligned}    
\end{equation}
Moreover, it holds that 
$$
\begin{aligned}
 \Pr_1\left\{ \frac{ \tau}{2 \rho_n} \sum_{(i,j) \in \Omega}\delta^2_{i,j}\I(F_1) \lesssim r_n\right\} \le&  \Pr_1\left\{ \frac{ \tau}{2 \rho_n} \sum_{(i,j) \in \Omega}\delta^2_{i,j}\I(F_1) \lesssim r_n, \I(F_1)=1\right\} + \Pr_1\left\{ F_1^c\right\}\\
 \le&  \Pr_1\left\{ \frac{ \tau}{2 \rho_n} \sum_{(i,j) \in \Omega}\delta^2_{i,j} \lesssim r_n\right\} + o(1) = o(1),
\end{aligned}
$$
where the last step above is due to $\epsilon_n$ satisfying \eqref{equ:upper_sep}. This entails that $$\frac{ \tau}{2 \rho_n} \sum_{(i,j) \in \Omega}\delta^2_{i,j}\I(F_1) \gtrsim r_n$$ with 
probability $1-o(1)$. Combining this with \eqref{equ:sep:05} and noting that $r_n \gg n$, we can obtain that 
$$
\begin{aligned}
A_{\Omega}^{(\tau)} \I(F_1) \gtrsim & r_n
\end{aligned}
$$
with probability $1-o(1)$. This concludes the proof for the dense scenario of Lemma \ref{lem:H1inequality}.

\section{Proofs of Theorems \ref{Thm:null}--\ref{Thm:power} and some key lemmas} \label{new.sec.B}

\subsection{Proof of Theorem \ref{Thm:null}} \label{new.sec.B.1}

We aim to derive the asymptotic null distribution of $\hat{A}^{(\tau)}_{\calS}$. To this end, let us write 
\begin{equation}\label{march15:equ01}
\begin{aligned}
    f_{i,j}^{(\tau)}&=\frac{q_n}{\sqrt{\tau}} \sum_{k=1}^K \left( \be_i^\top \tilde{\bW}^{(\tau, 2)} \frac{d_k}{t_k}\bv_k v_k(j) + \be_j^\top \tilde{\bW}^{(\tau, 2)} \frac{d_k}{t_k}\bv_k v_k(i) \right) \\
    &= \frac{q_n}{\sqrt{\tau}} \left( \be_i^\top \tilde{\bW}^{(\tau, 2)} \bb_j +  \be_j^\top \tilde{\bW}^{(\tau, 2)} \bb_i\right),
    \end{aligned}
\end{equation}
where $\tilde{\bW}^{(\tau, 2)}$ is defined in \eqref{def:tbW_lr} and $\bb_j = \sum_{k=1}^K \frac{d_k}{t_k} \bv_k v_k(j)$ with $t_k$ defined in \eqref{new.eq.FL011}. The conditions of Theorem \ref{Thm:null} ensure that the conditions of Lemma \ref{lem:gen1} in Section \ref{new.sec.D} are satisfied. Thus, by Lemma \ref{lem:gen1} and the fact that $|\calT|$ is a constant, it holds that 
\begin{equation}\label{equ:may11:01}
 \begin{aligned}
	\hat{w}_{i,j}^{(\tau)} = &\bar{w}_{i,j}^{(\tau, 1)} + \theta_{i,j}^{(\tau, 1)} - \theta_{i,j}^{(\tau, 2)} + \theta_{i,j}^{(\tau, 2)}-   \htheta_{i,j}^{(\tau, 2)}\\
	=& \bar{w}_{i,j}^{(\tau, 1)} - \frac{q_n}{\sqrt{\tau}} \sum_{k=1}^K \left( \be_i^\top \tilde{\bW}^{(\tau, 2)} \frac{d_k}{t_k}\bv_k v_k(j) + \be_j^\top \tilde{\bW}^{(\tau, 2)} \frac{d_k}{t_k}\bv_k v_k(i) \right) + { O}(\frac{q}{n \sqrt{\tau}})\\
	:=& \bar{w}_{i,j}^{(\tau,1)} -  f_{i,j}^{(\tau)} + { O}(\frac{q_n}{n \sqrt{\tau}})
\end{aligned}   
\end{equation}
uniformly over $(\tau, i, j) \in \calT \times [n] \times [n]$ with asymptotic probability one.

Similarly,  let us define 
$$
\begin{aligned}
\dot{f}_{i,j}^{(\tau)} & =\frac{q_n}{\sqrt{\tau}} \sum_{k=1}^K \left( \be_i^\top \tilde{\dot{\bW}}^{(\tau, 2)} \frac{d_k}{t_k}\bv_k v_k(j) + \be_j^\top \tilde{\dot{\bW}}^{(\tau, 2)} \frac{d_k}{t_k}\bv_k v_k(i) \right)\\
&= \frac{q_n}{\sqrt{\tau}} \left( \be_i^\top \tilde{\dot{\bW}}^{(\tau, 2)} \bb_j +  \be_j^\top \tilde{\dot{\bW}}^{(\tau, 2)} \bb_i\right),
\end{aligned}
$$
where $\tilde{\dot{\bW}}^{(\tau, 2)}$ is defined in \eqref{def:tbW_lr} with all $\bW^{(t)}$'s replaced by their i.i.d. copies $\dot{\bW}^{(t)}$'s. Then we see that event 
\begin{equation}\label{equ:may11:02}
 \begin{aligned}
	E_w := \left\{\hat{\dot{w}}_{i,j}^{(\tau)} = \bar{\dot{w}}_{i,j}^{(\tau,1)} -  \dot{f}_{i,j}^{(\tau)} + { O}(\frac{q_n}{n \sqrt{\tau}}) \text{ uniformly over}  (\tau, i, j) \in \calT \times [n] \times [n] \right\}
\end{aligned}   
\end{equation}
holds with asymptotic probability one.

In light of (\ref{equ:may11:01}) and (\ref{equ:may11:02}), we can decompose $\hA_{\calS}^{(\tau)}$ as 
\begin{equation}\label{Feb13:equ:A}
	\begin{aligned}
		\hA^{(\tau)}_{\calS} &= \sum_{(i,j) \in \calS} \hat{w}_{i,j}^{(\tau)} \hat{\dot{w}}_{i,j}^{(\tau)} \\
		&=\sum_{(i,j) \in \calS} \left( \bar{w}^{(\tau,1)}_{i,j} -  f^{(\tau)}_{i,j} + { O}(\frac{q_n}{n \sqrt{\tau}}) \right) \left(  \bar{\dot{w}}^{(\tau,1)}_{i,j} - \dot{f}^{(\tau)}_{i,j} + { O}(\frac{q_n}{n \sqrt{\tau}})\right)\\
		&=\sum_{(i,j) \in \calS} \bar{w}^{(\tau,1)}_{i,j}  \bar{\dot{w}}^{(\tau,1)}_{i,j}  - \sum_{(i,j) \in \calS} (\bar{w}^{(\tau,1)}_{i,j} \dot{f}^{(\tau)}_{i,j} + \bar{\dot{w}}^{(\tau,1)}_{i,j} f^{(\tau)}_{i,j}) \\
        & \quad+ \sum_{(i,j) \in \calS} (\bar{w}^{(\tau,1)}_{i,j}  + \bar{\dot{w}}^{(\tau,1)}_{i,j}) { O}(\frac{q_n}{n \sqrt{\tau}}) +\sum_{(i,j) \in \calS} f^{(\tau)}_{i,j}  \dot{f}^{(\tau)}_{i,j} \\
		&  \quad + { O}(\frac{q_n}{n \sqrt{\tau}}) \sum_{(i,j) \in \calS} (f^{(\tau)}_{i,j} + \dot{f}^{(\tau)}_{i,j} ) +\sum_{(i,j) \in \calS} { O}(\frac{q_n^2}{n^2 \tau}) \\
         &:= \sum_{(i,j) \in \calS} \bar{w}^{(\tau,1)}_{i,j}  \bar{\dot{w}}^{(\tau,1)}_{i,j} + (\nu_{11}^{(\tau)}+\nu_{12}^{(\tau)}) +  \nu_{2}^{(\tau)} +  \nu_3^{(\tau)} +  \nu_4^{(\tau)} + { O}\left(\frac{\rho_n |S|}{nT} \right)\\
		&:= \Upsilon_n^{(\tau)}  + \nu_{1}^{(\tau)} +  \nu_{2}^{(\tau)} +  \nu_3^{(\tau)} +  \nu_4^{(\tau)} + { O}\left(\frac{\rho_n |S|}{nT}\right) 
	\end{aligned}
\end{equation}
uniformly over $(\tau, i, j) \in \calT \times [n] \times [n]$ with asymptotic probability one.
With the aid of Lemma \ref{lem:reminder} in Section \ref{new.sec.B.3}, the variance of each remainder term satisfies that 
 $$
 {\Var(\nu^{(\tau)}_l)} = O\left(\frac{\rho^2_n {|\calS|}}{T^2 {n}}\right)
 $$
 for $l=1,\ldots, 4$.
 
Let $\xi_{n,\calS} := \rho_n^2|\calS|/(\eta_n^2 n \sigma_{\calS}^2) \to 0$ with some $\eta_n \to 0$, and { $\gamma_{\tau}:=\tau^{-2}$}. Note that $\E(\nu_l)=0$ with $l \in [4]$. An application of the Markov inequality gives that 
\begin{equation}\label{equ:march14:01}
\begin{aligned}
\Pr_0\left\{ \nu_l  \ge \eta_n \sqrt{\gamma_{\tau} \sigma_{\calS}^2} \right\} & \le  \Pr_0\left\{ |\nu_l - \E(\nu_l)|  \ge \eta_n \sqrt{\gamma_{\tau} \sigma_{\calS}^2} \right\} \\
& = O\left(\frac{\rho_n^2|\calS|}{ \eta_n^2 n \sigma_{\calS}^2}\right)= O(\xi_{n,\calS}),    
\end{aligned}
\end{equation}
where we have used the fact of $\gamma_{\tau} \asymp T^{-2}$. Then it follows that for each $t \in \R$,
\begin{equation}\label{Feb24:equ:01}
\begin{aligned}
   \Pr_0\left\{ \frac{\hA^{(\tau)}_{\calS}}{\sqrt{\gamma_{\tau} \sigma_{\calS}^2}} \ge t\right\} \le \Pr_0\left\{ \frac{\Upsilon^{(\tau)}_n}{\sqrt{\gamma_{\tau} \sigma_{\calS}^2}} \ge t -\eta_n \right\}  + O(\xi_n) + { \Pr_0\left\{ E_{w}^c\right\}}. 
\end{aligned}
\end{equation}
Further, by Lemma \ref{lem:asymptotic} in Section \ref{new.sec.B.3} and condition \eqref{con:nu1}, we can deduce that 
\begin{equation}\label{Feb13:equ:06}
\Pr_0\left\{ \frac{\Upsilon^{(\tau)}_n}{\sqrt{\gamma_{\tau} \sigma_{\calS}^2}} \ge t -\eta_n \right\} \le \Phi(t-\eta_n)+ { o(1)}.
\end{equation}
Using the anti-concentration property of Gaussian random variables (see, e.g., Theorem 2.1 in \cite{gotze2019large}), it holds that 
\begin{equation}\label{Feb24:equ:02}
    \Phi(t-\eta_n) \le  \Phi(t) + O(\eta_n^2).
\end{equation}
Hence, combining \eqref{Feb24:equ:01}--\eqref{Feb24:equ:02} yields that for each $t \in \R$,
\begin{equation*}
\begin{aligned}
   \Pr_0\left\{ \frac{\hA^{(\tau)}_{\calS}}{\sqrt{\gamma_{\tau} \sigma_{\calS}^2}}  \ge t\right\} - \Phi(t) = O\left(\xi_{n,\calS} +
   \eta_n^2 \right) + o(1).
\end{aligned}
\end{equation*}

For the other side of the inequality, similar arguments as for \eqref{equ:march14:01} and \eqref{Feb24:equ:01} are applicable to show that 
\begin{equation*}
\begin{aligned}
   \Pr_0\left\{ \frac{\hA^{(\tau)}_{\calS}}{\sqrt{\gamma_{\tau} \sigma_{\calS}^2}}  \ge t - \eta_n\right\} \ge \Pr_0\left\{ \frac{\Upsilon^{(\tau)}_n}{\sqrt{\gamma_{\tau} \sigma_{\calS}^2}} \ge t \right\}  - O(\xi_{n,\calS}) - o(1). 
\end{aligned}
\end{equation*}
This entails that 
\begin{equation*}
\begin{aligned}
   \Pr_0\left\{ \frac{\hA^{(\tau)}_{\calS}}{\sqrt{\gamma_{\tau} \sigma_{\calS}^2}}  \ge t \right\} \ge \Pr_0\left\{ \frac{\Upsilon^{(\tau)}_n}{\sqrt{\gamma_{\tau} \sigma_{\calS}^2}} \ge t +\eta_n\right\}  - O(\xi_{n,\calS}) - o(1). 
\end{aligned}
\end{equation*}
Consequently, similar to \eqref{Feb13:equ:06} and \eqref{Feb24:equ:02}, we can obtain that 
\begin{equation}\label{equ:march14:02}
\begin{aligned}
   \Pr_0\left\{ \frac{\hA^{(\tau)}_{\calS}}{\sqrt{\gamma_{\tau} \sigma_{\calS}^2}} \ge t \right\} & \ge \Phi(t+\eta_n)- O(\xi_{n,\calS})  \\
   &\ge  \Phi(t)  - O(\xi_{n,\calS} + \eta_n^2) - o(1). 
\end{aligned}
\end{equation}

Since inequalities \eqref{Feb24:equ:02} and \eqref{equ:march14:02} hold uniformly over $t \in \R$, we have that 
\begin{equation*}
\begin{aligned}
   \sup_{t \in \R}\left|\Pr_0\left\{ \frac{\hA^{(\tau)}_{\calS}}{\sqrt{\gamma_{\tau} \sigma_{\calS}^2}}  \le t\right\} - \Phi(t) \right| = O\left( \xi_{n,\calS}  + \eta_n^2\right) + o(1) = o(1), 
\end{aligned}
\end{equation*}
where we take $\eta_n \to 0$ such that $\xi_{n,\calS} \to 0$. Such sequence exists thanks to the condition that there exists a sequence $\eta_n \to 0$ satisfying $\xi_{n,\calS} = \rho_n^2 |\calS| / (\eta_n^2 n \sigma_{\calS}^2) \to 0$.  This completes the proof of Theorem \ref{Thm:null}.



\subsection{Proof of Theorem \ref{Thm:power}} \label{new.sec.B.2}

We will separate the proof into two parts.

\medskip

\textbf{Part I (Size control)}. We first show the size control of the empirical MOSAIC test statistic. In view of \eqref{equ:b6:5}, we can show that $$\Pr_0\left\{ \hcalS = \emptyset \right\} = 1- o(1)$$
and 
\begin{equation}\label{equ:may11:011}
\Pr_0\left\{ \hA \ge z_{\alpha/(2|\calT|)} \right\} \le  \sum_{\tau \in \calT}\Pr\left\{ \frac{|\hA^{(\tau)}_{\Omega}|}{\sqrt{\gamma_{\tau} \hsigma^2_{\Omega}}} \ge z_{\alpha/(2|\calT|)} \right\} + o(1).   
\end{equation}
By condition \ref{Jan5:1:omega} and with the aid of Lemma \ref{lem:svd_var} in Section \ref{new.sec.B.3}, we can obtain that for some $\zeta_n \to 0$, 
$$
\begin{aligned}
 \eqref{equ:may11:01} \le   \sum_{\tau \in \calT} \Pr\left\{ \frac{|\hA^{(\tau)}_{\Omega}|}{\sqrt{\gamma_{\tau} \sigma_{\Omega}^2}} \ge z_{\alpha/(2|\calT|)}(1-\zeta_n)^{1/2} \right\} + o(1).
\end{aligned}
$$
Let $\Phi(x)$ be the upper cumulative distribution function of the standard normal distribution. By noting that for edge set $\Omega$, the condition \eqref{Jan5:1:omega} holds, which implies that the moment conditions in \eqref{con:nu1} of Theorem \ref{Thm:null} hold.
      Besides, since there exists a sequence $\eta_n \to 0$ such that $\rho_n^2 n\ll\eta_n^2 \sigma^2_{\Omega}$, then it follows from Theorem \ref{Thm:null} and the continuity of the normal distribution that 
$$
\begin{aligned}
 \eqref{equ:may11:01} \le&  \sum_{\tau \in \calT}\Phi\left((1-\zeta_n)^{1/2}z_{\alpha/(2|\calT|)} \right) + O\left(\xi_n + \omega_n^{-1/5} + \eta_n^2 \right)+o(1)\\
 =& \alpha +o(1),
\end{aligned}
$$
where the last inequality is from the condition that there exists a sequence $\eta_n \to 0$ such that $\rho_n^2 n\ll\eta_n^2 \sigma^2_{\Omega}$, which implies there exists a sequence $\eta_n \to 0$ such that $\xi_{n} = \xi_{n,\Omega} = \rho_n^2|\Omega|/(\eta_n^2 n \sigma_{\Omega}^2) \to 0$. This establishes the size part.

\medskip

\textbf{Part II (Power)}. Note that for each given $\tau$ with $\tau \le \min(\tau^*,T - \tau^*) \le 2\tau$, the observations within the ranges $\{1,\ldots, \tau\}$ and $\{T-\tau+1, \ldots, T\}$ are stationary. Let us first consider the case when $s^* \ll n$. By definition, it holds that 
$$
\hat{A} \ge |\hat{A}_{\calS}| \ge \frac{|\hat{A}^{(\tau)}_{\hcalS}|}{\sqrt{\gamma_{\tau} \hsigma^2_{\hcalS}}}.
$$
Denote by 
\begin{equation}\label{def:s_out}
\tilde{\calS} = \{(i,j): |\hddz^{(hT)}_{i,j}| > \cd_n, (i,j) \in \Omega\},    
\end{equation}
which is defined similarly as $\hcalS$ with the sample replaced by the independent copies of $\hz_{i,j}$. Consequently, $\tilde{\calS}$ is independent of $w_{i,j}^{(\tau, l)}$ and $\dot{w}_{i,j}^{(\tau, l)}$ for all $(i,j) \in \Omega$. 
Based on $\gamma_{\tau} \asymp \tau^{-2}$ and Lemma \ref{lem:bridge} in Section \ref{new.sec.B.3}, we have that 
\begin{equation}\label{equ:may12:06}
   |\hat{A}^{(\tau)}_{\tilde{\calS}}-\hat{A}^{(\tau)}_{\hcalS}| = O_p\left(  \frac{|\calS_3| \rho_n \log(n)}{T} \right).
\end{equation}
Such $\hA^{(\tau)}_{\tilde{\calS}}$ serves as a bridge to prove the power of $\hA^{(\tau)}_{\hcalS}$.

We now investigate the asymptotic behavior of $\hA_{\tilde{\calS}}$. Since the Condition \ref{con:upper} of Theorem \ref{Thm:power} ensure the conditions of Lemma \ref{lem:gen1}, an application of Lemma \ref{lem:gen1} gives that 
\begin{equation}\label{equ:june22:thm4:01}
 \begin{aligned}
	\hat{w}_{i,j}^{(\tau)} = &\bar{w}_{i,j}^{(\tau, 1)} + \theta_{i,j}^{(\tau, 1)} - \theta_{i,j}^{(\tau, 2)} + \theta_{i,j}^{(\tau, 2)} -   \htheta_{i,j}^{(\tau, 2)}\\
	=& \bar{w}_{i,j}^{(\tau, 1)} + \delta_{i,j} - \frac{q_n}{\sqrt{\tau}} \sum_{k=1}^K \left( \be_i^\top \tilde{\bW}^{(\tau, 2)} \frac{d^{(2)}_k}{t^{(2)}_k} \bv^{(2)}_k v^{(2)}_k(j) + \be_j^\top \tilde{\bW}^{(\tau, 2)} \frac{d^{(2)}_k}{t^{(2)}_k}\bv^{(2)}_k v^{(2)}_k(i) \right) + { O}(\frac{q_n}{n \sqrt{\tau}})\\
    =& \bar{w}_{i,j}^{(\tau, 1)} + \delta_{i,j} - \frac{q_n}{\sqrt{\tau}} \left( \be_i^\top \tilde{\bW}^{(\tau, 2)} \bb^{(2)}_j + \be_j^\top \tilde{\bW}^{(\tau, 2)} \bb^{(2)}_i\right) + { O}(\frac{q_n}{n \sqrt{\tau}})\\
	:=& \bar{w}_{i,j}^{(\tau,1)} + \delta_{i,j} -  f_{i,j}^{(\tau)} + { O}(\frac{q_n}{n \sqrt{\tau}})
\end{aligned}   
\end{equation}
uniformly over $(i,j) \in \Omega$ with asymptotic probability one,
where $\bb^{(2)}_j = \sum_{k=1}^K \frac{d^{(2)}_k}{t^{(2)}_k}\bv^{(2)}_k v^{(2)}_k(j)$. Similarly, we can deduce that 
$$
\begin{aligned}
 \hat{w}_{i,j}^{(\tau)} =&\bar{\dot{w}}_{i,j}^{(\tau, 1)} + \delta_{i,j} - \frac{q_n}{\sqrt{\tau}} \left( \be_i^\top \tilde{\dot{\bW}}^{(\tau, 2)} \bb^{(2)}_j + \be_j^\top \tilde{\dot{\bW}}^{(\tau, 2)} \bb^{(2)}_i\right) + { O}(\frac{q_n}{n \sqrt{\tau}})\\
 =&\bar{\dot{w}}_{i,j}^{(\tau,1)} + \delta_{i,j} -  \dot{f}_{i,j}^{(\tau)} + { O}(\frac{q_n}{n \sqrt{\tau}}).   
\end{aligned}
$$
uniformly over $(i,j) \in \Omega$ with asymptotic probability one.

Since $\tilde{\calS}$ that is independent of $\hat{w}_{i,j}^{(\tau)}$, we can decompose $\hA^{(\tau)}_{\tilde{\calS}}$ as
\begin{equation}\label{Feb13:equ:07}
	\begin{aligned}
		\hA^{(\tau)}_{\tilde{\calS}} =& \sum_{(i,j) \in \tilde{\calS}} \hat{w}_{i,j}^{(\tau)} \hat{\dot{w}}_{i,j}^{(\tau)} \\
		=& \sum_{(i,j) \in \tilde{\calS}}  \delta_{i,j}^2 + \sum_{(i,j) \in \tilde{\calS}} \bar{w}^{(\tau,1)}_{i,j}  \bar{\dot{w}}^{(\tau,1)}_{i,j}  + \sum_{(i,j) \in \tilde{\calS}}   \delta_{i,j}(\bar{w}^{(\tau,1)}_{i,j} + \bar{\dot{w}}^{(\tau,1)}_{i,j})\\
        &-\sum_{(i,j) \in \tilde{\calS}}   \delta_{i,j}(f^{(\tau)}_{i,j} + \dot{f}^{(\tau)}_{i,j}) + { O}\left( \sum_{(i,j) \in \tilde{\calS}} \frac{q_n \delta_{i,j}}{n \sqrt{\tau}}\right) + O\left(\frac{\rho_n \sqrt{|\tilde{\calS}|}}{T \sqrt{n}}\right)\\
		:=&  \sum_{(i,j) \in \tilde{\calS}} \delta_{i,j}^2+ r_1 +  r_2^{(\tau)}  + r_3^{(\tau)} +  r_4^{(\tau)}   + O\left(\frac{\rho_n \sqrt{|\tilde{\calS}|}}{T \sqrt{n}}\right)\\
	\end{aligned}
\end{equation}
with asymptotic probability one. Here, the second equality above has used similar calculations as in \eqref{Feb13:equ:A} and Lemma \ref{lem:reminder}. 
In particular, in the decomposition (\ref{Feb13:equ:07}) above, the first term on the right-hand side is the signal part and the remaining four terms are the noise parts. By invoking Lemma \ref{lem:H1:reminder} in Section \ref{new.sec.B.3}, we see that all remaining terms are bounded by $O_p(\rho_n \sqrt{\log(en)} \sqrt{|\tilde{\calS}|\vee s^*} /T)$. Then combining these results yields that 
\begin{equation}\label{equ:may12:01}
\begin{aligned}
   \hA^{(\tau)}_{\tilde{\calS}} 
   &= \sum_{(i,j) \in \tilde{\calS}} \delta_{i,j}^2  + O_p\left(\frac{\rho_n \sqrt{\log(en)} s^*}{T}\right),
   \end{aligned}
\end{equation}
since $\Pr_1\left\{ \tilde{\calS} \subset \calS^*\right\} = \Pr_1\left\{\tilde{\calS} \cap (\calS^*)^c = \emptyset \right\} \to 1$ by Lemma \ref{lem:selection} and $|\calS^*| = (s^*)^2$. 

We are now ready to establish the power of the empirical MOSAIC test statistic. In light of condition \eqref{July:24:omega} and Lemma \ref{lem:svd_var}, we can deduce that 
\begin{equation}\label{march19:equ1}
 \begin{aligned}
     & \Pr_1\left\{\hat{A} \ge z_{\alpha/(2|\calT|)} \right\} \ge  \Pr_1\left\{ \frac{|\hA^{(\tau)}_{\hcalS}|}{\sqrt{\gamma_{\tau} \hsigma^2_{\hcalS}}} \ge z_{\alpha/(2|\calT|)} \right\}\\
     &\ge  \Pr_1\left\{ \frac{|\hA^{(\tau)}_{\hcalS}|}{\sqrt{\gamma_{\tau} \hsigma^2_{\calS^*}}} \ge z_{\alpha/(2|\calT|)} \right\}\\
     &\ge  \Pr_1\left\{ \frac{|\hA^{(\tau)}_{\hcalS}|}{\sqrt{\gamma_{\tau} \sigma^2_{\calS^*}}} \ge (1-\zeta_n)^{1/2}z_{\alpha/(2|\calT|)} \right\} - o(1)\\
     &\ge  \Pr_1\left\{ \hA^{(\tau)}_{\hcalS} \ge (1-\zeta_n)^{1/2}z_{\alpha/(2|\calT|)} \sqrt{\gamma_{\tau} \sigma^2_{\calS^*}} \right\}  - o(1)\\
     &\ge \Pr_1\left\{ \hA^{(\tau)}_{\tilde{\calS}} \ge (1-\zeta_n)^{1/2}z_{\alpha/(2|\calT|)} \sqrt{\gamma_{\tau} \sigma^2_{\calS^*}} +  O\left(  \frac{|\calS_3| \rho_n \log(n)}{T} \right) \right\}  - o(1),
\end{aligned}   
\end{equation}
where we have used $\Pr_1\left\{ \hsigma^2_{\hcalS} \le \hsigma^2_{\calS^*}\right\} \to1$ as $\Pr_1\left\{ \hcalS \subset  \calS^* \right\} \to 1$ by Lemma \ref{lem:selection} in the second last step above, Lemma \ref{lem:svd_var} in the third step above, and \eqref{equ:may12:06} in the last step above. 


By the definitions of $\calS_1$, $\calS_2$, $\calS_3$, and $s^* \ll n$, it holds that for some constant $c_1 > 0$, 
\begin{equation}\label{equ:may12:02}
 \begin{aligned}
 \sum_{(i,j) \in \calS_1} \delta_{i,j}^2 & \ge  \sum_{(i,j) \in \calS_1} \delta_{i,j}^2 =\sum_{(i,j) \in \calS^*} \delta_{i,j}^2 -  \sum_{(i,j) \in \calS_2 \cup \calS_3} \delta_{i,j}^2\\
  &\ge  \frac{\overline{\epsilon}^2_n}{T}  - c_1 \frac{\rho_n (s^*)^2 \log(en)}{Tn}\ge \frac{\overline{\epsilon}^2_n}{2T}  \\
  &\ge z_{\alpha/(2|\calT|)} T^{-1}s^* \rho_n \log(en).
\end{aligned}   
\end{equation}
Using (\ref{equ:may12:02}), \eqref{equ:may12:01}, and $\gamma_{\tau} = O(T^{-2})$,
we can deduce that 
$$
\begin{aligned}
 \eqref{march19:equ1} &\ge \Pr_1\left\{  \sum_{(i,j) \in \tilde{\calS}} \delta_{i,j}^2 \ge (1-\zeta_n)^{1/2}z_{\alpha/(2|\calT|)} \sqrt{\gamma_{\tau} \sigma^2_{\calS^*}} \right. \\
 &\quad \left. + O\left(\frac{\rho_n (|\calS_3|\log(n) + s^*\sqrt{\log(en)})}{T} \right) \right\} -o(1)\\
&\ge \Pr_1\left\{\sum_{(i,j) \in \calS_1} \delta_{i,j}^2 \ge o\left(T^{-1} s^* \rho_n \log(en)\right)\right\} -o(1) \to 1,
\end{aligned}
$$
since Condition \ref{con:modset} ensures that  $|\calS_3|\ll s^*$. Hence, we obtain the desired conclusion for the case when $s^* \ll n$.

It remains to examine the case when $s^* \asymp n$. By similar arguments as for \eqref{Feb13:equ:07} and \eqref{march19:equ1}, we can show that 
\begin{equation*}
 \begin{aligned}
     &\Pr_1\left\{\hat{A} \ge z_{\alpha/(2|\calT|)} \right\} 
     \ge \Pr_1\left\{ \frac{\hA_{\Omega}}{\sqrt{\gamma_{\tau} \sigma^2_{\Omega}}} \ge (1-\zeta_n)^{1/2}z_{\alpha/(2|\calT|)} \right\}\\
     &\ge \Pr_1\left\{ {\sum_{(i,j) \in \Omega} \delta_{i,j}^2} \ge (1-\zeta_n)^{1/2}z_{\alpha/(2|\calT|)}\sqrt{\gamma_{\tau} \sigma^2_{\Omega}} + o\left({T^{-1}\rho_n n \log(en)}\right)\right\} -o(1)\\
&\ge  \Pr_1\left\{ \overline{\epsilon}_n^2 \gtrsim  o\left(\rho_n n\log(en)\right)\right\} - o(1) \to 1.
\end{aligned}   
\end{equation*}
This concludes the proof of Theorem \ref{Thm:power}.

\subsection{Lemmas \ref{lemma:rho}--\ref{lem:bridge} and their proofs} \label{new.sec.B.3}

The lemma below establishes the consistency of estimate $\hat{\rho}_n$ for $\rho_n$.

\begin{lemma}
	\label{lemma:rho}
	Under Condition \ref{con:upper} and the time span condition $T q_n^2 \gg \log(n)$, for any small constant $0 < \eta < 1$ we have that 
    \begin{itemize}
        \item[(1)] Under $H_0(\rho_n, K^*)$,
     $$
	\Pr_{0}\left\{ \left|\frac{\rho_n}{\hat{\rho}_n} - 1\right| \ge \eta \right\} \to 0.
     $$
     \item[(2)] Under $H_1(\rho_n, K^*, s^*,  \bar{\epsilon}_n)$, 
    $$
   \Pr_{1}\left\{ \left|\frac{\rho_n}{\hat{\rho}_n} - 1\right| \ge \eta \right\} \to 0.
    $$
    \end{itemize}  
\end{lemma}

\noindent\textit{Proof}. We focus on proving the result under $H_1(\rho_n, K^*, s^*,  \bar{\epsilon}_n)$. Since the proof for $H_0(\rho_n, K^*)$ is nearly identical, we will omit the details here. Denote by $\bb^{(l)}_j = \sum_{k=1}^K \frac{d_k^{(l)}}{t_k^{(l)}}\bv^{(l)}_k v^{(l)}_k(j), j \in [n]$ and $\tilde{\bW}^{(hT,l)}$ as defined in \eqref{def:tbW_lr} with $\tau = hT$. Let us define an event
   \begin{equation}\label{event:F_3}
   \begin{aligned} 
       F_3  & :=\left\{\htheta^{(hT,l)}_{i,j} = \theta_{l,i,j} +\frac{q_n}{\sqrt{hT}} \left(\be_i^\top \tbW^{(hT,l)} \bb_j^{(l)} + \be_j^\top \tbW^{(hT,l)} \bb_i^{(l)}\right) \right.\\
       &\quad \left. + { O}\left(\frac{q_n}{n\sqrt{T}}\right); (i,j) \in \Omega, l=1,2 \right\}.
    \end{aligned} 
   \end{equation}
    From Lemma \ref{lem:gen1}, it holds that 
    \begin{equation}\label{equ:july:01}
       \Pr_1\left\{F_3^c\right\}=1- \Pr_1\left\{F_3\right\} \to 0.
    \end{equation}
    Then applying  the union bound gives that for any $\eta > 0$,  
    \begin{equation}\label{Feb13:equ:01}
      \Pr_1\left\{ \left|\frac{\rho_n}{\hat{\rho}_n} - 1\right| \ge \eta\right\}  \le \Pr_1\left\{ \frac{\rho_n}{\hat{\rho}_n} \ge 1+ \eta ,  F_3\right\} + \Pr_1\left\{ \frac{\rho_n}{\hat{\rho}_n} \le 1- \eta ,  F_3\right\} + \Pr_1\left\{F_3^c\right\}.
    \end{equation} 
We will bound the two terms on the right-hand side of \eqref{Feb13:equ:01} above separately. For the first term, let us define $$(l^*, i^*, j^*) = \arg\max_{(l,i,j) \in [2] \times [n] \times [n]} \theta_{l,i,j}.$$ 
We can deduce that 
\begin{align*}
    \Pr_1\left\{ \frac{\rho_n}{\hat{\rho}_n} \ge 1+ \eta, F_3\right\} =&	\Pr_1\left\{  \frac{\rho_n}{ \htheta^{(hT,l^*)}_{i^*,j^*}}\frac{ \htheta^{(hT,l^*)}_{i^*,j^*}}{\hat{\rho}_n} \ge 1 + \eta, F_3 \right\} \\
    \le & \Pr_1\left\{  \frac{\rho_n}{ \htheta^{(hT,l^*)}_{i^*,j^*}} \ge 1 + \eta,F_3 \right\}\\
    = & \Pr_1\left\{  \htheta^{(hT,l^*)}_{i^*,j^*} -\rho_n \le \frac{-\eta}{1 + \eta}\rho_n, F_3\right\},
\end{align*}
where the inequality above is due to  $\htheta^{(hT,l^*)}_{i^*,j^*}/{\hat{\rho}_n} \le 1$ and ${\rho_n},{\htheta_{l^*,i^*,j^*}} \ge 0$.

Further, under event $F_3$, we have that 
\begin{equation}\label{equ:lem6:june24:01}
 \begin{aligned}
 &\Pr_1\left\{   \htheta^{(hT,l^*)}_{i^*,j^*} -\rho_n \le \frac{-\eta}{1 + \eta}\rho_n, F_3 \right\}\\
 &= \Pr_1\left\{   \htheta^{(hT,l^*)}_{i^*,j^*} - \theta_{l^*,i^*,j^*}\le \frac{-\eta}{1 + \eta}\rho_n, F_3 \right\} \\
 &\le  \sum_{k=1}^K \left\{ \Pr\left\{  \left|n\be_i^\top \tilde{\bW}^{(hT,l^*)} \bb^{(l^*)}_j \right|\ge \frac{\eta}{2(1 + \eta)} \frac{\sqrt{T}n\rho_n}{q_n} \right\}\right.\\
 &\quad\left.+ \Pr\left\{  \left|n\be_j^\top \tilde{\bW}^{(hT,l^*)} \bb^{(l^*)}_i\right| \ge \frac{\eta}{2(1 + \eta)}\frac{\sqrt{T}n\rho_n}{q_n}\right\}\right\}\\
 &\le 2K \exp\left\{- c_{\eta} T q^2_n \right\}
\end{aligned}   
\end{equation}
for some constant $c_{\eta} > 0$. Here, the last inequality above has utilized \eqref{add:lem:equ1} and $\sqrt{T} n \rho_n/q_n = \sqrt{T}q_n \gg 1$. 

For the second term on the right-hand side of \eqref{Feb13:equ:01}, we can show that 
\begin{align*}
    \Pr_1\left\{ \frac{\rho_n}{\hat{\rho}_n} \le 1- \eta, F_3\right\} =& \Pr_1\left\{ \max_{l,i,j \in [2]\times[n]\times[n]} \frac{ \htheta^{(hT,l)}_{i,j}}{\rho_n} \ge \frac{1}{1-\eta}, F_3\right\}\\
    \le& \sum_{l,i,j \in [2]\times[n]\times[n]} \Pr_1\left\{  \htheta^{(hT,l)}_{i,j} - \theta_{l,i,j} \ge \frac{\rho_n}{1-\eta} -\theta_{k,i,j}, F_3\right\} \\
    \le& \sum_{l,i,j \in [2]\times[n]\times[n]} \Pr_1\left\{   \htheta^{(hT,l)}_{i,j} - \theta_{l,i,j} \ge \frac{\eta \rho_n}{1-\eta}, F_3\right\} \\
    \le& 2n^2 \exp\left\{- c_{\eta} T q_n^2 \right\}, 
\end{align*}
where the first inequality above is from the union bound, the second inequality above is due to $\theta_{k,i,j} \le \rho_n$ for all $(l,i,j) \in [2]\times[n]\times[n]$, and the third inequality above has used similar deduction as for \eqref{equ:lem6:june24:01}. Therefore, we can obtain that 
$$
\begin{aligned}
\Pr_1\left\{ \left|\frac{\rho_n}{\hat{\rho}_n} - 1\right| \ge \eta \right\} \le& \Pr_1\left\{ \left|\frac{\rho_n}{\hat{\rho}_n} - 1\right| \ge \eta, F_3 \right\} + \Pr_1\left\{F_3^c \right\}\\
\le& \exp\left\{- 2c_{\eta} T q_n^2 + 2\log(n)\right\} = o(1),    
\end{aligned}
$$
where we have used $T q_n^2  \gg \log(n)$, and \eqref{equ:july:01}. This completes the proof of Lemma \ref{lemma:rho}.

The lemma below establishes the consistency of estimate $\hsigma_{\calS}^2$. Observe that $$\sigma_{\calS}^2 = \sum_{(i,j) \in \calS} \theta^2_{i,j}$$ under null $H_0(\rho_n, K^*)$, and $$\sigma_{\calS}^2 = \sum_{(i,j) \in \calS} (\theta^2_{1,i,j}+ \theta^2_{2,i,j})/2$$ under alternative $H_1(\rho_n, K^*, s^*,  \bar{\epsilon}_n)$. Denote by $\bar{\theta}^2_{i,j} = (\theta^2_{1,i,j} + \theta^2_{2,i,j})/2$.

\begin{lemma}
	\label{lem:svd_var}
   Under Condition \ref{con:upper}, the time span condition $T q_n^2 \gg \log(n)$, and
    \begin{equation}\label{Jan5:1}
          \rho_n \log^{2}(n)\max\left\{|\calS|, n\log(n) \right\} \ll  nT \sigma^2_{\calS},
    \end{equation}
    we have that 
\begin{itemize}
    \item[(1)] Under $H_0(\rho_n, K^*)$, 
    $$
    \Pr\left\{\left|\hat{\sigma}^2_{\calS}/\sigma^2_{\calS}-1\right|\ge \log^{-1}(n) \right\} \to 0.
    $$
    \item[(2)] Under $H_1(\rho_n, K^*, s^*, \bar{\epsilon}_n)$, 
    $$
    \Pr\left\{\left|\hat{\sigma}^2_{\calS}/\sigma^2_{\calS}-1\right|\ge \log^{-1}(n) \right\} \to 0.
    $$
\end{itemize}
\end{lemma}

\noindent\textit{Proof}. The proofs under $H_0(\rho_n, K^*)$ and $H_1(\rho_n, K^*, s^*,  \bar{\epsilon}_n)$ are almost identical so we will present only the proof under $H_1(\rho_n, K^*, s^*, \bar{\epsilon}_n)$ here.
    Under $H_1(\rho_n, K^*, s^*,  \bar{\epsilon}_n)$ and Condition \ref{con:upper}, we see from Lemma \ref{lem:gen1} in Section \ref{new.sec.D} that event $F_3$ defined in \eqref{event:F_3} holds with probability one.
    Under event $F_3$, we can deduce that 
	$$
	\begin{aligned}
		(\htheta_{i,j}^{(hT,l)})^2 =& (\theta_{l,i,j})^2 + 2 {O}\left(\frac{q_n}{n\sqrt{T}}\right)\theta_{l,i,j} + \frac{2q_n}{\sqrt{hT}} \theta_{l,i,j} \left( \be_i^\top \tilde{\bW}^{(hT,l)} \bb^{(l)}_j +   \be_j^\top \tilde{\bW}^{(hT,l)} \bb^{(l)}_i\right)\\
		&+ {O}\left(\frac{q_n^2}{n^2T}\right) + \left(  \frac{q_n}{\sqrt{hT}} \left( \be_i^\top \tilde{\bW}^{(hT,l)} \bb^{(l)}_j +   \be_j^\top \tilde{\bW}^{(hT,l)} \bb^{(l)}_i\right) \right)^2 \\
		&+ 2 {O}\left(\frac{q^2_n}{nT \sqrt{h}}\right)  \left( \be_i^\top \tilde{\bW}^{(hT,l)} \bb^{(l)}_j +   \be_j^\top \tilde{\bW}^{(hT,l)} \bb^{(l)}_i\right). 
	\end{aligned}
	$$
	Consequently, under event $F_3$, we can decompose the variance estimator as
	\begin{equation}
		\label{varsvd:1}
			\begin{aligned}
			&\hsigma^2_{\calS} = \frac{1}{2}\sum_{(i,j) \in \calS}\sum_{l=1}^2(\htheta_{i,j}^{(hT,l)})^2 \\
            &= \sigma_{\calS}^2 +  {O}\left(\frac{q_n}{n\sqrt{T}}\right) \sum_{(i,j) \in \calS} \sum_{l=1}^2 \theta_{l,i,j} \\
            &\quad+ \frac{q_n}{\sqrt{hT}} \sum_{(i,j) \in \calS} \sum_{l=1}^2 \theta_{l,i,j} \left( \be_i^\top \tilde{\bW}^{(hT,l)} \bb^{(l)}_j +   \be_j^\top \tilde{\bW}^{(hT,l)} \bb^{(l)}_i\right)\\
		&\quad+ {O}\left(\frac{|\calS|q_n^2}{n^2T}\right) + \sum_{(i,j) \in \calS}\sum_{l=1}^2 \left(  \frac{q_n}{\sqrt{hT}} \left( \be_i^\top \tilde{\bW}^{(hT,l)} \bb^{(l)}_j +   \be_j^\top \tilde{\bW}^{(hT,l)} \bb^{(l)}_i\right) \right)^2 \\
		&\quad+  { O}\left(\frac{q_n^2}{nT \sqrt{h}}\right)  \sum_{(i,j) \in \calS} \sum_{l=1}^2\left( \be_i^\top \tilde{\bW}^{(hT,l)} \bb^{(l)}_j +   \be_j^\top \tilde{\bW}^{(hT,l)} \bb^{(l)}_i\right)\\
        &:= \sigma_{\calS}^2 + { O}(\log^{-1}(n)\sigma_{\calS}^2) + \sum_{l=1}^2 v^{(l)}_1 + o(\log^{-1}(n)\sigma_{\calS}^2) +\sum_{l=1} v^{(l)}_2+ \sum_{l=1}v^{(l)}_3\\
			&:= \sigma_{\calS}^2 + { O}(\log^{-1}(n)\sigma_{\calS}^2) + v_1 + { O}(\log^{-1}(n)\sigma_{\calS}^2) + v_2+v_3.
		\end{aligned}
	\end{equation}
	
	Note that the second last bound in (\ref{varsvd:1}) above is due to the following facts. First, it follows from the Cauchy--Schwarz inequality and condition \eqref{Jan5:1} that 
	$$
	{O}\left(\frac{q_n}{n\sqrt{T}}\right) \sum_{(i,j) \in \calS}\theta_{l,i,j} =  { O}\left(\frac{q_n|\calS|^{1/2}}{nT^{1/2}}\right) \sqrt{\sum_{(i,j) \in \calS}(\theta_{l,i,j})^2}  = { O}\left(\log^{-1}(n)\sigma_{\calS}^2\right).
	$$
   Second, \eqref{Jan5:1} directly implies that  $$|\calS|q_n^2/(n^2T) = o(\log^{-1}(n)\sigma_{\calS}^2).$$ 
   We make a claim that for all $r \in [3]$,
    \begin{equation}\label{Feb26:equ01}
        \Pr\left\{\frac{|v_r|}{\sigma_{\calS}^2} \gtrsim \frac{1}{\log(n)} \right\} = O\left( \frac{1}{\log(n)}\right).
    \end{equation}
    Then it holds that 
    $$
    \begin{aligned}
    \Pr\left\{ \left|\frac{\hsigma^2_{\calS}}{ \sigma^2_{\calS}} - 1 \right| \gtrsim \frac{1}{\log(n)}\right\}  \le&  \Pr\left\{ \left|\frac{\hsigma^2_{\calS}}{ \sigma^2_{\calS}} - 1 \right| \gtrsim \frac{1}{\log(n)},F_3\right\} + \Pr\left\{F^c_3\right\}  \\
    \le&\sum_{r=1}^3 \Pr\left\{\frac{|v_r|}{\sigma_{\calS}^2} \gtrsim \frac{1}{\log(n)} \right\} + o(1) =o(1),
    \end{aligned}
    $$
    which yields the desired conclusion.
    
    It remains to prove claim  \eqref{Feb26:equ01} above. We begin with investigating term $v_1$. In view of part 1) of Lemma \ref{lemma:moment_arbitrary_S} in Section \ref{new.sec.D}, we can show that 
    $$
    \Var(v^{(l)}_1) = O\left(\frac{q_n^2}{n T} \sum_{(i,j)\in \calS} \theta^2_{l,i,j}  \right) = O\left(\frac{\rho_n}{T} \sigma^2_{\calS} \right).
    $$
 Then it follows that 
 $$
 \begin{aligned}
  \Pr\left\{v_1 \gtrsim \log^{-1}(n)\sigma_{\calS}^2 \right\} \le& 2\Pr\left\{v^{(l)}_1 \gtrsim \log^{-1}(n)\sigma_{\calS}^2 \right\} \\
  \le& 2\Pr\left\{|v^{(l)}_1 - \E(v^{(l)}_1)| \gtrsim \log^{-1}(n)\sigma_{\calS}^2 \right\}\\
  \le& \frac{2\log^2(n)}{ (\sigma^2_{\calS})^2} \cdot \frac{\rho_n}{T} \sigma^2_{\calS} \\
  \lesssim& \frac{\rho_n}{T \sigma^2_{\calS}} \log^2(n) = o(\log^{-1}(n)),   
 \end{aligned}
 $$
  where the last bound above is from \eqref{Jan5:1}.
We next turn to term $v_2$. Observe that $T q_n^2 \gg \log(n)$ and 
\eqref{add:lem:equ1} lead to 
\begin{equation}\label{Feb26:equ02}
 \Pr\left\{\max_{1 \le i < j  \le n} |\be_i^\top \tilde{\bW}^{(hT,l)} \bb^{(l)}_j| \gtrsim \frac{  \log(n)}{n}\right\} = O\left( \log^{-1}(n) \right).
\end{equation}
This fact together with \eqref{Jan5:1} yields that 
$$
\begin{aligned}
\Pr\left\{v_2 \ge \log^{-1}(n)\sigma_{\calS}^2\right\}
    \le& \sum_{l=1}^2\Pr\left\{v^{(l)}_2 \ge \frac{\rho_n |\calS|\log^{2}(n)}{T n}\right\}\\
\le&\sum_{l=1}^2\Pr\left\{\sum_{(i,j) \in \calS} {(\be_i^\top \tilde{\bW}^{(hT,l)} \bb^{(l)}_j +\be_j^\top \tilde{\bW}^{(hT,l)} \bb^{(l)}_i)^2} \ge \frac{|\calS| \log^{2}(n)}{n^2}  \right\}\\
\le&\sum_{l=1}^2 \Pr\left\{\max_{1 \le i < j  \le n} |\be_i^\top \tilde{\bW}^{(hT,l)} \bb^{(l)}_j| \ge \frac{  \log(n)}{n}\right\} =O\left( \log^{-1}(n) \right),
\end{aligned}
$$
where the last inequality  above follows from \eqref{Feb26:equ02} and the worst case of $|\calS|=1$.  

Finally, for term $v_3$, we can obtain from part 1) of Lemma \ref{lemma:moment_arbitrary_S} that 
	$$
    \begin{aligned}
	\sqrt{\Var(v^{(l)}_3)} & = {O}\left(\frac{q_n^2}{Tn}\right)\sum_{l=1}^2\sqrt{ \Var\left\{ \sum_{(i,j) \in \calS}  \be_i^\top \tilde{\bW}^{(hT,l)} \bb^{(1)}_j \right\}}
    = {O}\left(\frac{\rho_n}{T} \cdot \sqrt{\frac{|\calS|}{n}}\right) ,
    \end{aligned}
	$$
where the last bound above is due to \eqref{Jan5:1}.  Therefore, it holds that 
$$
\Pr\left\{ v_3 \gtrsim \log^{-1}(n) \sigma_{\calS}^2\right\} \le \sum_{l=1}^2\Pr\left\{ v^{(l)}_3 \gtrsim \log^{-1}(n) \sigma_{\calS}^2\right\} = O\left( \log^{-1}(n) \right).
$$
This concludes the proof of Lemma \ref{lem:svd_var}.

The lemma below gives a uniform upper bound for the remaining terms in \eqref{Feb13:equ:A}.

\begin{lemma}\label{lem:reminder}
Under Condition \ref{con:upper}, we have 
$$\Var(\nu^{(\tau)}_l) = O(\rho_n^2|\calS|/(T^2 n))$$ 
for $l =1,\ldots, 4$, where $\nu^{(\tau)}_l$'s with $ l \in [4]$ are defined in \eqref{Feb13:equ:A}. We emphasize that the results hold under both the null hypothesis $H_0(\rho_n, K^*)$ and the alternative hypothesis $H_1(\rho_n, K^*, s^*, \bar{\epsilon}_n)$.
\end{lemma}

\noindent\textit{Proof}. In this proof, we set $\bb_j^{(2)} = \bb_j$ under null hypothesis $H_0(\rho_n, K^*)$.
 We start with analyzing $\nu_1^{(\tau)} = \nu_{11}^{(\tau)} + \nu_{12}^{(\tau)}$, where $\nu_{11}^{(\tau)}$ and $ \nu_{12}^{(\tau)}$ are defined in \eqref{Feb13:equ:A}. Clearly, the variance of $\nu_{12}^{(\tau)}$ is equal to that of $\nu_{11}^{(\tau)}$ by symmetry. Thus, we need only to bound $\Var(\nu_{11}^{(\tau)})$. Denote by $$\calF:=\sigma\left( \{\bar{w}_{i,j}^{(\tau,1)}, (i,j), t \in [T]\}\right)$$ the $\sigma$-algebra generated by random variables $\bar{w}_{i,j}^{(\tau,1)}$.  Using the law of total variance, it holds that 
$$
\begin{aligned}
\Var(\nu_{11}^{(\tau)}) & = \E(\Var(\nu_{11}^{(\tau)} \mid \calF)) + \Var\left( \E(\nu_{11}^{(\tau)} \mid \calF )\right)  \\
&= \E(\Var(\nu_{11}^{(\tau)} \mid \calF)),
\end{aligned}
$$
since $\E(\nu_{11}^{(\tau)} \mid \calF ) = 0$. From the definitions in \eqref{march15:equ01}, $K = O(1)$, and the Cauchy--Schwarz inequality, we can bound the conditional variance as 
\begin{equation}\label{march15:equ02}
\begin{aligned}
   \E\left\{\Var(\nu_{11}^{(\tau)} \mid \calF)\right\} & \lesssim \frac{q^2_n}{\tau}  \E\left\{ \Var\left(\sum_{(i,j) \in \calS}  \bar{w}^{(\tau,1)}_{i,j}\be_i^\top \tdbW^{(\tau,2)} \bb^{(2)}_j\mid \calF\right) \right\} \\
   &\quad + \frac{q^2_n}{\tau}\E\left\{\Var\left(\sum_{(i,j) \in \calS}  \bar{w}^{(\tau,1)}_{i,j}\be_j^\top \tdbW^{(\tau,2)} \bb^{(2)}_i\mid \calF\right)\right\}.   
\end{aligned}
\end{equation}

For the first term on the right-hand side of (\ref{march15:equ02}) above, we can further decompose the conditional variance as
	\begin{equation}\label{march15:equ03}
	   \begin{aligned}
	&\E\left\{\Var\left(\sum_{(i,j) \in \calS}  \bar{w}^{(\tau,1)}_{i,j}\be_i^\top \tdbW^{(\tau,2)} \bb^{(2)}_j\mid \calF\right)\right\} \\
		=&E\left\{ \sum_{(i_1,j_1) \in \calS}  \sum_{(i_1,j_2) \in \calS}  \bar{w}^{(\tau,1)}_{i_1, j_1} \bar{w}^{(\tau,1)}_{i_1, j_2}\Cov \left(  \be_{i_1}^\top \tdbW^{(\tau,2)} \bb^{(2)}_{j_1},  \be_{i_1}^\top \tdbW^{(\tau,2)} \bb^{(2)}_{j_2}\right) \right\}\\
		&+\E\left\{ \sum_{(i_1,j_1)\in \calS} \sum_{(i_2,j_2) \in \calS; i_1 \ne i_2} \bar{w}^{(\tau,1)}_{i_1, j_1} \bar{w}^{(\tau,1)}_{i_2, j_2} \Cov \left(  \be_{i_1}^\top \tdbW^{(\tau,2)} \bb^{(2)}_{j_1},  \be_{i_2}^\top \tdbW^{(\tau,2)} \bb^{(2)}_{j_2}\right)\right\}\\ :=& \E(z_1)+ \E(z_2).
	\end{aligned} 
	\end{equation}
	Denote by $b^{(2)}_j(l)$ the $l$th component of $\bb^{(2)}_{j}$. To analyze term $\E(z_1)$ on the right-hand side of (\ref{march15:equ03}) above, let us first calculate the covariance term 
\begin{equation*}
	\begin{aligned}
    & \Cov \left\{  \be_{i_1}^\top \tdbW^{(\tau,2)} \bb^{(2)}_{j_1},  \be_{i_1}^\top \tdbW^{(\tau,2)} \bb^{(2)}_{j_2}\right\} =\E\left( \be_{i_1}^\top \tilde{\dot{\bW}}^{(\tau,2)} \bb^{(2)}_{j_1}\be_{i_1}^\top \tilde{\dot{\bW}}^{(\tau,2)} \bb^{(2)}_{j_2}\right) \\
    &= \E\left( \left(\sum_{l \in [n]}\tilde{\dot{w}}^{(\tau,2)}_{i_1,l} b^{(2)}_{j_1}(l)  \right) \left(\sum_{l \in [n]}\tilde{\dot{w}}^{(\tau,2)}_{i_1,l} b^{(2)}_{j_2}(l) \right)\right)\\
	&=\E\left( \sum_{l \in [n]} (\tilde{\dot{w}}^{(\tau,2)}_{i_1,l})^2 b^{(2)}_{j_1}(l)b^{(2)}_{j_2}(l)\right).
		\end{aligned}
\end{equation*}
Then it holds that 
\begin{equation*}
	\begin{aligned}
		z_1 = \sum_{(i_1, j_1) \in \calS} \sum_{(i_1, j_2) \in \calS} \left\{  \bar{w}^{(\tau,1)}_{i_1, j_1}\bar{w}^{(\tau,1)}_{i_1, j_2} \E\left( \sum_{l \in [n]} (\tilde{\dot{w}}^{(\tau,2)}_{i_1,l})^2 b^{(2)}_{j_1}(l)b^{(2)}_{j_2}(l)\right) \right\}.
	\end{aligned}
\end{equation*}
Using the independence of the elements $\bar{w}_{i,j}^{(\tau,1)}$'s, we can deduce that 
\begin{equation}\label{equ:Jan6:1}
	\begin{aligned}
\E(z_1) =&  \sum_{(i_1,j_1) \in \calS} \E(\bar{w}^{(\tau,1)}_{i_1, j_1})^2\E\left( \sum_{l \in [n]} (\tilde{\dot{w}}^{(\tau,2)}_{i_1,l})^2 (b^{(2)}_{j_1}(l))^2\right) \\
=&\frac{1}{n^2} \sum_{(i_1,j_1) \in \calS} \E(\bar{w}^{(\tau,1)}_{i_1, j_1})^2 \le \frac{|\calS| \rho_n}{T n^2},
	\end{aligned}
\end{equation}
where the second last step above has used $\|\bb^{(2)}_j\|_{\infty} \lesssim 1/n, j \in [n]$ and $\Var(\tilde{\dot{w}}^{(\tau,2)}_{i,j}) \lesssim 1/n$ for $(i,j) \in \Omega$, and the last step above follows from  $\E(\bar{w}^{(\tau,1)}_{i, j})^2 \le \rho_n/T$.

To examine term $\E(z_2)$, we calculate the covariance term
	\begin{equation*}
		\begin{aligned}
        & \Cov \left\{  \be_{i_1}^\top \tdbW^{(\tau,2)} \bb^{(2)}_{j_1},  \be_{i_2}^\top \tdbW^{(\tau,2)} \bb^{(2)}_{j_2}\right\} = \E\left( \be_{i_1}^\top \tilde{\dot{\bW}}^{(\tau,2)} \bb^{(2)}_{j_1}\be_{i_2}^\top \tilde{\dot{\bW}}^{(\tau,2)} \bb^{(2)}_{j_2}\right) \\
        &= \E\left( \left(\sum_{l \in [n]}\tilde{\dot{w}}^{(\tau,2)}_{i_1,l} b^{(2)}_{j_1}(l)  \right) \left(\sum_{l \in [n]}\tilde{\dot{w}}^{(\tau,2)}_{i_2,l} b^{(2)}_{j_2}(l) \right)\right)\\
			&=\E\left(  (\tilde{\dot{w}}^{(\tau,2)}_{i_1,i_2})^2 b^{(2)}_{j_1}(i_1)b^{(2)}_{j_2}(i_2)\right)
		\end{aligned}
	\end{equation*}
	for $i_1 \ne i_2$. 
Then it follows that 
	 \begin{equation*}
	 	z_2 =  \sum_{(i_1,j_1)\in \calS}  \sum_{(i_2,j_2) \in  \calS; i_1 \ne i_2} \left\{\bar{w}^{(\tau,1)}_{i_1, j_1} \bar{w}^{(\tau,1)}_{i_2, j_2} \E\left(  (\tilde{\dot{w}}^{(\tau,2)}_{i_1,i_2})^2 b^{(2)}_{j_1}(i_1)b^{(2)}_{j_2}(i_2)\right) \right\}.
	 \end{equation*}
Hence, in view of the independence of the elements $\bar{w}_{i,j}^{(\tau,1)}$'s, we can obtain that 
\begin{equation}\label{equ:Jan6:2}
	\begin{aligned}
\E(z_2) \le &  \sum_{(i_1,j_1) \in \calS} \E(\bar{w}^{(\tau,1)}_{i_1, j_1})^2\E\left(  (\tilde{\dot{w}}^{(\tau,2)}_{i_1,i_1})^2 (b^{(2)}_{j_1}(i_1))^2\right) \\
=&\frac{1}{n^3} \sum_{(i_1,j_1) \in \calS} \E(\bar{w}^{(\tau,1)}_{i_1, j_1})^2 = o\left( \frac{|\calS| \rho_n}{T n^2}\right),
	\end{aligned}
\end{equation}
where the last step above has used $\|\bb^{(2)}_j\|_{\infty} \lesssim 1/n, j \in [n]$ and $\Var(\tilde{\dot{w}}^{(\tau,2)}_{i,j}) \lesssim 1/n$ for $(i,j) \in \Omega$.  

Combining \eqref{march15:equ03}, \eqref{equ:Jan6:1}, and \eqref{equ:Jan6:2} above leads to 
	\begin{equation*}
    \begin{aligned}
        		\E\left\{\Var\left(\sum_{(i,j) \in \calS}  \bar{w}^{(\tau,1)}_{i,j}\be_i^\top \tilde{\bW}^{(\tau,2)} \bb^{(2)}_j \mid \calF\right) \right\}  = O\left( \frac{|\calS| \rho_n}{T n^2}\right).
                    \end{aligned}
	\end{equation*}
Similar derivations yield that 
$$
\E\left\{\Var\left(\sum_{(i,j) \in \calS}  \bar{w}^{(\tau,1)}_{i,j}\be_j^\top \tilde{\bW}^{(\tau,2)} \bb^{(2)}_i \mid \calF\right) \right\} = O\left( \frac{|\calS| \rho_n}{T n^2}\right).
$$
Then it holds that 
$$
\begin{aligned}
\Var(v_{11}^{(\tau)}) =& O\left(\frac{q^2_n}{T}\right)\E\left(\Var\left\{\sum_{(i,j) \in \calS}  \bar{w}_{i,j}^{(\tau,1)}\be_i^\top \tilde{\dot{\bW}}^{(\tau,2)} \bb^{(2)}_j \mid \calF\right\}\right)\\
=& O\left(  \frac{q_n^2}{T n^2} \cdot \frac{|\calS| \rho_n}{T n^2} \right)= O\left( \frac{\rho_n^2|\calS|}{n T^2}\right),
\end{aligned}
$$
since $q_n = \sqrt{n \rho_n}$. Thus, we have that 
\begin{equation}\label{equ:Jan6:3}
 \Var(\nu_1^{(\tau)}) = O\left(\frac{\rho_n^2 |\calS|}{T^2n}\right). 
\end{equation}

We next turn to term $\nu_2^{(\tau)}$. It can be seen that 
$$
 \Var\left(  \sum_{(i,j) \in \calS} (\bar{w}_{i,j}^{(\tau,1)} + \bar{\dot{w}}_{i,j}^{(\tau,1)})\right) = O\left( \frac{|\calS| \rho_n}{T}\right)
$$
and thus
\begin{equation}\label{equ:Jan6:4}
   \Var(\nu_2^{(\tau)}) = o\left(\frac{q^2_n}{n^2 T} \cdot \frac{|\calS| \rho_n}{T}\right)= o\left(\frac{\rho_n^2 |\calS|}{T^2 n}\right).
\end{equation}
We now focus on term $\nu^{(\tau)}_3$. An application of part 2) of Lemma \ref{lemma:moment_arbitrary_S} leads to 
$$
\Var\left(\sum_{(i,j) \in \calS} \be_i^\top \tilde{\bW}^{(\tau,2)} \bb^{(2)}_j \be_i^\top \tilde{\dot{\bW}}^{(\tau,2)} \bb^{(2)}_j\right) = O\left(\frac{|\calS|}{n^3}\right),
$$
and the same bound can also be derived for 
$$\Var\left(\sum_{(i,j) \in \calS} \be_i^\top \tilde{\bW}^{(\tau,2)} \bb^{(2)}_j \be_j^\top \tilde{\dot{\bW}}^{(\tau,2)} \bb^{(2)}_i\right),$$ $$\Var\left(\sum_{(i,j) \in \calS} \be_j^\top \tilde{\bW}^{(\tau,2)} \bb^{(2)}_i \be_i^\top \tilde{\dot{\bW}}^{(\tau,2)} \bb^{(2)}_j\right),$$ 
and $$\Var\left(\sum_{(i,j) \in \calS} \be_j^\top \tilde{\bW}^{(\tau,2)} \bb^{(2)}_i \be_j^\top \tilde{\dot{\bW}}^{(\tau,2)} \bb^{(2)}_i\right).$$ Then it follows that 
$$
\begin{aligned}
& \Var(\nu^{(\tau)}_3) = \Var(\sum_{(i,j) \in \calS}f_{i,j}^{(\tau)}\dot{f}_{i,j}^{(\tau)}) \\
&\lesssim   \frac{q_n^4}{ T^2}\Var\left(\sum_{(i,j) \in \calS} \be_i^\top \tilde{\bW}^{(\tau,2)} \bb^{(2)}_j \be_i^\top \tilde{\dot{\bW}}^{(\tau,2)} \bb^{(2)}_j\right) \\
&= O\left(\frac{\rho_n^2|\calS|}{T^2n}\right),
\end{aligned}
$$
where we have used the fact that $\Cov\left(X, Y\right) \le \sqrt{\Var(X) \Var(Y)} \le  \Var(X) + \Var(Y)$. Combining these results gives that 
$$
\Var(\nu_3^{(\tau)}) = O\left(\frac{ \rho_n^2 |\calS|}{T^2n}\right).
$$

Further, by resorting to part 1) of Lemma \ref{lemma:moment_arbitrary_S}, we can show that 
$$
\Var\left(\sum_{(i,j) \in \calS} \be_i^\top \tilde{\bW}^{(\tau,2)} \bb^{(2)}_j\right)  = O\left(\frac{|\calS|}{n}\right).
$$
This entails that 
$$
\begin{aligned}
\Var\left(\sum_{(i,j) \in \calS} f_{i,j}^{(\tau)}\right) & \lesssim  \frac{q_n^2}{T}\Var\left(\sum_{(i,j) \in \calS} \be_i^\top \tilde{\bW}^{(\tau,2)} \bb^{(2)}_j\right)  \\
&\quad+ \frac{q_n^2}{T}\Var\left(\sum_{(i,j) \in \calS} \be_i^\top \tilde{\bW}^{(\tau,2)} \bb^{(2)}_j\right) \\
&= O\left(\frac{q_n^2|\calS|}{Tn}\right).
\end{aligned}
$$
Therefore, we can obtain that 
$$
\Var(\nu_4^{(\tau)}) = o\left(\frac{q_n^2}{n^2 T}  \Var\left(\sum_{(i,j) \in \calS} f_{i,j}^{(\tau)}\right)\right)= o\left(\frac{\rho_n^2|\calS|}{T^2 n} \right).
$$
This completes the proof of Lemma \ref{lem:reminder}.


The lemma below provides the asymptotic distribution of the main term in \eqref{Feb13:equ:03}. Let us define $\bSigma_{\calS} :=  \diag\{\theta_{i,j}(1-\theta_{i,j}), (i,j) \in \calS\} \in \R^{|\calS| \times |\calS|}$.

\begin{lemma}\label{lem:asymptotic}
Assume that there exists a divergent sequence $\omega_n = O(T)$ such that 
    \begin{align}
        &\sum_{(i,j)\in \calS} \theta_{i,j}^2 = O\left( \omega_n^{-1} T^2 (\sum_{(i,j) \in \calS}\theta_{i,j}^2)^2 \right), \label{con:Feb21:01}\\
        &\sum_{(i,j) \in \calS}\theta_{i,j}^3 = O\left( \omega_n^{-1} T(\sum_{(i,j) \in \calS} \theta_{i,j}^2)^2 \right), \label{con:Feb21:02}\\  
        &\tr\left( \bSigma_{\calS}^4\right) = O\left( \omega_n^{-1} \left( \tr\left( \bSigma_{\calS}^2\right)\right)^2 \right). \label{con:Feb21:03}
    \end{align}
    Then $\Upsilon^{(\tau)}_n$ defined in \eqref{Feb13:equ:03} satisfies that  
    $$
\sup_{t \in \R}\left| \Pr\left\{ \frac{\Upsilon^{(\tau)}_n}{\sqrt{\gamma_{\tau} \sigma_{\calS}^2}} \le t \right\} - \Phi(t)\right| = O(\omega_n^{-1/5}),
$$
   where $\sigma_{\calS}^2 = \sum_{(i,j)\in\calS} f(\theta_{i,j})$ with $f(x) = x^2 - 2x^3 + x^4$ and $\gamma_{\tau} = \tau^{-2}$.
\end{lemma}

\noindent\textit{Proof}. To simplify the notation, let us redefine 
$$w^{(t)}_{i,j} := a^{(t)}_{i,j} \ \text{ and } \  \dot{w}^{(t)}_{i,j}:=b^{(t)}_{i,j}.$$ 
For $\tau \in \calT$, we introduce weights
\begin{equation}\label{def:weight}
    c_{\tau}^{(t)} := \begin{cases}
    \frac{1}{\tau} \ \text{ if } 1 \le t \le \tau,\\
    0  \ \text{ otherwise.}
\end{cases}
\end{equation}
Then we have 
$$\bar{w}_{i,j}^{(\tau,1)} = \sum_{t=1}^T c_{\tau}^{(t)}a^{(t)}_{i,j} \ \text{ and } \  \bar{\dot{w}}_{i,j}^{(\tau,1)} = \sum_{t=1}^T c_{\tau}^{(t)}b^{(t)}_{i,j}. $$
We can rewrite
$$
\Upsilon_{n}^{(\tau)} =  ( \sum_{t=1}^T c_{\tau}^{(t)}\ba^{(t)}_{\calS},  \sum_{t=1}^T c_{\tau}^{(t)}\bb^{(t)}_{\calS}):= (L_a(\tau), L_b(\tau)),
$$
where $\ba^{(t)}_{\calS}:= (w_{i,j}^{(t)}, (i,j) \in \calS)^\top \in \R^{|\calS|}$, $\bb^{(t)}_{\calS}:=(\dot{w}_{i,j}^{(t)}, (i,j) \in \calS)^\top \in \R^{|\calS|}$, $L_a(\tau):= \sum_{t=1}^T c_{\tau}^{(t)}\ba^{(t)}_{\calS}$,  $L_b(\tau):=\sum_{t=1}^T c_{\tau}^{(t)}\bb^{(t)}_{\calS}$, and $(\ba, \bb) := \ba^\top \bb$ for two vectors $\ba, \bb$. We will separate the proof into two parts. \textit{Part I} below calculates the mean and variance of $\Upsilon^{(\tau)}_n$. 
\textit{Part II} below derives the asymptotic distribution of $\Upsilon^{(\tau)}_n/\sqrt{\Var(\Upsilon^{(\tau)}_n)}$ via the Berry--Esseen central limit theorem \citep{hall1980martingale, haeusler1988rate, mourrat2013rate}.

\medskip

\textbf{Part I}. We calculate the mean and variance of $\Upsilon^{(\tau)}_n$. With slight abuse of notation, let us define a $\sigma$-algebra $\mathcal{F} := \left\{\bb^{(t)}_{\calS},t=1,\ldots T \right\}$. By the law of total expectation, we have that 
	\begin{equation} 
        \label{equ:con_expectation}
        \begin{aligned}
		& \E(\Upsilon^{(\tau)}_n) =\E\left\{\E(\Upsilon^{(\tau)}_n\mid \mathcal{F})\right\}\\
        &=  \sum_{t=1}^T \sum_{t^\prime = 1}^T c_{\tau}^{(t)} c_{\tau}^{(t^\prime)} \E\left\{\E\left\{(\ba_{\calS}^{(t)})^\top \bb_{\calS}^{(t^\prime)}\mid \bb_{\calS}^{(t)}\right\} \right\}\\
        &=0.
        \end{aligned}
	\end{equation}
  We next calculate the variance of $\Upsilon^{(\tau)}_n$. It follows from the law of iterated variance and \eqref{equ:con_expectation} that 
	\begin{equation}
		\label{equ:nullvar}
        \begin{aligned}
			\Var(\Upsilon^{(\tau)}_n)
			&= \E\left\{  \Var\left(\Upsilon^{(\tau)}_n \mid \mathcal{F} \right) \right\}+ \Var\left\{\E(\Upsilon^{(\tau)}_n \mid \mathcal{F})\right\}\\ 
			&=  \E\left\{  \Var\left(\Upsilon^{(\tau)}_n \mid \mathcal{F} \right) \right\}.
            \end{aligned}
	\end{equation}  

Notice that $$\bSigma_{\calS} = \Var(\ba^{(1)}_{\calS}) = \diag\{\theta_{i,j}(1-\theta_{i,j}), (i,j) \in \calS\} \in \R^{|\calS| \times |\calS|},$$ where $\theta_{i,j}$ is the $(i,j)$th entry of $\bTheta$. Then it holds that 
	\begin{equation}\label{equ:c3:2}
		\begin{aligned}
			\Var\left(\Upsilon^{(\tau)}_n \mid \mathcal{F} \right) =& \Var\left\{  \sum_{t=1}^T  c_{\tau}^{(t)}  (\ba_{\calS}^{(t)})^\top L_b(\tau) \mid \mathcal{F}\right\} \\
			=&  \sum_{t=1}^T  \sum_{t^\prime=1}^T \Cov\left\{ c_{\tau}^{(t)}  (\ba_{\calS}^{(t)} )^\top L_b(\tau), c_{\tau}^{(t^\prime)}  (\ba_{\calS}^{(t^\prime)})^\top L_b(\tau)  \mid \mathcal{F} \right\} \\
                =&   \sum_{t=1}^T (c_{\tau}^{(t)})^2  L_b(\tau)^\top \Cov\left\{ \ba_{\calS}^{(t)} ,  \ba_{\calS}^{(t)}  \mid \mathcal{F} \right\} L_b(\tau)\\
			=& \sum_{t=1}^T (c_{\tau}^{(t)})^2 L_b(\tau)^\top \bSigma_{\calS} L_b(\tau),
		\end{aligned}
	\end{equation}
	where the third equality above is because of the joint independence of $\ba^{(t)}_{\calS}$'s across $t$, and the last equality above is because of $ \Cov\left\{ \ba^{(t)}_{\calS} ,  \ba^{(t)}_{\calS}  \mid \mathcal{F} \right\} =  \Cov\left\{ \ba^{(t)}_{\calS}  , \ba^{(t)}_{\calS}   \right\} = \bSigma_{\calS}$ due to the independence between $\{\ba^{(t)}\}$'s and $\{\bb^{(t)}_{\calS}\}$'s. 
    
    Further, it follows from the joint independence of $\{\bb^{(t)}_{\calS}\}_{t=1,\ldots,T}$ that 
	\begin{equation}\label{equ:c3:3}
		\begin{aligned}
			\E\left\{\Var\left(\Upsilon_n \mid \mathcal{F}  \right)\right\} =& \sum_{t=1}^T (c_{\tau}^{(t)})^2  \E\left\{ L_b(\tau)^\top \bSigma_{\calS} L_b(\tau) \right\}	\\
			=&\sum_{t=1}^T (c_{\tau}^{(t)})^2   \E\left\{ \left( \sum_{t=1}^T c_{\tau}^{(t)} \bb^{(t)}_{\calS}\right)^\top \bSigma_{\calS}  \left( \sum_{t=1}^T c_{\tau}^{(t)} \bb^{(t)}_{\calS} \right) \right\}\\
			=& \sum_{t=1}^T \sum_{t^\prime=1}^T  (c_{\tau}^{(t)})^2  (c_{\tau}^{(t^\prime)})^2  \E\left\{ \left( \bb^{(t^\prime)}_{\calS}\right)^\top \bSigma_{\calS}   \bb^{(t^\prime)}_{\calS}  \right\} \\
                =&  \left\{\sum_{t=1}^T  (c_{\tau}^{(t)})^2 \sum_{t^\prime=1}^T(c_{\tau}^{(t^\prime)})^2\right\}\tr\left(\bSigma^2_{\calS}\right)\\
			=&\tau^{-2} \tr\left(\bSigma^2_{\calS}\right) := \gamma_{\tau} \tr\left(\bSigma^2_{\calS}\right),
		\end{aligned}
	\end{equation}
	where the fourth equality above is due to 
 $$
 \begin{aligned}
    \E\left\{ \left( \bb^{(t^\prime)}_{\calS}\right)^\top \bSigma_{\calS}   \bb^{(t^\prime)}_{\calS}  \right\}=& \E\left\{ \tr\left\{ \bSigma_{\calS}   \bb^{(t^\prime)}_{\calS}\left( \bb^{(t^\prime)}_{\calS}\right)^\top \right\} \right\}\\
    =& \tr\left\{\bSigma_{\calS} \E\left\{  \bb^{(t^\prime)}_{\calS} \left( \bb^{(t^\prime)}_{\calS}\right)^\top  \right\} \right\}\\
    =&\tr\left(\bSigma^2_{\calS}\right),  
 \end{aligned}
 $$
 and the last equality above is from direct calculation.
 
 Hence, combining \eqref{equ:nullvar}, \eqref{equ:c3:2}, and \eqref{equ:c3:3}, we can obtain that 
	\begin{equation}
        \label{equ:true_var}
	    \begin{aligned}
		\Var(\Upsilon^{(\tau)}_n) &= \gamma_T \tr\left(\bSigma^2_{\calS}\right) =  \gamma_{\tau} \sum_{(i,j) \in \calS} f(\theta_{i,j}) 
        := \gamma_{\tau} \sigma_{\calS}^2,
	\end{aligned}
	\end{equation}
	where $f(x) := x^2 - 2x^3 + x^4$. This completes the mean and variance calculations for $\Upsilon^{(\tau)}_n$.  

\medskip

\textbf{Part II}. We now aim to prove the asymptotic normality with an explicit convergence rate by applying the Berry--Esseen martingale central limit theorem (see, e.g., Theorem 2 of \cite{haeusler1988rate} and Theorem 1.1 of \cite{mourrat2013rate}). To this end, we first show that $\Upsilon^{(\tau)}_n$ has the representation as summation of martingale differences. Let us define  $$\zeta_{t,l}=   c_{\tau}^{(t)} c_{\tau}^{(l)} (\ba_{\calS}^{(t)} )^\top \bb^{(l)}_{\calS} := d_{t,l} (\ba_{\calS}^{(t)} )^\top \bb_{\calS}^{(l)} $$ with $d_{t,l}  :=   c_{\tau}^{(t)} c_{\tau}^{(l)} $.  Then it holds that $$d_{t,l} \asymp T^{-2}$$ in light of \eqref{def:weight} ($\tau \asymp T$ for all $\tau \in \calT$). Further, define  
 $$v_t  = \sum_{l=1}^T \zeta_{t,l} = (\ba_{\calS}^{(t)} )^\top \big(\sum_{l=1}^Td_{t,l} \bb_{\calS}^{(l)}\big).
 $$ 
We can obtain the following representation of $\Upsilon^{(\tau)}_n $ 
	\begin{equation}\label{equ:c3:1}
    \begin{aligned}
	    \Upsilon^{(\tau)}_n & :=  \sum_{t=1}^T \sum_{l=1}^T c_{\tau}^{(t)} c_{\tau}^{(l)} (\ba_{\calS}^{(t)} )^\top \bb_{\calS}^{(l)} = \sum_{t=1}^T \sum_{l=1}^T    c_{\tau}^{(t)} c_{\tau}^{(l)} (\ba_{\calS}^{(t)} )^\top \bb_{\calS}^{(l)} \\
        &= \sum_{t=1}^T\sum_{l=1}^T\zeta_{t,l}=\sum_{t=1}^Tv_t.
        \end{aligned}
	\end{equation}
Let us introduce the filtrations $\mathcal{F}_{t} := \sigma\left\{ \bb^{(1)}_{\calS}, \ldots, \bb^{(T)}_{\calS}, \ba^{(1)}_{\calS}, \ldots, \ba^{(t)}_{\calS}\right\}$, $t=1,\cdots, T$. Then we have that 
	$$
	\E\left( v_{t}  \mid\mathcal{F}_{t-1} \right) =  \big(\sum_{l=1}^Td_{t,l}\bb^{(l)}_{\calS}\big)^\top \E\left\{ \ba^{(t)}_{\calS}   \mid \mathcal{F}_{t-1}\right\} = 0.
	$$
	This shows that $\Upsilon^{(\tau)}_n = \sum_{t=1}^T v_t$ is indeed a summation of martingale differences.
    
To prove the Berry--Esseen bound, let us define two quantities
\begin{equation*}
    M_n :=  \sum_{t=1}^T \Var(\Upsilon^{(\tau)}_n)^{-2}\E\left( v_t^4 \right) \ \text{ and } \ V_n:= \sum_{t=1}^T \Var(\Upsilon^{(\tau)}_n)^{-1} \E(v_t^2 \mid \mathcal{F}_{t-1}).
\end{equation*}
Specifically, we will verify the following two conditions
\begin{equation}\label{equ:lindeberg}
       M_n = O( \omega_n^{-1})
\end{equation}
and
\begin{equation}\label{equ:v_con}
   \E\left(V_n-1\right)^2=O( \omega_n^{-1}).
\end{equation} 
Then the Berry--Esseen martingale central limit theorem (see, e.g., Theorem 2 of \cite{haeusler1988rate}) is applicable to show that 
$$
\sup_{t \in \R}\left| \Pr\left\{ \frac{\Upsilon^{(\tau)}_n}{\sqrt{\Var(\Upsilon^{(\tau)}_n)}} \le t \right\} - \Phi(t)\right| = O(\omega_n^{-1/5}).
$$
It remains to verify the two conditions \eqref{equ:lindeberg} and \eqref{equ:v_con} above. 
  
We begin with verifying condition \eqref{equ:lindeberg}. Let us first calculate $\sum_{t=1}^T \E(v_t^4)$. In view of $\E(\zeta_{t,l_1} \zeta^3_{t,l_2}) = 0$,  $\E(\zeta_{t,l_1} \zeta_{t,l_2} \zeta^2_{t,l_3}) = 0$, and $\E(\zeta_{t,l_1} \zeta_{t,l_2} \zeta_{t,l_3} \zeta_{t,l_4}) = 0$ for $l_1 \ne l_2 \ne l_3 \ne l_4$, it holds that 
\begin{equation}
 \label{equ:c3:4}
		\begin{aligned}
			&\sum_{t=1}^T \E(v_t^4) =  \sum_{t=1}^T\E\left\{ \left( \sum_{l=1}^T \zeta_{t,l}\right)^4\right\}\\
                &= \sum_{t=1}^T \sum_{l=1}^T \E(\zeta_{t,l}^4) + 3\sum_{t=1}^T  \sum_{l=1}^T  \sum_{l^\prime=1, l^\prime \ne l}^T \E\left( \zeta_{t,l}^2 \zeta_{t,l^\prime}^2\right)\\
                &:= v_1 + 3v_2. 
		\end{aligned}
\end{equation}
We first calculate term $v_1$ on the right-hand side of (\ref{equ:c3:4}) above. Since $(\ba^{(t)}_{\calS}, \bb^{(l)}_{\calS})$ is identically distributed as $(\ba^{(1)}_{\calS},\bb^{(1)}_{\calS})$, we have that 
\begin{equation} \label{equ:c3:5}
    \E(\zeta_{t,l}^4) = d_{t,l}^4 \E\left((\ba^{(t)}_{\calS} )^\top \bb^{(l)}_{\calS}  \right)^4 =  d_{t,l}^4 \E\left((\ba^{(1)}_{\calS} )^\top \bb^{(1)}_{\calS}  \right)^4.
\end{equation}
By the joint independence of the elements of $\{a^{(1)}_{i,j}, (i,j) \in \calS\}$ and $\{b^{(1)}_{i,j}, (i,j)\in \calS\}$, and using similar arguments as for \eqref{equ:c3:4}, we can deduce that 
\begin{equation} \label{equ:c3:6}
\begin{aligned}
    & \E\left((\ba^{(1)}_{\calS} )^\top \bb^{(1)}_{\calS} \right)^4 =  \sum_{(i,j) \in \calS} \E(a^{(1)}_{i,j} b^{(1)}_{i,j} )^4 \\
    &\quad+ 3\sum_{\substack{(i_1,j_1), (i_2,j_2) \in \calS\\ (i_1, j_1) \ne (i_2, j_2)}} \E\left\{(a^{(1)}_{i_1,j_1} b^{(1)}_{i_1,j_1} )^2(a^{(1)}_{i_2,j_2} b^{(1)}_{i_2,j_2})^2\right\}\\
    &= O\left(\sum_{(i,j) \in \calS}\theta_{i,j}^2 + \sum_{\substack{(i_1,j_1), (i_2,j_2) \in \calS\\ (i_1,j_1) \ne (i_2,j_2)}} \theta_{i_1,j_1}^2 \theta_{i_2,j_2}^2\right).
\end{aligned}
\end{equation}
In light of \eqref{equ:c3:5} and \eqref{equ:c3:6}, and noting that $d_{t,l} \asymp T^{-2}$ by \eqref{def:weight}, we can obtain that  
\begin{equation}\label{equ:c6:v1}
    v_1 = O\left(T^{-6} \left(\sum_{(i,j) \in \calS}\theta_{i,j}^2 + \sum_{\substack{(i_1,j_1), (i_2,j_2) \in \calS\\ (i_1,j_1) \ne (i_2,j_2)} }\theta_{i_1,j_1}^2 \theta_{i_2,j_2}^2\right)\right).
\end{equation}

We now focus on term $v_2$ above. Observe that
\begin{equation}\label{equ:c3:7}
\begin{aligned}
        \E\left( \zeta_{t,l}^2 \zeta_{t,l^\prime}^2\right) &= d_{t,l}^2 d_{t,l^\prime}^2 \E\left\{\E\left\{\left( (\ba^{(t)}_{\calS} )^\top \bb^{(l)}_{\calS} \right)^2\left((\ba^{(t)}_{\calS})^\top \bb^{(l^\prime)}_{\calS}  \right)^2 \mid \ba^{(t)}_{\calS}\right\} \right\}\\
        &=d_{t,l}^2 d_{t,l^\prime}^2 \E\left\{ \E\left\{\left( (\ba^{(t)}_{\calS})^\top \bb^{(l)}_{\calS} \right)^2 \mid \ba^{(t)}_{\calS}\right\}\E\left\{\left( (\ba^{(t)}_{\calS})^\top \bb^{(l^\prime)}_{\calS} \right)^2 \mid \ba^{(t)}_{\calS}\right\}\right\}\\
        &=O\left(T^{-8}\E\left\{( \ba^{(t),\top}_{\calS} \bSigma_{\calS} \ba^{(t)}_{\calS})^2\right\}\right)\\
        &= O\left(T^{-8}\E\left\{\left( (\ba^{(1)}_{\calS})^\top \bSigma_{\calS} \ba^{(1)}_{\calS}\right)^2\right\}\right),
\end{aligned}
\end{equation}
where the third step above has used $ \E\left\{\left( (\ba^{(t)}_{\calS})^\top \bb^{(l)}_{\calS} \right)^2 \mid \ba^{(t)}_{\calS}\right\}= (\ba^{(t)}_{\calS})^\top \bSigma_{\calS} \ba^{(t)}_{\calS}$, and the last step above is because  $\ba^{(t)}_{\calS}$ is identically distributed as $\ba^{(1)}_{\calS}$.
Since $\bSigma_{\calS} = \diag\{\theta_{i,j}(1-\theta_{i,j}) , (i,j) \in \calS\}$, it holds that 
\begin{equation}\label{equ:c3:8}
    \begin{aligned}
    &\E\left\{( (\ba^{(1)}_{\calS})^\top \bSigma \ba^{(1)}_{\calS})^2\right\} = \E\left(\sum_{(i,j) \in \calS} \theta_{i,j}(1-\theta_{i,j}) (a_{i,j}^{(1)})^2 \right)^2\\
    &=\sum_{(i,j)\in \calS}\theta_{i,j}^2(1-\theta_{i,j})^2\E(a_{i,j}^{(1)})^4 \\
    &\quad+  \sum_{\substack{(i_1,j_1), (i_2,j_2) \in \calS\\ (i_1,j_1) \ne (i_2,j_2)}} \theta_{i_1,j_1}(1-\theta_{i_1,j_1}) \theta_{i_2,j_2}(1-\theta_{i_2,j_2}) \E(a_{i_1,j_1}^{(1)})^2\E(a_{i_2,j_2}^{(1)})^2\\
    &=O\left(\sum_{(i,j) \in \calS}\theta_{i,j}^3 + \sum_{\substack{(i_1,j_1), (i_2,j_2) \in \calS\\ (i_1,j_1) \ne (i_2,j_2)}}\theta_{i_1,j_1}^2 \theta_{i_2,j_2}^2\right).
    \end{aligned}
\end{equation}

Hence, combining \eqref{equ:c3:7} and \eqref{equ:c3:8} above yields that 
\begin{equation}\label{equ:c6:v2}
\begin{aligned}
    &v_2:=\sum_{t=1}^T  \sum_{l=1}^T  \sum_{l^\prime=1, l^\prime \ne l}^T \E\left( \zeta_{t,l}^2 \zeta_{t,l^\prime}^2\right) \\
    &= O\left( T^{-5}\left(\sum_{(i,j) \in \calS}\theta_{i,j}^3 + \sum_{\substack{(i_1,j_1), (i_2,j_2) \in \calS\\ (i_1,j_1) \ne (i_2,j_2)}}\theta_{i_1,j_1}^2 \theta_{i_2,j_2}^2\right)\right).
    \end{aligned}
\end{equation}
This result along with \eqref{equ:c3:4} and \eqref{equ:c6:v1} leads to 
\begin{equation}\label{eq:v-4th-moment}
    \begin{aligned}
        \sum_{t=1}^T \E(v_t^4) 
        &= T^{-4}O\left( T^{-2} \left(\sum_{(i,j) \in \calS}\theta_{i,j}^2 + \sum_{\substack{(i_1,j_1), (i_2,j_2) \in \calS\\ (i_1,j_1) \ne (i_2,j_2)} }\theta_{i_1,j_1}^2 \theta_{i_2,j_2}^2\right)\right. \\
        &\quad \left.+ T^{-1}\left(\sum_{(i,j) \in \calS}\theta_{i,j}^3 + \sum_{\substack{(i_1,j_1), (i_2,j_2) \in \calS\\ (i_1,j_1) \ne (i_2,j_2)}}\theta_{i_1,j_1}^2 \theta_{i_2,j_2}^2\right)\right). 
    \end{aligned}
\end{equation}

We now point out two facts. From \eqref{con:Feb21:01}, we can show that 
\begin{equation}\label{Feb13:equ:04}
   \sum_{(i,j) \in \calS} \theta_{i,j}^2 + \sum_{\substack{(i_1,j_1), (i_2,j_2) \in \calS\\ (i_1,j_1) \ne (i_2,j_2)}} \theta_{i_1,j_1}^2 \theta_{i_2,j_2}^2= O\left(\omega_n^{-1} T^2(\sum_{(i,j)\in \calS} \theta_{i,j}^2)^2\right). 
\end{equation}
Moreover, it follows from \eqref{con:Feb21:02} and $\omega_n = O(T)$ that 
\begin{equation}\label{Feb13:equ:05}
  \sum_{(i,j) \in \calS}\theta_{i,j}^3 + \sum_{\substack{(i_1,j_1), (i_2,j_2) \in \calS\\ (i_1,j_1) \ne (i_2,j_2)}}\theta_{i_1,j_1}^2 \theta_{i_2,j_2}^2 = O\left(\omega_n^{-1}T(\sum_{(i,j) \in \calS} \theta_{i,j}^2)^2 \right).
\end{equation}
These results together with \eqref{equ:true_var}, \eqref{eq:v-4th-moment}, \eqref{Feb13:equ:04}, and \eqref{Feb13:equ:05} yield that 
\begin{equation*}
    \begin{aligned}
        \sum_{t=1}^T \E(v_t^4) = O\left( \omega_n^{-1}T^{-4}(\sum_{(i,j) \in \calS} \theta_{i,j}^2)^2\right) = O\left(\omega_n^{-1}\Var(\Upsilon_n)^2\right),
    \end{aligned}
\end{equation*}
which establishes condition \eqref{equ:lindeberg}.

It remains to verify condition \eqref{equ:v_con}. To this end, we will make a claim that 
 \begin{equation}\label{equ:con_v1}
     \E\left\{ \sum_{t=1}^T \E(v_t^2 \mid \calF_{t-1}) \right\} = \Var(\Upsilon_n)
 \end{equation}
and
 \begin{equation}\label{equ:con_v2}
     \E\left\{\sum_{t=1}^T \E(v_t^2 \mid \calF_{t-1}) \right\}^2 \to \Var(\Upsilon_n)^2(1+\omega_n^{-1}).
 \end{equation}
Then combining the above two results gives that 
\begin{equation*}
    \E\left(V_n-1\right)^2 = \E\left\{\frac{\sum_{t=1}^T \E(v_t^2 \mid \calF_{t-1})}{\Var(\Upsilon_n)} - 1\right\}^2 \to O(\omega_n^{-1}),
\end{equation*}
which establishes condition \eqref{equ:v_con}. It remains to prove (\ref{equ:con_v1}) and (\ref{equ:con_v2}) in the claim above.

We first establish \eqref{equ:con_v1}. Let us write
	\begin{equation*}
		\begin{aligned}
			&U_t:= \E(v_t^2 \mid \calF_{t-1}) \\
            &= \E\left\{ \sum_{l=1}^T \sum_{l^\prime = 1}^T \left\{d_{t,l} d_{t,l^\prime} (\ba_{\calS}^{(t)} )^\top\bb_{\calS}^{(l)}  (\ba_{\calS}^{(t)} )^\top\bb_{\calS}^{(l^\prime)} \right\}\mid \calF_{t-1}\right\}\\
			&=\sum_{l=1}^T \sum_{l^\prime=1}^T d_{t,l} d_{t,l^\prime} (\bb_{\calS}^{(l)} )^\top \bSigma_{\calS} \bb_{\calS}^{(l^\prime)}.
		\end{aligned}
	\end{equation*}
    Then we can deduce that 
	\begin{equation}
		\label{null:10}
		\begin{aligned}
			&\sum_{t=1}^T\E(U_t) = \sum_{t=1}^T \sum_{l=1}^T \sum_{l^\prime=1}^Td_{t,l} d_{t,l^\prime} \E\left\{  (\bb_{\calS}^{(l)})^\top \bSigma \bb_{\calS}^{(l^\prime)} \right\}\\
			&=\sum_{t=1}^T \sum_{l=1}^Td_{t,l}^2\E\left\{  (\bb_{\calS}^{(l)} )^\top \bSigma \bb_{\calS}^{(l)}\right\} \\
			&=\sum_{t=1}^T \sum_{l=1}^Td_{t,l}^2 \tr\left( \bSigma_{\calS}^2\right) = \Var(\Upsilon_n),
		\end{aligned}
	\end{equation}
	where the last equality above follows from
	\begin{equation}
		\label{null:11}
		\begin{aligned}
        \sum_{t=1}^T \sum_{l=1}^Td_{t,l}^2 =\gamma_{\tau}.
		\end{aligned}
	\end{equation}
	This completes the proof for \eqref{equ:con_v1}. 
    
    Finally, it remains to establish \eqref{equ:con_v2}. From the joint independence of $\{\bb^{(t)}, t=1,\ldots,T \}$, we can show that 
	\begin{equation} \label{new.eq.FL007}
		\begin{aligned}
			&\E\left\{\left(\sum_{t=1}^T U_t\right)^2 \right\} = \E\left(\sum_{t=1}^T \sum_{t^\prime=1}^T U_t U_{t^\prime}\right)\\
			&=\sum_{t=1}^T \sum_{t^\prime=1}^T  \sum_{l_1=1}^T\sum_{l_1^\prime=1}^T\sum_{l_2=1}^T\sum_{l_2^\prime=1}^T d_{t,l_1}d_{t,l_1^\prime}d_{t^\prime,l_2}d_{t^\prime,l_2^\prime} \E\left\{ (\bb^{(l_1)}_{\calS} )^\top \bSigma_{\calS} \bb^{(l_1^\prime)}_{\calS}  (\bb^{(l_2)}_{\calS})^\top  \bSigma_{\calS} \bb_{\calS}^{(l_2^\prime)} \right\}\\
		  &:= r_{1} + r_2 + r_3 + r_4,
		\end{aligned}
	\end{equation}
	where $r_1$ corresponds to the summation in $\{(t, t^\prime, l_1, l_1^\prime, l_2, l_2^\prime):l_1 = l_1^\prime, l_2 = l_2^\prime, l_1 \ne l_2\}$, $r_2$ corresponds to the summation in $\{(t, t^\prime, l_1, l_1^\prime, l_2, l_2^\prime):l_1 = l_2, l_1^\prime = l_2^\prime, l_1 \ne l_1^\prime\}$, $r_3$ corresponds to the summation in $\{(t, t^\prime, l_1, l_1^\prime, l_2, l_2^\prime):l_1 = l_2^\prime, l_1^\prime = l_2, l_1^\prime \ne l_2\}$, and $r_4$ corresponds to the summation in $\{(t, t^\prime, l_1, l_1^\prime, l_2, l_2^\prime):l_1 = l_1^\prime= l_2 = l_2^\prime\}$. Other types of summations are all equal to zero by independence.
    
    For term $r_1$ on the right-hand side of (\ref{new.eq.FL007}) above, it holds that 
	\begin{equation}\label{equ:c2:01}
		\begin{aligned}
			r_1 =& \sum_{t=1}^T \sum_{t^\prime=1}^T\sum_{l_1=1}^T \sum_{l_2=1,l_1 \ne l_2}^T d_{t,l_1}^2 d_{t^\prime,l_2}^2 \E\left\{ (\bb_{\calS}^{(l_1)})^\top \bSigma_{\calS} \bb_{\calS}^{(l_1)}  (\bb_{\calS}^{(l_2)})^\top  \bSigma_{\calS} \bb_{\calS}^{(l_2)} \right\}\\
			=&\gamma^2_{\tau} (\tr\left( \bSigma_{\calS}^2\right))^2 = \Var(\Upsilon_n)^2.
		\end{aligned}
	\end{equation}
	For term $r_2$ above, note that $d_{t,l_1} \asymp T^{-1}$ and
        { \begin{equation*}
            \begin{aligned}
                &\E\left\{ (\bb^{(l_1)}_{\calS} )^\top \bSigma_{\calS} \bb_{\calS}^{(l_1^\prime)}  (\bb_{\calS}^{(l_1)})^\top  \bSigma_{\calS} \bb_{\calS}^{(l_1^\prime)}  \right\} \\
                =&\E\left\{ (\bb^{(l_1^\prime)}_{\calS})^\top \bSigma_{\calS} \bb^{(l_1)}_{\calS}  (\bb^{(l_1)}_{\calS})^\top  \bSigma_{\calS} \bb^{(l_1^\prime)}_{\calS} \right\} \\
                =& \tr \left \{\E\left\{ \bSigma_{\calS} \bb^{(l_1)}_{\calS}  (\bb^{(l_1)}_{\calS} )^\top  \bSigma_{\calS} \bb_{\calS}^{(l_1^\prime)} (\bb^{(l_1^\prime)}_{\calS})^\top \right\} \right\}\\
                =&\tr \left \{\E\left\{ \bSigma_{\calS} \bb^{(l_1)}  (\bb^{(l_1)}_{\calS})^\top\right\} \E \left\{ \bSigma_{\calS} \bb^{(l_1^\prime)_{\calS}} (\bb^{(l_1^\prime)}_{\calS})^\top  \right\} \right\} \\
                =& \tr \left \{ \bSigma_{\calS}^4 \right\},
            \end{aligned}
        \end{equation*}}
        where the first equality above is because $\bSigma_{\calS}$ is diagonal. Moreover, by \eqref{con:Feb21:03}, we have that 
        $$
        \tr\left( \bSigma_{\calS}^4\right) = O\left( \omega_n^{-1} \left( \tr\left( \bSigma_{\calS}^2\right)\right)^2 \right).
        $$
        Hence, it follows that 
	\begin{equation}\label{equ:c2:02}
		\begin{aligned}
			r_2 &= \sum_{t=1}^T \sum_{t^\prime=1}^T\sum_{l_1=1}^T \sum_{l_1^\prime=1,l_1 \ne l_1^\prime}^T d_{t,l_1} d_{t,l_1^\prime}d_{t^\prime,l_1}d_{t^\prime,l_1^\prime}\E\left\{ (\bb^{(l_1)}_{\calS} )^\top \bSigma_{\calS} \bb^{(l_1^\prime)}_{\calS}  (\bb^{(l_1)}_{\calS})^\top  \bSigma_{\calS} \bb_{\calS}^{(l_1^\prime)}  \right\}\\
			&= O(T^{-4}  \tr\left( \bSigma_{\calS}^4\right)) = O(\omega_n^{-1}T^{-4}  \left(\tr\left( \bSigma_{\calS}^2\right)\right)^2 ) \\
            &= O(\omega_n^{-1}\Var(\Upsilon_n)^2).
		\end{aligned}
	\end{equation}
    
	Similarly, for  term $r_3$, we can show that 
	$$r_3 = O(\omega_n^{-1}\Var(\Upsilon_n)^2).$$ We now focus on term $v_4$. Notice that 
    $$
    \begin{aligned}
    \E\left\{(\bb_{\calS}^{(1),\top} \bSigma \bb_{\calS}^{(1)})^2\right\} & = \sum_{(i,j) \in \calS}\theta_{i,j}^3 + \sum_{\substack{(i_1,j_1),(i_2,j_2) \in \calS\\(i_1,j_1) \ne (i_2,j_2)}}\theta_{i_1,j_1}^2 \theta_{i_2,j_2}^2 \\
    & = O\left(\omega_n^{-1}T(\sum_{(i,j)\in \calS} \theta^2_{i,j})^2 \right),
    \end{aligned}
    $$ 
    where we have used \eqref{con:Feb21:02} and \eqref{equ:c3:8} to obtain the bound in the last step above. Then we can obtain that 
	\begin{equation}\label{equ:c2:03}
		\begin{aligned}
			r_4 &=\sum_{t=1}^T \sum_{t^\prime=1}^T \sum_{l=1}^T d_{t,l}^4 \E\left\{ ((\bb^{(l)}_{\calS})^\top \bSigma_{\calS} \bb^{(l)}_{\calS})^2    \right\} \\
   &= O\left(T^{-5} \E\left\{(\bb^{(1),\top}_{\calS} \bSigma_{\calS} \bb^{(1)}_{\calS})^2\right\}\right) \\
   &= O\left(T^{-4} \omega_n^{-1} (\sum_{(i,j)\in \calS} \theta_{i,j})^2 \right) \\
   &= O(\omega_n^{-1}\Var(\Upsilon_n)^2).
		\end{aligned}
	\end{equation}
	 Therefore, combining \eqref{equ:c2:01}, \eqref{equ:c2:02}, and \eqref{equ:c2:03} proves \eqref{equ:con_v2}. This concludes the proof of Lemma \ref{lem:asymptotic}.

The lemma below provides a uniform upper bound for the remaining terms in \eqref{Feb13:equ:07}.

\begin{lemma}\label{lem:H1:reminder}
Under Condition \ref{con:upper}, and letting $\calS \subset \{(i,j): 1 \le i < j \le n\}$ be a nonrandom set, we have 
$$
r^{(\tau)}_l = O_p\left(\frac{\rho_n\sqrt{\log(en) (|\calS| \vee s^*)}}{T}\right)
$$
for $l =1,\ldots, 4$, where $r^{(\tau)}_l $'s with $l \in [4]$ are defined in \eqref{Feb13:equ:07}.    
\end{lemma}

\noindent\textit{Proof}. Observe that all of $r_l^{(\tau)}$ with $ l=1,\ldots 4$ are zero mean random variables. Thus it suffices to calculate the variances of the terms and apply the Markov inequality. Here, we use the notation introduced in the proof of Lemma \ref{lem:asymptotic}, where $\ba^{(t)}_{\calS}:= (w_{i,j}^{(t)}, (i,j) \in \calS)^\top \in \R^{|\calS|}$ and $\bb^{(t)}_{\calS}:=(\dot{w}_{i,j}^{(t)}, (i,j) \in \calS)^\top \in \R^{|\calS|}$.
For $r_1^{(\tau)}$, let us define $\calF = \{\bb^{(t)}, t=1,\ldots,T\}$, $\bSigma^{(t)} := \Var(\ba^{(t)}) = \Var(\bb^{(t)}) = \diag\{\theta^{(t)}_{i,j}(1-\theta^{(t)}_{i,j}), (i,j) \in \calS\} \in \R^{|\calS| \times |\calS|}$, and $\bar{\bSigma} := \sum_{t=1}^T \bSigma^{(t)}/T$. In view of \eqref{equ:nullvar}, the variance of $r_1^{(\tau)}$ is given by 
	\begin{equation*}
		\begin{aligned}
			\Var(r_1^{(\tau)}) &= \E\left\{ \Var(r_1^{(\tau)} \mid \calF)\right\} = \E\left\{ \Var\left(\sum_{t=1}^T\sum_{t^\prime=1}^T w_{\tau}^{(t)} w_{\tau}^{(t^\prime)}(\ba^{(t)}_{\calS}, \bb^{(t^\prime)}_{\calS} )\mid \calF\right)\right\}\\
			&= \E\left\{ \sum_{t=1}^T\sum_{t^\prime=1}^T \sum_{l=1}^T\sum_{l^\prime=1}^Tw_{\tau}^{(t)} w_{\tau}^{(t^\prime)} w_{\tau}^{(l)} w_{\tau}^{(l^\prime)}  \Cov\left( (\ba^{(t)}_{\calS}, \bb^{(t^\prime)}_{\calS}), (\ba^{(l)}_{\calS}, \bb^{(l^\prime)}_{\calS})\mid \calF\right) \right\}\\
			&= \sum_{t=1}^T\sum_{l^\prime=1}^T \sum_{t^\prime=1}^T (w_{\tau}^{(t)})^2 w_{\tau}^{(t^\prime)}  w_{\tau}^{(l^\prime)}  \E\left\{ (\bb^{(t^\prime)}_{\calS})^\top \bSigma^{(t)} (\bb^{(l^\prime)}_{\calS}) \right\}\\
			&= \sum_{t=1}^T\sum_{t^\prime=1}^T (w_{\tau}^{(t)})^2 (w_{\tau}^{(t^\prime)})^2   \tr\left( \bSigma^{(t)}_{\calS} \bSigma^{(t^\prime)}_{\calS}\right) \\
            &\lesssim T^{-2}  \tr(\bar{\bSigma}^2_{\calS}),
		\end{aligned}
	\end{equation*}
 where the third and fourth equalities above are due to the joint independence of $\{\ba^{(t)}, t \in [T]\}$ and $\{\bb^{(t)}, t \in [T]\}$, and the last inequality above is from $w_{\tau}^{(t)} \lesssim T^{-1}$ for $t \le \tau$ (see \eqref{def:weight}). Moreover, since $\tr(\bar{\bSigma}_{\calS}^2) \le \bar{\btheta}^\top\bar{\btheta} \le |\calS| \rho_n^2$, an application of the Markov inequality yields that 
	\begin{equation}
		\label{power:v1}
        \begin{aligned}
		 r_1^{(\tau)} & = O_p(\sqrt{\Var(r_1^{(\tau)})}) = O_p\left(T^{-1} \sqrt{\tr(\bar{\bSigma}^2_{\calS})}\right)  \\
         &= O_p\left(\frac{\sqrt{|\calS|} \rho_n}{T}\right).
         \end{aligned}
	\end{equation}

For $r_2^{(\tau)}$, it holds that 
$$
\begin{aligned}
\Var(r_2^{(\tau)}) &= \sum_{(i,j) \in \calS}  \delta_{i,j}^2 \left( \E(\bar{w}_{i,j}^{(\tau,1)})^2 + \E(\bar{\dot{w}}_{i,j}^{(\tau,1)})^2\right) \\
&\lesssim \sum_{(i,j) \in \calS} \delta_{i,j}^2 \frac{\rho_n}{T} \lesssim \frac{\rho_n \bar{\epsilon}^2_n}{T^2} \lesssim \frac{\rho^2_n s^* \log(en)}{T^2},
\end{aligned}
$$
which entails that 
$$
r_2^{(\tau)} = O_p\left(\frac{\rho_n \sqrt{s^*\log(en)}}{T}\right).
$$
For $r_3^{(\tau)}$, it follows from part 1) of Lemma \ref{lemma:moment_arbitrary_S} that  $$\Var(r_3^{(\tau)}) \lesssim \sum_{(i,j) \in \calS} \delta_{i,j}^2 \frac{q_n^2}{Tn}  = \frac{\rho_n}{T}\sum_{(i,j) \in \calS} \delta_{i,j}^2. $$ Then we have that 
$$
r_3^{(\tau)} = O_p\left(\sqrt{ \frac{\rho_n}{T} \sum_{(i,j) \in \calS} \delta_{i,j}^2}\right) = O_p\left(\frac{\rho_n \sqrt{s^* \log(en)}}{T}\right).
$$

We now focus on $r_4^{(\tau)}$. Some direct calculations lead to 
$$
\begin{aligned}
r_4^{(\tau)} & = O\left(\sqrt{\frac{q_n^2 |\calS|}{n^2 T} \sum_{(i,j)\in \calS} \delta^2_{i,j}} \right)= O\left(\sqrt{\frac{\rho^2_n |\calS|}{n T^2} s^* \log(en)} \right) \\
& = O\left(\frac{\rho_n \sqrt{|\calS| \log(en)}}{T}\right).
\end{aligned}
$$
This completes the proof of Lemma \ref{lem:H1:reminder}.

The lemma below provides the property of the candidate sets in $\hcalS$ and $\tilde{\calS}$ defined in \eqref{def:s_out}. In particular, this lemma shows that under the alternative, the strong signal set $\calS_1$ is contained in the candidates sets with probability tending to one. Meanwhile, the weak signal set $\calS_2$ is ruled out with probability tending to one simultaneously. 

\begin{lemma}\label{lem:selection}
    Under $H_1(\rho, K^*, s^*,  \bar{\epsilon}_n)$ with $s^* \ll n$, Condition \ref{con:upper}, and the time span condition $T q_n^2 \gg \log(n)$, we have $\Pr_1\{ \calS_1 \subset \hat{\calS}, \, \calS_2 \cap \hat{\calS} = \emptyset\} \to 1$ and $\Pr_1\left\{ \hcalS \subseteq \calS^*\right\} \to 1$. This result also holds when replacing $\hcalS$ with $\tilde{\calS}$ defined in \eqref{def:s_out}.
\end{lemma}



\noindent\textit{Proof}. The proofs for $\tilde{\calS}$ and $\hcalS$ are the same so we only provide the proof for $\tilde{\calS}$ here. For any small constant $1/2 > \eta > 0$, let us define an event
\begin{equation}\label{def:E1}
    E_1 := \{|\rho_n/\hat{\rho}_n - 1| \le \eta \} \cap F_1,
\end{equation}
where event $F_1$ is defined in \eqref{event:F_1}. By invoking Lemmas \ref{lemma:rho} and \ref{lem:gen1}, we see that 
\begin{equation}\label{equ:b6:01}
\Pr_1\{E_1\} \to 1.
\end{equation}
Let us further define another event
\begin{equation}\label{def:E2}
    E_2 = \{\calS_1 \subseteq \tilde{\calS}, \, \calS_2 \cap \tilde{\calS} = \emptyset\}.
\end{equation}
We make a claim that 
\begin{equation}\label{equ:b6:02}
	\Pr_1\{E_2^c, E_1\} \to 0.
\end{equation}
Then it holds that 
$$
\begin{aligned}
\Pr_1\{E_2^c\} & = \Pr_1\{E_2^c\cap E_1\} + \Pr_1\{E_2^c\cap E_1^c\} \\
&\le \Pr_1\{E_2^c\cap E_1\} + \Pr_1\{E_1^c\} \to 0.
\end{aligned}
$$ 

Hence, to prove the lemma, it remains to establish the claim in \eqref{equ:b6:02} above. To this end, we can decompose $E_2^c$ as 
$$ E_2^c  = \{  \calS_1  \not\subseteq \tilde{\calS}\} \cup \{\tilde{\calS} \cap \calS_2 \ne \emptyset \},$$ 
where $\emptyset$ denotes the empty set. Then we have that 
\begin{equation}\label{equ:b6:03}
\begin{aligned}
	\Pr_1\{ E_2^c,E_1\}  & \le  \Pr_1\left\{\{ \calS_1  \not\subseteq \tilde{\calS} \} , E_1\right\}  +  \Pr_1\left\{ \{\tilde{\calS} \cap \calS_2 \ne \emptyset \}, E_1\right\} \\
    &:= v_1 + v_2.
    \end{aligned}
\end{equation} 
We will analyze terms $v_1$ and $v_2$ above separately. 

We first investigate term $v_1$ on the right-hand side of (\ref{equ:b6:03}) above. Let us define 
$$\calS_1^{+} := \{(i,j)\in \calS_1: \delta_{i,j} > 0\}$$ and 
$$\calS_1^{-} := \{(i,j)\in \calS_1: \delta_{i,j} < 0\}.$$ 
By the union bound, it holds that 
\begin{equation}\label{equ:b6:v1}
    \begin{aligned}
	v_1 & =  \Pr_1\left\{\cup_{(i,j) \in \calS_1} \{|{\hddz}_{i,j}^{(hT)}| < \cd_n\}, E_1  \right\}\\
	& \le \sum_{(i,j) \in \calS_1}\Pr_1\{|{\hddz}_{i,j}^{(hT)}| < \cd_n, E_1\}  \\
	& = \sum_{(i,j) \in \calS_1^+}\Pr_1\{|{\hddz}_{i,j}^{(hT)}| < \cd_n, E_1\}  + \sum_{(i,j) \in \calS_1^-}\Pr_1\{|{\hddz}_{i,j}^{(hT)}| < \cd_n,E_1\} \\
    & := v_{11} + v_{12}. 
\end{aligned}
\end{equation}
For term $v_{11}$ on the right-hand side of (\ref{equ:b6:v1}) above, let 
$$\bar{e}_{i,j}^{(\tau)} =  \sqrt{\frac{n}{2}}\sum_{l=1}^2 (-1)^{l+1}(\be_i^\top \tilde{\bW}^{(\tau,l)} \bb^{(l)}_j + \be_j^\top \tilde{\bW}^{(\tau,l)} \bb^{(l)}_i),$$ 
where the definitions of related quantities can be found above \eqref{event:F}. Define $c_{\eta} = 2-(1-\eta)^{-1/2}$. 

For any positive constant $c$, we can deduce that 
\begin{equation}\label{equ:b6:3}
	\begin{aligned}
		v_{11} & \le  \sum_{(i,j) \in \calS_1^+} \Pr_1\left\{ |\hddz_{i,j}^{(hT)}| \le \cd_n, E_1\right\}\\
		& \le  \sum_{(i,j) \in \calS_1^+}\Pr_1\left\{ -(1-\eta)^{-1/2}\cd_n \le \sqrt{\frac{hT}{2\rho_n}} \delta_{i,j} + \bar{e}_{i,j}^{(hT)} \le (1-\eta)^{-1/2} \cd_n\right\} \\
		& \le \sum_{(i,j) \in \calS_1^+} \Pr_1\left\{   {\bar{e}_{i,j}^{(hT)}} \le -\sqrt{\frac{hT}{2\rho_n}} \delta_{i,j} + (1-\eta)^{-1/2} \cd_n\right\} \\
		& \le  \sum_{(i,j) \in \calS_1^+}  \Pr_1\left\{   -\bar{e}_{i,j}^{(hT)} \ge c_{\eta}\cd_n,E_1 \right\} \\
         & =  \sum_{(i,j) \in \calS_1^+}  \Pr_1\left\{   -\sqrt{n}\bar{e}_{i,j}^{(hT)} \ge 4c_{\eta} { c_d}\log(n) \right\} \\
		& \le 4 |\calS_1^+| \exp\left\{-\frac{16 c_{\eta}^2}{(2+c)}\log^2(n) \right\}\\
        & \le 4\exp\left\{-\frac{16 c_{\eta}^2}{(2+c)}\log^2(n) - 2\log(s^*) \right\} \to 0,
	\end{aligned} 
\end{equation}
where the first inequality above is from event $E$ and the fact that $\ddot{z}_{i,j}^{(hT)}$ is an i.i.d. copy of $z_{i,j}^{(hT)}$, the fourth inequality above is because $\sqrt{\frac{hT }{{2 {\rho}_n}}}\delta_{i,j} \ge  2\cd_n$ for $(i,j) \in \calS_1^{+}$, the fifth inequality above is due to the definition of $\cd_n$, the second last inequality above is based on \eqref{add:lem:equ1}, and in the last inequality above, we choose $\eta$ such that $c_{\eta} < 2$. Similarly, for term $v_{12}$ we can obtain that 
\begin{equation}\label{equ:b6:4}
\begin{aligned}
    v_{12} & \le  \sum_{(i,j) \in \calS_1^-} \Pr_1\left\{ |z_{i,j}^{(\tau)}| \le (1-\eta_n)^{-1/2}\cd_n, E_1\right\} \\
    & \le \sum_{(i,j) \in \calS_1^-}  \Pr_1\left\{ \bar{e}_{i,j}^{(\tau)} \ge { c_{\eta}} \cd_n \right\} \to 0.
    \end{aligned}
\end{equation}
Hence, combining (\ref{equ:b6:v1}), (\ref{equ:b6:3}), and (\ref{equ:b6:4}) yields that $v_1 \to 0$.

On the other hand, for any $1/2 < c^\prime < 1$, it holds that 
\begin{equation}\label{equ:b6:5}
	\begin{aligned}
		v_2 &= \Pr_1\left\{\cup_{(i,j) \in \calS_2} \{|{\hddz}_{i,j}^{(hT)}| \ge \cd_n\} , E_1   \right\}  \\
		&\le \sum_{(i,j) \in \calS_2}\Pr_1\{|{\hddz}_{i,j}^{(hT)}| \ge \cd_n, E_1\} \\
		&\le \sum_{(i,j) \in \calS_2}\Pr_1\left\{  \sqrt{\frac{hT}{2\rho_n}} \delta_{i,j} + {\bar{e}_{i,j}^{(hT)}} \ge \cd_n, E_1\right\}  \\
        &\quad+ \sum_{(i,j) \in \calS_2}\Pr_1\left\{  \sqrt{\frac{hT}{2\rho_n}} \delta_{i,j} + {\bar{e}_{i,j}^{(hT)}} \le -\cd_n, E_1\right\} \\
		&\le \sum_{(i,j) \in \calS_2}\Pr_1\left\{  \bar{e}_{i,j}^{(hT)} \ge c^\prime \cd_n, E_1\right\}  + \sum_{(i,j) \in \calS_2}\Pr_1\left\{  \bar{e}_{i,j}^{(hT)} \le -c^\prime\cd_n, E_1\right\}  \\
        &\le \sum_{(i,j) \in \calS_2}\Pr_1\left\{  \sqrt{n}\bar{e}_{i,j}^{(hT)} \ge 4c^\prime { c_d}\log(n)\right\}  + \sum_{(i,j) \in \calS_2}\Pr_1\left\{  \sqrt{n}\bar{e}_{i,j}^{(hT)} \le -4c^\prime { c_d}\log(n) \right\}  \\
		&\le 4|\calS_2| \exp\left\{-\frac{16(c^\prime)^2}{3} \log^2(n) \right\} \le 4\exp\left\{-\log^2(n) \right\} \to 0,
	\end{aligned}
\end{equation}
where the third inequality above is due to $\sqrt{\frac{hT}{2\rho_n}} \delta_{i,j}\ll \cd_n$ for $(i,j) \in \calS_2$, and the last inequality above is because of \eqref{add:lem:equ1}. Therefore, combining the above result with (\ref{equ:b6:03}) and (\ref{equ:b6:5}) gives the claim in \eqref{equ:b6:02}.

To prove $\Pr_1\left\{ \hcalS \subseteq \calS^*\right\} \to 1$, we only need to show
$$
\Pr_1\left\{ \hcalS \cap (\calS^*)^c = \emptyset\right\} \to 1
$$
Based on \eqref{equ:b6:5}, we can similarly show
$$
\Pr_1\left\{ \hcalS \cap (\calS^*)^c = \emptyset\right\} \le 4n^2 \exp\left\{-\frac{16(c^\prime)^2}{3} \log^2(n) \right\} \le 4\exp\left\{-\log^2(n) \right\} \to 0.
$$
This concludes the proof of Lemma \ref{lem:selection}.

\begin{lemma}\label{lem:bridge}
    Under $H_1(\rho, K^*, s^*,  \bar{\epsilon}_n)$, Condition \ref{con:upper}, the time span condition  $T\rho_n \gg \log(n)$, and letting $\tau \le \min( \tau^*,T-\tau^*) \le 2\tau$, we have
   $$
\begin{aligned}
\left| \hA^{(\tau)}_{\hcalS} - \hA^{(\tau)}_{\tilde{\calS}}\right|  = O_p\left(  \frac{|\calS_3| \rho_n \log(n)}{T} \right).
\end{aligned}
$$
\end{lemma}

\noindent\textit{Proof}. Without loss of generality, we assume that $\tau^* \le T/2$. Let us first make a claim that 
   \begin{equation}\label{equ:may12:04}
  \Pr_1\left\{ \hcalS \Delta \tilde{\calS} \subset \calS_3 \right\} \to 1
\end{equation}
and 
\begin{equation}\label{equ:may12:03}
   \max_{\tau \in \calT}\max_{(i,j)\in \calS_3} |\hat{w}_{i,j}^{(\tau)}| =  O_p\left( \sqrt{\frac{\rho_n \log(n)}{T}} \right), \  \max_{\tau \in \calT}\max_{(i,j)\in \calS_3}|\hat{\dot{w}}_{i,j}^{(\tau)}| =  O_p\left( \sqrt{\frac{\rho_n \log(n)}{T}} \right).
\end{equation}
Then using (\ref{equ:may12:04}) and (\ref{equ:may12:03}), we can obtain that 
$$
\begin{aligned}
    \left| \hA_{\hcalS} - \hA_{\tilde{\calS}} \right| & = \sum_{\tau \in \calT} \sum_{(i,j) \in \hcalS \Delta \tilde{\calS}} \left| \hat{w}_{i,j}^{(\tau)} \hat{\dot{w}}_{i,j}^{(\tau)} \right| \\
    & = O_p\left( \sum_{\tau \in \calT} \sum_{(i,j) \in \calS_3} \left| \hat{w}_{i,j}^{(\tau)} \hat{\dot{w}}_{i,j}^{(\tau)} \right| \right) \\
    & = O_p\left( \frac{|\calS_3| \rho_n \log(n)}{T} \right),
\end{aligned}
$$
which gives the desired conclusion of the lemma. 

It remains to establish \eqref{equ:may12:04} and \eqref{equ:may12:03} in the claim above. We will first prove \eqref{equ:may12:04}. To this end, we need to show that 
$$
\Pr_1\left\{ \hcalS \Delta \tilde{\calS} \subset \calS_3 \right\} = 1-\Pr_1\left\{ \hcalS \Delta \tilde{\calS} \cap (\calS_1 \cup \calS_2) \ne \emptyset \right\} \to 1.
$$  
Observe that 
\begin{equation}\label{equ:june22:lem12:05}
  \Pr_1\left\{ \hcalS \Delta \tilde{\calS} \cap (\calS_1 \cup \calS_2) \ne \emptyset \right\} \le \Pr_1\left\{ \hcalS \Delta \tilde{\calS} \cap \calS_1 \ne \emptyset \right\} + \Pr_1\left\{ \hcalS \Delta \tilde{\calS} \cap \calS_2 \ne \emptyset \right\}.  
\end{equation}  
For the first term on the right-hand side of (\ref{equ:june22:lem12:05}) above, it holds that 
$$
\begin{aligned}
    &\Pr_1\left\{ \hcalS \Delta \tilde{\calS} \cap \calS_1 \ne \emptyset \right\} \\
    &= \Pr_1\left\{ \sum_{(i,j) \in \calS_1} \left[\I\left(\hddz^{(hT)}_{i,j} \ge d_n, \hz_{i,j}^{(hT)} < d_n\right) + \I\left(\hddz^{(hT)}_{i,j} < d_n, \hz_{i,j}^{(hT)} \ge d_n\right)\right] \ge 1 \right\} \\
    &\le \Pr_1\left\{ \sum_{(i,j) \in \calS_1} \left[\I\left(\hz_{i,j}^{(hT)} < d_n\right) + \I\left(\hddz^{(hT)}_{i,j} < d_n\right)\right] \ge 1 \right\} \\
    &\le \sum_{(i,j) \in \calS_1} \left( \Pr_1\left\{ \hz_{i,j}^{(hT)} < d_n \right\} + \Pr\left\{ \hddz_{i,j}^{(hT)} < d_n \right\} \right) \to 0,
\end{aligned}
$$  
where the last step above follows similar arguments as for \eqref{equ:b6:3}. 

Similarly, for the second term on the right-hand side of (\ref{equ:june22:lem12:05}) above, we can deduce that 
$$
\begin{aligned}
    &\Pr_1\left\{ \hcalS \Delta \tilde{\calS} \cap \calS_2 \ne \emptyset \right\} \\
    &= \Pr_1\left\{ \sum_{(i,j) \in \calS_2} \left[\I\left(\hddz^{(hT)}_{i,j} \ge d_n, \hz_{i,j}^{(hT)} < d_n\right) + \I\left(\hddz^{(hT)}_{i,j} < d_n, \hz_{i,j}^{(hT)} \ge d_n\right)\right] \ge 1 \right\} \\
    &\le \Pr_1\left\{ \sum_{(i,j) \in \calS_2} \left[\I\left(\hz_{i,j}^{(hT)} \ge d_n\right) + \I\left(\hddz^{(hT)}_{i,j} \ge d_n\right)\right] \ge 1 \right\} \\
    &\le \sum_{(i,j) \in \calS_2} \left( \Pr_1\left\{ \hz_{i,j}^{(hT)} \ge d_n \right\} + \Pr_1\left\{ \hddz_{i,j}^{(hT)} \ge d_n \right\} \right) \to 0,
\end{aligned}
$$  
where the last step above uses similar arguments as for \eqref{equ:b6:5}. Consequently, we see that both terms on the right-hand side of \eqref{equ:june22:lem12:05} go to zero, which establishes (\ref{equ:may12:04}).

We now focus on proving \eqref{equ:may12:03}. In light of \eqref{equ:june22:thm4:01}, it holds that 
\begin{equation}\label{equ:june22:lem12:03}
    \hat{w}^{(\tau,1)}_{i,j} = \bar{w}_{i,j}^{(\tau,1)} + \delta_{i,j} -  f_{i,j}^{(\tau)} + o(\frac{q_n}{n \sqrt{\tau}})
\end{equation}
uniformly over $(i,j) \in \Omega$ with asymptotic probability one.
Similarly, we can show that 
\begin{equation}\label{equ:june22:lem12:04}
  \hat{\dot{w}}^{(\tau,1)}_{i,j} = \bar{\dot{w}}_{i,j}^{(\tau,1)} + \delta_{i,j} -  \dot{f}_{i,j}^{(\tau)} + o(\frac{q_n}{n \sqrt{\tau}})  
\end{equation}
uniformly over $(i,j) \in \Omega$ with asymptotic probability one.
It follows from \eqref{add:lem:equ1} and $T q_n^2 \gg \log(n) $ that 
\begin{equation}\label{equ:june22:lem12:01}
    \max_{(i,j) \in \Omega}|f_{i,j}^{(\tau)}| = O_p\left(\frac{q_n \sqrt{\log(n)}}{n \sqrt{\tau}}\right)  \text{ and }  \max_{(i,j) \in \Omega}|\dot{f}_{i,j}^{(\tau)}| = O_p\left(\frac{q_n \sqrt{\log(n)}}{n \sqrt{\tau}}\right).
\end{equation}

Moreover, an application of Bernstein's inequality leads to 
$$
\begin{aligned}
\Pr\left\{| \bar{w}^{(\tau,1)}_{i,j}| \ge 2\sqrt{\frac{\rho_n \log(n)}{\tau}} \right\} & \le \exp\left\{-\frac{ 2\tau \rho_n \log(n) }{ \tau \rho_n + \frac{2 \sqrt{\tau \rho_n \log(n)}}{3}} \right\} \\
& \lesssim \exp\{-2 \log(n)\}
\end{aligned}
$$
as $\tau \rho_n \ge hT \rho_n \gg \log(n)$. This shows that 
\begin{equation}\label{equ:june22:lem12:02}
 \max_{(i,j) \in \Omega} |\bar{w}^{(\tau,1)}_{i,j}| = O_p\left( \sqrt{\frac{\rho_n \log(n)}{\tau}} \right) \text{ and } \max_{(i,j) \in \Omega} |\bar{\dot{w}}^{(\tau,1)}_{i,j}| = O_p\left( \sqrt{\frac{\rho_n \log(n)}{\tau}} \right).   
\end{equation} 
Therefore, plugging \eqref{equ:june22:lem12:01} and \eqref{equ:june22:lem12:02} into \eqref{equ:june22:lem12:03} and \eqref{equ:june22:lem12:04}, and noting that $\delta_{i,j} = o\left(\sqrt{\rho_n \log(n) / T}\right)$ for $(i,j) \in \calS_3$ establish \eqref{equ:may12:03}. This completes the proof of Lemma \ref{lem:bridge}.

\section{Proofs of Theorems \ref{thm:minimax_lower}--\ref{thm:minimax_upper} and some key lemmas} \label{new.sec.C}

\subsection{Proof of Theorem \ref{thm:minimax_lower}} \label{new.sec.C.1}


Our proof of the lower bound without the low-rank structure is motivated by \cite{liu2021minimax}. The main idea is similar to that in the proof of Theorem \ref{low:minimax_lower} in Section \ref{proof:lower}. To simplify the technical analysis, we will repeat some claims from the proof of Theorem \ref{low:minimax_lower} with a different parameter space. The major goal is to construct a suitable prior distribution $\nu$ on the alternative hypothesis parameter space $\mathcal{A}(\rho_n, m^*, \underline{c}^\prime \epsilon)$, and show that the worst-case power can be upper bounded by the total variation distance between the null distribution and the mixture alternative distribution induced by prior distribution $\nu$.

Let us consider a single parameter $\bTheta^0_0 \in \calN(\rho_n)$ in the null space. We also consider a parameter $h:=\bTheta^0_1$ sampled from a prior distribution $\nu$ on the alternative parameter space, where $\supp(\nu) \subset \calA(\rho_n, m^*, \underline{c}^\prime\epsilon)$. The joint distribution of dynamic networks with parameters in $\bTheta_0^0$ is denoted as $\Pr_0$, and we use $\E_{0}$ to represent the expectation under $\Pr_0$. Given a single parameter $h \sim \nu$ sampled from the alternative space, the joint distribution of dynamic networks with parameters in $h$ is defined as $\Pr_h$, and we use $\E_{h}$ to denote the expectation under $\Pr_h$. For a given prior distribution $\nu$ and a parameter $h$ sampled from $\nu$, denote by $\bar{\Pr} := \E_{h \sim \nu}(\Pr_h) $ the induced average probability measure, and $\bar{\E}$ the expectation under such average probability measure. 

For any test $\psi_0$, similar to \eqref{new.eq.FL002}, we can show that 
	$$
	\inf_{h \in \calA(\rho,m^*,\underline{c}^\prime\epsilon)} \Pr_h\left\{\psi_0 \text{ rejects } H_{0}\right\} = \inf_{h \in \calA(\rho,m^*,\underline{c}^\prime\epsilon)}\E_h(\psi_0) \le \alpha + \|\mathbb{P}_0 - \bar{\mathbb{P}} \|_{TV}.
	$$
    In view of the proof of Theorem \ref{low:minimax_lower} in Section \ref{proof:lower}, to prove that for any given $\eta > \alpha$ and some constant $c > 0$, no test of significance level $\alpha$ can be rejected, it suffices to show that 
	\begin{equation}
        \label{june:equ:diver}
	    \E_{(\gamma, \zeta) \sim \nu \otimes \nu } \left\{\exp\left(\sum_{t=1}^T \sum_{i < j} \frac{\gamma_{i,j}^{(t)} \zeta_{i,j}^{(t)}}{\rho_n(1-\rho_n)}  \right)\right\} \le 1 + \xi.
	\end{equation}
Thus, the problem reduces to constructing prior distribution $\nu$ with $\supp(\nu) \subset \calA(\rho_n, m^*, \underline{c}^\prime\epsilon_n)$ 
such that \eqref{june:equ:diver} is satisfied. The sparsity 
parameter $m^*$ plays a key role in the technical analysis. We will separate the remaining proof into three parts according to the sparsity level. Denote by $p_n = n(n-1)/2$.

\medskip

\textbf{Part I. Dense alternative with $m^* \ge \sqrt{p_n} \sim n$}. Let us define the signal strength parameter 
$$(\underline{c}_1^\prime)^2\epsilon_n^2 = 2^{-1}m^*\rho_n \beta^2 $$ with $\beta = \beta(n,T,m^*) = (4^{-1}(m^*)^{-2}p_n)^{1/4} <1$. This implies that $$(\underline{c}_1^\prime)^2\epsilon_n^2 = (\underline{c}_1^\prime)^2\rho_n \sqrt{p_n}$$ with $\underline{c}_1^\prime=1/2$. We will construct prior distribution $\nu$ in the following way such that each element in $\boldsymbol{\Theta}^0 \in \calA(\rho_n,m^*,\underline{c}_1^\prime\epsilon_n)$ ($\epsilon_n$ as defined in \eqref{equ:minimax_rate}) is independently generated from the sampling process below:
	\begin{enumerate}
		\item[1)] Uniformly sample a subset $S \subset \{(i,j): 1 \le i < j \le n\}$ of cardinality $m^*$.
        
		\item[2)] Independent of $S$, generate $k \sim \operatorname{Unif}\left\{{ \log_2(h_{\xi}T), \log_2(h_{\xi}T) + 1 },\ldots, \log_2(T/2)\right\}$, where the bandwidth parameter $h_{\xi}$ is a function of $\xi$ that will be specified later.
        
		\item[3)] Independent of $(S, k)$, sample an upper triangular matrix $\bU = (u_{i,j}) \in \R^{n \times n}$, where $u_{i,j} \stackrel{i.i.d.}{\sim} \operatorname{Unif}(\{-1,1\})$ for $i < j$.
        
		\item[4)] Given the sampled triplet $(S, k, \bU)$, 
        define $\theta_{i,j}^{(t)} = \theta_{j,i}^{(t)} := \rho_n+ \gamma_{i,j}^{(t)} $ for all $\{(i,j), t\} \in \{(i,j): 1 \le i < j \le n\} \times [T]$, where i) $ \gamma_{i,j}^{(t)} = \rho_n^{1/2} \frac{\beta}{\sqrt{2^k}} u_{i,j} $ for $\{(i,j) ,t\} \in S \times [2^k]$, ii)  $\gamma_{i,j}^{(t)}=0$ for $\{(i, j),t\}\in \{S^c \times [2^k]\} \cup \left\{  \{(i,j): 1 \le i < j \le n\} \times [T] \backslash [2^k]\right\} $, and iii) all the diagonal entries $\theta_{i,i}^{(t)} = 0$ for all $i \in [n], t \in [T]$. We will refer to $\gamma_{i,j}^{(t)}$ as the jump size hereafter.
	\end{enumerate}
    
	Note that for any $\bTheta^0$ sampled from the above construction process, we have 
    $$\tau^* = 2^k.$$ 
    Then we see that the corresponding $\bDelta = \bTheta_1-\bTheta_2$ of $\bTheta^0$ satisfies that 
	$$
    \begin{aligned}
	\frac{\tau^*(T-\tau^*)}{T}\|\vech(\bDelta) \|_2^2 & =\frac{2^k(T- 2^k)}{T} \frac{\beta^2 \rho_n}{2^k} \sum_{(i,j) \in S} u_{i,j}^2 \\
    & = m^* \beta^2 \rho_n\frac{T-2^k}{T} \\
    & \ge \frac{m^*\beta^2 \rho_n}{2} =(\underline{c}_1^\prime)^2 \epsilon_n^2.
    \end{aligned}
	$$
Further, it holds that 
$$\left|\rho_n^{1/2} \frac{\beta}{\sqrt{2^k}} u_{i,j}\right| \lesssim \rho_n^{1/2} T^{-1/2} (m^*)^{-1/2} p_n^{1/4} \lesssim \rho_n$$ in light of the time span condition of $T  \gtrsim \rho_n^{-1} p_n^{1/2} (m^*)^{-1}$ in \eqref{equ:time_span_con} for the dense alternative. These two conditions ensure that $$\supp(\nu) \subset \calA^{(2^k)}(\rho_n, m^*, \underline{c}^\prime_1\epsilon_n).$$ 
For two independently sampled triplets $(S, k, \bU)$ and $(T, l, \bV)$ from the construction process described above, denote by $\{\gamma_{i,j}^{(t)}\}_{t \in [T], 1 \le i<j\le n}$ and  $\{\zeta_{i,j}^{(t)}\}_{t \in [T], 1 \le i<j\le n}$ the corresponding jump sizes in Step 4, respectively. We aim to show that \eqref{june:equ:diver} holds with such $\{\gamma_{i,j}^{(t)}\}_{t \in [T], 1 \le i<j\le n}$ and  $\{\zeta_{i,j}^{(t)}\}_{t \in [T], 1 \le i<j\le n}$. 

Observe that 
	$$
    \begin{aligned}
	\sum_{t=1}^T \sum_{i < j}\gamma_{i,j}^{(t)} \zeta_{i,j}^{(t)} & = (2^k \land 2^l) \frac{\beta^2 \rho_n}{\sqrt{2^{k+l}}} \sum_{(i,j) \in S \cap T} u_{i,j} v_{i,j}  \\
    & = \frac{\beta^2 \rho_n}{2^{|l-k|/2}} \sum_{(i,j) \in S \cap T} u_{i,j} v_{i,j}. 
    \end{aligned}
	$$
	Then it follows that 
	$$
	\begin{aligned}
		& \E_{(\gamma, \zeta) \sim \nu \otimes \nu } \left\{\exp\left(\sum_{t=1}^T \sum_{i < j} \frac{\gamma_{i,j}^{(t)} \zeta_{i,j}^{(t)}}{\rho_n(1-\rho_n)}  \right)\right\} \\
        &= \E \left\{\exp\left( \frac{\beta^2 \rho_n }{2^{|l-k|/2} \rho_n(1-\rho_n)} \sum_{(i,j) \in S \cap T} u_{i,j} v_{i,j}  \right) \right\}\\
		&\le \E \left\{\exp\left( \frac{2\beta^2 }{2^{|l-k|/2}} \sum_{(i,j) \in S \cap T} u_{i,j} v_{i,j}  \right)\right\},
	\end{aligned}
	$$
	where the expectation is over the joint distribution of $(S,k,\bU,T,l,\bV)$, and we have used $\rho_n<1/2$ for sufficient large $n$ in the last step above. In view of Step 3 above, we have $$u_{i,j} \stackrel{i.i.d.}{\sim} \operatorname{Unif}(\{-1,1\}) \ \text{ and } \ v_{i,j} \stackrel{i.i.d.}{\sim}\operatorname{Unif}(\{-1,1\}),$$ 
    where all $u_{i,j}$'s and $v_{i,j}$'s are independent of each other, leading to $u_{i,j} v_{i,j} \stackrel{i.i.d.}{\sim} \operatorname{Unif}(\{-1,1\})$. Hence, we can deduce that 
	$$
	\begin{aligned}
		& \E \left\{\exp\left( \frac{2\beta^2 }{2^{|l-k|/2}} \sum_{(i,j) \in S \cap T} u_{i,j} v_{i,j}  \right) \right\} \\
        & = \E \left\{\prod_{(i,j) \in S \cap T} \left( \frac{1}{2} e^{2\beta^2/2^{|l-k|/2}} + \frac{1}{2} e^{-2\beta^2/2^{|l-k|/2}}\right)\right\} \\
		& \le \E \left\{\exp\left(  |S \cap T| \frac{4\beta^4 }{2^{|l - k|}} \right)\right\},
	\end{aligned}
	$$
	where the last inequality above has used the fact that $(e^x + e^{-x})/2 \le e^{x^2/2}$. 
	
	Notice that $|S \cap T|$ is distributed as the hypergeometric distribution $\operatorname{Hyp}(p_n,m^*,m^*)$. Recall the fact that the hypergeometric distribution $\operatorname{Hyp}(p_n,m^*,m^*)$ is no larger than the binomial distribution $$Y:=\operatorname{Bin}(m^*,m^*/p_n):= Y$$ in the convex ordering sense (see, e.g., Theorem 4 in {\cite{hoeffding1963probability}}). Let us write 
    $$Y = \sum_{i=1}^{m^*} X_i$$ with independent Bernoulli random variables $X_i \stackrel{i.i.d.}{\sim} \operatorname{Ber}(m^*/p_n)$.  Then it holds that 
    \begin{equation}\label{june:eq:004}
    \begin{aligned}
      \E\left\{\exp\left(  |S \cap T| \frac{4\beta^4 }{2^{|l - k|}} \right) \right\} \le& \E\left\{\exp\left(  Y \frac{4\beta^4 }{2^{|l - k|}} \right) \right\} \\
      = &\E\left\{\prod_{i=1}^{m^*}\exp\left(  X_i \frac{4\beta^4 }{2^{|l - k|}} \right) \right\} \\
       =& \E \left\{\left( 1- \frac{m^*}{p_n} + \frac{m^*}{p_n} e^{4\beta^4/2^{|l-k|}}\right)^{m^*} \right\}\\
       =& \E \left\{\left( 1 + \frac{m^*}{p_n} \big(e^{4\beta^4/2^{|l-k|}}-1\big)\right)^{m^*} \right\}\\
	\le& \E\left\{\left( 1 + \frac{m^*}{p_n} \frac{4\beta^4}{2^{|l-k|} }e^{4\beta^4/2^{|l-k|}}\right)^{m^*}\right\} =: \E(L(l,k)),
    \end{aligned} 
    \end{equation}
    where the last inequality above is due to $e^x - 1 \le x e^x$ for all $x \ge 0$. Hence, combining the above results, 
    we see that to prove \eqref{june:equ:diver}, it remains to establish bound 
    \begin{equation} \label{new.eq.FL008}
    \E\left\{L(l,k)\right\} \le1+ \xi
    \end{equation}
    with $\xi = (\eta-\alpha)^2$. 

	Since $\beta = (4^{-1} (m^*)^{-2} p_n )^{1/4}$, the condition $m^* \ge \sqrt{ p_n}$ entails that $\beta^4 \le 1/4$ and thus 
    $$e^{4\beta^4/2^{|l-k|}} \le e.$$ 
    Since $l$ and $k$ are generated independently from $\operatorname{Unif}\{{ \log_2(h_{\xi}T), \log_2(h_{\xi}T) + 1 },\ldots, \log_2(T/2)\}$ in view of Step 2 above, it holds that 
	\begin{equation}\label{june:eq:002}
    \begin{aligned} 
	    	\Pr\{|l-k| = x\} & =  \E \left( \Pr\{k= x \pm l \mid l = l\} \right) \\
            & \le \E\left(\frac{2}{\log_2((2h_{\xi})^{-1})}\right) \\
            & \le \frac{2}{\log_2((2h_{\xi})^{-1})}
            \end{aligned} 
	\end{equation}
	for each $x \in \{0,1,2, \ldots, \log_2\{(2h_{\xi})^{-1}\}\}$. It follows from the basic inequality $(1+1/x)^x \le e $, the definition of $\beta$, and $e^{4\beta^4/2^{|l-k|}} \le e$ that 
	$$
	\begin{aligned}
		\left( 1 + \frac{m^*}{p_n} \frac{4\beta^4}{2^{|l-k|} }e^{4\beta^4/2^{|l-k|}}\right)^{m^*} \leq & \left(1+ \frac{1}{e^{-1} m^* 2^{|l-k|}} \right)^{m^*}\\
		\le & \left\{ \left(1+ \frac{1}{ e^{-1}m^* 2^{|l-k|}}\right)^{e^{-1} s 2^{|l-k|}}\right\}^{\frac{1}{e^{-1}2^{|l-k|}}}\\ \le&   e^{ e  2^{-|l-k|} }.
	\end{aligned}
	$$
	For each given $\xi > 0$, there exist some sufficiently small bandwidth parameter $h_{\xi}$ and integer $m_0$ such that 1) $\lceil \log_2\big\{e \log^{-1}(1+\xi/2) \big\} \rceil \le m_0 \le   \log_2^{1/2}\{(2h_{\xi})^{-1}\} $ and 2) $4e^e/\xi \le  \log_2^{1/2}\{(2h_{\xi})^{-1}\}$. From 1), we see that $$e^{e2^{-x}} \le 1 + \xi/2$$ for $x \ge m_0$. A combination of 1) and 2) yields that $$\frac{m_02e^e}{\log_2\{(2h_{\xi})^{-1}\}}\leq 2^{-1}\xi.$$ 
	
	Thus, combining the above 
    results with 2), the definition of $L(l,k)$, and \eqref{june:eq:002}, we can obtain that 
	$$
	\begin{aligned}
		\E\left\{ L(l,k) \right\} \le& \sum_{x = 1}^{\log_2\{(2h_{\xi})^{-1}\}} e^{e 2^{-x}} \Pr\{|l-k| = x\} \\
		= & \sum_{x = 1}^{m_0} e^{e 2^{-x}} \Pr\{|l-k| = x\}  +  \sum_{x = m_0+1}^{\log_2\{(2h_{\xi})^{-1}\}} e^{e 2^{-x}} \Pr\{|l-k| = x\}  \\
		\le& m_0\cdot e^e \cdot\frac{2}{\log_2\{(2h_{\xi})^{-1}\}} + (1+ \frac{\xi}{2}) \sum_{x = m_0+1}^{\log_2\{(2h_{\xi})^{-1}\}} \Pr\{|l-k| = x\}\\
   \le& \frac{m_02e^e}{\log_2\{(2h_{\xi})^{-1}\}} + (1+ \frac{\xi}{2})  \le 1 + \xi.
	\end{aligned}
	$$
This along with (\ref{new.eq.FL008}) completes the proof of \eqref{june:equ:diver} for the dense case with $m^* \ge \sqrt{p_n} \sim n$.

\medskip

\textbf{Part II. Sparse alternative with $m^* < \sqrt{p_n}$ and $\log\left(\frac{c^\prime_{\xi}p_n}{(m^*)^2}\right) \ge 0$}. We now consider the sparse case with $m^* < \sqrt{p_n}$ and $\log\left(\frac{c^\prime_{\xi}p_n}{(m^*)^2}\right) \ge 0$ for a  small constant $c_{\xi}^\prime \in (0,1)$ whose dependence on $\xi$ will be made clear later. It is easy to see that $$m^* \leq \sqrt{c^\prime_{\xi}p_n}.$$ In this regime, we set $$\beta = \sqrt{\frac{1}{2}\log\left(\frac{c^\prime_{\xi}p_n}{(m^*)^2}\right)}$$ for sufficient small $c^\prime_{\xi} \le \log(1+\xi)$. Then it holds that  $$\epsilon_n^2 = (\underline{c}_2^\prime)^2\rho_n m^*\log\left(\frac{c^\prime_{\xi}p_n}{s^2}\right)$$ with $\underline{c}_2^\prime = 1/2$.

Let us consider the same prior distribution $\nu$ as specified in the dense case, with the exception that in the third step therein, we set $u_{i,j} = 1$ for all $(i, j) \in S$. We can verify that 
$$\supp(\nu) \subseteq \calA^{(\tau^*)}(\rho_n, m^*, \underline{c}_2^\prime\epsilon_n).$$ 
Similar to \eqref{june:eq:004}, we can deduce that 
	$$
	\begin{aligned}
		& \E_{(\gamma, \zeta) \sim \nu \otimes \nu } \left\{\exp\left(\sum_{t=1}^T \sum_{i < j} \frac{\gamma_{i,j}^{(t)} \zeta_{i,j}^{(t)}}{\rho_n(1-\rho_n)}  \right)\right\} \\
        & = \E \left\{\exp\left(|S \cap T | \frac{2\beta^2}{2^{|l-k|/2 }} \right) \right\}\\
		&\le \E\left\{ 1 - \frac{m^*}{p_n} + \frac{m^*}{p_n} \exp\left(\frac{2\beta^2}{2^{|l-k|/2}} \right) \right\}^{m^*}\\
		&\le\E\left(1 +\frac{m^*}{p_n} e^{2\beta^2/2^{|l-k|/2}}\right)^{m^*} =:\E\{R(l,k)\}.
	\end{aligned}
	$$  
	With the choice of $\beta = \sqrt{\frac{1}{2}\log(\frac{c^\prime_{\xi}p_n}{(m^*)^2})}$ for sufficient small $ c^\prime_{\xi} \le \log(1+\xi)/2$,   
    it holds that 
	$$
	\begin{aligned}
		\left(1 +\frac{m^*}{p_n} e^{2\beta^2/2^{|l-k|/2}}\right)^{m^*} \le& 	\left(1 +\frac{m^*}{p_n} e^{2\beta^2}\right)^{m^*} \le 	\left(1 +\frac{m^*}{p_n} \left( \frac{c^\prime_{\xi}p_n}{(m^*)^2}\right)\right)^{m^*}\\
        \le& \left(1+\frac{c^\prime_{\xi}}{m^*}\right)^{m^*} \le e^{c^\prime_{\xi}} \le 1+\xi.
	\end{aligned}
	$$
Hence, we have that 
	$$\E\left\{R(l,k)\right\} \le 1+\xi.$$ 
    This proves \eqref{june:equ:diver} for the sparse case with $m^* < \sqrt{p_n}$ and $\log\left(\frac{c^\prime_{\xi}p_n}{s^2}\right) \ge 0$.

\medskip

\textbf{Part III. Sparse alternative with $m^* < \sqrt{p_n}$ and $\log\left(\frac{c^\prime_{\xi}p_n}{(m^*)^2}\right) < 0$}. We finally examine the sparse case with $m^* < \sqrt{p_n}$ and $\log\left(\frac{c^\prime_{\xi}p_n}{(m^*)^2}\right) < 0$. For the scenario of $m^* < \sqrt{p_n}$ and $\log\left(\frac{c^\prime_{\xi}p_n}{(m^*)^2}\right) < 0$, the lower bound derived in \textit{Part II} above becomes negative, which is useless now. For a given $m^*$, let us consider an arbitrary $p_n > m^*$ to obtain a uniform 
lower bound for the sparse alternative. Here, we set the signal strength parameter 
$$\epsilon_n^2 = (\underline{c}^\prime_3)^2\rho_n := 2^{-1}c^\prime_{\xi} \rho_n$$ for $c^\prime_{\xi} \le 2^{-1} \log(1+\xi)$, and fix a nonrandom set $S \subset \{(i,j):1 \le i < j \le n\}$. 

We will construct the prior distribution $\nu$ in the following way: 
1) sample a $k \sim \operatorname{Unif}\{\log_2(h_{\xi}T),$ $ \log_2(h_{\xi}T) + 1, \ldots, \log_2(T/2)\}$, and 2) set $\theta^{(t)}_{i,j} = \theta^{(t)}_{j,i}$ for $\{(i,j),t\} \in S \times [2^{k}]$ as $\rho_n +\sqrt{\frac{c^\prime_{\xi}\rho_n}{2^{k} m^*}} := \rho_n + \gamma_{i,j}^{(t)}$, all other $\theta_{i,j}^{(t)} = \theta_{j,i}^{(t)}$ as $\rho_n$ for $i\neq j$, and all the diagonal entries $\theta^{(t)}_{l,l} = 0$ for all $t$ and $l \in [n]$. Hence, the signal strength is given by 
$$
\begin{aligned}
\frac{\tau^*(T-\tau^*)}{T}\|\vech(\bDelta) \|_2^2 & =\frac{2^{k}(T-2^{k})}{T} \frac{c^\prime_{\xi}\rho_n}{2^{k} m^*} |S|  \\
& \ge \frac{c^\prime_{\xi}\rho_n}{2} = (\underline{c}_3^\prime)^2\epsilon_n^2, 
\end{aligned}
$$
and $$\theta_{i,j}^{(t)} \lesssim \rho_n + \sqrt{\rho_n (T(m^*))^{-1}} \lesssim \rho_n$$ by the time span condition of $T \gtrsim \rho_n^{-1} (m^*)^{-1}$ in \eqref{equ:time_span_con}. It follows that 
$$\supp(\nu) \subset \calA^{(\tau^*)}(\rho_n, m^*, \underline{c}_3^\prime\epsilon_n)$$ for $m^* < \sqrt{p_n}$. 

With such sampling scheme, it holds that 
$$
\begin{aligned}
\sum_{t=1}^T \sum_{i < j} \gamma^{(t)}_{i,j} \zeta^{(t)}_{i,j} & = \sum_{t=1}^T \sum_{(i,j) \in S} \gamma^{(t)}_{i,j} \zeta^{(t)}_{i,j} = (2^{k} \land 2^{l}) \frac{c^\prime_{\xi}\rho_n}{\sqrt{2^{k+l}} m^*} |S| \\
& = \frac{c^\prime_{\xi}\rho_n}{2^{|l-k|/2}}.
\end{aligned}
$$
Hence, we can obtain that 
	$$
	\begin{aligned}
		&\E_{(\gamma, \zeta) \sim \nu \otimes \nu } \left\{\exp\left(\sum_{t=1}^T  \sum_{i < j}\frac{\gamma^{(t)}_{i,j} \zeta^{(t)}_{i,j}}{\rho_n(1-\rho_n)}  \right)\right\} \\
        & \le \E \left\{\exp\left(\frac{2c^\prime_{\xi}}{2^{|l-k|/2}}\right) \right\} \le e^{2c^\prime_{\xi}} \\
        & \le 1+\xi,
	\end{aligned}
	$$  
	since $c^\prime_{\xi} \le  2^{-1}\log(1+\xi)$. This gives \eqref{june:equ:diver} for the sparse case with $m^* < \sqrt{p_n}$ and $\log\left(\frac{c^\prime_{\xi}p_n}{(m^*)^2}\right) < 0$.

Therefore, combining the above three parts and choosing 
$$
{
\begin{aligned}
 \underline{c}^\prime:=& \underline{c}^\prime_1 \I\left( m^* \ge \sqrt{p_n}\right)+\underline{c}^\prime_2 \I\left( m^* < \sqrt{c^\prime_{\xi} p_n}\right)+\underline{c}^\prime_3 \I\left( \sqrt{c^\prime_{\xi} p_n}
\le m^* \le \sqrt{p_n} \right)\\
\leq & 2^{-1}\sqrt{\log(1+\xi)}
\end{aligned}
}
$$ 
with $c_{\xi}^\prime \le 2^{-1}\log(1+\xi)$ yield the desired conclusion. This concludes the proof of Theorem \ref{thm:minimax_lower}.

\subsection{Proof of Theorem \ref{thm:minimax_upper}} \label{new.sec.C.2}


Throughout this proof, the subscript of $\Pr_l$ with $l=0,1$ indicates the probability measure evaluated under $H_0(\rho_n)$ or $H_1(\rho_n, m^*, \bar{c}^\prime\epsilon_n)$, respectively. We will consider the dense and sparse cases separately. To simplify the notation, denote by $B_{\calS} := B_{a_n}$, where we define $a_n:= d(m^*)$ which is the threshold in $\calS$. We will show that with the separating rate defined in \eqref{equ:minimax_rate1}, the test statistic $\psi_n$ introduced in \eqref{def:phi_test} has size bounded by $\alpha$, and power $\eta >\alpha$. The size analysis is relatively straightforward and follows directly from some basic probability inequalities. 

For the power analysis, let us assume that both the sparsity parameter $m^*$ and degree parameter $\rho_n$ are known.
The main idea of the proof has two steps. First, by the construction of $\calT$, there exists some $\tau \in \calT$ such that $\tau^*/2 < \tau < \tau^*$, where $\tau^*$ is the true change point.  We aim to show that 
\begin{align}
    & r_n\leq 2^{-1}\E(B_{a_n}^{(\tau)}), \label{eq:power-1a} \\
    & \frac{4\Var(B_{a_n}^{(\tau)})}{\E^2(B_{a_n}^{(\tau)} )}\leq 1-\eta.\label{eq:power-1b}
\end{align}
 Then it follows from (\ref{eq:power-1a}) and (\ref{eq:power-1b}) that 
\begin{equation*}
    \begin{aligned}
	\Pr_1(\psi_1 = 1) =& \Pr_1\{\max_{\tau \in \calT} B_{a_n}^{(\tau)} \ge r_n \} \ge \Pr_1\{B_{a_n}^{(\tau)} \ge r_n\}\\
 =& \Pr_1\{ \E(B_{a_n}^{(\tau)} ) - B_{a_n}^{(\tau)}  \le \E(B_{a_n}^{(\tau^*)} ) - r_n  \} \\
 \ge& \Pr_1\{ \E(B_{a_n}^{(\tau)} ) - B_{a_n}^{(\tau)}  \le \frac{1}{2}\E(B_{a_n}^{(\tau)} )\} \\
 = &  1-\Pr_1\{ \E(B_{a_n}^{(\tau)} ) - B_{a_n}^{(\tau)}  > \frac{1}{2}\E(B_{a_n}^{(\tau)} )\} \\
	\ge&  1 - \frac{4\Var(B_{a_n}^{(\tau)})}{\E^2(B_{a_n}^{(\tau)} )} \geq \eta.
\end{aligned}
\end{equation*}
Hence, to prove the power result, it remains to establish \eqref{eq:power-1a} and \eqref{eq:power-1b}. We will separate the remaining proof into two parts.

\medskip

\textbf{Part I: Dense case with $m^* \ge \sqrt{p_n}$}. In this regime, the threshold parameters of test statistic $\psi_n$ defined in \eqref{def:phi_test} become $a_n = 0$ and $r_n = c_{\alpha}\sqrt{p_n}$, where $c_{\alpha}\ge \sqrt{2\alpha^{-1}\log_2\{(2h)^{-1}\}}$ is a positive constant.
We also set the lower bound 
$$\epsilon^2_n = \bar{c}^\prime\rho_n\sqrt{p_n} \geq 4c_{\eta} \rho_n\sqrt{p_n}$$ with $c_{\eta}\geq 2c_{\alpha}\vee \sqrt{8(1-\eta)^{-1}}$. 
We will first show that the test statistic $\psi_n$ controls the size under $H_0(\rho_n)$. Applying Lemma \ref{lemma:june:inequality} in Section \ref{new.sec.C.3} with $a=0$ and using the fact $c_{\alpha}\ge \sqrt{2\alpha^{-1}\log_2\{(2h)^{-1}\}}$, we can deduce that 
$$
\begin{aligned}
\Pr_0\{\psi_n  =1\} & \le \sum_{\tau \in \calT} \Pr_0\{ B_{0}^{(\tau)} > r_n \} = \sum_{\tau \in \calT} \Pr_0\{ B_{0}^{(\tau)} > c_{\alpha}\sqrt{p_n} \} \\
& \le \frac{2 |\calT|}{c_{\alpha}^2} \le \frac{2\log_2\{(2h)^{-1}\}}{c_{\alpha}^2} \le \alpha.
\end{aligned}
$$

We next prove the power result. As discussed at the beginning of the proof, we need only to show \eqref{eq:power-1a} and \eqref{eq:power-1b}. 
Note that we have the decomposition
\begin{equation}
\label{app:equ_z}
\begin{aligned}
   { e}_{i,j}^{(\tau)} & =  \frac{\tau\delta_{i,j} }{\sqrt{2\tau\rho_n}} + \frac{ \sum_{t=1}^{\tau}(w_{i,j}^{(t)} - w_{i,j}^{(T-t+1)})}{\sqrt{2\tau\rho_n}} \\
   & := \frac{\tau\delta_{i,j} }{\sqrt{2\tau\rho_n}} + \bar{e}_{i,j}^{(\tau)},
   \end{aligned}
\end{equation}
where $\delta_{i,j} := \bDelta_{i,j}$. Then it holds that 
$$\E(\bar{e}_{i,j}^{(\tau)})^2\leq 1.$$ Meanwhile, we have that 
$$
\begin{aligned}
\dot{{ e}}_{i,j}^{(\tau)} & =  \frac{\tau\delta_{i,j} }{\sqrt{2\tau\rho_n}} + \frac{ \sum_{t=1}^{\tau}(\dot{w}_{i,j}^{(t)} - \dot{w}_{i,j}^{(T-t+1)})}{\sqrt{2\tau\rho_n}} \\
& := \frac{\tau\delta_{i,j} }{\sqrt{2\tau\rho_n}} + \bar{\dot{e}}_{i,j}^{(\tau)}.
\end{aligned}
$$
In view of the condition $\tau(T-\tau)T^{-1}\|\vech(\bDelta)\|_2^2\geq \bar{c}^\prime \epsilon_n^2$ in the definition of alternative space $\mathcal A^{(\tau^*)}(\rho_n,m^*, \bar{c}^\prime\epsilon_n)$, and noting that $\tau>\tau^*/2$, we can deduce that 
\begin{equation}
\label{equ:expect}
\begin{aligned}
	\E(B_{0}^{(\tau)}) & =\sum_{i < j} \E \left( { e}^{(\tau)}_{i,j} \dot{{ e}}^{(\tau)}_{i,j} \right) = \sum_{i < j} \frac{{\tau} \delta_{i,j}^2 }{2\rho_n} \\
    & \ge \sum_{i < j} \frac{{\tau^*} \delta_{i,j}^2 }{4\rho_n}  \ge \frac{ (\bar{c}^\prime)^2\epsilon_n^2}{4\rho_n} \ge c_{\eta}\sqrt{p_n}.
\end{aligned}
\end{equation}
 Thus, it follows that 
 $$r_n = c_{\alpha}\sqrt{p_n} \leq c_{\eta}^{-1}c_{\alpha}\mathbb E(B_0^{(\tau)}) \leq 2^{-1}\mathbb E(B_0^{(\tau)})$$ since $c_{\eta}\geq 2c_{\alpha}$.  This establishes \eqref{eq:power-1a}. 

 It remains to prove \eqref{eq:power-1b}. Notice that $\bar{\dot{e}}^{(\tau)}_{i,j}$ is defined similarly as $\bar{e}_{i,j}^{(\tau)}$. Then $\bar{\dot{e}}^{(\tau)}_{i,j}$ is an i.i.d. copy of $\bar{e}_{i,j}^{(\tau)}$ and $$\dot{{ e}}_{i,j}^{(\tau)} = \frac{\tau\delta_{i,j} }{\sqrt{2\tau\rho_n}} + \bar{\dot{e}}_{i,j}^{(\tau)}.$$ 
 By the joint independence of $\{{ e}^{(\tau)}_{i,j} \dot{{ e}}^{(\tau)}_{i,j}, 1 \le i < j \le n\}$, for any $c_{2} > 0$, we have that for large enough $p_n$,
\begin{equation}
\label{equ:var_upper}
    \begin{aligned}
    \Var\left( B_{0}^{(\tau)}\right)=& \sum_{i < j} \Var\left( { e}^{(\tau)}_{i,j} \dot{{ e}}^{(\tau)}_{i,j}\right)\\
    =& \sum_{i <j} \left( \E({ e}^{(\tau)}_{i,j})^2  \E(\dot{{ e}}^{(\tau)}_{i,j})^2 - \left(\E({ e}^{(\tau)}_{i,j}) \E(\dot{{ e}}^{(\tau)}_{i,j}) \right)^2\right) \\
    =& \sum_{i <j}  \left( \frac{\tau \delta_{i,j}^2}{2 \rho_n} + \E (\bar{e}_{i,j}^{(\tau)})^2\right) \left( \frac{\tau \delta_{i,j}^2}{2 \rho_n} + \E (\bar{\dot{e}}_{i,j}^{(\tau)})^2\right) - \sum_{i <j} \left(\frac{\tau \delta_{i,j}^2}{2\rho_n}\right)^2  \\
    =&   2 \sum_{i < j} \frac{\tau \delta_{i,j}^2}{2\rho_n}\E (\bar{e}_{i,j}^{(\tau)})^2 + \sum_{i < j} \E^2 (\bar{e}_{i,j}^{(\tau)})^2\\
    \le& 2 \sum_{i < j} \frac{\tau \delta_{i,j}^2}{2\rho_n} + p_n =2\E(B_{0}^{(\tau)}) + p_n \le 2c_{\eta}^{-2} \left( \E(B_{0}^{(\tau)})\right)^2,
\end{aligned}
\end{equation}
where that last step above is due to \eqref{equ:expect}. Since $c_{\eta}^2 \ge 8/(1-\eta)$, it holds that 
$$
\frac{4\Var(B_{a_n}^{(\tau)})}{\E^2(B_{a_n}^{(\tau)} )}\leq \frac{8}{c_{\eta}^2} \leq 1-\eta.
$$
This establishes \eqref{eq:power-1b} and thus completes the proof for the power analysis.


\medskip

\textbf{Part II: Sparse case with $m^* < \sqrt{p_n}$}. We choose the threshold $$r_n = 3c_{\alpha} m^*\log(\frac{e p_n}{(m^*)^2})$$ and define $$a^2_n :=  3\log(\frac{e p_n}{(m^*)^2 }),$$ where $c_{\alpha} \ge \sqrt{2 (\alpha e) ^{-1}\log_2\{(2h)^{-1}\}}$. It holds that $$r_n =c_{\alpha} m^* a^2_n\ge c_{\alpha} m^*$$ as $a^2_n  \ge 3\log(e) >1$. Observe that $(m^*)^2 =  e^{1-a^2_n/3}p_n$. We also set the lower bound $$\epsilon^2_n = (\bar{c}^\prime)^2 m^* \rho_n a_n^2 \ge 2cm^* \rho_n a_n^2 $$ for $c \ge 64+ c_{\eta}$, where $c_{\eta} \ge 16 c_{\alpha} \vee \sqrt{260(1-\eta)^{-1}}$. The time span condition of $T \ge 6 h^{-1}\rho_n^{-1} \log(\frac{e p_n}{(m^*)^2}) = 2 h^{-1}\rho_n^{-1} a_n^2$ in \eqref{equ:time_span_con1} entails that $$\tau \rho_n \ge h T \rho_n  \ge 2 a_n^2$$ for all $\tau \in \calT$. Then an application of Lemma \ref{lemma:june:inequality} gives that under $H_{0}(\rho_n)$, 
\begin{align*}
     \Pr_0\{\psi_n  = 1\} &= \Pr_0\{\max_{\tau \in \calT} B_{a_n}^{(\tau)} \ge r_n\} \le \sum_{\tau \in \calT}  \Pr_0\{ B_{a_n}^{(\tau)} \ge c_{\alpha}m^* \} \\
     & \le \sum_{\tau \in \calT}  \Pr_0\left\{ B_{a_n}^{(\tau)} \ge \sqrt{ p_n e^{-a^2_n/3}} \cdot \sqrt{e}c_{\alpha}\right\}  \leq \frac{2|\calT|}{c_{\alpha}^2 e }  \\
     & \leq \frac{2\log_2\{(2h)^{-1}\}}{c_{\alpha}^2 e } \le \alpha,
\end{align*}
since $c_{\alpha}^2 \ge 2\log_2\{(2h)^{-1}\}/(\alpha e)$.

We next show that the test statistic $\psi_n$ has power $\eta$ under alternative $H_1(\rho_n, m^*, \bar{c}^\prime\epsilon_n)$.
Similar to \textit{Part I}, we need only to establish \eqref{eq:power-1a} and \eqref{eq:power-1b}. We will divide the signal coordinate set $\calG := \{1 \le i < j \le n: \delta_{i,j} \ne 0 \}$ into a strong signal set $$\calG_1:= \left\{(i,j) \in \calG: |\delta_{i,j}| \ge \sqrt{\frac{64 \rho_n}{\tau^*}} a_n\right\}$$ and a weak signal set $$\calG_2 := \left\{(i,j) \in \calG: |\delta_{i,j}| < \sqrt{\frac{64  \rho_n}{\tau^*}} a_n\right\}.$$ Then it holds that 
$$\calG = \calG_1 \cup \calG_2.$$ 
In light of the condition $\tau^*(T-\tau^*)T^{-1}\|\vech(\bDelta)\|_2^2\geq (\bar{c}^\prime)^2\epsilon_n^2$, we have that 
\begin{equation}\label{eq:june:004}
    \tau^* \sum_{(i,j) \in \calG} \delta^2_{i,j} \geq \frac{1}{2}(\bar{c}^\prime)^2 \epsilon_n^2 = c m^* \rho_n a^2_n.
\end{equation} 
We can deduce that 
\begin{equation}\label{eq:june:002}
\begin{aligned}
   \sum_{(i,j) \in \calG_1} \delta^2_{i,j} & =  \sum_{(i,j) \in \calG} \delta^2_{i,j}  - \sum_{(i,j) \in \calG_2} \delta^2_{i,j}  \\
   & \ge c\frac{m^* \rho_n a^2_n}{\tau^*} -  \frac{64 |\calG_2|\rho_n a^2_n}{\tau^*}  \\
   & \ge c_{\eta}\frac{m^* \rho_n a^2_n}{\tau^*}  
   \end{aligned}
\end{equation}
as $c \ge 64 + c_{\eta}$.
Consequently, by Chebyshev's inequality, we can obtain that for each $(i,j) \in \calG_1$ with $\delta_{i,j} > 0$,
$$
\begin{aligned}
	\Pr\left\{ |\ddot{{ e}}^{(\tau)}_{i,j}| \ge a_n \right\} & \ge \Pr\left\{   \bar{e}_{i,j}^{(\tau)} \ge -\sqrt{{\frac{\tau}{{2 \rho_n}}}} \delta_{i,j} + a_n\right\}
	\ge \Pr\left\{   -\bar{e}_{i,j}^{(\tau)} \le 3a_n\right\} \\
    & \ge 1 - \frac{\Var(\bar{e}_{i,j}^{(\tau^*)})}{9a^2_n}
	\ge \frac{1}{2}, 
\end{aligned}
$$
where the second inequality above is due to $\delta_{i,j} \ge \sqrt{\frac{64\rho_n}{\tau}} a_n \ge \sqrt{\frac{32\rho_n}{\tau^*}} a_n$ and the last inequality above is from $\Var(\bar{e}_{i,j}^{(\tau)} ) \le 1$ and $a^2_n \ge 1$. 

Similar lower bound can be obtained for $(i,j) \in \calG_1$ with $\delta_{i,j} < 0$. Meanwhile, from \eqref{app:equ_z} it holds that 
$$\E \left({ e}_{i,j}^{(\tau)}\dot{{ e}}_{i,j}^{(\tau)} \right)= {\tau \delta_{i,j}^2 }/{(2\rho_n)} \geq {\tau^* \delta_{i,j}^2 }/{(4\rho_n)}  > 0$$ for all $(i,j) \in \calG$. Combining the previous 
results and noting that $\tau > \tau^*/2$ yield that 
\begin{equation}\label{eq:june:003}
   \begin{aligned}
   \E(B_{a_n}^{(\tau)}) =& \sum_{i < j} \E \left( { e}_{i,j}^{(\tau^*)}\dot{{ e}}_{i,j}^{(\tau)}  \right) \Pr\left\{ |\ddot{{ e}}^{(\tau)}_{i,j}| \ge a_n \right\} \\
   \ge& \sum_{(i,j) \in \calG_1} \E \left( { e}_{i,j}^{(\tau)}\dot{{ e}}_{i,j}^{(\tau)}  \right) \Pr\left\{ |\ddot{{ e}}^{(\tau^*)}_{i,j}| \ge a_n \right\} \\
   \ge& \sum_{(i,j) \in \calG_1} \frac{\tau^* \delta_{i,j}^2}{8\rho_n} \ge \frac{c_{\eta} m^* a_n^2}{8},
\end{aligned} 
\end{equation}
where the last step above has used \eqref{eq:june:002}. Hence, it follows that 
$$r_n = c_{\alpha} m^*a_n^2 \le 8c_{\alpha} c_{\eta}^{-1}\E(B_{a_n}^{(\tau)}) \le 2^{-1}\E(B_{a_n}^{(\tau)})$$ since $c_{\eta} \ge 16c_{\alpha}$. This establishes \eqref{eq:power-1a}. 

Finally, it remains to prove \eqref{eq:power-1b}. Note that
$$
\begin{aligned}
	\Var(B_{a_n}^{(\tau)}) & = \sum_{(i,j) \in \calG} \Var\left( { e}^{(\tau)}_{i,j} \dot{{ e}}^{(\tau)}_{i,j}  \I_{\{|\ddot{{ e}}^{(\tau)}_{i,j}| \ge a_n\}}\right) + \sum_{(i,j) \in \calG^c} \Var\left({ e}^{(\tau)}_{i,j} \dot{{ e}}^{(\tau)}_{i,j}  \I_{\{|\ddot{{ e}}^{(\tau)}_{i,j}| \ge a_n\}}\right) \\
    & := v_1 + v_2.
\end{aligned}
$$
Since all ${ e}^{(\tau)}_{i,j}$ and $ \dot{{ e}}^{(\tau)}_{i,j}$ with $(i,j) \in \calG^c$ have mean zero, using similar arguments as for \eqref{equ:var_bound} and noting that $a^2_n =3  \log(ep_n/(m^*)^2)$ and $m^* \le (8\E(B_{a_n}^{(\tau)}))/c_{\eta}$ (i.e., \eqref{eq:june:003}), we can obtain that 
$$
v_2 \le 2p_n e^{-a_n^2/3} \le 2e^{-1}(m^*)^2 \le\frac{64}{c_{\eta}^2} \left(\E(B_{a_n}^{(\tau)}) \right)^2.
$$

We now consider term $v_1$ above. Similar to \eqref{eq:june:003}, using the fact that $\tau>\tau^*/2$ and \eqref{eq:june:004}, we can show that 
$$
\begin{aligned}
\E(B_{a_n}^{(\tau)}) & \geq \frac{1}{2}\sum_{(i,j)\in \calG}\tau^*\delta_{i,j}^2/\rho_n \geq \epsilon_n^2/(4\rho_n) \\
& > 4^{-1}c m^*a_n^2 \geq 4^{-1}cm^*.
\end{aligned}
$$ 
An application of similar variance calculation as in \eqref{equ:var_upper} yields that for sufficient large $n$, 
$$
\begin{aligned}
v_1 & \le 2 \sum_{(i,j)\in \calG} \frac{\tau \delta_{i,j}^2}{2\rho_n} + m^* \le (4+\frac{4}{c})\E(B_{a_n}^{(\tau)}) \\
& \le \frac{1}{c_{\eta}^2} \left(\E(B_{a}^{(\tau)})\right)^2,
\end{aligned}
$$
where the last step above is because  $\left(\E(A_{a}^{(\tau)})\right)^2\gg \E(A_{a_n}^{(\tau)}) \rightarrow \infty$ (see \eqref{eq:june:003}). 
Thus, it follows that 
$$\Var(B_{a_n}^{(\tau)}) \le 65\left(\E(B_{a_n}^{(\tau)})\right)^2/c_{\eta}^2$$ and $$
\frac{4\Var(B_{a_n}^{(\tau)})}{\E^2(B_{a_n}^{(\tau)} )} \le \frac{260}{c_{\eta}^2} \le 1-\eta,
$$
where the last inequality above is due to $c_{\eta}^2 \ge 260/(1-\eta)$. This establishes \eqref{eq:power-1b} and concludes the proof under the sparse alternative, which completes the proof of Theorem \ref{thm:minimax_upper}.

\subsection{Lemma \ref{lemma:june:inequality} and its proof} \label{new.sec.C.3}

Denote by $\bW^{(t)} = \bX^{(t)} - \bTheta^{(t)} := (w_{i,j}^{(t)})_{1 \le i,j \le n}$ the 
noise matrix of $\bX^{(t)}$. 
\begin{lemma}
	\label{lemma:june:inequality}
  Under $H_{0}(\rho_n)$, if $\tau \rho_n \ge 2c^{-1}a^2$ for some $c>0$, we have that for any set $\mathcal{R} \subset \{(i,j): 1 \le i < j \le n\}$ and $x > 0$, 
	 $$
	\Pr\left\{ \sum_{(i,j) \in \mathcal{R}} { e}_{i,j}^{(\tau)}\dot{{ e}}_{i,j}^{(\tau)}  \I_{\{|\ddot{{ e}}^{(\tau)}_{i,j}| \ge a\}} \geq \sqrt{|\mathcal{R}| e^{- a^2/(2+c)}} x \right\} \leq \frac{2}{x^2}.
	$$
For the specific case of $a=0$, the above inequality holds for all $\tau \in (0, T/2]$. 
\end{lemma}

\noindent\textit{Proof}. Since $\E ({ e}_{i,j}^{(\tau)}) =0$, $\E({ e}_{i,j}^{(\tau)})^2 \le 1$, and $\dot{{ e}}_{i,j}^{(\tau)}$ and $\ddot{{ e}}_{i,j}^{(\tau)}$ are independent copies of ${ e}_{i,j}^{(\tau)}$, it holds that 
	\begin{equation}\label{eq:001}
        \begin{aligned}
            \Var\left(\sum_{(i,j) \in \mathcal{R}} { e}_{i,j}^{(\tau)}\dot{{ e}}_{i,j}^{(\tau)}  \I_{\{|\ddot{{ e}}^{(\tau)}_{i,j}| \ge a\}} \right) =& \sum_{(i,j) \in \mathcal{R}} \Var\left({ e}_{i,j}^{(\tau)}\dot{{ e}}_{i,j}^{(\tau)}  \I_{\{|\ddot{{ e}}^{(\tau)}_{i,j}| \ge a\}}\right)\\
            =&\sum_{(i,j) \in \mathcal{R}} \E\left({ e}_{i,j}^{(\tau)}\dot{{ e}}_{i,j}^{(\tau)}\right)^2 \Pr\left\{|\ddot{{ e}}^{(\tau)}_{i,j}| \ge {a} \right\}\\
            \le&\sum_{(i,j) \in \mathcal{R}} \Pr\left\{|\ddot{{ e}}^{(\tau)}_{i,j}| \ge {a} \right\}.
        \end{aligned}	    
	\end{equation}
Moreover, in view of $\E(w_{i,j}^{(t)}/\sqrt{\rho_n})^2 \le 1$ and $|w_{i,j}^{(t)}/\sqrt{\rho_n}| \le 3/\sqrt{\rho_n}$, an application of Bernstein's inequality shows that for an arbitrarily small constant $\epsilon>0$ and any  $0\leq a^2\le 2^{-1}c\tau\rho_n$,
\begin{equation}
\label{equ:berns}
    \begin{aligned}
    \Pr\left\{|\ddot{{ e}}^{(\tau)}_{i,j}| \ge a \right\} =& \Pr\left\{ \left| \frac{\sum_{t=1}^{\tau} ({w}^{(t)}_{i,j} - {w}^{(T-t+1)}_{i,j})}{\sqrt{\rho_n}}  \right|  \ge \sqrt{2\tau}a \right\}\\
    \le& 2\exp\left\{-\frac{ a^2}{2+ (\sqrt{2}a)/\sqrt{\tau\rho_n}} \right\} \leq 2\exp\left\{  -\frac{a^2}{2+c}\right\}.
\end{aligned}
\end{equation}

Then plugging \eqref{equ:berns} into \eqref{eq:001} yields that 
	\begin{equation}
		\label{equ:var_bound}
		\Var\left( \sum_{(i,j) \in \mathcal{R}} { e}^{(\tau)}_{i,j} \dot{{ e}}^{(\tau)}_{i,j}\I_{\{|\ddot{{ e}}^{(\tau)}_{i,j}| \ge a\}} \right)  \leq  2|\mathcal{R}|e^{- a^2/(2+c)}.
	\end{equation}
Therefore, it follows from (\ref{equ:var_bound}) and Chebyshev's inequality that for any $x > 0$,
	$$
	\Pr\left\{  \sum_{(i,j) \in \mathcal{R}} { e}_{i,j}^{(\tau)} \dot{{ e}}_{i,j}^{(\tau)} \I_{\{|\ddot{{ e}}^{(\tau)}_{i,j}| \ge a\}}  \geq  \sqrt{|\mathcal{R}| e^{- a^2/(2+c)} } x\right\} \leq \frac{2}{x^2}.
	$$
This concludes the proof of Lemma \ref{lemma:june:inequality}.

\section{Asymptotic expansions of empirical spiked eigenvalues and eigenvectors, and additional technical results} \label{new.sec.D}

This section provides the asymptotic expansions of the empirical spiked eigenvalues and eigenvectors, based on the recent developments in \cite{fan2022asymptotic}, \cite{fan2022simple}, and \cite{han2019universal}, as well as some additional technical results. Assume that we observe $T$ stationary adjacency matrices 
\begin{equation} \label{new.eq.FL009}
\bX^{(t)} =\bTheta + \bW^{(t)}, \ t = 1, \dots, T,
\end{equation}
where $\bTheta = (\theta_{i,j})_{1 \le i,j \le n} = \bV \bD \bV^\top$ is a deterministic low-rank latent mean matrix, $\bV = (\bv_1, \ldots, \bv_{K^*})$ is an $n \times K^*$ orthonormal matrix of population eigenvectors $\bv_k$'s with $\bV^\top \bV = \bI_{K^*}$, and $\bD = \diag\left(d_1,\ldots, d_{K^*} \right)$ is a diagonal matrix of population eigenvalues $d_k$'s with $|d_1| \ge \ldots \ge |d_{K^*}| > 0$. The matrices $\bW^{(t)} = (w^{(t)}_{i,j})_{1 \le i,j\le n}$ with $ t \in [T]$ are symmetric random noise matrices that are independent and identically distributed across $t$, with zero mean $\E(w_{i,j}^{(t)}) = 0$, variance profile $\sigma_{i,j} = \E(w_{i,j}^{(t)})^2$ for $ t \in [T]$, and bounded entries $\max_{t \in [T], \, 1 \le i,j \le n} |w_{i,j}^{(t)}| \le 1$. Since there are $T$ independent adjacency matrices $\bX^{(t)}$'s given in (\ref{new.eq.FL009}), we can take the average over the time span to reduce the variance
\begin{equation}\label{gen:x}
\begin{aligned}
	\bar{\bX} & = \frac{\sum_{t=1}^{T} \bX^{(t)}}{T} =    \bTheta  + \frac{\sum_{t=1}^{T} \bW^{(t)}}{T} \\
    & := \bTheta+ \bar{\bW}.
    \end{aligned}
\end{equation}
We then perform a spectral decomposition on $\bar{\bX}$ in (\ref{gen:x}) to obtain the empirical eigenvalues $\hd_1,\ldots, \hd_n$ with $|\hd_1| \geq \ldots |\hd_n| \geq 0$ and the corresponding eigenvectors $\hbv_1, \ldots, \hbv_n$.  The mean matrix $\bTheta = (\theta_{i,j})_{1 \le i,j \le n}$ in (\ref{new.eq.FL009}) and (\ref{gen:x}) is estimated as 
\begin{equation} \label{new.eq.FL010}
\widehat{\bTheta} = (\htheta_{i,j})_{1 \le i,j \le n} 
= \sum_{k=1}^K \hd_k \hbv_k \hbv_k^\top,
\end{equation}
where $K \geq 1$ is some given integer. We refer to $\hd_k$'s with  $1 \leq k \leq K$ as the empirical spiked eigenvalues, and $\hbv_k$'s with $1 \leq k \leq K$ as the empirical spiked eigenvectors.

More specifically, a major goal of this section is to characterize the asymptotic behavior of $\htheta_{i,j} - \theta_{i,j}$ as formalized in Lemma \ref{lem:gen1} later, based on the asymptotic expansions of the empirical spiked eigenvalues and eigenvectors. To this end, we will first derive some precise bounds on the estimation errors of the spiked eigenvalues and eigenvectors. To facilitate the technical analysis, we impose the following regularity conditions, most of which are directly taken from \cite{han2019universal}. Recall that $\rho_n = \max_{t \in [T], \, 1 \le i< j \le n} \theta_{i,j}^{(t)}$ and $q_n = \sqrt{n \rho_n}$. Denote by $\alpha_n^2 := \max_{i \in [n]} \sum_{j=1}^n \sigma_{i,j}^2$ and  $\bar{\alpha}_n^2 := \max_{i \in [n]} \sum_{j=1}^n \Var(\bar{w}_{i,j}) = \alpha_n^2/T$.

\begin{condition}\label{con:expansion}
For some integer $K$ with $1 \le K \le K^*$ and some positive constants {$\epsilon_0$}, $\epsilon_0^\prime$, and $c_0$, the following conditions are satisfied:
    \begin{enumerate}
        \item[1)] (Spiked eigenvalues) It holds that 
        $$|d_{K}| \gtrsim \max\left(\frac{q_n
        \log^{3+2\epsilon_0}(n)}{\sqrt{T}}, \frac{q_n^2}{T(\log\log(n))^{\epsilon_0^\prime}}\right)
        $$
        with $\sqrt{T} q_n \gtrsim \log^4(n)$ and $n \ge \log^8(n)$. Moreover, we have $\max_{k \ge K+1}|d_{k}| = o(\frac{q_n}{\sqrt{T}})$.
    
        \item[2)] (Eigengap) For $1 \le k \le K-1$, if $|d_k| \ne |d_{k+1}|$, it holds that $\frac{|d_{k}|}{|d_{k+1}|} \ge 1+c_0$,
        where we do not require eigengaps for smaller eigenvalues $|d_k|$ with $K+1 \le k \le K^*$. 
        
        \item[3)] (Eigenvectors) It holds that $\| \bV_{K^*}\|_{\infty} \lesssim \frac{1}{\sqrt{n}}$,  where $\| \bV\|_{\infty}:= \max_{1 \le i,j \le n} |v_{i,j}|$. 
    \end{enumerate}
\end{condition}


The first part of 1) and 2)--3) in Condition \ref{con:expansion} above are directly adapted from Conditions (1)--(3) in \cite{han2019universal}, which are standard for analyzing the asymptotic behavior of spiked eigenvalues and eigenvectors. Notably, we do \textit{not} require any eigen-structure for $k > K$, allowing for weak signals as in \cite{han2019universal} and \cite{fan2022simplerc}. The second part of 1) in Condition \ref{con:expansion} ensures that misspecification of the working rank $K$ does not distort the estimation asymptotically. Such condition is mild as $T$ typically scales with a power of $\log(n)$. 


Let $\be_i \in \R^{n}$ be the basis vector with the $i$th component being $1$ and all other components being zero, $v_k(i)$ and $\hv_k(i)$ the $i$th components of $\bv_k$ and $\hbv_k$, respectively, and $t_k \in \mathbb{R}$ the unique solution to equation (10) defined in \cite{fan2022asymptotic}. From Lemma 3 of \cite{fan2022asymptotic}, it holds that for $1 \leq k \leq K$, 
\begin{equation} \label{new.eq.FL011}
d_k/t_k \to 1
\end{equation}
as $n \to \infty$. For random (or deterministic) variables $\zeta_n(i)$ and $\xi_n(i)$ dependent on $1 \leq i \leq n$, we say that $\zeta_n = O_{p_u}(\xi_n)$ if for any positive constants $D$ and $\epsilon_1$, there exists some positive integer $n_0(D,\epsilon_1)$ depending only on $D$ and $\epsilon_1$ such that for all $n \ge n_0(D,\epsilon_1)$, 
\begin{equation} \label{new.eq.FL012}
\begin{aligned}
& \Pr\left\{
\text{there exists }
1 \le i \le n \text{ such that } |\xi_n(i)| > \log^{\epsilon_1}(n)|\zeta_n(i)| \right\} = { o(1)}.
\end{aligned}
\end{equation}
Essentially, the $O_{p_u}(\cdot)$ notation introduced above gives a uniform upper bound for all $ \xi_n(i), \, i \in [n]$, with high probability. Here, we put a constraint that $\epsilon_1 \le 1/3$.

The results in the lemma below are consequences of Lemma 4.3 of \cite{erdHos2013spectral} and (S.30) of \cite{han2019universal}. 

\begin{lemma} \label{add:lem:fan}
	Under Condition \ref{con:expansion}, we have that 
\begin{equation}\label{prop:d}
	\hd_k- d_k  =  { O_{p}\left(\frac{q_n}{\sqrt{T}}\right)}
\end{equation}
and 
\begin{equation}\label{prop:v}
\begin{aligned}
  \hv_k(i) - v_k(i) =& \frac{\be_i^\top \bar{\bW} \bv_k }{t_k} +  {O_{p_u}\left( \frac{q_n}{ \log(n)\sqrt{nT} |d_k|} \right)}
\end{aligned}
\end{equation}
with $1 \leq k \leq K$ and $1 \leq i \leq n$. 
\end{lemma}

\noindent\textit{Proof}. Let us first prove \eqref{prop:d}. 
 Denote by $\tilde{\bW} = (\tilde{w}_{i,j})_{1 \le i, j \le n} := \sqrt{T} \bar{\bW}/q_n$. We make a claim that
\begin{equation}\label{equ:july:08:01}
  \E (\tilde{w}_{i,j})^2 \lesssim \frac{1}{n} 
  \ \text{ and } \ \E|\tilde{w}_{i,j}|^k \lesssim \frac{1}{n \tilde{q}^{k-2}}
\end{equation}
with $k \ge 3$ for some $\tilde{q}$ satisfying that 
\begin{equation}\label{equ:july:08:02}
   \sqrt{n} \gtrsim \tilde{q} \ge \log^4(n).
\end{equation}
Then $\tilde{\bW}$ satisfies Definition 1 of \cite{erdHos2013spectral}. By resorting to Lemma 4.3 of \cite{erdHos2013spectral}, we have that 
$$
\|\tilde{\bW}\|_2 = O_p(1).
$$
This entails that 
$$
\| \bar{\bW}\|_2 = O_p\left(\frac{q_n}{\sqrt{T}}\right).
$$
By Weyl's inequality for eigenvalues (See equation (1.63) in \cite{tao2012topics}), we have
$$
|\hat d_k - d_k| \le \| \bar{\bW}\|_2 = O_p\left(\frac{q_n}{\sqrt{T}}\right).
$$
This proves \eqref{prop:d}.

It remains to establish \eqref{equ:july:08:01} and \eqref{equ:july:08:02} in the claim above. The second moment result in \eqref{equ:july:08:01} is trivial so we will only verify the $k$th absolute moment bound for $k > 3$. To see this, notice that 
$$
\begin{aligned}
 \E|\tilde{w}_{i,j}|^k =& \left(\frac{1}{\sqrt{T}q_n}\right)^k \E\left|\sum_{t=1}^T w^{(t)}_{i,j}\right|^k   \\
 \lesssim &\left(\frac{1}{\sqrt{T}q_n}\right)^k \left( T\E|w_{i,j}^{(1)}|^k + (T \E(w^{(1)}_{i,j})^2 )^{k/2}\right)\\
 \lesssim& n^{-1} \left(\frac{1}{\sqrt{Tq_n^2}}\right)^k T q_n^2 + \left(\frac{1}{\sqrt{Tn \rho_n}}\right)^k T^{k/2}  \rho_n^{k/2}\\
 =& n^{-1}  (\sqrt{Tq^2_n})^{-(k-2)} + n^{-1} (\sqrt{n})^{-(k-2)},
\end{aligned}
$$
where the second step above has utilized Lemma 19 of \cite{Fan2025asymptotic}, and the third step above is due to $\E|w_{i,j}^{(1)}|^k \lesssim \rho_n$. Let us separately consider two regimes. If $Tq_n^2 > n$, it holds that 
$$
\E|\tilde{w}_{i,j}|^k \lesssim  n^{-1} (\sqrt{n})^{-(k-2)}:= \frac{1}{n \tilde{q}^{(k-2)}}
$$
with $\tilde{q} := \sqrt{n}$. Such $\tilde{q}$ directly satisfies \eqref{equ:july:08:02} as $n \ge \log^8(n)$.  
On the other hand, if $Tq_n^2 \le n$, i.e, $T \rho_n = O(1)$, we have that 
$$
\E|\tilde{w}_{i,j}|^k \lesssim n^{-1}  (\sqrt{Tq^2_n})^{-(k-2)} := \frac{1}{n \tilde{q}^{k-2}}
$$
with $\tilde{q} := \sqrt{T q_n^2}$. Then it follows from conditions $q_n \gtrsim \log^4(n)/\sqrt{T}$ and $T \rho_n = O(1)$ that 
$$\sqrt{n} \gtrsim \tilde{q} \ge \log^4(n).$$

We now focus on showing \eqref{prop:v}. Similar to (S.30) in \cite{han2019universal} with $\bW$ replaced by $\bar{\bW}$ and $\alpha_n$ by $\bar{\alpha}_n$, it holds that 
$$
\begin{aligned}
 v_k(i) - \hat{v}_k(i) & = \frac{\be_i \bar{\bW} \bv_k }{t_k} + \frac{O_{p_u}(\bar{\alpha}^2_n)}{\sqrt{n}t_k^2} + \frac{O_{p_u}(\bar{\alpha}^2_n \log^2(n)) + O(\bar{\alpha}_n^2)}{\sqrt{n}t_k^2}  \\
& \quad+ O_{p_u}\left(\frac{\bar{\alpha}_n}{n |t_k|}\right)+ \frac{O_{p_u}(\bar{\alpha}_n^3 \log^3(n))}{\sqrt{n} |t_k|^3} \\
&= \frac{\be_i \bar{\bW} \bv_k }{t_k}  +O_{p_u}\left(\frac{(\bar{\alpha}_n \log(n))^2}{\sqrt{n} |d_k|^2} + \frac{q_n}{\sqrt{T}n |d_k|}\right)\\
&= \frac{\be_i \bar{\bW} \bv_k }{t_k}  + { O_{p_u}\left(\frac{\bar{\alpha}_n}{\sqrt{n} |d_k| \log(n)} + \frac{q_n}{\sqrt{T}n |d_k|}\right)}\\
&=\frac{\be_i \bar{\bW} \bv_k }{t_k}  + {O_{p_u}\left( \frac{q_n}{ \log(n)\sqrt{nT} |d_k|} \right)}, 
\end{aligned}
$$
where the second last step above has used part 1) of Condition \ref{con:expansion}. This establishes \eqref{prop:v} and completes the proof of Lemma \ref{add:lem:fan}. 


The lemma below provides a uniform entrywise asymptotic expansion of the estimated mean matrix 
$\widehat{\bTheta} = (\htheta_{i,j})_{1 \le i,j \le n}$ defined in (\ref{new.eq.FL010}). Denote by $\bb_i = \sum_{k=1}^K \frac{d_k}{t_k} \bv_k v_k(i)$ with $1 \leq i \leq n$.

\begin{lemma}\label{lem:gen1}
	Under Condition \ref{con:expansion}, we have that  
	\begin{equation}\label{Oct18:1}
    \begin{aligned}
      \htheta_{i,j}
      =&\theta_{i,j} +   \be_i^\top \bar{\bW}\bb_j + \be_j^\top \bar{\bW} \bb_i  + { O}\left(\frac{q_n}{n\sqrt{T}}\right)
    \end{aligned}
	\end{equation}
	uniformly in $1 \le i,j\le n$ with asymptotic probability one. 
\end{lemma}

\noindent\textit{Proof}. Let us define 
$$\Delta(d_k) := \hd_k- d_k \ \text{ and } \ \Delta(v_k(i)) := \hv_k(i) - v_k(i)$$
for $1 \leq k \leq K$ and $1 \leq i \leq n$. 
Then we can decompose the entrywise estimation error as
	\begin{equation} \label{gen:deltaz}
		\begin{aligned}
			& \htheta_{i,j} -\theta_{i,j} = \sum_{k=1}^{K}\left[ \hd_k \hat{v}_k(i) \hat{v}_k(j) - d_k v_k(i) v_k(j)\right] - \sum_{k = K+1}^{K^*}d_k v_k(i) v_k(j)\\
			&= \sum_{k=1}^{K}\Big\{ \left[d_k+\Delta(d_k)\right] \left[v_k(i) + \Delta\left(v_k(i)\right)\right] \left[v_k(j) +\Delta\left(v_k\left(j\right)\right)\right]- d_k v_k(i) v_k(j)\Big\}\ \\
            &\quad- \sum_{k = K+1}^{K^*}d_k v_k(i) v_k(j)\\
			&= \sum_{k=1}^{K}\Big\{d_k v_k(i) \Delta(v_k(j))  + d_k \Delta(v_k(i))v_k(j) + d_k\Delta(v_k(i))\Delta(v_k(j))\\
            &\quad+ \Delta(d_k) v_k(i) v_k(j)+ \Delta(d_k) v_k(i) \Delta(v_k(j)) + \Delta(d_k) \Delta(v_k(i)) v_k(j) \\
            &\quad+ \Delta(d_k) \Delta(v_k(i)) \Delta(v_k(j)) \Big\} - \sum_{k = K+1}^{K^*}d_k v_k(i) v_k(j)
		\end{aligned}
	\end{equation}
	for $1 \leq i, j \leq n$. We will bound each term on the right-hand side above with the aid of Lemma \ref{add:lem:fan}.

    For the first term on the right-hand side of (\ref{gen:deltaz}) above, in view of \eqref{prop:d} and part 3) of Condition \ref{con:expansion}, 
    we can deduce that 
    \begin{equation}\label{gen:delta:v}
		\begin{aligned}
			d_k  v_k(i)\Delta(v_k(j)) = &d_k v_k(i) \left( \frac{1}{t_k} \be_j^\top \bar{\bW}\bv_k + O_{p_u}\left(\frac{q_n}{\log(n) |d_k| \sqrt{T n}}\right) \right) \\
			= & \frac{d_k}{t_k}\be_j^\top \bar{\bW} \bv_k v_k(i) + O_{p_u}\left(\frac{q_n}{\log(n) n\sqrt{T}} \right).	
		\end{aligned}
	\end{equation} 
Similarly, it holds that 
\begin{equation}\label{gen:delta:v2}
d_k  \Delta(v_k(i))v_k(j) = \frac{d_k}{t_k}\be_i^\top \bar{\bW} \bv_k v_k(j) + O_{p_u}\left(\frac{q_n}{\log(n) n\sqrt{T}} \right).    
\end{equation}
Moreover, we can show that for any constant $D > 0$,
\begin{equation*}
\begin{aligned}
    &\Pr\left\{ \max_{i \in [n]}|\be_i^\top \bar{\bW} \bv_k| \gtrsim \frac{\alpha_n \log^{(1+\epsilon_0)}(n)}{\sqrt{Tn}} \right\} \\
    &\le \sum_{i \in [n]} \Pr\left\{|\be_i^\top \bar{\bW} \bv_k| \gtrsim \frac{\bar{\alpha}_n \log^{(1+\epsilon_0)}(n)}{\sqrt{n}} \right\}\\
    &\le n \exp\left\{ -\frac{1}{3}(\log(n))^{1+\epsilon_0} \right\} = o(1), 
\end{aligned}
\end{equation*}
where the 
last inequality above is from Theorem S.1 in \cite{han2019universal}. 

It follows that 
\begin{equation*}
\begin{aligned}
   \Delta(v_k(i)) & = O_{p_u}\left( \frac{\alpha_n \log^{(1+\epsilon_0)}(n)}{|t_k|\sqrt{Tn}} \right) + O_{p_u}\left( \frac{q_n}{ \log^{\epsilon_0}(n)\sqrt{nT} |d_k|} \right) \\
   &= O_{p_u} \left(\frac{q_n \log^{(1+\epsilon_0)}(n)}{|d_k|\sqrt{nT}} \right),
   \end{aligned}
\end{equation*}
where we have utilized $|d_k| \asymp |t_k|$, $\alpha_n \le q_n$, and $v_k(i) \lesssim 1/\sqrt{n}$. Combining this result with \eqref{prop:d} and \eqref{prop:v} leads to 
\begin{equation}\label{equ:june17:01}
\begin{aligned}
   & d_k \Delta(v_k(i)) \Delta(v_k(j)) = O_{p_u}\left(\frac{q_n^2\log^{(2+2\epsilon_0)}(n)}{|d_k| n T}\right) = O_{p_u}\left( \frac{q_n}{\log(n) n\sqrt{T}}\right), \\
   & \Delta(d_k) \Delta(v_k(i)) v_k(j) ={O_{p_u}\left(\frac{q^2_n \log^{(1+\epsilon_0)}(n)}{ |d_k|Tn } \right)} = O_{p_u}\left( \frac{q_n}{\log(n) n\sqrt{T}}\right),\\
\end{aligned}
\end{equation}
where the last step above is due to {$d_k \gtrsim  q_n\log^{(3+2\epsilon_0)}(n)/\sqrt{T}$} in Part 1) of Condition \ref{con:expansion}. 

Meanwhile, it holds that 
\begin{equation}\label{equ:june17:02}
    \begin{aligned}
        \Delta(d_k) v_k(i) v_k(j) & = { O_p\left( \frac{q_n}{n \sqrt{T}}\right),} \\
        \Delta(d_k) \Delta(v_k(i)) \Delta(v_k(j)) & = O_{p_u}\left( d_k \Delta(v_k(i)) \Delta(v_k(j)) \right)  \\
        & = O_{p_u}\left( \frac{q_n}{\log(n) n \sqrt{T}}\right).
    \end{aligned}
\end{equation}
Plugging \eqref{gen:delta:v}--\eqref{equ:june17:02} into \eqref{gen:deltaz}, and in view of the definition of $O_{p_u}(\cdot)$, 
it follows that 
$$
\begin{aligned}
\htheta_{i,j} - \theta_{i,j} & = \sum_{k=1}^K \left(\be_i^\top \bar{\bW} \bv_k \frac{d_k}{t_k} v_k(j) +\be_j^\top \bar{\bW} \bv_k \frac{d_k}{t_k} v_k(i)  \right) \\
&\quad+  O\left( \frac{q_n}{\log(n) n\sqrt{T}}\right) + { O\left( \frac{q_n}{n\sqrt{T}}\right) } - \sum_{k = K+1}^{K^*}d_k v_k(i) v_k(j)
\end{aligned}
$$
holds uniformly in $1 \le i,j\le n$ with probability $1 - o(1)$. Since part 1) of Condition \ref{con:expansion} ensures that $d_k v_k(i) v_k(j) = o(q_n/(n\sqrt{T}))$ for $K+1 \le k \le K^*$, we can obtain that 
$$
\begin{aligned}
 \htheta_{i,j} - \theta_{i,j} &= \sum_{k=1}^K \left(\be_i^\top \bar{\bW} \bv_k \frac{d_k}{t_k}v_k(j) +\be_j^\top \bar{\bW} \bv_k \frac{d_k}{t_k}v_k(i)  \right) \\
 &\quad+ O\left( \frac{q_n}{\log(n) n\sqrt{T}}\right) + { O}\left(\frac{q_n}{ n\sqrt{T}}\right)  \\
 &= \be_i^\top \bar{\bW} \bb_j +\be_j^\top \bar{\bW} \bb_i  + { O}\left(\frac{q_n}{ n\sqrt{T}}\right)
\end{aligned}
$$
holds uniformly in $1 \le i,j\le n$ with asymptotic probability one. This concludes the proof of Lemma \ref{lem:gen1}.

The uniform entrywise asymptotic expansion given in Lemma \ref{lem:gen1} above characterizes precisely the asymptotic behavior of $\htheta_{i,j}$, which can be further extended to analyze the asymptotic behavior of $z_{i,j}^{(\tau)}$. Recall that 
\begin{equation}\label{def:barbX}
    \bar{\bX}^{(\tau, 1)} = \frac{\sum_{t=1}^{\tau} \bX^{(t)}}{\tau} \ \text{ and } \ \bar{\bX}^{(\tau, 2)} = \frac{\sum_{t=1}^{\tau} \bX^{(T-t+1)}}{\tau}.
\end{equation}
Let us define
\begin{equation}\label{def:barbW_lr}
    \bar{\bW}^{(\tau, 1)} = \frac{\sum_{t=1}^{\tau} \bW^{(t)}}{\tau}, \ \bar{\bW}^{(\tau, 2)} = \frac{\sum_{t=1}^{\tau} \bW^{(T-t+1)}}{\tau},
\end{equation}
and 
\begin{equation}\label{def:tbW_lr}
  \tilde{\bW}^{(\tau, 1)} = \frac{\sqrt{\tau}}{q_n}\bar{\bW}^{(\tau, 1)}, \ \tilde{\bW}^{(\tau, 2)}= \frac{\sqrt{\tau}}{q_n} \bar{\bW}^{(\tau, 2)}.
\end{equation}
Denote by 
\begin{equation}\label{def:tbW}
    \tilde{\bW}^{(\tau)} := \tilde{\bW}^{(\tau, 1)} - \tilde{\bW}^{(\tau, 2)}.
\end{equation}

Based on the definitions in (\ref{def:barbX})--(\ref{def:tbW}) above, we can rewrite
\begin{equation} \label{new.eq.FL014}
z^{(\tau)}_{i,j} =\sqrt{\frac{\tau}{2\rho_n}}ED\left( \bar{\bX}^{(\tau,1)}\right)_{i,j}  - \sqrt{\frac{\tau}{2\rho_n}}ED\left( \bar{\bX}^{(\tau,2)}\right)_{i,j}
\end{equation}
defined in (\ref{low:zij}). If the data points within the time spans $\{1,\ldots, \tau\}$ and $\{T-\tau+1,\ldots, T\}$ are stationary with mean matrices $\bTheta_1$ and $\bTheta_2$, respectively, we can apply Lemma \ref{lem:gen1} to each term on the right-hand side of (\ref{new.eq.FL014}) above. This yields an asymptotic characterization of the accuracy of $z_{i,j}^{(\tau)}$ in approximating the entrywise signal $\delta_{i,j} = \theta_{1,i,j} - \theta_{2,i,j}$. Specifically, $\bTheta_l = \bV^{(l)} \bD^{(l)} (\bV^{(l)})^\top$ is a deterministic low-rank latent mean matrix, $\bV^{(l)} = (\bv^{(l)}_1, \ldots, \bv^{(l)}_{K^*})$ is an $n \times K_{l}^*$ orthonormal matrix of population eigenvectors $\bv^{(l)}_k$'s with $(\bV^{(l)})^\top \bV^{(l)} = \bI_{K_l^*}$, and $\bD^{(l)} = \diag\left(d^{(l)}_1,\ldots, d^{(l)}_{K_l^*} \right)$ is a diagonal matrix of population eigenvalues $d^{(l)}_k$'s with $|d^{(l)}_1| \ge \ldots \geq |d^{(l)}_{K_{l}^*}| > 0$ for $l =1,2$. 

\begin{lemma}\label{lem:F}
Assume that the eigenvalues and eigenvectors of $\bTheta_1$ and $\bTheta_2$ satisfy Condition \ref{con:expansion}, and time span $T$ satisfies $\tau \le \min(\tau^*,T-\tau^*) \le T/2$. Then we have that 
\begin{itemize}
    \item[1)] Under $H_0(\rho_n, K^*)$ with $\bTheta_1 = \bTheta_2$, 
    $$\Pr\left( F_0 \right) \ge 1-{ o(1)},$$
    where we define event 
 \begin{equation}\label{event:F}
 \begin{aligned}
    F_0 & := \left\{ z^{(\tau)}_{i,j} = \sqrt{\frac{n}{2}}\left(\be_i^\top \tilde{\bW}^{(\tau)} \bb_j + \be_j^\top \tilde{\bW}^{(\tau)}\bb_i \right) \right.\\
    &\quad\left.+ { O}\left(\frac{1}{\sqrt{n}}\right), \, 1 \le i, j \le n, \tau \in \calT \right\}	
    \end{aligned}
    \end{equation}    
    with $\bb_i = \sum_{k=1}^K \frac{d^{(1)}_k}{t^{(1)}_k} \bv^{(1)}_k v^{(1)}_k(i)$ (or equivalently, $\bb_i = \sum_{k=1}^K \frac{d^{(2)}_k}{t^{(2)}_k} \bv^{(2)}_k v^{(2)}_k(i)$).
    
    \item[2)] Under $H_1(\rho_n, K^*, s^*, \epsilon)$ with  $\bTheta_1 \ne \bTheta_2$, 
$$\Pr\left( F_1 \right) \ge 1-{ o(1)}, $$
    where we define event 
    \begin{equation}\label{event:F_1}
    \begin{aligned}
    F_1 & := \left\{ z^{(\tau)}_{i,j} = \sqrt{\frac{\tau}{2\rho_n}}\delta_{i,j} + \sqrt{\frac{n}{2}}\sum_{l=1}^2(-1)^{l+1}\left(\be_i^\top \tilde{\bW}^{(\tau,l)} \bb^{(l)}_j + \be_j^\top \tilde{\bW}^{(\tau,l)}\bb^{(l)}_i \right)\right. \\
    &\quad \left.+ { O}\left(\frac{1}{\sqrt{n}}\right) , 1 \le i, j \le n \right\}	
    \end{aligned}
    \end{equation}
    with $\bb^{(l)}_i = \sum_{k=1}^K \frac{d^{(l)}_k}{t^{(l)}_k} \bv^{(l)}_k v^{(l)}_k(i)$ for $l =1,2$ and any $\tau$ with $\tau \asymp T$.
\end{itemize}
\end{lemma}


\noindent\textit{Proof}. Since \textit{Part 1)} is a direct consequence of \textit{Part 2)}, we prove only \textit{Part 2)} here. An application of Lemma \ref{lem:gen1} to $ED\left( \bar{\bX}^{(\tau,1)}\right)_{i,j}$ and $ED\left( \bar{\bX}^{(\tau,2)}\right)_{i,j}$ yields that 
     \begin{equation}\label{Oct18:2}
\begin{aligned}
z^{(\tau)}_{i,j} &= \sqrt{\frac{\tau}{2\rho_n}} \left( \theta_{1,i,j} +   \be_i^\top \bar{\bW}^{(\tau,1)}\bb^{(1)}_j + \be_j^\top \bar{\bW}^{(\tau,1)} \bb^{(1)}_i  + { O}\left(\frac{q_n}{n\sqrt{\tau}}\right) \right)\\
&\quad-  \sqrt{\frac{\tau}{2\rho_n}} \left( \theta_{2,i,j} +   \be_i^\top \bar{\bW}^{(\tau,2)}\bb^{(2)}_j + \be_j^\top \bar{\bW}^{(\tau,2)} \bb^{(2)}_i  + { O}\left(\frac{q_n}{n\sqrt{\tau}}\right) \right)\\
&=\sqrt{\frac{\tau}{2\rho_n}}\delta_{i,j} + \sqrt{\frac{n}{2}}\sum_{l=1}^2 (-1)^{l+1}\left(\be_i^\top \tilde{\bW}^{(\tau,l)} \bb^{(l)}_j + \be_j^\top \tilde{\bW}^{(\tau,l)}\bb^{(l)}_i \right) \\
&\quad+ { O}\left(\frac{1}{\sqrt{n}}\right)
\end{aligned}
\end{equation}
uniformly in $(i,j) \in \Omega$ with asymptotic probability one. This completes the proof of Lemma \ref{lem:F}.

Based on the facts given in Lemma \ref{lem:F} above, we will provide in the lemma below some concentration inequalities and moment results related to $z^{(\tau)}_{i,j}$. These results are extensively used in our technical analyses. For a subset $\calG \subset [n]$ with $|\calG| = s$, denote by $$\calS := \calG \times \calG.$$ Let $ \boldsymbol{\Pi} = \{\mu_{i,j}\}_{1 \le i, j \le n }$ be a symmetric matrix with $ \mu_{i,j} = \mu_{j,i}$ and zero diagonal entries $\mu_{i,i} = 0, \, i \in [n]$. Recall the definitions of events $F_0$ and $F_1$ given in (\ref{event:F}) and (\ref{event:F_1}), respectively. 

\begin{lemma}\label{lem:gen2}
For any $\epsilon > 0$ and some constant $c_0 > 0$, the following statements hold: 
	\begin{enumerate}
		\item[1)] If $\delta_{i,j} = 0$ (under null), for any $ \frac{1}{\sqrt{n}} { \ll }  d \ll \sqrt{\tau\rho_n}$, we have 
		\begin{equation*}
			{ \E\left( \I(|z_{i,j}^{(\tau)}| \ge  d) \I(F_0)\right)} = \Pr_0\left\{|z_{i,j}^{(\tau)}| \ge  d, { F_0} \right\}\le 4 \exp\left\{ - \frac{(1-\epsilon)^2 }{8c_0(1+\epsilon)} nd^2\right\}.
		\end{equation*}
		
		\item[2)] If $\delta_{i,j} = 0$ (under null) and $\frac{1}{q}\sqrt{\frac{\tau}{n}} =o(d)$, for any $ \frac{1}{\sqrt{n}} { \ll }   d \ll \sqrt{\tau\rho_n}$, we have 
	\begin{equation*}
		\Pr_0\left\{ |z_{i_1,j_1}^{(\tau)}| \ge  d, \,  |z_{i_2,j_2}^{(\tau)}| \ge d, { F_0} \right\}\le 16 \left( \exp\left\{ - \frac{(1-\epsilon)^2 }{8c_0(1+\epsilon)} n d^2\right\} \right)^2
	\end{equation*}
    for $i_1 \ne i_2$ and $j_1 \ne j_2$.

		\item[3)] 
        \begin{itemize}
            \item [3.1)] If $\delta_{i,j} = 0$ (under null), $s\ll n$, $d(s) \in [c_d\sqrt{1/n}a(s) ,c_d\sqrt{1/s}a(s)]$ for a constant $c_d \ge \sqrt{\frac{8c_0 (1+\epsilon)}{(1-\epsilon)^2}}$  with $a^2(s)\ge c_1 \log(en/s)$, and $T$ satisfies 
        $Ts \rho_n \gg a^2(s)$, we have
		$$
		\Var\left( \sum_{(i,j) \in \calS}  \mu_{i,j} \be_i^\top \tbW^{(\tau)} \bb_j \I\left(|\ddot{z}^{(\tau)}_{i,j}| > d(s)\right){ \I(F_0)}\right) = O\left(\frac{s^2}{n^3} \sum_{(i,j) \in \calS} \mu_{i,j}^2\right)
		$$
        and 
        $$
		\Var\left( \sum_{1 \le i< j\le n}  \mu_{i,j} \be_i^\top \tbW^{(\tau)} \bb_j \I\left(|\ddot{z}^{(\tau)}_{i,j}| > d(s)\right){ \I(F_0)}\right) = O\left(\frac{s}{n^2} \sum_{1 \le i< j\le n} \mu_{i,j}^2\right).
		$$
        \item[3.2)] If { (1) $s \sim n$ with $\delta_{i,j} = 0$ (under null)} or (2) $\delta_{i,j} \ne 0$ (under alternative), we have a conservative bound 
        $$
        \begin{aligned}
            &\Var\left( \sum_{(i,j) \in \calS}  \mu_{i,j} \be_i^\top \tbW^{(\tau,l)} \bb^{(l)}_j \I\left(|\ddot{z}^{(\tau)}_{i,j}| > d(s)\right){ \I(F_*)}\right)  = O\left(\frac{s}{n^2} \sum_{(i,j) \in \calS} \mu_{i,j}^2\right),\\
            &\Var\left( \sum_{(i,j) \in \calS}  \mu_{i,j} \be_i^\top \tbW^{(\tau,l)} \bb^{(l)}_j\right) = O\left(\frac{s}{n^2} \sum_{(i,j) \in \calS} \mu_{i,j}^2\right)
        \end{aligned}
        $$
        for $l=1,2$, and { $F_* = F_0$ under null and $F_* = F_1$ under alternative.}
        \end{itemize}

		\item[4)] 
        \begin{itemize}
            \item [4.1)]If $\delta_{i,j} = 0$ (under null), $s\ll n$, $d(s) \in [c_d \sqrt{1/n} a(s), c_d\sqrt{1/s}a(s)]$ for a constant $c_d \ge \sqrt{\frac{8c_0 (1+\epsilon)}{(1-\epsilon)^2}}$ with $a^2(s)\ge  \log(en/s)$, and $T$ satisfies 
        $Ts \rho_n \gg a^2(s)$, we have
		$$
		\Var\left\{\sum_{(i,j)\in \calS} \be_i^\top \tilde{\bW}^{(\tau)} \bb_j \be_i^\top \tilde{\dot{\bW}}^{(\tau)} \bb_j \I\left(|\ddot{z}_{i,j}^{(\tau)}| > d(s)\right){ \I(F_0)}\right\}  = O\left(\frac{s^4}{n^5}\right)
		$$
        and
        $$
		\Var\left\{\sum_{1 \le i< j\le n} \be_i^\top \tilde{\bW}^{(\tau)} \bb_j \be_i^\top \tilde{\dot{\bW}}^{(\tau)} \bb_j \I\left(|\ddot{z}_{i,j}^{(\tau)}| > d(s)\right){ \I(F_0)}\right\}  = O\left(\frac{s}{n^2}\right).
		$$
        \item[4.2)]If {(1) $s \sim n$ with $\delta_{i,j} = 0$ (under null)} or (2) $\delta_{i,j} \ne 0$ (under alternative), we have 
        $$
        \begin{aligned}
        &\Var\left( \sum_{(i,j)\in \calS} \be_i^\top \tilde{\bW}^{(\tau,l)} \bb^{(l)}_j \be_i^\top \tilde{\dot{\bW}}^{(\tau,l)} \bb^{(l)}_j\I\left(|\ddot{z}^{(\tau)}_{i,j}| > d(s)\right) { \I(F_*)} \right)= O\left(\frac{s^3}{n^4}\right),\\ 
          &\Var\left( \sum_{(i,j)\in \calS} \be_i^\top \tilde{\bW}^{(\tau,l)} \bb^{(l)}_j \be_i^\top \tilde{\dot{\bW}}^{(\tau,l)} \bb^{(l)}_j \right) = O\left(\frac{s^3}{n^4}\right)  
        \end{aligned}
        $$
        and
        $$
        \begin{aligned}
        & \Var\left( \sum_{1 \le i<j\le n} \be_i^\top \tilde{\bW}^{(\tau,l)} \bb^{(l)}_j \be_i^\top \tilde{\dot{\bW}}^{(\tau,l)} \bb^{(l)}_j\I\left(|\ddot{z}^{(\tau)}_{i,j}| > d(s)\right){ \I(F_*)} \right)= O\left(\frac{1}{n}\right),\\
          & \Var\left( \sum_{1 \le i<j\le n} \be_i^\top \tilde{\bW}^{(\tau,l)} \bb^{(l)}_j \be_i^\top \tilde{\dot{\bW}}^{(\tau,l)} \bb^{(l)}_j \right) = O\left(\frac{1}{n}\right)  
        \end{aligned}
        $$
        for $l =1,2$, and { $F_* = F_0$ under null and $F_* = F_1$ under alternative.}
        \end{itemize}
    \end{enumerate}
\end{lemma}

\noindent\textit{Proof}. 
\textbf{Part 1)}. Let us first show that 
	 for any $x > 0$ and some constant $c_0 > 0$, 
	\begin{equation}\label{add:lem:equ1}
	\Pr\left\{ \left|n\be_i^\top \tilde{\bW}^{(\tau)} \bb_j \right| \ge  x \right\} \le 2 \exp\left\{ - \frac{ x^2}{4c_0+ (2c_0 x)/(3\sqrt{\tau}q_n)}\right\}. 
	\end{equation}
	Indeed, writing $\tilde{w}^{(t)}_{i,j}$ as the $(i,j)$th entry of $\tilde{\bW}^{(t)}$ and invoking Bernstein's inequality, it holds that 
	\begin{equation*}
		\begin{aligned}
			\eqref{add:lem:equ1} = &	\Pr\left\{  \left| \sum_{l \in [n]} \sum_{t \in [\tau]}  \frac{n}{q_n \sqrt{\tau}} (w^{(t)}_{i,l} - w^{(T-t+1)}_{i,l}) b_j(l) \right| \ge x\right\}\\
			\le& 2\exp\left\{-\frac{ x^2/2}{c_0(2 + \frac{x}{3q_n\sqrt{\tau}})} \right\},
		\end{aligned}
	\end{equation*}
	where the inequality above is due to $\| \bb_j\|_{\infty} \lesssim 1/n$. 
    
    Then it follows from \eqref{add:lem:equ1} and $d  \gg 1/\sqrt{n}$ that for some small positive constant $\epsilon >0$, 
	\begin{equation*}
		\begin{aligned}
			\Pr_0\left\{ |{z}^{(\tau)}_{i,j}| > d, { F_0}\right\} 
            & \le \Pr\left\{ \sqrt{\frac{n}{2}}\left| \be_i^\top \tilde{\bW}^{(\tau)} \bb_j + \be_j^\top \tilde{\bW}^{(\tau)} \bb_i \right| \ge \left(1- \epsilon\right) d \right\}\\
			&\le \Pr\left\{ n\left|  \be_i^\top \tilde{\bW}^{(\tau)} \bb_j +\be_j^\top \tilde{\bW}^{(\tau)} \bb_i \right| \ge (1-\epsilon)\sqrt{2n} d \right\} \\
			&\le \Pr\left\{ \left|n \be_i^\top \tilde{\bW}^{(\tau)} \bb_j \right| \ge (1-\epsilon)   \sqrt{\frac{ n}{2}} d\right\}  \\
            &\quad+ \Pr\left\{ \left|n \be_j^\top \tilde{\bW}^{(\tau)} \bb_i \right| \ge (1-\epsilon)   \sqrt{\frac{ n}{2}} d \right\}  \\
			&\le 4 \exp\left\{ - \frac{(1-\epsilon)^2 n d^2/2}{4 c_0 + \frac{\sqrt{2} c_0 (1-\epsilon)\sqrt{n}d}{3\sqrt{\tau}q_n}} \right\} \le 4\exp\left\{ - \frac{(1-\epsilon)^2 n d^2}{8 c_0(1+\epsilon)} \right\} ,
		\end{aligned}
	\end{equation*}
where the last inequality above is from $\sqrt{ n} d/(\sqrt{\tau}q_n) =d/ \sqrt{\tau\rho_n}  = o(1)$.

\medskip

	\textbf{Part 2)}. Let us define two quantities $$X_{i_1,j_1}^{(\tau)} := \sqrt{n/2} \left( \be_{i_1}^\top \tilde{\bW}^{(\tau)} \bb_{j_1} +\be_{j_1}^\top \tilde{\bW}^{(\tau)} \bb_{i_1} \right)$$ and $$X_{i_2, j_2}^{(\tau)}:= \sqrt{n/2} \left( \be_{i_2}^\top \tilde{\bW}^{(\tau)} \bb_{j_2} +\be_{j_2}^\top \tilde{\bW}^{(\tau)} \bb_{i_2} \right).$$ 
    In view of $d \gg 1/\sqrt{n}$, it holds that 
	\begin{equation}\label{add:lem:equ2}
			\Pr_0\left\{ |z_{i_1,j_1}^{(\tau)}| \ge  d, |z_{i_2,j_2}^{(\tau)}| \ge  d, { F_0} \right\} \le \Pr\left\{ |X_{i_1, j_1}^{(\tau)}| \ge (1-\epsilon)d,  |X_{i_2, j_2}^{(\tau)}| \ge (1-\epsilon)d\right\}.
	\end{equation}
 Further denote by 
 $$X_{i_1,j_1}^{(\tau, -(i_2, j_2))} := \sqrt{n/2} \left( \be_{i_1}^\top \tilde{\bW}^{(\tau)} \bb_{j_1}^{-(i_2, j_2)} +\be_{j_1}^\top \tilde{\bW}^{(\tau)} \bb_{i_1}^{-(i_2, j_2)} \right)$$ and $$X_{i_2, j_2}^{(\tau,-(i_1, j_1))}:= \sqrt{n/2} \left( \be_{i_2}^\top \tilde{\bW}^{(\tau)} \bb_{j_2}^{-(i_1, j_1)} +\be_{j_2}^\top \tilde{\bW}^{(\tau)} \bb_{i_2}^{-(i_1, j_1)} \right),$$ 
 where $\bb_{j}^{-(k,l)}$ is obtained by setting the $k$th and $l$th components of $\bb_j$ as $0$. It is easy to see that $X_{i_1,j_1}^{(\tau, -(i_1, j_1))}$ and $X_{i_1,j_1}^{(\tau, -(i_2, j_2))}$ are 
 independent of each other. 
 
 Moreover, we can deduce that 
	\begin{equation*}
		\begin{aligned}
			|X_{i_1, j_1}^{(\tau)}| \le& |X_{i_1,j_1}^{(\tau, -(i_2, j_2))}| + \sqrt{n/2}\max_{j}\|\bb_j \|_{\infty} (\tilde{w}^{(\tau)}_{i_1,i_2} + \tilde{w}^{(\tau)}_{i_1,j_2})\\
            \le&	|X_{i_1,j_1}^{(\tau, -(i_2, j_2))}| + O\left(\frac{1}{q_n}\sqrt{\frac{\tau}{n}} \right)\\
			=& 	|X_{i_1,j_1}^{(\tau, -(i_2, j_2))}| +o\left(d\right),
		\end{aligned}
	\end{equation*}
	where the second inequality above is due to  
	 $|\tilde{w}^{(\tau)}_{i,j}| = \frac{\sqrt{2\tau}}{q_n} \bar{w}_{i,j}^{(\tau)} \le \frac{\sqrt{2\tau}}{q_n}$ since $ \bar{w}_{i,j}^{(\tau)} \le 1$ and $\|\bb_{j_1} \|_{\infty} \lesssim 1/n$, and the last step above is from the condition of $\frac{1}{q_n}\sqrt{\frac{2\tau}{n}} =o(d) $.
     Similarly, we can show that 
     $$|X_{i_2, j_2}^{(\tau)}| \le |X_{i_2,j_2}^{(\tau, -(i_1, j_1))}| +o(d).$$ 
     
     Combining the above results and \eqref{add:lem:equ2} leads to 
	\begin{equation}
		\begin{aligned}
		&\Pr_0\left\{ |z_{i_1,j_1}^{(\tau)}| \ge d, \, |z_{i_2,j_2}^{(\tau)}| \ge  d, { F_0}\right\}\\
		 \le& \Pr\left\{ |X_{i_1, j_1}^{(\tau,  -(i_2, j_2))}| \ge (1-\epsilon)d, \, |X_{i_2, j_2}^{(\tau,-(i_1, j_1))}| \ge (1-\epsilon)d\right\}\\
		=&\Pr\left\{ |X_{i_1, j_1}^{(\tau,  -(i_2, j_2))}| \ge (1-\epsilon)d \right\} \Pr\left\{ |X_{i_2, j_2}^{(\tau,-(i_1, j_1))}| \ge (1-\epsilon)d\right\}\\
		 \le& 16 \left( \exp\left\{ - \frac{(1-\epsilon)^2 n d^2}{8c_0(1+\epsilon)}\right\} \right)^2,
		\end{aligned}
	\end{equation}
	where the equality above is due to the independence of $X_{i_1, j_1}^{(\tau,  -(i_2, j_2))}$ and $X_{i_2, j_2}^{(\tau,  -(i_1, j_1))}$, and the last inequality above follows from similar arguments as for \eqref{add:lem:equ1}.

\medskip
    	
	\textbf{Part 3)}. We will first show the case of \textit{3.1)} $\delta_{i,j} = 0$. 
    In light of $s \ll n$ and $T s\rho_n \gg  a^2(s)$, it holds that 
	\begin{equation}\label{Oct21:1}
		\frac{1}{\sqrt{n}}  \ll d(s) \ll \sqrt{\rho_n\tau}.
	\end{equation}
	 We now focus on proving the first result and note that the variance can be written as
	$$
	\begin{aligned}
		&\Var\left\{\sum_{(i,j) \in \calS}  \mu_{i,j}\be_i^\top \tbW^{(\tau)} \bb_j \I\left(|\ddot{z}^{(\tau)}_{i,j}| > d(s)\right){ \I(F_0)}\right\} \\
		&\le  \sum_{i_1 \in \calG, j_1 \in \calG, j_2 \in \calG}   \Cov \left\{ \mu_{i_1, j_1} \be_{i_1}^\top \tbW^{(\tau)} \bb_{j_1}\I\left(|\ddot{z}_{i_1,j_1}^{(\tau)}| > d(s)\right){ \I(F_0)}, \right. \\
        &\quad \left. \mu_{i_1, j_2} \be_{i_1}^\top \tbW^{(\tau)} \bb_{j_2} \I\left(|\ddot{z}^{(\tau)}_{i_1,j_2}| > d(s)\right){ \I(F_0)}\right\}\\
		&\quad+\sum_{j_1 \in \calG, i_1 \in \calG, i_2 \in \calG} \Cov \left\{ \mu_{i_1, j_1} \be_{i_1}^\top\tbW^{(\tau)} \bb_{j_1}\I\left(|\ddot{z}_{i_1,j_1}^{(\tau)}| > d(s)\right){ \I(F_0)}, \right. \\
        &\quad \left. \mu_{i_2, j_1} \be_{i_2}^\top \tbW^{(\tau)} \bb_{j_1} \I\left(|\ddot{z}^{(\tau)}_{i_2,j_1}| > d(s)\right){ \I(F_0)}\right\}\\
		&\quad+ \sum_{(i_1,j_1)\in \calS} \sum_{(i_2,j_2) \in \calS; i_1 \ne i_2, j_1 \ne j_2} \Cov \left\{ \mu_{i_1, j_1} \be_{i_1}^\top \tbW^{(\tau)} \bb_{j_1}\I\left(|\ddot{z}_{i_1,j_1}^{(\tau)}| > d(s)\right){ \I(F_0)}, \right. \\
        &\quad \left. \mu_{i_2, j_2} \be_{i_2}^\top \tbW^{(\tau)} \bb_{j_2} \I\left(|\ddot{z}^{(\tau)}_{i_2,j_2}| > d(s)\right){ \I(F_0)}\right\}\\ 
        &:= v_1+v_2 + v_3.
	\end{aligned}
	$$
	We will bound terms $v_1$, $v_2$, and $v_3$ on the right-hand side above separately.
    
	We first analyze term $v_1$ above. Observe that 
	\begin{equation}\label{equ:iandj}
		\begin{aligned}
			& \Cov \left\{ \mu_{i_1, j_1}\be_{i_1}^\top \tbW^{(\tau)} \bb_{j_1} \I\left(|\ddot{z}^{(\tau)}_{i_1,j_1}| > d(s)\right){ \I(F_0)}, \, \mu_{i_1, j_2} \be_{i_1}^\top \tbW^{(\tau)} \bb_{j_2}\I\left(|\ddot{z}_{i_1,j_2}^{(\tau)}| > d(s)\right){ \I(F_0)}\right\}\\
			& \le |\mu_{i_1, j_1} \mu_{i_1, j_2} |\E\left( \be_{i_1}^\top \tbW^{(\tau)} \bb_{j_1}\be_{i_1}^\top \tbW^{(\tau)} \bb_{j_2}\right) \Pr_0\left\{ |\ddot{z}_{i_1,j_1}^{(\tau)}| > d(s), { F_0}\right\}.
		\end{aligned}
	\end{equation}
	We will bound the last two terms on the right-hand side of (\ref{equ:iandj}) above. Since \eqref{Oct21:1} entails that threshold $d(s)$ satisfies the conditions in \textit{Part 1)}, an application of \textit{Part 1)} yields that 
	\begin{equation}\label{add:lem1:equ3}
    \begin{aligned}
		\Pr_0\left\{ |\ddot{z}_{i_1,j_1}^{(\tau)}| > d(s),{ F_0}\right\}  \le & \Pr_0\left\{ |\ddot{z}_{i_1,j_1}^{(\tau)}| >c_d \frac{a(s)}{\sqrt{n}}, { F_0}\right\}\\ \le& 4 \exp\left\{ -\frac{(1-\epsilon)^2}{8c_0(1+\epsilon)} c^2_d a^2(s) \right\} 
         \le 4 \exp\left\{ - \log(en/s) \right\}  \lesssim \frac{s}{n}.
        \end{aligned}
	\end{equation}
Moreover, it follows from the independence between the row elements of $\tilde{\bW}^{(\tau)}$ that 
\begin{equation}\label{add:lem1:equ4}
	\begin{aligned}
	\E\left( \be_{i_1}^\top \tilde{\bW}^{(\tau)} \bb_{j_1}\be_{i_1}^\top \tilde{\bW}^{(\tau)} \bb_{j_2}\right) =& \E\left( \left(\sum_{l \in [n]}\tilde{w}^{(\tau)}_{i_1,l} b_{j_1}(l)  \right) \left(\sum_{l \in [n]}\tilde{w}^{(\tau)}_{i_1,l} b_{j_2}(l) \right)\right)\\
	=& \sum_{l \in [n]} \E\left( (\tilde{w}^{(\tau)}_{i_1,l})^2 b_{j_1}(l)b_{j_2}(l)\right) \lesssim \frac{1}{n^2},
		\end{aligned}
\end{equation}
where the last inequality above has used $\|\bb_j\|_{\infty} \lesssim 1/n$ and $\E(\tilde{w}^{(\tau)}_{i,j})^2 \lesssim 1/n$. 

Then combining \eqref{equ:iandj}, \eqref{add:lem1:equ3}, and \eqref{add:lem1:equ4} leads to 
\begin{equation*}
	\begin{aligned}
	& \Cov \left\{ \mu_{i_1, j_1}\be_{i_1}^\top \tilde{\bW}^{(\tau)} \bb_{j_1} \I(|\ddot{z}^{(\tau)}_{i_1,j_1}| > d(s)){ \I(F_0)}, \, \mu_{i_1, j_2} \be_{i_1}^\top \tilde{\bW}^{(\tau)} \bb_{j_2}\I(|\ddot{z}_{i_2,j_2}^{(\tau)}| > d(s)){ \I(F_0)}\right\}\\
	 & \lesssim |\mu_{i_1, j_1} \mu_{i_1, j_2} |\cdot \frac{1}{n^2} \cdot \frac{s}{n} = |\mu_{i_1, j_1} \mu_{i_1, j_2}|  \frac{s}{n^3}.
		\end{aligned}
\end{equation*}
Using this result and $|\calG| = s$, we can obtain that 
\begin{equation}\label{add:lem1:equv1}
	\begin{aligned}
		v_1 \lesssim & \frac{s}{n^3} \sum_{i_1 \in \calG} \sum_{j_1 \in \calG} \sum_{j_2 \in \calG} |\mu_{i_1, j_1} \mu_{i_1, j_2}|  \\
		\le & \frac{s}{2n^3} \sum_{i_1 \in \calG} \sum_{j_1 \in \calG} \sum_{j_2 \in \calG} (\mu^2_{i_1, j_1} + \mu^2_{i_1, j_2}) \\
        \le & \frac{ s}{2n^3} \sum_{i_1 \in \calG} s\left(\sum_{j_1 \in \calG} \mu_{i_1, j_1}^2 +  \sum_{j_2 \in \calG}  \mu_{i_1, j_2}^2\right) \\
		\lesssim& \frac{s^2}{n^3} \sum_{(i,j)\in\calS}\mu^2_{i,j},
	\end{aligned}
\end{equation}
where the second inequality above is due to the fact that $2|\mu_{i_1,j_1} \mu_{i_2, j_2}| \le \mu^2_{i_1,j_1} +\mu^2_{i_2, j_2}$.

	 We now turn to term $v_2$ above. For $i_1 \ne i_2$, it holds that 
	 \begin{equation}\label{equ:iorj}
	 	\begin{aligned}
	 		& \Cov \left\{ \mu_{i_1, j_1}\be_{i_1}^\top \tilde{\bW}^{(\tau)} \bb_{j_1} \I\left(|\ddot{z}^{(\tau)}_{i_1,j_1}| > d(s)\right){ \I(F_0)}, \, \mu_{i_2, j_1} \be_{i_2}^\top \tilde{\bW}^{(\tau)} \bb_{j_1}\I\left(|\ddot{z}_{i_2,j_1}^{(\tau)}| >d(s)\right){ \I(F_0)}\right\}\\
	 		& \le \mu_{i_1, j_1} \mu_{i_2, j_1} \E\left( \be_{i_1}^\top \tilde{\bW}^{(\tau)} \bb_{j_1}\be_{i_2}^\top \tilde{\bW}^{(\tau)} \bb_{j_1}\right) \Pr_0\left\{ |\ddot{z}_{i_1,j_1}^{(\tau)}| > d(s), { F_0}\right\}\\
	 		& \le  \frac{s}{n}\mu_{i_1, j_1} \mu_{i_2, j_1} \E\left( \be_{i_1}^\top \tilde{\bW}^{(\tau)} \bb_{j_1}\be_{i_2}^\top \tilde{\bW}^{(\tau)} \bb_{j_1}\right), 
	 	\end{aligned}
	 \end{equation}
	 where the last inequality above is from \eqref{add:lem1:equ3}. By the independence of $\tilde{w}^{(\tau)}_{i_1,j_1}$ and $\tilde{w}^{(\tau)}_{i_2,j_2}$ when $(i_1, j_1) \ne (j_2, i_2)$, we can deduce that 
	\begin{equation}\label{add:lem1:equv1:1}
		\begin{aligned}
			\E\left( \be_{i_1}^\top \tilde{\bW}^{(\tau)} \bb_{j_1}\be_{i_2}^\top \tilde{\bW}^{(\tau)} \bb_{j_1}\right) =& \E\left( \left(\sum_{l \in [n]}\tilde{w}^{(\tau)}_{i_1,l} b_{j_1}(l)  \right) \left(\sum_{l \in [n]}\tilde{w}^{(\tau)}_{i_2,l} b_{j_1}(l) \right)\right)\\
			=&\E\left(  (\tilde{w}^{(\tau)}_{i_1,i_2})^2 b_{j_1}(i_1)b_{j_1}(i_2)\right) \lesssim \frac{1}{n^3}.
		\end{aligned}
	\end{equation}
    Then an application of similar arguments as for \eqref{add:lem1:equv1} yields that 
	 \begin{equation}\label{add:lem1:equv2}
     \begin{aligned}
	 	v_2 & \lesssim  \frac{s}{n^4} \sum_{j_1 \in \calG, i_1 \in \calG, i_2 \in \calG} |\mu_{i_1, j_1} \mu_{i_2, j_1}| \lesssim  \frac{s^2}{n^4} \sum_{(i,j)\in \calS} \mu_{i,j}^2 \\
        & = o\left(\frac{ s^2 }{n^3} \sum_{(i,j)\in \calS}\mu^2_{i,j}\right).
        \end{aligned}
	 \end{equation}
 
	 It remains to examine term $v_3$ above. Since $i_1 \ne i_2$, the independence of $\tilde{w}^{(\tau)}_{i_1,j_1}$ and $\tilde{w}^{(\tau)}_{i_2,j_2}$ when $(i_1, j_1) \ne (j_2, i_2)$ entails that 
	\begin{equation}\label{equ:inej}
		\begin{aligned}
			& \Cov \left\{ \mu_{i_1, j_1}\be_{i_1}^\top \tilde{\bW}^{(\tau)} \bb_{j_1} \I\left(|\ddot{z}^{(\tau)}_{i_1,j_1}| > d(s)\right){ \I(F_0)}, \, \mu_{i_2, j_2} \be_{i_2}^\top \tilde{\bW}^{(\tau)} \bb_{j_2}\I\left(|\ddot{z}_{i_2,j_2}^{(\tau)}| >d(s)\right) { \I(F_0)}\right\}\\
			& \le |\mu_{i_1, j_1} \mu_{i_2, j_2}| \E\left( \be_{i_1}^\top \tilde{\bW}^{(\tau)} \bb_{j_1}\be_{i_2}^\top \tilde{\bW}^{(\tau)} \bb_{j_2}\right) \Pr_0\left\{ |\ddot{z}^{(\tau)}_{i_1,j_1}| > d(s), |\ddot{z}^{(\tau)}_{i_2,j_2}| > d(s), { F_0}\right\}\\
			& \le |\mu_{i_1, j_1} \mu_{i_2, j_2} |\frac{16s^2}{n^2} \E\left( (\tilde{W}^{(\tau)}_{i_1,i_2})^2 b_{j_2}(l)b_{j_2}(l)\right) \\
            & \le  \frac{16 s^2}{n^5} |\mu_{i_1, j_1} \mu_{i_2, j_2}|,
		\end{aligned}
	\end{equation}
	where the last inequality above is due to 
    $$
    \begin{aligned}
      \Pr_0\left\{ |\ddot{z}^{(\tau)}_{i_1,j_1}| > d(s), |\ddot{z}^{(\tau)}_{i_2,j_2}| > d(s), { F_0}\right\} \le& \Pr_0\left\{ |\ddot{z}^{(\tau)}_{i_1,j_1}| > c_d \frac{a(s)}{\sqrt{n}}, |\ddot{z}^{(\tau)}_{i_2,j_2}| >  c_d \frac{a(s)}{\sqrt{n}}, { F_0}\right\} \lesssim (\frac{s}{n})^2 
    \end{aligned}
    $$
    from \textit{Part 2)} and \eqref{Oct21:1}. This result leads to 
	\begin{equation}\label{add:lem1:equv3}
		\begin{aligned}
		v_3 \lesssim&  \frac{s^2}{n^5} \sum_{(i_1,j_1)\in \calS}  \sum_{(i_2,j_2) \in  \calS; i_1 \ne i_2, j_1 \ne j_2}| \mu_{i_1, j_1} \mu_{i_2, j_2}|\\	
        \lesssim&  \frac{s^2}{n^5} \sum_{(i_1,j_1)\in \calS}  \sum_{(i_2,j_2) \in  \calS}(\mu^2_{i_1, j_1} + \mu^2_{i_2, j_2})\\	
		\le&  \frac{16 s^2 }{n^5} \cdot s^2 \sum_{(i,j) \in \calS} \mu_{i,j}^2 = O\left(\frac{ s^2}{n^3} \sum_{(i,j)\in \calS}\mu^2_{i,j}\right).
		\end{aligned}
	\end{equation}
Hence, combining \eqref{add:lem1:equv1}, \eqref{add:lem1:equv2}, and \eqref{add:lem1:equv3} yields that 
	\begin{equation*}
		\Var\left\{\sum_{(i,j) \in \calG}  \delta_{i,j}\be_i^\top \tilde{\bW}^{(\tau)} \bb_j \I\left(|\ddot{z}^{(\tau)}_{i,j}| > d(s)\right) { \I(F_0)}\right\} = O\left( \frac{s^2 }{n^3} \sum_{(i,j) \in \calG \times \calG} \delta_{i,j}^2\right).
	\end{equation*}

The proof of the second result is similar. We can decompose the variance as
	$$
	\begin{aligned}
		& \Var\left\{\sum_{1 \le i < j \le n}  \mu_{i,j}\be_i^\top \tbW^{(\tau)} \bb_j \I\left(|\ddot{z}^{(\tau)}_{i,j}| > d(s)\right){ \I(F_0)}\right\} \\
		& \le  \sum_{i_1 \in [n], j_1 \in [n], j_2 \in [n]}   \Cov \left\{ \mu_{i_1, j_1} \be_{i_1}^\top \tbW^{(\tau)} \bb_{j_1}\I\left(|\ddot{z}_{i_1,j_1}^{(\tau)}| > d(s)\right){ \I(F_0)}, \right.\\
        &\quad \left. \mu_{i_1, j_2} \be_{i_1}^\top \tbW^{(\tau)} \bb_{j_2}\I\left(|\ddot{z}^{(\tau)}_{i_1,j_2}| > d(s)\right){ \I(F_0)}\right\}\\
	&\quad +\sum_{j_1 \in [n], i_1 \in [n], i_2 \in [n]} \Cov \left\{ \mu_{i_1, j_1} \be_{i_1}^\top\tbW^{(\tau)} \bb_{j_1}\I\left(|\ddot{z}_{i_1,j_1}^{(\tau)}| > d(s)\right){ \I(F_0)}, \right.\\
    &\quad \left. \mu_{i_2, j_1} \be_{i_2}^\top \tbW^{(\tau)} \bb_{j_1}\I\left(|\ddot{z}^{(\tau)}_{i_2,j_1}| > d(s)\right){ \I(F_0)}\right\}\\
		&\quad + \sum_{(i_1,j_1)\in [n] \times [n]} \sum_{(i_2,j_2) \in [n] \times [n]; i_1 \ne i_2, j_1 \ne j_2} \Cov \Big\{ \mu_{i_1, j_1} \be_{i_1}^\top \tbW^{(\tau)} \bb_{j_1}\I\left(|\ddot{z}_{i_1,j_1}^{(\tau)}| > d(s)\right){ \I(F_0)}, \\
        &\quad \mu_{i_2, j_2} \be_{i_2}^\top \tbW^{(\tau)} \bb_{j_2}\I\left(|\ddot{z}^{(\tau)}_{i_2,j_2}| > d(s)\right){ \I(F_0)}\Big\}\\ :=& r_1+r_2 + r_3.
	\end{aligned}
	$$
	We will bound terms $r_1$, $r_2$, and $r_3$ on the right-hand side above separately.  
    
    For term $r_1$ above, similar to \eqref{add:lem1:equv1}, it holds that 
    \begin{equation}\label{equ:june16:01}
	\begin{aligned}
		r_1 \lesssim & \frac{s}{n^3} \sum_{i_1 \in [n]} \sum_{j_1 \in [n]} \sum_{j_2 \in [n]} |\mu_{i_1, j_1} \mu_{i_1, j_2}|  \\
		\le & \frac{s}{2n^3} \sum_{i_1 \in [n]} \sum_{j_1 \in [n]} \sum_{j_2 \in [n]} (\mu^2_{i_1, j_1} + \mu^2_{i_1, j_2}) \\
        \le & \frac{ s}{2n^3} \sum_{i_1 \in [n]} n\left(\sum_{j_1 \in [n]} \mu_{i_1, j_1}^2 +  \sum_{j_2 \in [n]}  \mu_{i_1, j_2}^2\right) \\
		\le & \frac{s}{n^2} \sum_{(i,j)\in [n] \times [n]}\mu^2_{i,j} = \frac{2s}{n^2} \sum_{1 \le i < j \le n}\mu^2_{i,j},
	\end{aligned}
\end{equation}
   where the last equality above is due to the symmetry of matrix $\boldsymbol{\Pi}$. 
   
   For term $r_2$ above, an application of similar arguments as for \eqref{add:lem1:equv2} leads to 
    \begin{equation}\label{equ:june16:02}
    \begin{aligned}
	 	r_2 & \lesssim  \frac{s}{n^4} \sum_{j_1 \in [n], i_1 \in [n], i_2 \in [n]} |\mu_{i_1, j_1} \mu_{i_2, j_1}| \lesssim  \frac{s^2}{n^3} \sum_{1 \le i < j \le n} \mu_{i,j}^2 \\
        & = o\left(\frac{s}{n^2} \sum_{1 \le i < j \le n}\mu^2_{i,j}\right).
        \end{aligned}
	 \end{equation}
For term $r_3$ above, similar to \eqref{add:lem1:equv3}, we can deduce that 
   \begin{equation}\label{equ:june16:03}
		\begin{aligned}
		r_3 \lesssim&  \frac{s^2}{n^5} \sum_{(i_1,j_1)\in [n] \times [n]}  \sum_{(i_2,j_2) \in  [n] \times [n]; i_1 \ne i_2, j_1 \ne j_2}| \mu_{i_1, j_1} \mu_{i_2, j_2}|\\	
        \lesssim&  \frac{s^2}{n^5} \sum_{(i_1,j_1)\in [n] \times [n]}  \sum_{(i_2,j_2) \in  [n] \times [n]; i_1 \ne i_2, j_1 \ne j_2}(\mu^2_{i_1, j_1} + \mu^2_{i_2, j_2})\\	
		\le&  \frac{16 s^2 }{n^5} \cdot n^2 \sum_{(i,j) \in [n] \times [n]} \mu_{i,j}^2 = O\left(\frac{s}{n^2} \sum_{1 \le i<j \le n}\mu^2_{i,j}\right).
		\end{aligned}
	\end{equation}
Thus, combining \eqref{equ:june16:01}, \eqref{equ:june16:02}, and \eqref{equ:june16:03} gives the second result.

The proof for the case of {$s \sim n$ with $\delta_{i,j}=0$ (under null)} and $\delta_{i,j} \ne 0$ (under the alternative) is similar to that for the case of $\delta_{i,j} = 0$ (under the null) outlined above. The only difference is that we provide a conservative upper bound for the probabilities $\Pr_1\left\{ |\ddot{z}_{i_1,j_1}^{(\tau)}| > d(s), { F_*}\right\}$, which are upper bounded by $1$ in \eqref{equ:iandj}, \eqref{equ:iorj}, and \eqref{equ:inej}. Such adjustment introduces no additional scaling factor of $s/n$. Therefore, similar arguments as in the proof of the first result are applicable to establish the third result.

\medskip

\textbf{Part 4)}. We will first focus on the case of $\delta_{i,j} = 0$. It holds that
	$$
	\begin{aligned}
		& \Var\left\{ \sum_{(i,j) \in \calS} \be_i^\top \tilde{\bW}^{(\tau)} \bb_j\be_i^\top \tilde{\dot{\bW}}^{(\tau)} \bb_j \I\left(|\ddot{z}_{i,j}^{(\tau)}| > d(s)\right){ \I(F_0)}\right\}\\
		& \le \sum_{(i,j) \in \calS}  \Var\left(\be_i^\top \tilde{\bW}^{(\tau)} \bb_j\be_i^\top \tilde{\dot{\bW}}^{(\tau)} \bb_j \I\left(|\ddot{z}_{i,j}^{(\tau)}| > d(s)\right){ \I(F_0)}\right) \\
		&\quad+ \sum_{(i_1,j_1) \in \calS} \sum_{(i_1,j_2) \in \calS, j_1 \ne j_2}\Cov\Big(\be_{i_1}^\top \tilde{\bW}^{(\tau)} \bb_{j_1} \be_{i_1}^\top \tilde{\dot{\bW}}^{(\tau)} \bb_{j_1}\I\left(|\ddot{z}_{i_1,j_1}^{(\tau)}| > d(s)\right){ \I(F_0)},\\
		& \quad \be_{i_1}^\top \tilde{\bW}^{(\tau)} \bb_{j_2}\be_{i_1}^\top \tilde{\dot{\bW}}^{(\tau)} \bb_{j_2} \I\left(|\ddot{z}_{i_2,j_2}^{(\tau)}| > d(s)\right){ \I(F_0)}\Big) \\
        &\quad+ \sum_{(i_1,j_1) \in \calS} \sum_{(i_2,j_1) \in \calS, i_1 \ne i_2}\Cov\Big(\be_{i_1}^\top \tilde{\bW}^{(\tau)} \bb_{j_1} \be_{i_1}^\top \tilde{\dot{\bW}}^{(\tau)} \bb_{j_1}\I\left(|\ddot{z}_{i_1,j_1}^{(\tau)}| > d(s)\right){ \I(F_0)},\\
		& \quad \be_{i_2}^\top \tilde{\bW}^{(\tau)} \bb_{j_1}\be_{i_2}^\top \tilde{\dot{\bW}}^{(\tau)} \bb_{j_1} \I\left(|\ddot{z}_{i_2,j_1}^{(\tau)}| > d(s)\right){ \I(F_0)}\Big) \\
		&\quad+ \sum_{(i_1, j_1) \in \calS} \sum_{(i_2, j_2) \in \calS, i_1 \ne i_2, j_1 \ne  j_2}\Cov\Big(\be_{i_1}^\top \tilde{\bW}^{(\tau)} \bb_{j_1} \be_{i_1}^\top \tilde{\dot{\bW}}^{(\tau)} \bb_{j_1}\I\left(|\ddot{z}_{i_1,j_1}^{(\tau)}| > d(s)\right){ \I(F_0)},\\
		& \quad \be_{i_2}^\top \tilde{\bW}^{(\tau)} \bb_{j_2}\be_{i_2}^\top \tilde{\dot{\bW}}^{(\tau)} \bb_{j_2} \I\left(|\ddot{z}_{i_2,j_2}^{(\tau)}| > d(s)\right){ \I(F_0)}\Big) \\
		&:= e_{1} + e_{2} + e_3 + e_4.
	\end{aligned}
	$$
	We will bound each of the four terms on the right-hand side above separately.
    
        For term $e_{1}$ above, note that
	$$
	\begin{aligned}
		& \Var\left(\be_i^\top \tilde{\bW}^{(\tau)} \bb_j\be_i^\top \tilde{\dot{\bW}}^{(\tau)} \bb_j \I\left(|\ddot{z}^{(\tau)}_{i,j}| > d(s)\right) \I(F_0) \right) \\
        & = \E\left(\be_i^\top \tilde{\bW}^{(\tau)} \bb_{j}\right)^2 \E\left( \be_i^\top \tilde{\dot{\bW}}^{(\tau)} \bb_j  \right)^2 \Pr_0\left\{ |\ddot{z}^{(\tau)}_{i_1,j_1}| > d(s), { F_0} \right\}.
	\end{aligned}
	$$
	The expectations on the right-hand side above can be written as
	\begin{equation}\label{march:old:1}
    \begin{aligned}
	 \E\left( \be_i^\top \tilde{\bW}^{(\tau)} \bb_{j} \right)^2 & = \E\left( \sum_{l \in [n]}\tilde{w}_{i,l}^{(\tau)}b_j(l)\right)^2 = \sum_{l \in [n]} \E\left( \tilde{w}_{i,l}^{(\tau)}b_{j}(l)\right)^2 \\
     & \lesssim \frac{1}{n^2}.   
     \end{aligned}
	\end{equation}
	Then it follows from (\ref{march:old:1}) and \textit{Part 1)} that 
	\begin{equation}\label{Oct19:3}
    \begin{aligned}
	e_{1} & =\sum_{(i,j) \in \calS}\Var\left(\be_i^\top \tilde{\bW}^{(\tau)} \bb_j\be_i^\top \tilde{\dot{\bW}}^{(\tau)} \bb_j \I\left(|\ddot{z}^{(\tau)}_{i,j}| > d(s)\right) { \I(F_0)}\right) \\
    & \lesssim s^2 \cdot \frac{1}{n^2} \cdot \frac{1}{n^2} \cdot \frac{s}{n} =\frac{s^3}{n^5} = o\left(\frac{s^4}{n^5}\right).
    \end{aligned}
	\end{equation}
    
We now turn to term $e_2$ above.  The summand in $e_2$ can be bounded by
	$$
	\begin{aligned}
		&\Cov\left(\be_{i_1}^\top \tilde{\bW}^{(\tau)} \bb_{j_1}\be_{i_1}^\top \tilde{\dot{\bW}}^{(\tau)} \bb_{j_1}\I\left(|\ddot{z}^{(\tau)}_{i_1,j_1}| > d(s)\right){ \I(F_0)}, \, \be_{i_1}^\top \tilde{\bW}^{(\tau)} \bb_{j_2}\be_{i_1}^\top \tilde{\dot{\bW}}^{(\tau)} \bb_{j_2} \right. \\
        &\quad \left. \times \I\left(|\ddot{z}^{(\tau)}_{i_1,j_2}| > d(s)\right){ \I(F_0)}\right)\\
		&\le \E\left( \be_{i_1}^\top \tilde{\bW}^{(\tau)} \bb_{j_1}\be_{i_1}^\top \tilde{\bW}^{(\tau)}\bb_{j_2} \right) \E \left( \be_{i_1}^\top \tilde{\dot{\bW}}^{(\tau)} \bb_{j_1}\be_{i_1}^\top \tilde{\dot{\bW}}^{(\tau)}\bb_{j_2}  \right)  \Pr_0\left\{|\ddot{z}^{(\tau)}_{i_1,j_1}| > d(s), { F_0} \right\}.
	\end{aligned}
	$$
	Moreover, it holds that 
	\begin{equation}\label{march:old:2}
	 \begin{aligned}
		\E\left( \be_{i_1}^\top \tilde{\bW}^{(\tau)} \bb_{j_1}\be_{i_1}^\top \tilde{\bW}^{(\tau)}\bb_{j_2} \right)=& \E\left( \left( \sum_{l \in [n]} \tilde{W}_{i_1,l}^{(\tau)}b_{j_1}(l) \right) \left( \sum_{l \in [n]} \tilde{W}^{(\tau)}_{i_1,l}b_{j_2}(l) \right)\right)\\
		\le & \sum_{l \in [n]} \E\left( (\tilde{W}_{i_1,l}^{(\tau)})^2 b_{j_1}(l) b_{j_2}(l)\right) \lesssim \frac{1}{n^2}.
	\end{aligned}   
	\end{equation}
    Combining (\ref{march:old:2}) and \textit{Part 1)} leads to 
	$$
    \begin{aligned}
	& \Cov\left(\be_{i_1}^\top \tilde{\bW}^{(\tau)} \bb_{j_1}\be_{i_1}^\top \tilde{\dot{\bW}}^{(\tau)} \bb_{j_2}\I\left(|\ddot{z}^{(\tau)}_{i_1,j_1}| > d(s)\right){ \I(F_0)}, \be_{i_1}^\top \tilde{\bW}^{(\tau)} \bb_{j_2}\be_{i_1}^\top \tilde{\dot{\bW}}^{(\tau)} \bb_{j_1} \right. \\
        &\quad \left. \times \I\left(|\ddot{z}^{(\tau)}_{i_2,j_2}| > d(s)\right) { \I(F_0)}\right)\\
    &= \frac{s}{n^5}.
    \end{aligned}
	$$
    Hence, we can obtain that 
	\begin{equation}\label{Oct19:1}
		e_2 \lesssim s^3 \frac{s}{n^5} =O\left(\frac{s^4}{n^5}\right),
	\end{equation}
    since the cardinality of $\{(i_1, j_1, j_2):(i_1, j_1) \in \calG \times \calG, (i_1, j_2) \in \calG \times \calG,  j_1 \ne j_2\}$ is at most of order $s^3$.
    Similarly, for term $e_3$ above, we can show that 
    \begin{equation} \label{new.eq.FL015}
    e_3 = O(\frac{s^4}{n^5}).
    \end{equation}
	
    For term $e_4$ above, 
    observe that 
	$$
	\begin{aligned}
		&\Cov\left(\be_{i_1}^\top \tilde{\bW}^{(\tau)} \bb_{j_1}\be_{i_1}^\top \tilde{\dot{\bW}}^{(\tau)} \bb_{j_1}\I\left(|\ddot{z}^{(\tau)}_{i_1,j_1}| > d(s)\right){ \I(F_0)}, \, \be_{i_2}^\top \tilde{\bW}^{(\tau)} \bb_{j_2}\be_{i_2}^\top \tilde{\dot{\bW}}^{(\tau)} \bb_{j_2} \right. \\
        &\quad \left. \times \I\left(|\ddot{z}^{(\tau)}_{i_2,j_2}| > d(s)\right){ \I(F_0)}\right)\\
		&\le \E\left( \be_{i_1}^\top \tilde{\bW}^{(\tau)} \bb_{j_1}\be_{i_2}^\top \tilde{\bW}^{(\tau)}\bb_{j_2} \right) \E \left( \be_{i_1}^\top \tilde{\dot{\bW}}^{(\tau)} \bb_{j_1}\be_{i_2}^\top \tilde{\dot{\bW}}^{(\tau)}\bb_{j_2}  \right)  \\
        &\quad \times \Pr_0\left\{|\ddot{z}^{(\tau)}_{i_1,j_1}| > d(s); |\ddot{z}^{(\tau)}_{i_2,j_2}| > d(s), { F_0}\right\}.
	\end{aligned}
	$$
	Further, it holds that 
	\begin{equation}\label{march:old:3}
	   \begin{aligned}
		\E\left( \be_{i_1}^\top \tilde{\bW}^{(\tau)} \bb_{j_1}\be_{i_2}^\top \tilde{\bW}^{(\tau)}\bb_{j_2} \right)=& \E\left( \left( \sum_{l \in [n]} \tilde{W}_{i_1,l}^{(\tau)}b_{j_1}(l) \right) \left( \sum_{l \in [n]} \tilde{W}^{(\tau)}_{i_2,l}b_{j_2}(l) \right)\right)\\
		\lesssim &  \E\left( (\tilde{W}^{(\tau)}_{i_1,i_2})^2 b_{j_1}(i_2) b_{j_2}(i_1)\right) = \frac{1}{n^3}.
	\end{aligned} 
	\end{equation}
    Combining (\ref{march:old:3}) and \textit{Part 2)} yields that 
	$$
    \begin{aligned}
	& \Cov\left(\be_{i_1}^\top \tilde{\bW}^{(\tau)} \bb_{j_1}\be_{i_2}^\top \tilde{\dot{\bW}}^{(\tau)} \bb_{j_2}\I\left(|\ddot{z}^{(\tau)}_{i_1,j_1}| > d(s)\right){ \I(F_0)}, \be_{i_1}^\top \tilde{\bW}^{(\tau)} \bb_{j_1}\be_{i_2}^\top \tilde{\dot{\bW}}^{(\tau)} \bb_{j_2} \right. \\
        &\quad \left. \times \I\left(|\ddot{z}^{(\tau)}_{i_2,j_2}| > d(s)\right){ \I(F_0)}\right) \\
    & = \frac{s^2}{n^8}.
    \end{aligned}
	$$
	Hence, it follows that 
	\begin{equation}\label{Oct19:2}
		e_4 \lesssim s^4 \frac{s^2}{n^8} = o(\frac{s^4}{n^5}).
	\end{equation}
	Therefore, combining \eqref{Oct19:3}, \eqref{Oct19:1}, (\ref{new.eq.FL015}), and \eqref{Oct19:2} gives the first result. 

     To show the second result, we can decompose the variance as 
     $$
	\begin{aligned}
		& \Var\left\{ \sum_{1 \le i < j \le n} \be_i^\top \tilde{\bW}^{(\tau)} \bb_j\be_i^\top \tilde{\dot{\bW}}^{(\tau)} \bb_j \I\left(|\ddot{z}_{i,j}^{(\tau)}| > d(s)\right){ \I(F_0)}\right\}\\
		& \le \sum_{(i,j) \in [n] \times [n]}  \Var\left(\be_i^\top \tilde{\bW}^{(\tau)} \bb_j\be_i^\top \tilde{\dot{\bW}}^{(\tau)} \bb_j \I\left(|\ddot{z}_{i,j}^{(\tau)}| > d(s)\right){ \I(F_0)}\right) \\
		&\quad+ \sum_{(i_1,j_1) \in [n] \times [n]} \sum_{(i_1,j_2) \in [n] \times [n], j_1 \ne j_2}\Cov\Big(\be_{i_1}^\top \tilde{\bW}^{(\tau)} \bb_{j_1} \be_{i_1}^\top \tilde{\dot{\bW}}^{(\tau)} \bb_{j_1}\I\left(|\ddot{z}_{i_1,j_1}^{(\tau)}| > d(s)\right){ \I(F_0)},\\
		& \quad \be_{i_1}^\top \tilde{\bW}^{(\tau)} \bb_{j_2}\be_{i_1}^\top \tilde{\dot{\bW}}^{(\tau)} \bb_{j_2} \I\left(|\ddot{z}_{i_2,j_2}^{(\tau)}| > d(s)\right){ \I(F_0)}\Big) \\
        &\quad+ \sum_{(i_1,j_1) \in [n] \times [n]} \sum_{(i_2,j_1) \in [n] \times [n], i_1 \ne i_2}\Cov\Big(\be_{i_1}^\top \tilde{\bW}^{(\tau)} \bb_{j_1} \be_{i_1}^\top \tilde{\dot{\bW}}^{(\tau)} \bb_{j_1}\I\left(|\ddot{z}_{i_1,j_1}^{(\tau)}| > d(s)\right){ \I(F_0)},\\
		& \quad \be_{i_2}^\top \tilde{\bW}^{(\tau)} \bb_{j_1}\be_{i_2}^\top \tilde{\dot{\bW}}^{(\tau)} \bb_{j_1} \I\left(|\ddot{z}_{i_2,j_1}^{(\tau)}| > d(s)\right){ \I(F_0)}\Big) \\
		&\quad+ \sum_{(i_1, j_1) \in [n] \times [n]} \sum_{(i_2, j_2) \in [n] \times [n], i_1 \ne i_2, j_1 \ne  j_2}\Cov\Big(\be_{i_1}^\top \tilde{\bW}^{(\tau)} \bb_{j_1} \be_{i_1}^\top \tilde{\dot{\bW}}^{(\tau)} \bb_{j_1}\I\left(|\ddot{z}_{i_1,j_1}^{(\tau)}| > d(s)\right){ \I(F_0)},\\
		& \quad \be_{i_2}^\top \tilde{\bW}^{(\tau)} \bb_{j_2}\be_{i_2}^\top \tilde{\dot{\bW}}^{(\tau)} \bb_{j_2} \I\left(|\ddot{z}_{i_2,j_2}^{(\tau)}| > d(s)\right){ \I(F_0)}\Big) \\
		&:= m_{1} + m_{2} + m_3 + m_4.
	\end{aligned}
	$$
    We will consider the four terms on the right-hand side above separately.
    
For term $m_1$ above, similar to \eqref{Oct19:3}, it holds that 
\begin{equation}\label{equ:june16:11}
\begin{aligned}
	m_{1} & =\sum_{(i,j) \in [n] \times [n]}\Var\left(\be_i^\top \tilde{\bW}^{(\tau)} \bb_j\be_i^\top \tilde{\dot{\bW}}^{(\tau)} \bb_j \I\left(|\ddot{z}^{(\tau)}_{i,j}| > d(s)\right) { \I(F_0)}\right) \\
    & \lesssim n^2 \cdot \frac{1}{n^2} \cdot \frac{1}{n^2} \cdot \frac{s}{n} =\frac{s}{n^3} = o\left(\frac{s}{n^2}\right).
    \end{aligned}
	\end{equation}
 For term $m_2$ above, an application of similar arguments as for \eqref{Oct19:1} leads to 
	\begin{equation}\label{equ:june16:12}
		m_2 \lesssim n^3 \frac{s}{n^5} =O\left(\frac{s}{n^2}\right),
	\end{equation}
    since the cardinality of $\{(i_1, j_1, j_2):(i_1, j_1) \in [n] \times [n], (i_1, j_2) \in [n] \times [n],  j_1 \ne j_2\}$ is at most of order $n^3$. 
Similarly, for term $m_3$ above, we have that 
\begin{equation} \label{new.eq.FL016}
m_3 = O\left(\frac{s}{n^2}\right). 
\end{equation}
For term $m_4$ above, similar to \eqref{Oct19:2}, we can deduce that 
	\begin{equation}\label{equ:june16:13}
		m_4 \lesssim n^4 \frac{s^2}{n^8} = o(\frac{s}{n^2}).
	\end{equation}
    Thus, combining \eqref{equ:june16:11}, \eqref{equ:june16:12}, (\ref{new.eq.FL016}), and \eqref{equ:june16:13} yields the second result.

The proofs of the results for the case of {$s \sim n$ with $\delta_{i,j}=0$ (under null)} and  $\delta_{i,j} \ne 0$ (under the alternative) are similar to those of the first and second results above, where we only need to replace a conservative probability bound with $1$. Such adjustment introduces no additional scaling factor of $s/n$. The remaining steps are similar. This concludes the proof of Lemma \ref{lem:gen2}.

\begin{lemma}\label{lemma:moment_arbitrary_S}
\begin{itemize}
For any subset $\calS \subset [n] \times [n]$, the following bounds hold for $l=1,2$:
    \item[1)] 
    $$
    \Var\left(\sum_{(i,j) \in \calS} \theta_{i,j} \be_i^\top \tilde{\bW}^{(\tau,l)} \bb^{(l)}_j\right)  = O\left(\frac{1}{n}\left(\sum_{(i,j) \in \calS} \theta^2_{i,j} \right)\right).
    $$
    \item[2)]
    $$
\Var\left(\sum_{(i,j) \in \calS} \be_i^\top \tilde{\bW}^{(\tau,l)} \bb^{(l)}_j \be_i^\top \tilde{\dot{\bW}}^{(\tau,l)} \bb^{(l)}_j\right) = O\left(\frac{|\calS|}{n^3}\right).
$$
\end{itemize}
\end{lemma}

\noindent\textit{Proof}.
\textbf{Part 1)}. 
For a given subset $\calS \subset [n] \times [n]$, define a projection set 
$$\calS_x := \{i \in [n]: (i,j) \in \calS\}.$$ 
For each given index $i$, denote by 
$$\calS[i]:=\{j \in [n]: (i,j) \in \calS\}.$$ 
Similar to \eqref{add:lem1:equ4} and \eqref{add:lem1:equv1:1}, we can deduce that 
 $$
 \begin{aligned}
   &\Var\left(\sum_{(i,j) \in \calS} \theta_{i,j} \be_i^\top \tilde{\bW}^{(\tau, l)} \bb^{(l)}_j\right) \\
    \lesssim&  \sum_{(i_1, j_1) \in \calS} \sum_{(i_1, j_2) \in \calS} \frac{\theta_{i_1, j_1} \theta_{i_1, j_2}}{n^2}  + \sum_{(i_1, j_1) \in \calS} \sum_{(i_2, j_2) \in \calS; i_1 \ne i_2} \frac{\theta_{i_1, j_1} \theta_{i_2, j_2}}{n^3}\\
    \le & \frac{1}{2n^2} \sum_{(i_1, j_1) \in \calS} \sum_{(i_1, j_2) \in \calS} (\theta^2_{i_1,j_1} +\theta^2_{i_1,j_2} ) + \frac{1}{2 n^3} \sum_{(i_1, j_1) \in \calS} \sum_{(i_2, j_2) \in \calS; i_1 \ne i_2} (\theta^2_{i_1,j_1} + \theta^2_{i_2,j_2})\\
    \le & \frac{1}{2n^2} \sum_{i_1 \in \calS_x} (\max_{i \in [n]}|\calS[i]|)(\sum_{j_1 \in \calS[i]} \theta^2_{i_1, j_1}+ \sum_{j_2 \in \calS[i]} \theta^2_{i_1, j_2}) + \frac{|\calS|}{n^3}\left(\sum_{(i,j) \in \calS} \theta^2_{i,j} \right)\\
     \le & \frac{1}{n} \sum_{i\in\calS_x} \sum_{j \in \calS[i]} \theta^2_{i,j} +  \frac{1}{n}\left(\sum_{(i,j) \in \calS} \theta^2_{i,j} \right)\\
    \le & \frac{1}{n} \sum_{(i,j)\in\calS} \theta^2_{i,j} +  \frac{1}{n}\left(\sum_{(i,j) \in \calS} \theta^2_{i,j} \right) \lesssim \frac{1}{n}\left(\sum_{(i,j) \in \calS} \theta^2_{i,j} \right),
 \end{aligned}
 $$
 where the second last step has utilized $\max |\calS[i]| \le n$ and $|\calS| \le n^2$. This establishes the first result.
 

\medskip

      \textbf{Part 2)}. Note that we can decompose the variance as
    \begin{equation}\label{march16:equ03}
     \begin{aligned}
		& \Var\left\{ \sum_{(i,j) \in \calS} \be_i^\top \tilde{\bW}^{(\tau,l)} \bb^{(l)}_j\be_i^\top \tilde{\dot{\bW}}^{(\tau,l)} \bb^{(l)}_j \right\} 
	    \le \sum_{(i,j) \in \calS}  \Var\left(\be_i^\top \tilde{\bW}^{(\tau,l)} \bb^{(l)}_j\be_i^\top \tilde{\dot{\bW}}^{(\tau,l)} \bb^{(l)}_j\right) \\
		&\quad + \sum_{(i_1,j_1) \in \calS} \sum_{(i_1,j_2) \in \calS, j_1 \ne j_2}\Cov\Big(\be_{i_1}^\top \tilde{\bW}^{(\tau,l)} \bb^{(l)}_{j_1} \be_{i_1}^\top \tilde{\dot{\bW}}^{(\tau,l)} \bb^{(l)}_{j_1}, \\
        &\quad \be_{i_1}^\top \tilde{\bW}^{(\tau,l)} \bb_{j_2} \be_{i_1}^\top \tilde{\dot{\bW}}^{(\tau,l)} \bb^{(l)}_{j_2} \Big) \\
        &\quad + \sum_{(i_1,j_1) \in \calS} \sum_{(i_2,j_1) \in \calS, i_1 \ne i_2}\Cov\Big(\be_{i_1}^\top \tilde{\bW}^{(\tau,l)} \bb^{(l)}_{j_1} \be_{i_1}^\top \tilde{\dot{\bW}}^{(\tau,l)} \bb^{(l)}_{j_1}, \\
        &\quad \be_{i_2}^\top \tilde{\bW}^{(\tau)} \bb^{(l)}_{j_1} \be_{i_2}^\top \tilde{\dot{\bW}}^{(\tau,l)} \bb^{(l)}_{j_1}\Big) \\
		&\quad + \sum_{(i_1, j_1) \in \calS} \sum_{(i_2, j_2) \in \calS, i_1 \ne  i_2, j_1 \ne j_2}\Cov\Big(\be_{i_1}^\top \tilde{\bW}^{(\tau,l)} \bb^{(l)}_{j_1} \be_{i_1}^\top \tilde{\dot{\bW}}^{(\tau,l)} \bb^{(l)}_{j_1}, \\
        &\quad \be_{i_2}^\top \tilde{\bW}^{(\tau,l)} \bb^{(l)}_{j_2} \be_{i_2}^\top \tilde{\dot{\bW}}^{(\tau,l)} \bb^{(l)}_{j_2}\Big).
	\end{aligned}   
    \end{equation}
    
Then with the aid of \eqref{march:old:1},  \eqref{march:old:2}, and \eqref{march:old:3}, we can obtain that  
 $$
 \begin{aligned}
     \eqref{march16:equ03} & = O\left(\sum_{(i,j) \in \calS} \frac{1}{n^4} + \sum_{(i_1,j_1) \in \calS} \sum_{(i_1,j_2) \in \calS, j_1 \ne j_2} \frac{2}{n^4} \right. \\
        &\quad \left. +  \sum_{(i_1, j_1) \in \calS} \sum_{(i_2, j_2) \in \calS, (i_1,j_1) \ne (i_2, j_2)}\frac{1}{n^6}\right)\\
     & = O\left( \frac{|\calS|}{n^4} + \frac{n|\calS|}{n^4} + \frac{|\calS|^2}{n^6}\right) = O\left( \frac{|\calS|}{n^3}\right).
 \end{aligned}
 $$
This establishes the second result and completes the proof of Lemma \ref{lemma:moment_arbitrary_S}.

\end{document}

%% file: definition.tex
\def\htau{\hat{\tau}}
\def\calI{\mathcal{I}}
\def\calT{\mathcal{T}}
\def\calB{\mathcal{B}}
\def\by{\mathbf{y}}
\def\bX{\mathbf{X}}
\def\beps{\boldsymbol{\epsilon}}
\def\btheta{\boldsymbol{\theta}}
\def\bM{\mathbf{M}}
\def\bI{\mathbf{I}}
\def\tbX{\tilde{\mathbf{X}}}
\def\bSigma{\boldsymbol{\Sigma}}
\def\bs{\mathbf{s}}
\def\bS{\mathbf{S}}
\def\tbU{\tilde{\mathbf{U}}}
\def\calG{\mathcal{G}}
\def\tbtheta{\tilde{\boldsymbol{\theta}}}
\def\bw{\mathbf{w}}
\def\bZ{\mathbf{Z}}
\def\bP{\mathbf{P}}
\def\hgamma{\hat{\gamma}}
\def\supp{\operatorname{supp}}
\def\calC{\mathcal{C}}
\def\hbSigma{\widehat{\boldsymbol{\Sigma}}}
\def\hsigma{\hat{\sigma}}
\def\hcalA{\widehat{\mathcal{A}}}
\def\calS{\mathcal{S}}
\def\calH{\mathcal{H}}
\def\hbbeta{\hat{\boldsymbol{\beta}}}
\def\tbbeta{\tilde{\boldsymbol{\beta}}}
\def\htbeta{\hat{\tilde{\boldsymbol{\beta}}}}
\def\tr{\operatorname{tr}}
\def\calA{\mathcal{A}}
\def\hcalS{\hat{\mathcal{S}}}
\def\httheta{\hat{\tilde{\boldsymbol{\theta}}}}
\def\hbtheta{\hat{\boldsymbol{\theta}}}
\def\hpi{\hat{\pi}}
\def\bX{\mathbf{X}}
\def\calZ{\mathcal{Z}}
\def\calY{\mathcal{Y}}
\def\bx{\mathbf{x}}
\def\calE{\mathcal{E}}
\def\calM{\mathcal{M}}
\def\bM{\mathbf{M}}
\def\bm{\mathbf{m}}
\def\bo{\mathbf{o}}
\def\be{\mathbf{e}}
\def\bu{\mathbf{u}}
\def\bv{\mathbf{v}}
\def\bvartheta{\boldsymbol{\vartheta}}
\def\bO{\mathbf{O}}
\def\bE{\mathbf{E}}
\def\bV{\mathbf{V}}
\def\bU{\mathbf{U}}
\def\tbz{\tilde{\mathbf{z}}}
\def\bGamma{\boldsymbol{\Gamma}}
\def\tbeps{\tilde{\boldsymbol{\epsilon}}}
\def\calV{\mathcal{V}}
\def\teps{\tilde{\epsilon}}
\def\calL{\mathcal{L}}
\def\bzeta{\boldsymbol{\zeta}}
\def\bc{\mathbf{c}}
\def\P{\operatorname*{P}}
\def\E{\mathbb{E}}
\def\Var{\operatorname*{Var}}
\def\Cov{\operatorname*{Cov}}
\def\Cor{\operatorname*{Cor}}
\def\sign{\operatorname*{sign}}
\def\R{\operatorname*{\mathbb{R}}}
\def\F{\operatorname*{\mathcal{F}}}
\def\Rank{\operatorname*{Rank}}
\def\diag{\operatorname*{diag}}
\def\swap{\operatorname*{swap}}
\def\FDR{\operatorname{FDR}}
\def\Power{\operatorname{Power}}
\def\Pr{\mathbb{P}}
\def\D{\operatorname*{D}}
\def\I{\mathbb{I}}
\def\FDP{\operatorname{FDP}}
\allowdisplaybreaks[4]
\def\red{\color{red}}
\def\blue{\color{blue}}
\def\cyan{\color{cyan}}
\newcommand{\bbeta}{\boldsymbol{\beta}}
\def\mb{\mathbf{b}}
\def\bD{\mathbf{D}}
\def\bd{\mathbf{d}}
\def\htheta{\hat{\theta}}
\def\ttheta{\tilde{\theta}}
\def\bmu{\boldsymbol{\mu}}
\def\tby{\tilde{\mathbf{y}}}
\def\ba{\mathbf{a}}
\def\bv{\mathbf{v}}
\def\bA{\mathbf{A}}
\def\bT{\mathbf{T}}
\def\tbM{\hat{\mathbf{M}}}
\def\hJ{\hat{J}}
\def\bE{\mathbf{E}}
\def\bW{\mathbf{W}}
\def\bgamma{\boldsymbol{\gamma}}
\def\hbv{\hat{\mathbf{v}}}
\def\bY{\mathbf{Y}}
\def\bz{\mathbf{z}}
\def\hbM{\hat{\mathbf{M}}}
\def\hG{\hat{G}}
\def\hV{\hat{V}}
\def\vec{\operatorname{vec}}
\def\bdel{\boldsymbol{\delta}}
\def\hK{\hat{K}}
\def\tG{\tilde{G}}
\def\PWER{\operatorname{PWER}}
\def\tbc{\tilde{\mathbf{c}}}
\def\bg{\mathbf{g}}
\def\PFER{\operatorname{PFER}}
\def\CUS{\operatorname{CUS}}
\def\FD{\operatorname{FD}}
\def\TD{\operatorname{TD}}
\def\hSigma{\operatorname{\hat{\Sigma}}}
\def\hGamma{\operatorname{\hat{\Gamma}}}
\def\CV{\operatorname{CV}}
\def\bQ{\mathbf{Q}}
\def\te{\operatorname{te}}
\def\bb{\mathbf{b}}
\def\vech{\operatorname{vech}}
\def\calN{\mathcal{N}}
\def\calG{\mathcal{G}}
\def\tbZ{\tilde{\mathbf{Z}}}
\def\tz{\tilde{z}}
\def\tbE{\tilde{\mathbf{E}}}
\def\htheta{\hat{\theta}}
\def\he{\hat{e}}
\def\hz{\hat{z}}
\def\htz{\hat{\tilde{z}}}